\documentclass{article}

\usepackage[letterpaper,margin=3cm,footskip=30pt]{geometry}
\usepackage{amsmath}
\usepackage{amsfonts}
\usepackage{amsthm}
\usepackage{paralist}
\usepackage{booktabs}
\usepackage{graphicx}
\usepackage{array}
\usepackage{multirow}
\usepackage{multicol}
\usepackage{xcolor}
\usepackage{xspace}
\usepackage{hyperref}
\usepackage{balance}
\usepackage[most]{tcolorbox}
\usepackage{caption}
\usepackage{float}
\usepackage{subcaption}
\usepackage{listings}
\usepackage{url}
\usepackage{authblk}
\usepackage{soul}
\usepackage{comment}
\usepackage{bbding}
\usepackage{amsmath}

\definecolor{lightred}{RGB}{255, 200, 200}
\definecolor{lightgreen}{RGB}{200, 255, 200}
\definecolor{deepviolet}{RGB}{96, 0, 96}
\newcommand{\highlightred}[1]{\sethlcolor{lightred}\hl{#1}}
\newcommand{\highlightgreen}[1]{\sethlcolor{lightgreen}\hl{#1}}
\tcbset{
  promptbox/.style={
    top=10pt,
    colback=white,
    colframe=violet,
    colbacktitle=deepviolet,
    fontupper=\itshape,
    enhanced,
    center,
    attach boxed title to top left={yshift=-0.1in,xshift=0.15in},
    boxed title style={boxrule=0pt,colframe=white},
  }
}
\newtcolorbox{prompt}[2][]{promptbox,title=#2,#1}
\newtheorem{example}{Example}

\newcommand{\myparagraph}[1]{\vspace{1ex}\noindent\underline{\it #1.}\xspace}

\newcommand{\set}[1]{\{#1\}\xspace}
\newcommand{\str}[1]{\textit{#1}}
\newcommand{\var}[1]{\texttt{\set{\upshape #1}}}
\newcommand{\group}[1]{\textsf{#1}\xspace}

\begin{document}

\title{Smart Agent-Based Modeling: On the Use of Large Language Models in Computer Simulations}







\newcommand{\osaka}{$^{1}$}
\newcommand{\umich}{$^{2}$}
\newcommand{\fordham}{$^{3}$}
\newcommand{\kyoto}{$^{4}$}
\newcommand{\osanag}{$^{1,5}$}
\newcommand{\osakyo}{$^{1,4}$}
\newcommand{\nagoya}{$^{5}$}

\author{
{\osakyo Zengqing Wu$^{\dag}$, \umich Run Peng, \fordham Xu Han,} \par
\vspace{0.3ex}
{\osakyo Shuyuan Zheng, \kyoto Yixin Zhang, \osanag Chuan Xiao \Envelope}\\
\small{\osaka Osaka University, \umich University of Michigan, \fordham Fordham University, \kyoto Kyoto University, \nagoya Nagoya University}\\
\vspace{1ex}
\small{wuzengqing@outlook.com, roihn@umich.edu, xhan44@fordham.edu, zheng@ist.osaka-u.ac.jp, zhangyx@dl.soc.i.kyoto-u.ac.jp, chuanx@nagoya-u.jp}
}



\date{}

\maketitle

\begin{figure}[H]
    \centering
    \includegraphics[width =.8\textwidth]{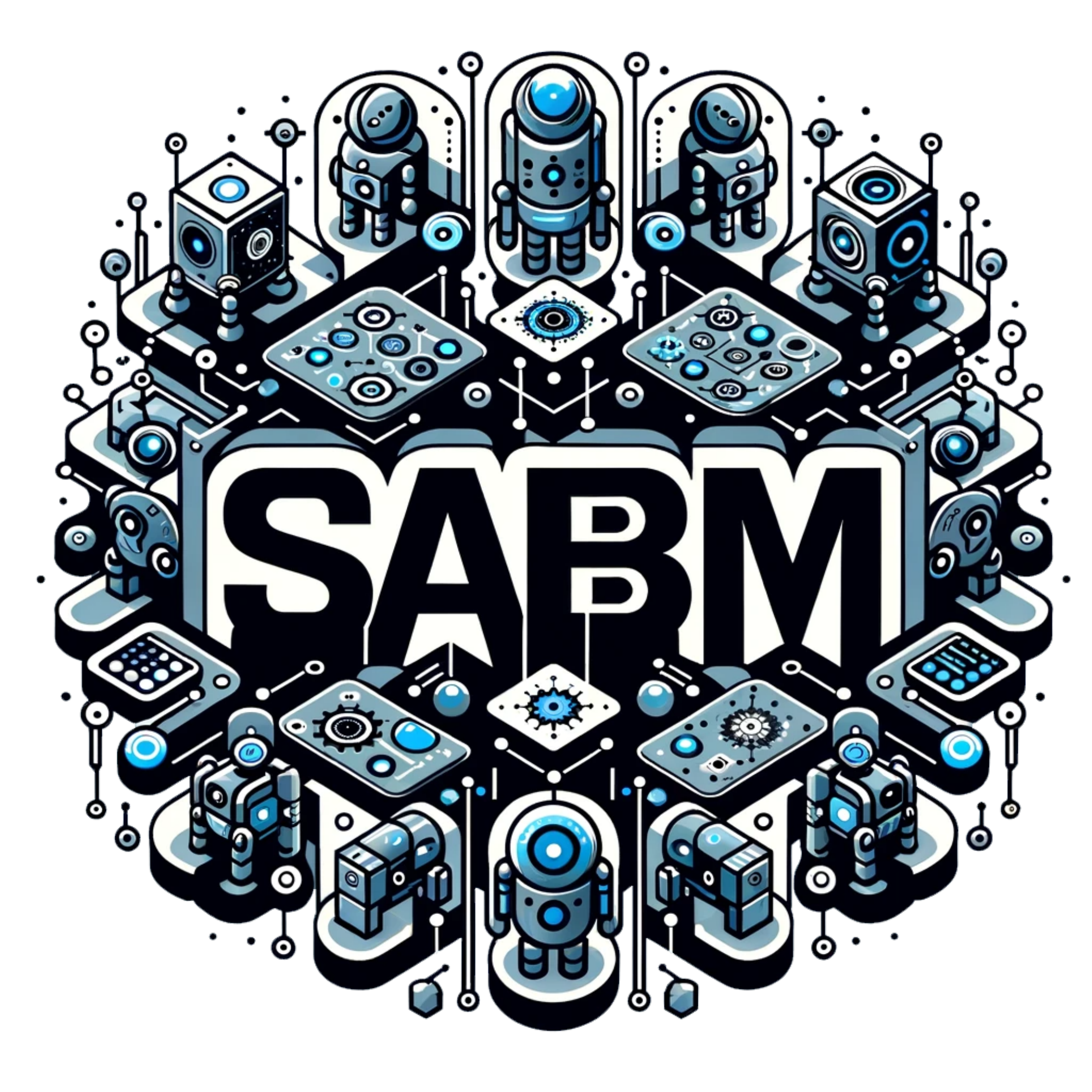}
    \captionsetup{labelformat=empty}
    \label{fig:logo}
\end{figure}

\def\thefootnote{$^{\dag}$}\footnotetext{Work done while studying at Osaka University.}
\newpage

\begin{abstract}
  Computer simulations offer a robust toolset for exploring complex systems across various disciplines. A particularly impactful approach within this realm is Agent-Based Modeling (ABM), which harnesses the interactions of individual agents to emulate intricate system dynamics. ABM's strength lies in its bottom-up methodology, illuminating emergent phenomena by modeling the behaviors of individual components of a system. Yet, ABM has its own set of challenges, notably its struggle with modeling natural language instructions and common sense in mathematical equations or rules. This paper seeks to transcend these boundaries by integrating Large Language Models (LLMs) like GPT into ABM. This amalgamation gives birth to a novel framework, Smart Agent-Based Modeling (SABM). Building upon the concept of smart agents -- entities characterized by their intelligence, adaptability, and computation ability -- we explore in the direction of utilizing LLM-powered agents to simulate real-world scenarios with increased nuance and realism. In this comprehensive exploration, we elucidate the state of the art of ABM, introduce SABM's potential and methodology, and present three case studies~\footnote{The source codes of our case studies are available at \url{https://github.com/Roihn/SABM} .}, demonstrating the SABM methodology and validating its effectiveness in modeling real-world systems. Furthermore, we cast a vision towards several aspects of the future of SABM, anticipating a broader horizon for its applications. Through this endeavor, we aspire to redefine the boundaries of computer simulations, enabling a more profound understanding of complex systems.
\end{abstract}




\hypersetup{linktocpage}
\tableofcontents
\newpage

\section{Introduction}
\label{sec:intro}

\begin{figure}[!t]
    \centering
    \includegraphics[width =.7\textwidth]{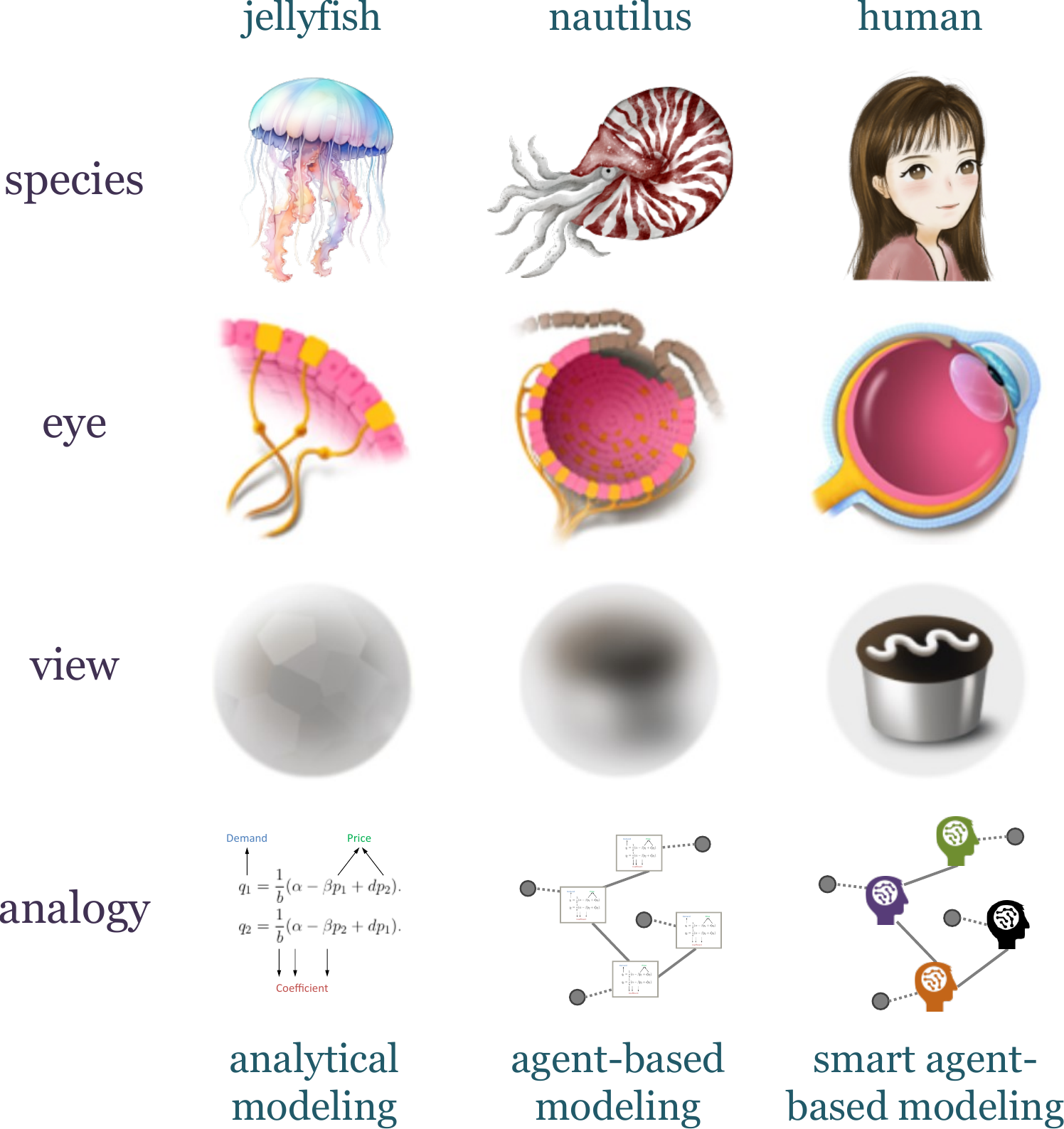}
    \caption{Evolution of vision. Jellyfish have the simplest 
    visual system, sensing only light patches. Nautilus has a more complex visual system, 
    seeing in low resolution. The human visual system is the most complex and able 
    to construct a high-resolution image. Perceiving the real world with the three visual systems 
    is similar to modeling real-world processes with analytical modeling, agent-based modeling, 
    and smart agent-based modeling, respectively, with the increasing fine-grained features as well as complexities of real-world interpretations. 
    Source of images: \cite{visual-capitalist}.}
    \label{fig:intro:keystone}
\end{figure}

\subsection{Keystone Story: From Sight to Insight -- Transcending Boundaries with Large Language Models in Agent-Based Modeling}

We perceive the world in different ways. Our individual perspectives -- shaped by experience, understanding, and the tools we possess to observe -- impact how we perceive, comprehend, and interact with the world. Just as an organism's visual system influences its interaction with the environment, the lens through which we view complex systems defines our understanding and ability to predict, control, and shape them. As depicted in Figure~\ref{fig:intro:keystone}, by drawing a parallel to the evolution of vision from a pigment spot to a comprehensive visual system, we can better understand the progression of modeling complex systems.

Pigment spot ocelli, as in jellyfish, sensing just light and dark and basically telling where the light comes from, resemble analytical models, which represent the system in terms of a set of mathematical equations. Much like how pigment spots can discern the presence and source of light, these models can capture a basic sense of the world, focusing on specific parameters. They are simple, abstract, and lack interactivity. They can describe general patterns and basic cause-and-effect relationships, but are inherently limited in their ability to describe the complexity and nuance of the real world.

Imagine a scenario where vision evolves to the stage of pinholes, much like a nautilus's, capable of perceiving low-resolution images. This parallels the development of agent-based modeling (ABM), which represents the system through the actions and interactions of autonomous agents specified by decision-making and learning rules. Like the light-sensitive cells in pinholes, each agent contributes to a part of the overall image. This approach allows for more nuanced interactions and a better sense of the system's structure, but the resultant image remains blurred and lacks details.

Now, envision the sophisticated human visual system -- complete with not just lens eyes, but a signal processing mechanism in the brain, translating the captured light into a high-definition, multi-dimensional, full-color image. This is where we draw an analogy to the use of large language models (LLMs) -- intelligent, adaptive, inherently sophisticated but easy to interact with -- in modeling agents and other ABM components. Just as the visual system makes sense of the nuances of light and color, turning them into a coherent visual representation, the use of LLMs in ABM has the potential to provide a rich, detailed simulation of the complex real world. By accounting for the interactivity and the emergent behaviors of the agents, the LLM-powered ABM offers a far more nuanced and accurate simulation of the world.

Through this paper, we hope to illustrate how LLM-powered ABM, much like the advanced human visual system, can provide a higher dimension, accuracy, and complexity in our understanding and simulations of the world. As we evolve from the pigment-spot eyes to comprehensive visual systems, our ability to comprehend, predict, and influence the complexities of the real world will increase significantly. By exploring and expanding the capabilities of LLMs in ABM, we aim to not just see the world but to truly understand it, to gain insight into the myriad complex systems that govern our world. And, in doing so, we hope to empower humanity to better navigate, predict, and shape our collective future.

\subsection{Motivation} 

The advent of computer simulation has revolutionized the way we approach problem-solving in various disciplines, providing a powerful tool for understanding complex systems. Among the various simulation methodologies, ABM has emerged as a particularly potent tool due to its unique ability to model complex systems through the interaction of individual agents specified by simple rules~\cite{bonabeau2002agent}. 
With this bottom-up construction approach, ABM allows researchers to observe and understand complex phenomena that emerge from these interactions,  
thereby facilitating the development of theories and models. 
ABM has found application in a diverse range of fields and has been instrumental in advancing our understanding of complex systems, and the importance of ABM in these disciplines is underscored by the extensive body of work that has been conducted using this methodology. 
For instance, ABM has been used to model the spread of diseases in epidemiology~\cite{perez2009agent}, to simulate market dynamics in economics~\cite{tesfatsion2002agent}, and to understand the behavior of ecosystems in ecology~\cite{grimm2005pattern}. These studies exemplify the breadth of problems that ABM can address, highlighting its significance in the realm of computer simulation.



While ABM has proven to be a powerful tool in the simulation of complex systems, it is not without its limitations. For example, when modeling advanced human behaviors that involve natural language descriptions and common sense that are hard to express in a formal language, ABM methods tend to ignore these natural language factors or simplify them by heuristics or learning rules. Due to the high complexity of these behaviors, such modeling often results in oversimplification, capturing only a subset of their properties, which can limit their accuracy and reduce the relevance of their findings to real-world scenarios. This is exacerbated by the soft factors of human behaviors that are difficult to quantify, calibrate, and justify~\cite{bonabeau2002agent}. Furthermore, ABM methods often rely heavily on predetermined parameters to define agent behaviors. This approach, while effective in certain scenarios, can introduce researcher bias and high uncertainty in model output, particularly when the parameter settings are not appropriately calibrated~\cite{grimm2005pattern}. 
Therefore, as a critical aspect of many real-world systems, the heterogeneity of agents, though available in ABM methods, demands researchers understand the nuanced implications of parameter settings. For many ABMs, uncovering actionable insight into a phenomenon depends on empirically-grounded heterogeneity specification~\cite{reeves2022structural}, posing considerable challenges in parameter tuning. 

The emergence of LLMs such as Generative Pre-trained Transformers (GPT) presents an opportunity to address these limitations. LLMs, with their ability to process and generate natural language, can be used to mimic the aforementioned advanced human behaviors. By prompt engineering~\cite{prompt-engineering} over LLMs, researchers can design models in natural language, with less effort in choosing parameters to define agent behaviors. 
Moreover, LLMs possess strong reasoning capabilities~\cite{gpt-4-not-just-statistics}, which can be leveraged to improve the model interpretability. By incorporating LLMs into ABM, researchers can create agents that not only follow prescribed rules but also make decisions by learning from hand-crafted examples~\cite{brown2020language} and reasoning instructions~\cite{kojima2022large}. This can enhance the realism of the simulation and provide deeper insights into the mechanisms driving the behavior of the system. By leveraging the learning capabilities of LLMs, computer simulations can be made more adaptive to changes in the system, providing more robust simulations. LLMs can also impersonate different roles~\cite{salewski2023context}, capturing the diversity of agent behaviors seen in real-world systems. Thus, incorporating LLMs into ABM has the potential to improve the performance of ABM methods in simulation. 


\subsection{Research Purpose}
The primary objective of this paper is to enhance the capability of agent-based approaches in formulating theories, hypotheses, and explanations by establishing a bottom-up, natural language description-based computer simulation framework. This paper proposes the modeling framework of \textbf{Smart Agent-Based Modeling (SABM)} -- an innovative agent-based approach that leverages the power of modern AI models, in particular, LLMs, for modeling real-world systems -- and introduces the methodological construction of computer simulations based on this framework. 


The notion of smart agents was proposed by Carley~\cite{carley2002smart} in the context of organizations of the future. In \cite{carley2002smart}, smart agents are defined as entities that are intelligent, adaptive, and computational, and human beings are the canonical smart agents. With the advent of modern AI models, especially LLMs -- imbued with remarkable language and reasoning abilities that emulate human behaviors -- we extend the concept of smart agents to the realm of ABM. We posit that these smart agents, powered by LLMs, can enhance our understanding of complex systems by simulating real-world problems in a more nuanced and realistic manner. 

Besides \emph{modeling in natural language}, SABM paves the way for \emph{a priori modeling}, where it is assumed that LLM-powered agents mimic human behaviors using their common sense and knowledge aligned to humans'~\cite{ji2023ai}. This obviates the need for prescribed rules or parameters, in contrast to ABM, where such rules and parameters are derived from observed human actions \emph{a posteriori}. Moreover, SABM enables a dichotomy of the simulation model into an ``engine'', represented by the LLM, and a ``chassis'', represented by the remainder of the model. As such, enhancing agents' performance in emulating human behaviors and constructing simulation models can be decoupled in the sense that AI researchers can keep improving the engine while researchers studying real-world systems can select an engine and work on the chassis. 

\subsection{Outline}
In Section~\ref{sec:sota}, we begin with an introduction to ABM, highlighting its role in modeling complex systems. A comprehensive review of the state of the art in ABM and its inherent limitations is presented. Additionally, we offer a brief survey of works focusing on LLMs and their uses in agent-based simulations and systems.

In Section~\ref{sec:sabm}, we present SABM, elucidating its salient features in various aspects including language ability, modeling paradigm, adaptability, and interpretability. The potential application domains of SABM are outlined, followed by an exposition of its limitations.

Section~\ref{sec:impl} is dedicated to elaborating the methodology of SABM implementation, spanning task specification, model setup, simulation process, and analysis of outcomes. Many of the methods introduced in this section, which emphasize implementation using natural language, harness the power of prompt engineering to specify agents and their interactions. We employ a simple number-guessing game as a tutorial to demonstrate these methods. 

In Section~\ref{sec:design}, we propose the methodology for designing SABM instances, encompassing phases like task definition, model design, implementation, simulation, and model validation. Notably, we advocate for drafting a fact sheet prior to design, initiating an preliminary design phase for LLM validation, and fragmenting the simulation process into incremental sub-tasks. A discussion on debugging techniques for SABM ensues. Recognizing the pivotal role of sensitivity analysis in evaluating model robustness, we introduce a series of prompt alteration techniques, which can be viewed as a natural language analogue to traditional sensitivity analysis centered on parameter impact.

In Sections~\ref{sec:case-evac} -- \ref{sec:case-firm}, we conduct three case studies -- emergency evacuation, plea bargaining, and firm pricing competition -- rooted in varied research domains. These studies, predicated on prior research either employing ABM or enlisting human participants, showcase the SABM methodology and collectively validate its effectiveness.

In Section~\ref{sec:future}, we discuss the future of SABM in various aspects across theoretical foundations, technological opportunities, and ethical concerns. We envision a paradigm shift with the advent of multimodal SABM, which assimilates visual, auditory, and other forms of information, poised to unleash the full potential of SABM and significantly broaden its application scope. Its applications in simulating civilizations could furnish invaluable insights into our social, cultural, psychological, and linguistic theories, and shed light on the cognitive capabilities of AI, extending even to realms of self-awareness and consciousness. Following these envisions, we conclude this paper by summarizing the key findings as well as limitations of this work in Section~\ref{sec:concl}. 

Our hope is that this paper will serve as a stepping stone towards the integration of LLMs into computer simulation approaches, thereby enhancing their realism, interpretability, and performance. By doing so, we aim to push the boundaries of what is possible in the realm of computer simulation, opening up new avenues for understanding and exploring complex systems. 
\section{State of the Art}
\label{sec:sota}

\subsection{Complex Systems and ABM}
\label{sec:sota:abm}

ABM has emerged as a powerful tool for understanding complex phenomena, particularly in the realm of complex systems~\cite{heath2010some}. We delve into the reasons behind the utility of ABM in dealing with complex systems and underscore the importance of addressing these systems.

\begin{figure}[!t]
    \centering
    \includegraphics[width = 0.9\textwidth]{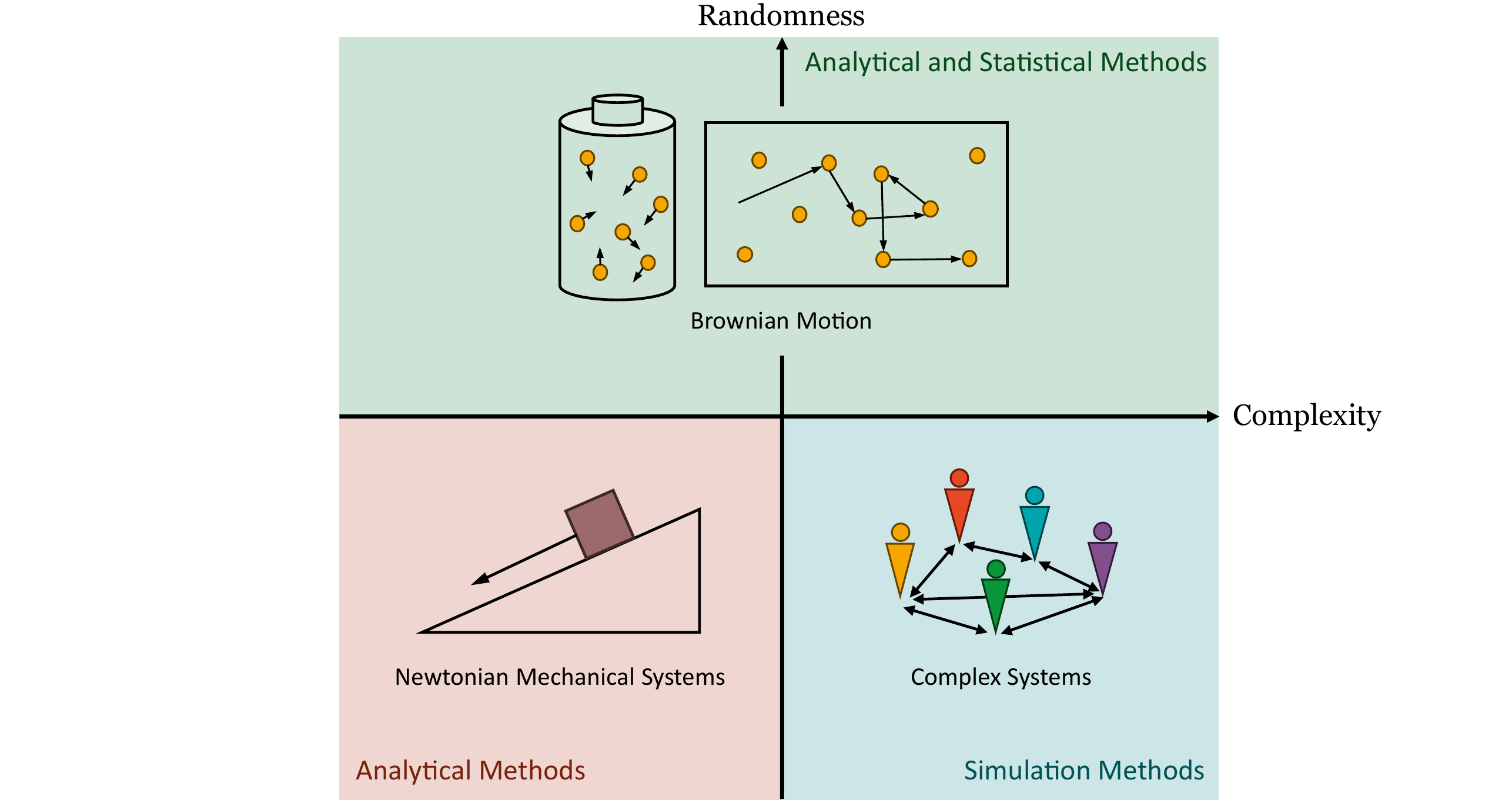}
    \caption{A comparison of systems and corresponding research methods.}
    \label{fig:sota:systems}
\end{figure}

Complex systems, by definition, are characterized by a moderate number of interacting entities or subjects, where the relationships between these subjects are not simple enough to be analyzed using analytical methods, such as Newtonian mechanics~\cite[pp. 1--22]{weinberg2001introduction}. Conversely, these systems also do not have a large enough number of subjects to allow for the behavior of the system to be regressed using statistical methods, e.g., overall behavior of the gas. Thus, in complex systems, understanding or predicting the behavior of the entire system based on individual behavior patterns and rules or statistics can be challenging. This is primarily due to the emergence of phenomena that are not apparent from the behavior of individual subjects but arise from the structure and processes of the system as a whole. For instance, the social structure of an ant colony, which emerges from the interactions between individual ants, cannot be understood by studying the behavior of a single ant in isolation~\cite{fletcher1995modelling}. Interactions between subjects and between subjects and their environment give rise to patterns of behavior in the system that emerge at a level higher than that of the individual subject. Another classic example is the formation of societal regularities, such as norms and price equilibria, which emerges from the interactions between individuals~\cite{epstein1999agent}. Examples of such systems abound in nature and society, from ant colonies to human societies and ecosystems to economies. Figure~\ref{fig:sota:systems} presents the differences between real-world systems and their research methods.

Existing research suggests that this complexity arises, in part, because the subjects involved in complex systems may be intelligent and/or adaptive. They are capable of modifying their behavioral rules and making decisions based on new and local information, thereby giving rise to complex patterns of behavior~\cite{remondino2010learning}. For example, in a market economy, individual buyers and sellers adjust their behavior based on changes in market conditions, leading to complex dynamics at the level of the market as a whole~\cite{tesfatsion2002agent}. Including this example, empirical research, particularly in the social sciences, often faces significant hurdles and reproduction problems due to the complexity of these systems, e.g., social structures and social practices are interrelated and difficult to find cause-and-effect relationships~\cite{giddens1986constitution, giddens1986sociology}. Social phenomena are typically difficult to isolate and define precisely, making research findings challenging to reproduce. These phenomena often involve non-linear, chaotic systems that are sensitive to initial conditions~\cite{gordon1992chaos}. It is hence difficult to deal with the problem using traditional research methods, e.g., top-down statistical methods, or methods that conduct investigations on a localized part of the system analytically. Differences in uncontrolled initial conditions can lead to vastly different results in empirical studies if the research theory does not accurately capture all relevant variables about a given social phenomenon, even if the most severe experimental control measures are taken~\cite{krantz1989social, phillips1972some}. That is, differences in sample selection, experimental procedures, measurement criteria, and other factors can lead to significant variations in results, thereby weakening the validity, generalizability, and reproducibility of the study~\cite{bloombaum1991influence}.

Despite the inherent challenges, understanding complex systems holds significant research and practical implications. As aforementioned, many phenomena that directly impact human well-being, such as socioeconomic issues, are inherently complex systems. To address these challenges, we need recordable comprehensive information about the system, not just parts of it. Meanwhile, this is extremely difficult for studies involving a moderate number of adaptive subjects. For instance, in natural science systems like ecosystems, it is challenging to capture all variables accurately to isolate individual causes and effects for countless interactions occurring simultaneously~\cite{wootton2002indirect}. In certain instances, the precise act of measuring variables can inadvertently alter the inherent properties of the system under observation. This is due to the interference or influence of the observer on the system's natural state. A classic illustration of this is the Hawthorne effect. This phenomenon describes how individuals might adjust or enhance their behavior simply because they are aware of being observed~\cite{sedgwick2015understanding}. In social science systems, the difficulty of experimentally controlling variables increases as the number of samples in the experiment rises, making large-scale social experiments challenging to conduct~\cite{morris2001experimental}. While recently developed computational social science methods have made significant breakthroughs in specific areas, such as social networks and opinion dynamics ~\cite{park2018strength, zhang2020data}, current research does not cover all complex system problems involved in traditional research areas.

\begin{figure}[!t]
    \centering
    \includegraphics[width = 0.5\textwidth]{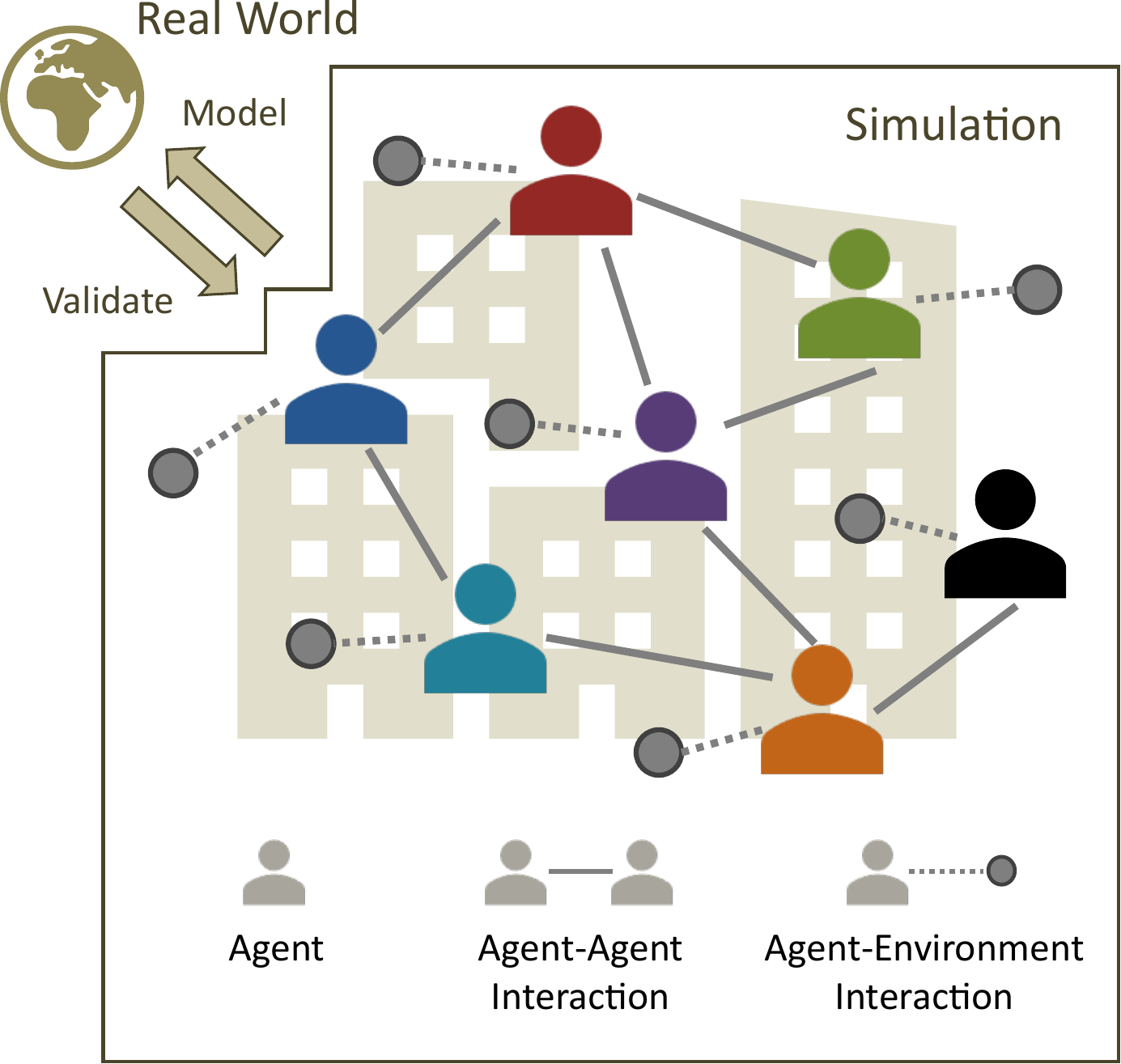}
    \caption{Illustration of ABM.}
    \label{fig:sota:abm}
\end{figure}

In such a context, computer simulation methods provide some effective tools for dealing with complex system problems. Among various computer simulation methods, ABM is a computational approach that simulates high-level phenomena by building bottom-up models for the interactions of the underlying subjects. By utilizing ABM, we can simulate the behaviors among adaptive individuals to comprehend the overall patterns of complex systems and model microphenomena that cannot be analyzed by top-down statistical approaches. Moreover, ABM can circumvent the previously mentioned issue where observers may influence the behavior of the observed for the sake of precise variable measurement. 

ABM, illustrated in Figure~\ref{fig:sota:abm}, simulates the intricate dynamics of complex systems by focusing on individual entities, known as agents~\cite{macal2005tutorial}. Each agent is endowed with unique attributes or states that depict their current circumstances, and they operate based on either fixed, adaptive, or learning-oriented behaviors. This allows them to make decisions, take actions, and potentially evolve over time~\cite{elsenbroich2014agent}. The environment, which can either remain static or evolve, presents conditions, evokes competition, establishes boundaries, and sometimes provides resources that influence the agents' behaviors~\cite{cipi2011simulation}. Interactions, as an important component of ABM, map out the relationships and communications between agents and the environment. These interactions can manifest in various forms: cooperative, where agents collaborate towards a shared goal; competitive, where they might work against each other; or even neutral, where their interactions do not significantly impede or enhance each other's objectives~\cite{elliott2002exploring}. Furthermore, the authenticity and precision of an ABM simulation largely hinge on the parameters and data integrated into it~\cite{squazzoni2005does, hassan2008stepping}. These elements, ranging from initial conditions to empirical datasets, serve as the foundational pillars, ensuring that the modeled scenarios are either reflective of real-world situations or are theoretically sound by characterizing the key components, agents, interactions, and environment of ABM. By detailing each component, ABM offers a bottom-up perspective, enabling researchers to discern macro-level outcomes that emerge from micro-level interactions, thereby providing a holistic understanding of complex systems.

\subsection{Nature of the Problems Targeted by ABM}
\label{sec:sota:scope}
\myparagraph{What System}
Currently, complex systems can be simulated using a variety of developed methods, each with its own strengths and limitations. These methods include microsimulation, system dynamics, and ABM. Microsimulation is a method that predicts the overall expected situation from a real sample. It is particularly useful in situations where individual-level data are available and can provide statistical results based on individual properties and a number of transition probabilities, while its primary limitation is that it does not take into account the interaction between agents~\cite[pp. 13, 58]{gilbert2005simulation}. System dynamics, on the other hand, is a method that focuses on the macroscopic set of variables. It is capable of dealing with large-scale systems within multiple feedback loops and can capture the dynamic cause-and-effect of the variables of the system over time~\cite{azar2012system}. However, as the whole system is the only one agent simulated in system dynamics, system dynamics does not distinguish differences between model subjects from a micro level, and thus difficult to perform individual behavior and adaptability~\cite[pp. 28--30]{gilbert2005simulation}. ABM is particularly useful in situations where the behavior of the system emerges from the interactions of individual agents. However, it also suffers from the difficulty in establishing agents and interactions that match reality, which will be elaborated in Section~\ref{sec:sota:limitations}. 
A comparison of these approaches is given in Table~\ref{tab:sota:cmp-simulation-methods}. Each of these methods has its application scope, and they complement each other in the study of complex systems.

\begin{table}[!t]
  \small
  \centering
  \caption{A comparison of simulation methods for studying complex systems.}
  \begin{tabular}{ c | c | c | c }
    \hline
    \textbf{Simulation Method} & \textbf{Research Scale} & \textbf{Individual Behavior} & \textbf{System State} \\
    \hline
    Microsimulation & Micro & Simple (statistics) & Static \\
    System dynamics & Macro & None (system-level behavior) & Dynamic \\
    ABM & Micro & Complex & Dynamic \\
    \hline
  \end{tabular}
  \label{tab:sota:cmp-simulation-methods}
\end{table}

It is not difficult to anticipate that microsimulation can achieve approximate results more economically in the presence of limited computing power, and that this degree of approximation may be sufficient for systems where the heterogeneity or otherwise of the behavior of the individuals under study is not important. For instance, in the realm of economics, microsimulation models have been used to analyze the impact of tax and policies on income distribution~\cite[pp. 42--90]{mitton2000microsimulation}. These models use detailed data on individual households to simulate the effects of policy changes, taking into account the heterogeneity of individual behaviors and circumstances.

In contrast, if one uses ABM to model such systems, one may introduce too many properties thus creating some kind of meaningless chaos and randomness, making it necessary to use a large number of simulations averaged to obtain a stable and usable result. For example, in the study of social-ecological systems, the inclusion of too many microscopic properties in ABM can lead to a high degree of complexity and nonlinearity, which can only be managed by averaging over a large number of simulations~\cite{van2019resilience}.

On the other hand, at the target scale of the study, the same problem may arise from the excessive microscopic properties involved in ABM systems when our study aims at questions about the macroscopic. Therefore, depending on the target scale of the study and whether the heterogeneity of individual behaviors significantly affects the results, we need to choose different methods for simulation. Using ABM is a good choice when we want to understand how the dynamic and complex interactions between individuals affect the performance of the whole system~\cite{foramitti2021agentpy}.

\myparagraph{Why Simulation}
Determining the problem nature targeted by the simulation is essential for the success of ABM. A careful examination of the system factors, the research question under consideration, and the suitability of simulation against other data analysis or experimental methods are the preliminary steps before initiating the modeling process. As per the spectrum of research methods, we can distinguish field studies that conduct experiments on real-world systems, theoretical studies that construct abstract models for theoretical predictions, and methods that employ lab experiments or simulations. In field experiments, variables are manipulated within a natural setting, offering high external validity, while controlling these variables can be challenging~\cite{pruitt2005field}. Lab experiments address this challenge by allowing researchers to manipulate independent variables in a controlled environment. A limitation is they might sometimes lack validity due to incorrect assumptions or the oversight of certain variables~\cite{oddy19809}. Simulation methods are employed to analyze complex systems and phenomena, especially when field or lab experimentation is prohibitive due to cost, danger, or feasibility~\cite{carle2012simulation, peck2004simulation}. Other than these experimental methods, theoretical studies rely on logical reasoning and mathematical models, eschewing direct empirical testing, to generate hypotheses and establish foundational insights~\cite{taborsky2008use}.

By comparing these experimental methods, it can be seen that simulations become a preferable empirical approach in scenarios where both field studies and reproducible lab experiments are challenging, where precise data are unable or hard to procure, e.g., limited data are available or involve ethical issues, or where hypothetical events are to be studied, e.g., forecasting results for public policy making. Simulations allow us to construct controlled virtual environments to manipulate various parameters, test different hypotheses, and derive outcomes preventing the Hawthorne effect or ethic problems we discussed above~\cite [pp. 4--10]{railsback2019agent}. Simulation methods may also be able to approximate experimental results in a virtual environment at a much smaller cost.

\begin{figure}[!t]
    \centering
    \includegraphics[width = \textwidth]{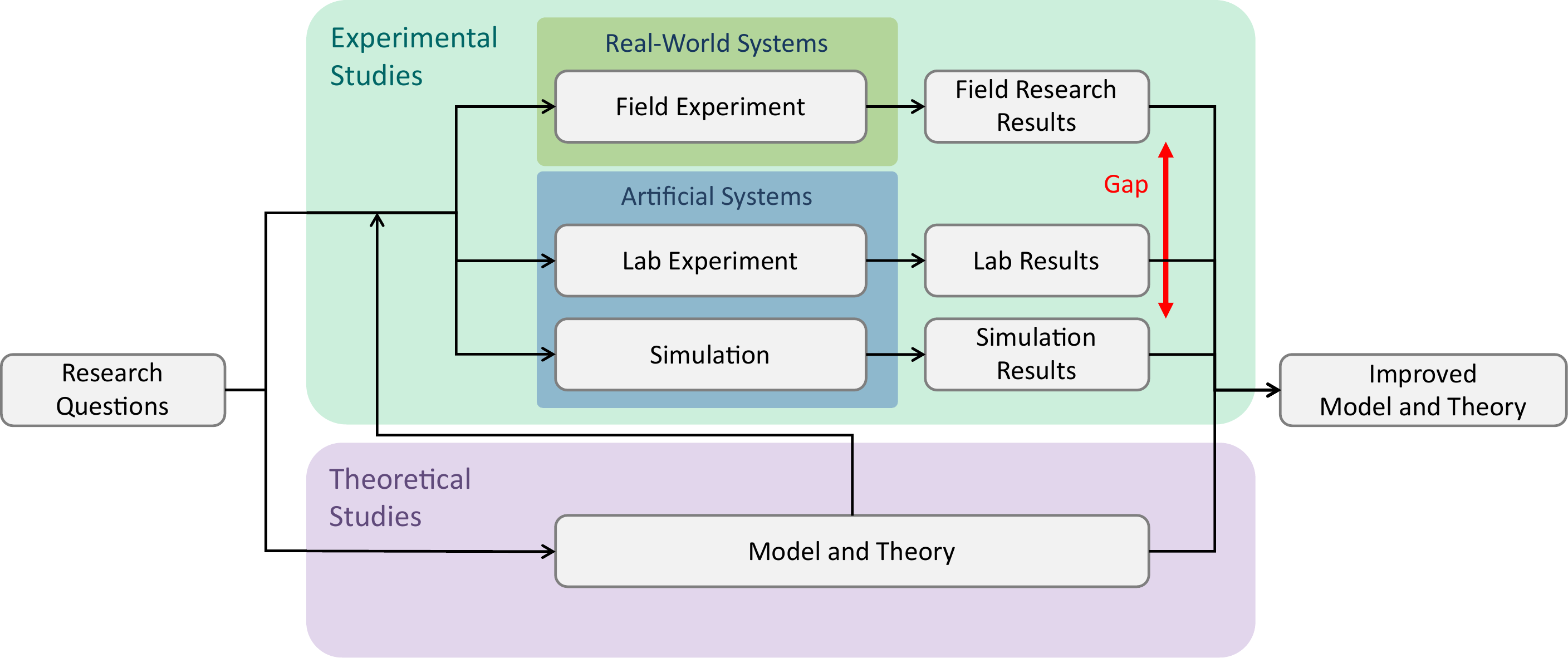}
    \caption{Interplay between experiment, simulation, and theory.}
    \label{fig:sota:interplay}
\end{figure}

Furthermore, although these experimental approaches all share the common goal of refining models and theories through the comparison of experimental data and theory, these methods often produce disparate results due to the inherent variability of real-world systems, the assumptions built into theoretical models, and the artificial environments of controlled simulations, e.g.,~\cite{hyman1982quasi, miller2007social, sterman1994learning, sterman2002all}. Therefore, discrepancies in the findings from field studies, lab experiments, and simulations often arise~\cite{balci1998verification, barber2002causes}. For instance, in studying the spread of infectious diseases, field studies might produce results that differ from theoretical predictions due to factors such as individual behavior variations, environmental differences, or unexpected mutation rates in the disease-causing agent~\cite{funk2010modelling, keeling2008modeling}. Similarly, lab experiments and simulations could diverge due to differences in the assumptions made for the disease spread rate, social contact networks, or intervention strategies~\cite{squazzoni2020computational}. In studies of pedestrian evacuation, real-world situations and experimental and simulated situations can be very different. In disaster preparedness drills, participants usually behave in an orderly manner because they know that it is not a real disaster, which is different from the real situation, but again we usually have difficulty in obtaining data for the real situation. It is also difficult to conduct experiments due to the complexities of experimental setup and ethical considerations~\cite{niu2021strategy}. It is of importance to understand the differences in results obtained from various approaches to improve the models and theories. Thus, the aim of this paper is to close the gap shown in Figure~\ref{fig:sota:interplay} by constructing a more realistic approach to simulation.

However, it is crucial to remember that while simulations can illuminate patterns and trends otherwise hidden in the complexity of real-world systems, they do not replace the need for field studies or lab experiments, as simulation studies have no agreed upon scientific norms or standards~\cite{winecoff2021simulation}. Instead, they serve as a complementary tool, enabling researchers to bridge the gap between abstract theoretical models and field studies while circumventing many of the experimental issues already mentioned. Moreover, the utility of simulations extends beyond mere reproduction of real-world phenomena. They serve as an exploratory tool that can generate new hypotheses and drive the discovery of emergent phenomena. For example, in a study of social networks, a simulation might reveal unexpected clustering patterns or influence dynamics that stimulate new research directions~\cite{squazzoni2020computational}. Work aimed at generating hypotheses may not necessarily require accuracy in simulations, as they are simply intuitive inferences about the relationships between observed variables.

In summary, the choice of using ABM simulation as a research method hinges on the nature of the problem at hand, the constraints of alternative methods, and the depth and breadth of insight that simulations can potentially offer. For instance, some common types of ABM work, such as opinion dynamics~\cite{motsch2014heterophilious}, behavioral simulation~\cite{wu2014intra}, disaster risk management~\cite{wang2016agent}, policy simulation~\cite{dalouglu2014development}, involve complex interactions between heterogeneous subjects, and the results of such interactions are difficult to predict accurately using analytical or statistical methods.

\subsection{Limitations of ABM}
\label{sec:sota:limitations}

Although ABM is theoretically capable of handling problems with the above characteristics, constrained by traditional models and computational power, ABM still has limitations in the actual practice. We begin the discussion with an example. 

\begin{figure}[!t]
    \centering
    \includegraphics[width =.5\textwidth]{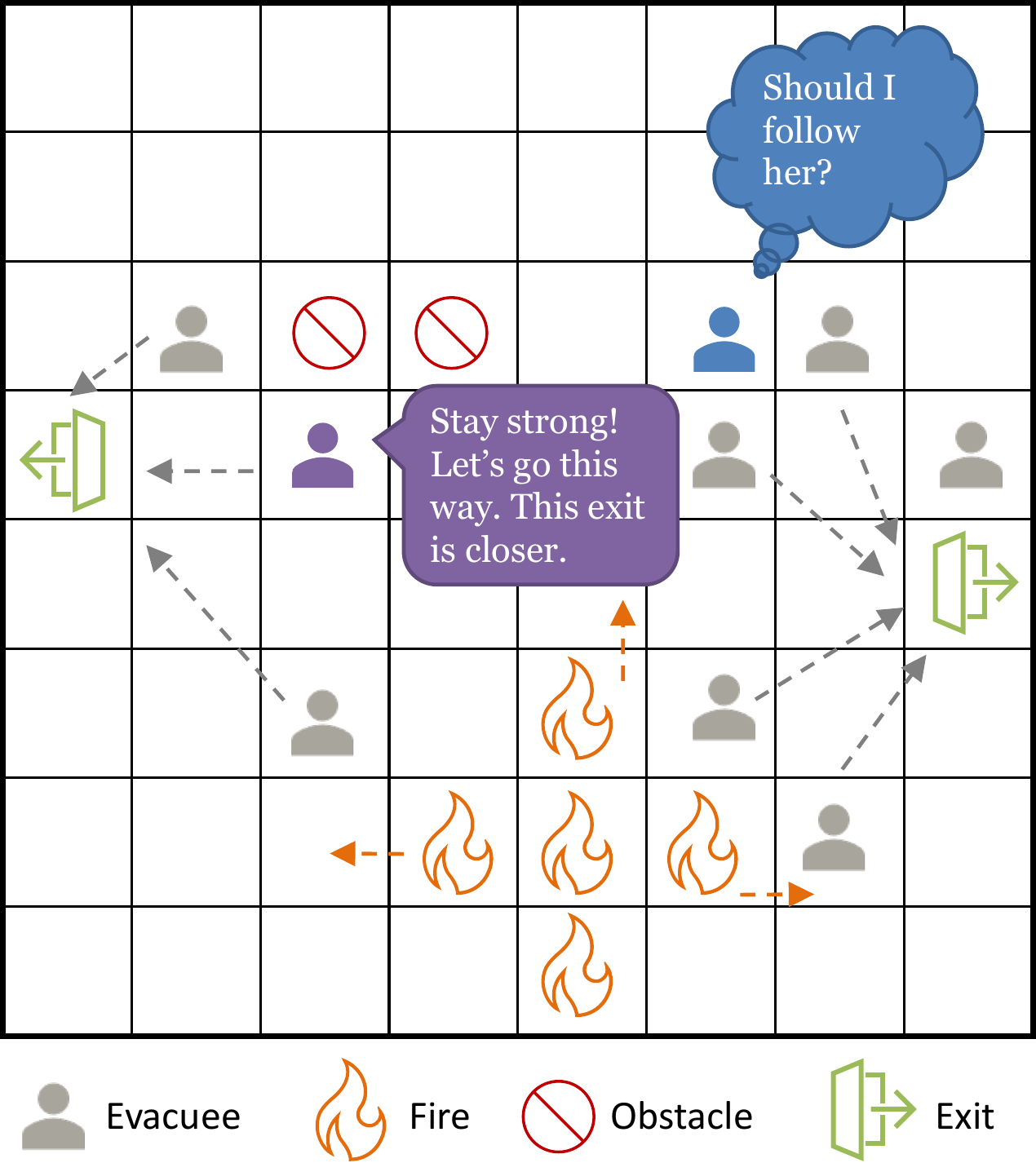}
    \caption{Fire evacuation simulation with factors difficult to model in ABM.}
    \label{fig:sota:evacuation}
\end{figure}

\begin{example}
    \label{ex:fire-evacuation}
    Figure~\ref{fig:sota:evacuation} depicts a simulation of fire evacuation, focusing on modeling the behaviors of evacuees. To capture the real-world dynamics of information sharing during evacuation, we include agent communication in the simulation. Some agents might naturally assume information spreader, conveying natural language messages to nearby agents. These agents then determine whether to follow the suggestions in these messages or not, based on the message and other factors such as their distances to the exits and the level of congestion. The process of simulating communication content requires natural language generation and produces a vast number of actions (i.e., all possible messages as output). On the other hand, simulating the decision-making of followers, factoring in communication, necessitates natural language understanding and processes a vast number of states (i.e., all possible messages as input). Together, these complexities surpass the current limits of ABM methods that rely on decision-making rules and learning algorithms. 
\end{example}

While the above example highlights a scenario in behavioral science, similar situations can be found in other areas. Take cartels, for instance, where independent market players collude to boost their profits. Current studies on simulating market competitions either rely on human volunteers~\cite{andres2023communication} or bypass inter-company communication, focusing instead on tacit collusion~\cite{calvano2020artificial}. In these examples, simulating communication based on natural language and common sense is crucial for modeling and understanding the research problems. We cannot simply map the influence of communication between these agents on their behaviors, where the scope of communication in real situations is not limited to a specific range. Although ABM approaches may utilize pre-trained language models for specific tasks (e.g.,~\cite{zhang2023simulating}), language is seldom part of the input/output of agents. The limitations of ABM make it difficult to model problems involving these complex behaviors. 

\begin{enumerate}
    \item \textbf{Model Complexity:} 
    When modeling many of the intricate behaviors observed in real-world systems -- especially those involving natural language descriptions and common-sense knowledge that are hard to express in formal language -- a model must possess significant complexity to encapsulate the system's core mechanisms. As shown in Example~\ref{ex:fire-evacuation}, modeling these behaviors is beyond the capabilities of ABM methods.
    Another challenge arises from the ``soft factors'' of human behavior~\cite{bonabeau2002agent}, such as irrational actions, subjective decisions, and intricate psychological dynamics, which are hard to quantify, calibrate, and justify. Even though ABM is arguably the only game in town to deal with such situations~\cite{bonabeau2002agent}, researchers are compelled to model these soft factors using predefined rules, which invariably leads to problem oversimplification. 
    
    \item \textbf{Choosing Parameters:} The modeling of agents in ABM needs parameter settings in its decision-making heuristics and learning rules. 
    This can introduce researcher bias and high uncertainty in model output, particularly when the parameter settings are not appropriately calibrated~\cite{grimm2005pattern}. 
    Moreover, the parameters are set based on researchers' experience and intuition, which if flawed or limited, may introduce biases, from researchers or the obtained data, into the ABM setting. For example, theories in social sciences often rely on data from small-scale studies, such as studies for WEIRD societies~\cite{henrich2010weirdest}, which may not generalize effectively to larger-scale group behavior due to the lack of extensive experimental data or theoretical foundations.
    
    \item \textbf{Finding the Medawar Zone:} While complexity theory has been applied to ABM~\cite{an2012modeling}, determining the optimal level of complexity for model construction remains a fundamental challenge. Following the above limitations in model complexity and predetermined parameters, in ABM, a simplistic model might overlook crucial system mechanisms, whereas an overly complex one can lead to intricate analyses that become mired in excessive detail. Thus, identifying an optimal range of model complexity, referred to as the ``Medawar zone'', is essential~\cite{grimm2005pattern}. However, this is not a easy task. The initial step in modeling involves identifying specific questions. These questions guide researchers in developing a conceptual model, which aids in determining which aspects and processes of the real-world system we should include or ignore. However, when dealing with complex systems, the question addressed by the model does not always pinpoint the Medawar zone, given that real-world systems have numerous degrees of freedom. Moreover, our conceptual model might be overly influenced by our viewpoint as external observers, encompassing our unique interests, beliefs, and perceptual scales. Thus, bias in the selection of modeling factors in ABM may lead to deviation of the results from the actual situation.
    
    \item \textbf{Agent Heterogeneity:} 
    As a critical aspect of many real-world systems, the heterogeneity of agents, though available in ABM methods, demands researchers understand the nuanced implications of parameter settings. For many ABM instances, uncovering actionable insight into a phenomenon depends on empirically-grounded heterogeneity specification, posing considerable challenges in parameter tuning. Inappropriate specification of heterogeneity can have the unintended effect of dampening other important features of the model, thereby preventing a reliable understanding of the observed phenomena~\cite{reeves2022structural}. This issue, combined with the dependence on predetermined parameters, renders it tricky to model the heterogeneity of agents in a complex system and compromises the robustness of the model.

    \item \textbf{Model Adaptability:} Many real-world systems are dynamic. In ABM, though agents can leverage learning algorithms for adaptation (e.g., by reinforcement learning~\cite{ale2023using}) to capture the dynamic nature of these systems, such learning ability is limited, and it often requires a sufficient amount of training data and time to become adapted to a complex and rapidly changing system. This is exacerbated by the fact that computer simulation is often used to study the case when real data are hard to obtain. 
    
    \item \textbf{Model Interpretability:} To interpret a model constructed by ABM, researchers must understand the rationale behind each rule in the model. Critics and concerns argue that when faced with obviously complex phenomena, researchers strive to keep their models simple, to an extent that is beyond any evident justification, while simplification should be only applied if and when the model and evidence justify this~\cite{edmonds2004kiss}. As such, without justification, the simplification tends to lack a realistic direct mapping and thus becomes uninterpretable. 
\end{enumerate}



\subsection{Agent-Based Simulations and Systems with LLMs}
\label{sec:sota:related}
Prior to presenting SABM, we review recent advancements on LLMs and their use in agent-based simulations and systems. Whereas the research on LLM-powered autonomous agents has recently received remarkable attention from the AI community~\cite{wang2023survey,xi2023rise}, these works mainly study and improve the behaviors and capabilities of LLM agents in solving practical or engineering problems in an autonomous manner. The aim of this paper, in line with the literature on ABM, is to study from the system science perspective and investigate the methodology of computer simulations for studying real-world phenomena, featuring more control over agents than in the works from the AI community. 

\myparagraph{LLMs and Capability Evaluations}
LLMs, namely, are language models characterized by large size, typically more than hundreds millions of parameters. They are trained on increasingly large human language corpora. Recent models, such as GPT-3 and GPT-4, can be prompt-engineered, i.e., communicated in a text-to-text manner, to solve natural language processing (NLP) tasks. Many of them are fine-tuned through reinforcement learning from human feedback (RLHF) \cite{ouyang2022training}, meaning that their output can be aligned to human preferences~\cite{ji2023ai}. As such, they acquire a variety of capabilities, including general knowledge, commonsense reasoning, and mathematical problem-solving. In addition, they can learn from few-shot examples~\cite{brown2020language}, reason from zero-shot~\cite{kojima2022large} and chain-of-thought (i.e., solving a problem as a series of intermediate steps) \cite{wei2022chain} instructions, and exhibit emergent abilities~\cite{wei2022emergent}. The development of LLMs can be found in a survey~\cite{zhao2023survey}. A benchmark has been proposed for evaluating LLMs as agents~\cite{liu2023agentbench}. Other evaluations of LLMs can be found in \cite{chang2023survey}. Results have demonstrated that ChatGPT, built upon GPT-3.5, outperforms humans in emotional awareness evaluations~\cite{elyoseph2023chatgpt} and the capability of GPT-4 in various aspects such as mathematics, coding, and psychology is close to human-level~\cite{bubeck2023sparks}. 

\begin{figure}[!t]
    \centering
    \includegraphics[width = 0.6\textwidth]{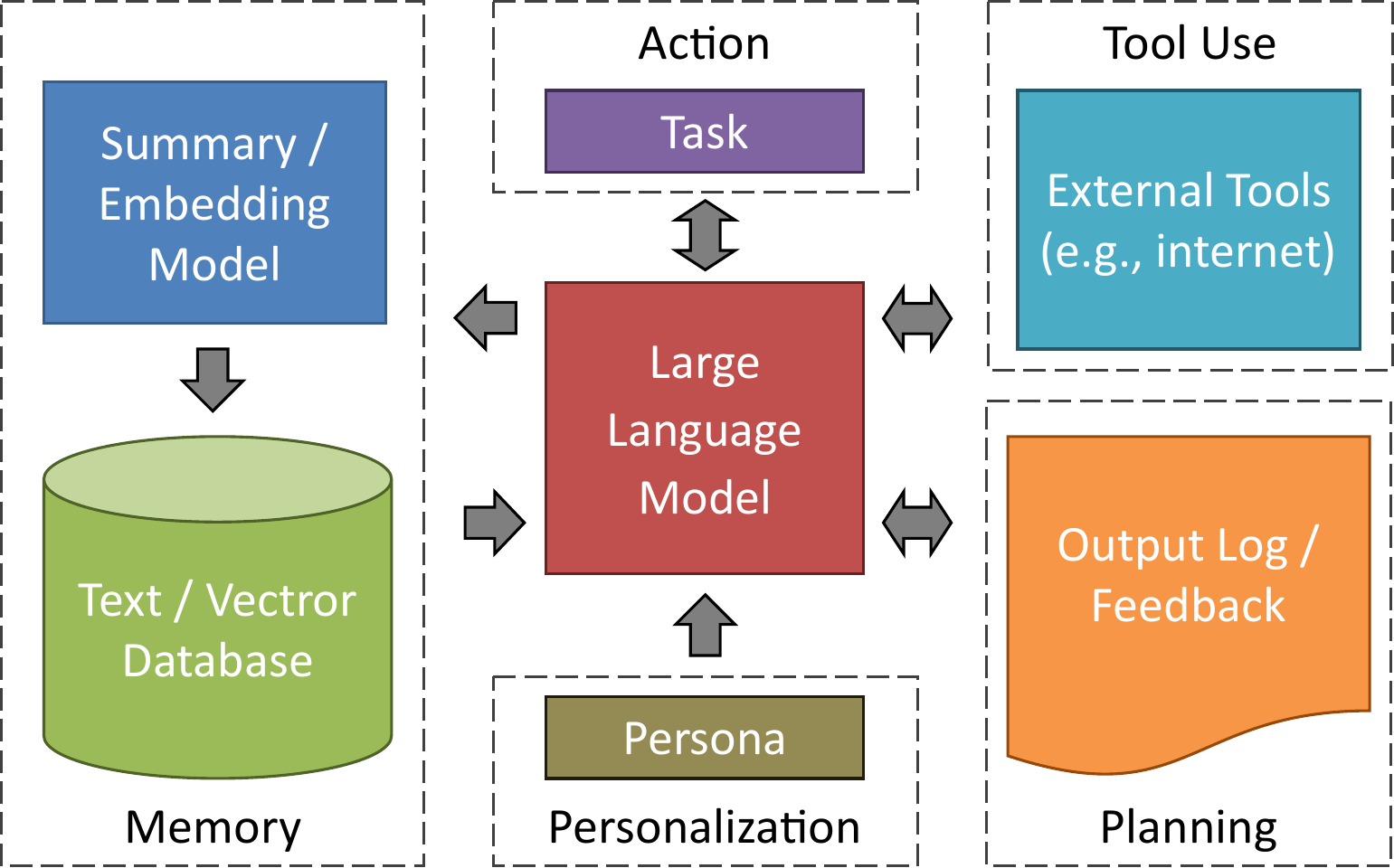}
    \caption{Overview of an LLM-powered agent.}
    \label{fig:sota:llmagent}
\end{figure}

\myparagraph{LLMs as Autonomous Agents}
Thanks to the outstanding capability of LLMs, many LLM-powered autonomous agents have been developed for task solving. The key components of an LLM agent are action, planning, memory, and tool use~\cite{llmagents}, as depicted in Figure~\ref{fig:sota:llmagent}. Action manages how an agent communicates with the user and solves the task. Planning decomposes a complex task into several smaller and simpler sub-tasks, and performs self-reflection over past actions to improve the performance of future actions. Memory equip LLMs with short-term memory, which is often implemented via prompt engineering, and long-term memory, which is often implemented using summarization~\cite{park2023generative} or text embedding~\cite{langchain}. Tool use enables an agent to call external APIs to acquire extra information. While developing autonomous agents, researchers noticed that LLMs can be personalized to improve the performance of task solving~\cite{salewski2023context,wang2023unleashing}. Hence personalization (a.k.a. profiling\cite{wang2023survey}) becomes an optional component in an LLM agent. A typical instance of an LLM agent is Auto-GPT~\cite{autogpt}, which employs GPT-3.5 or GPT-4 for solving a wide range of tasks. The core technology is breaking the task into sub-tasks and addressing them with internet and other tools. In addition, recent attempts created reusable tools for task-solving with GPT-4~\cite{cai2023large} and task planning and tool usage with ChatGPT and Claude~\cite{ruan2023tptu}. Besides general task solving, LLM agents were also developed for specific tasks, e.g., playing Minecraft, in which an agent explores the virtual world, acquires skills, and makes discoveries~\cite{wang2023voyager,zhu2023ghost}. We refer readers to a survey~\cite{wang2023survey} for recent progress. 

\myparagraph{LLMs in Multi-Agent Simulations and Systems}
For the case of multi-agent simulations and systems, a sandbox environment was developed for interactive simulation of human behavior using GPT-3.5 agents~\cite{park2023generative}. Another environment was proposed in \cite{liu2023training} to simulate human society and train socially-aligned LLMs. Other works on sandbox environments for multi-agent simulation with LLMs include \cite{lin2023agentsims,giabbanelli2023gpt,gao2023s}. We refer readers to a survey~\cite{xi2023rise} for recent works on simulated societies of AI agents. In addition to simulation, multi-party conversations between agents, where each participant has an assigned character to role-play, have been studied in \cite{wei2023multi}. Role-playing~\cite{li2023camel} is a framework where LLMs serve as chat agents and conversational data are collected for studying their behaviors and capabilities. For multi-agent collaboration, a framework was proposed in \cite{talebirad2023multi} for enhancing the capabilities of LLMs completing complex tasks in a collaborative manner. Another framework multi-agent collaboration is AgentVerse~\cite{chen2023agentverse}, featuring the emergence of agents' social behaviors during collaborative task accomplishment. In ChatDev~\cite{qian2023communicative} and MetaGPT~\cite{hong2023metagpt}, LLM agents work in a collaborative manner for software development. In addition to collaboration, there are also studies on multi-agent negotiation~\cite{fu2023improving} and debate~\cite{du2023improving,liang2023encouraging}, demonstrating that the agents can improve their performance in a negotiation or debating game by playing and reflecting. 

\myparagraph{Applications of Modeling with LLM Agents}
Since LLMs can cope with both classification and generation tasks, the use of LLMs for labeling and free-form coding in computational science was discussed~\cite{ziems2023can}. For sociology, the notion of generative agent-based modeling, which resembles SABM but with more emphasis on agents generating reasoning and decisions, was proposed in \cite{ghaffarzadegan2023generative} for studying social system dynamics, where GPT-3.5 agents were used in a simple model of diffusion of norms. Following this, the use of GPT-4 agents for studying disinformation in social networks was discussed~\cite{pastor2023generative}. For behavioral science, the emergent behaviors of GPT-3.5 agents in social dilemmas (e.g., Prisoner's Dilemma) were investigated~\cite{phelps2023investigating}. GPT-3.5 agents were also used for simulating a two-party negotiation and a six-party murder mystery game~\cite{junprung2023exploring}. For economics, GPT-3 agents, termed as homo silicus, were used in a set of experiments on social preferences, fairness, status quo bias, and minimum wages~\cite{horton2023large}. Besides, GPT-3.5 agents are prompt-engineered for human-like decision-making and used in simulating macroeconomic activities such as inflation and unemployment~\cite{li2023large}. For jurisprudence, GPT-2 agents were employed to impersonate the justices of the Supreme Court of the US, and a correlation was found between model accuracy with respect to individual justices and their alignment between legal conservatism and liberalism~\cite{hamilton2023blind}. For education, fine-tuned LLaMa~\cite{touvron2023llama} was used for simulating K-12 students for studying sentence reading efficiency~\cite{zelikman2023generating}. The applications of LLMs in academic knowledge creation (e.g., writing, editing, reviewing, dataset creation and curation) for management research were envisioned~\cite{williams2023algorithmic}. In addition, we refer readers to a series of vision papers for the use of LLM agents in social science~\cite{grossmann2023ai,bail2023can}, behavioral science~\cite{mills2023ai}, and education~\cite{kasneci2023chatgpt}. 
\section{Concepts of Smart Agent-Based Modeling}
\label{sec:sabm}

\begin{figure}[!t]
    \centering
    \includegraphics[width = \textwidth]{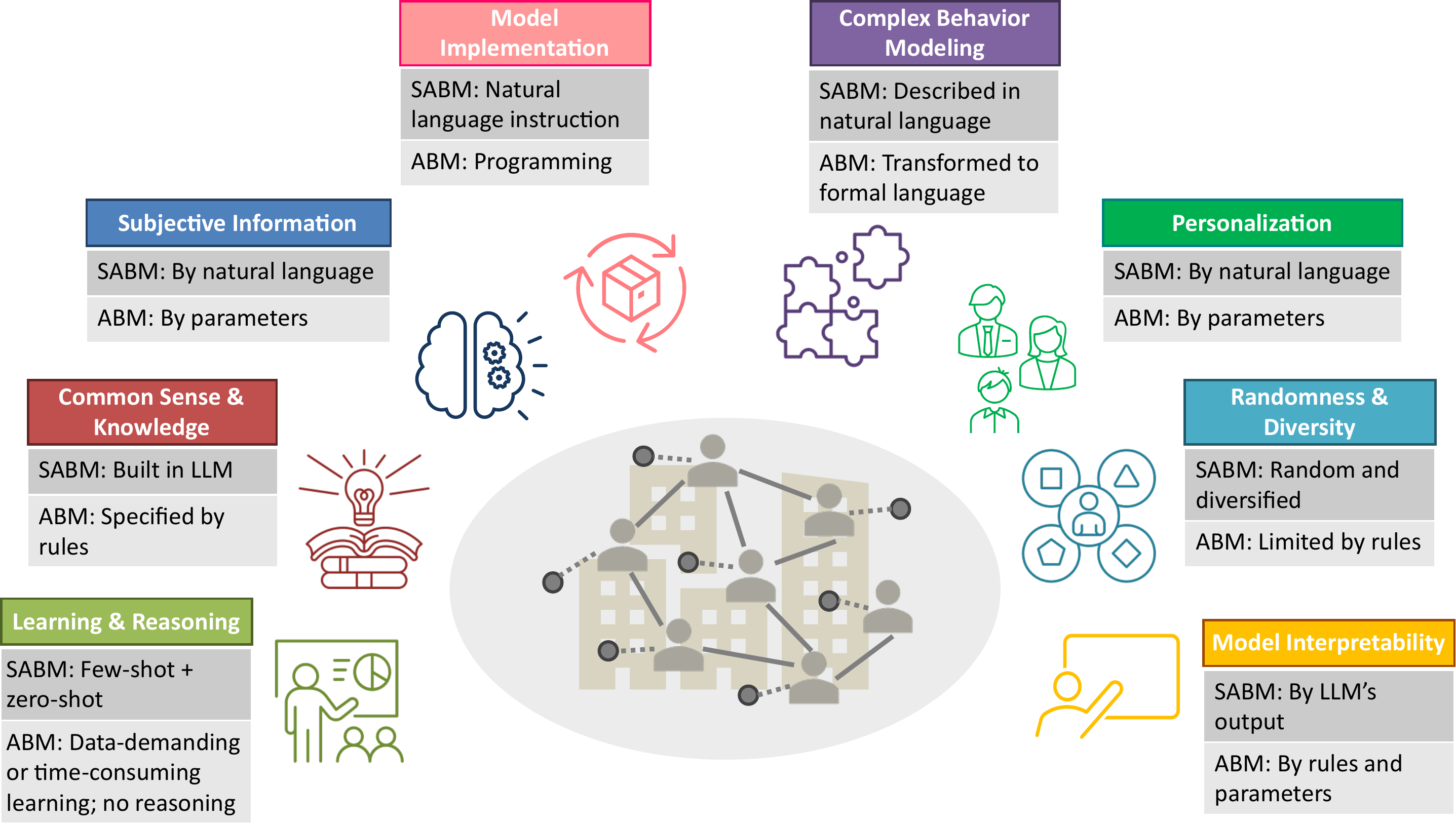}
    \caption{Comparison of SABM and ABM features.}
    \label{fig:sabm:features}
\end{figure}




A fundamental aspect of an ABM's simulation comes from its individual components: agents, environmental, interactions, and the parameters and data characterizing these components~\cite{macal2005tutorial}. The simulation's quality and relevance heavily rely on the accurate representation of these elements, impacting the model's overall performance. Given the  limitations summarized in Section~\ref{sec:sota:limitations}, there is a pressing need for methods that can address these limitations and extend the application scope of computer simulation. Due to LLMs' superb capabilities of understanding natural language and emulating human behaviors~\cite{bubeck2023sparks}, one such potential avenue for improvement is the integration of LLMs into ABM's components. We propose smart agent-based modeling (SABM), which extends the modeling approach from ABM by employing LLMs and formulating its components in natural language. For the name of this new modeling approach, we borrow the term ``smart agents'' from Carley~\cite{carley2002smart}, who defined smart agents as entities that are intelligent, adaptive, and computational. These characteristics are exactly what the new modeling approach is endowed with by utilizing LLMs. 

Figure~\ref{fig:sabm:features} illustrates the features of SABM and compares them to ABM features. The ensuing discussion will explore the features of SABM, considering aspects such as language ability, modeling paradigm, adaptability, and interpretability. By analyzing these facets, we aim to determine the potential enhancements that SABM could bring to computer simulations and the potential scope of application of SABM. 

\subsection{Reshaping Agents with Natural Language Ability}
\label{sec:sabm:language}

SABM gives rise to agents that can handle natural language as both input and output, thereby crafting more nuanced and realistic simulations, which subsequently extend its applicability. The effectiveness of handling natural language will be illustrated with respect to modeling and comprehension capabilities.

\myparagraph{Modeling in Natural Language} 
With an LLM as its core, an agent has the ability to process natural language for both input and output. Instead of relying on empirical rules or mathematical formulas to define agent attributes, the textual description given in natural language serves as the primary reference.

\begin{example}[Modeling Ant Behavior]
    Gordon leverages ABM to model ant behavior~\cite{gordon2010ant}. In this study, the conduct of individual ants was modeled based on a precise set of mathematical rules, designed and fine-tuned to mirror observed ant behavior.
\end{example}

On the other side of the mathematical approaches, SABM presents the potential to model ant behavior rooted in qualitative descriptions of the ants' decision-making process. Researchers can offer natural language descriptions detailing how ants decide on foraging locations, how they communicate food source locations with fellow ants, and how they adapt their behavior in response to environmental changes. Such natural language descriptions are intuitively easier to comprehend than mathematical representations, offering an accessible way to model ant behaviors. Also, LLMs can utilize these natural language descriptions to generate appropriate behaviors for the ant agents within the model. For example, the model might indicate that ants are more likely to forage where other ants have located food. This information can then be used to direct the behavior of the ant agents in the simulation as the prompt to the agents.
Moreover, this approach of avoiding the use of parameters to express rules also contributes to the reproducibility and comparability of simulation experiments: for research problems with the same type of background, we can use similar prompts to describe their background settings and add problem-specific elements on top of this same basis.

\myparagraph{Modeling Complex Behaviors} 
The omission of natural language from ABM makes it difficult to accurately model descriptions hard to express in a formal language and simulate systems wherein natural language is critical, without losing detail in the information exchange~\cite{muller2014standardised, nissen2004agent}. By leveraging LLMs' capability of understanding natural language input and creating contextually relevant and coherent output~\cite{radford2019language}, SABM offers a solution to this issue. 


\begin{example}[Public Health]
    Consider the field of public health where ABM has been leveraged to predict the spread of diseases and evaluate the impact of various interventions. Marshall et al.~\cite{marshall2012complex} used ABM to model HIV transmission in high-risk populations by using a complex systems approach, which revealed valuable insights into the potential collective benefits of integrated prevention interventions. In such models, agents represent individuals whose behaviors and interactions are based on prescribed rules. However, they often struggle to incorporate the complexity and nuances of human behavior, particularly concerning communication and decision-making processes which are typically described in natural language. 
\end{example}

SABM allows agents to handle natural language instructions (or we say \emph{prompt} in the context of LLMs) and simulate more complex behaviors and interactions described in natural language. For instance, agents can be instructed to discuss their health status, share disease prevention information, or decide to seek tests or treatments based on these discussions. This results in a more authentic simulation of information dissemination within a community and its consequent impact on behavior. 
Beyond enhancing the realism of simulations, SABM could provide fresh avenues for intervention. Public health officials, for instance, could intervene within the simulation by imparting new information or resources to an agent and then observing how this influences the disease's spread. This approach could help identify effective strategies for real-world interventions and facilitate the simulation of such interventions within a reliable simulated environment. Simulation can provide validation that the development of new regulations or policies will have the desired effect.


\subsection{Alignment for A Priori Modeling}
\label{sec:sabm:alignment}
Many LLMs are fine-tuned using RLHF~\cite{ouyang2022training}, which aims to align the AI model's behavior with human values and knowledge by incorporating human feedback into the learning process. This can not only enhance the model's performance but also increase the ethical and moral alignment of its responses~\cite{liu2023perspectives, ganguli2023capacity}. It is particularly beneficial for SABM as it allows for simulations that more closely resemble human interactions, thereby enhancing the accuracy of simulation outcomes. 


In Carley's definition~\cite{carley2002smart} of smart agents, humans stand as the canonical smart agents. On the other hand, it is argued that GPT-4 and other LLMs would probably now pass the Turing test, in that they can fool many people, at least for short conversations~\cite{biever2023chatgpt}. Evaluations also showed that GPT-3.5/4 outperforms or is on a par with humans in emotional awareness~\cite{elyoseph2023chatgpt}, mathematics, coding, and other aspects~\cite{bubeck2023sparks}. Given that LLMs exhibit these capabilities and they are aligned with human values and behavior, it allows us to adopt an a priori modeling paradigm within SABM, with the assumption that smart agents, powered by LLMs, can mimic human behavior. Conversely, ABM employs a posteriori modeling paradigm where behavioral rules and parameters are deduced or inferred from observed human actions.

To delineate the distinctions between these two paradigms, consider the realm of consumer behavior. In ABM, researchers examine real-world consumption patterns and distill them into a set of mathematical representations or formulas to depict the nuances of how a consumer selects goods or services. However, in reality, individuals' purchasing decisions are governed more by innate reasoning or common sense -- assessing necessity, personal preference, and perceived value -- rather than any explicit formula. Within the SABM framework, we can emulate such consumption behaviors from this foundational viewpoint, leveraging the innate common sense of LLMs aligned to humans', obviating the need for rules or parameters to encapsulate such actions. In addition, LLMs can learn with few-shot examples, meaning that the actions can be further calibrated to human data, which will be elaborated in Section~\ref{sec:sabm:adaptability}.  

Thanks to the alignment with human values and knowledge, SABM broadens the horizons of a priori modeling, which has traditionally been reserved for formal science or theoretical studies that do not rely on real data (e.g., elementary cellular automata~\cite[pp. 23--50]{wolfram2002new}). Moreover, in SABM, a model can be divided into two components: the ``engine'', represented by the LLM, and the ``chassis'', which encompasses the remainder of the model. As AI researchers work towards enhancing the engine, thereby solidifying the foundation of a priori modeling (i.e., the aforementioned assumption), those utilizing SABM can concentrate on crafting the chassis and using the most fitting engine for their specific endeavors. This practice significantly reduces the difficulty in modeling and helps researchers quickly find the Medawar zone of modeling~\cite{grimm2005pattern}. When a more advanced LLM is available, researchers may simply replace the engine for better performance. In ABM, enhancing agents' performance in emulating human behavior and constructing simulation models are highly coupled, and thus researchers have to manage both procedures, which inevitably calls for more expertise and modeling effort.  

\myparagraph{Common Sense}
As previously discussed, a remarkable feature of LLMs is their ability to exhibit common sense reasoning~\cite{zhao2023survey}. This capability stem from the extensive and diverse training data these models are exposed to, which encompass a broad array of topics and everyday knowledge. 
Common sense reasoning can facilitate simpler model setups in SABM, thereby reducing the need for complex configurations. For example, agents can utilize this feature to make sense of their environments, interact with other agents, and adapt their behaviors based on context, all without needing explicit programming of these abilities. 

\begin{example}[Misinformation Propagation]
    The application of common sense reasoning is valuable in scenarios such as the simulation of misinformation propagation in social networks. 
    In ABM approaches to this problem~\cite{tseng2020agent}, agents might pass misinformation to others based on their distances. This method, while simple, may not accurately capture the nuanced behaviors of individuals in real-world social networks who possess and apply common sense in such situations.
\end{example}

By leveraging LLMs with built-in common sense, we can transform these probability parameters into more specific and realistic behaviors. For instance, agents can question the validity of information, seek additional sources, or attempt to refute misinformation based on their common sense. Such detailed behaviors align more closely with human responses and enable more accurate and comprehensive simulations of misinformation propagation. 

\myparagraph{Built-in Knowledge} 
ABM is often confined by the domain knowledge explicitly coded into the agents, which inherently limits the depth and diversity of their behaviors. In contrast, SABM holds the promise to access a vast reservoir of domain knowledge, owing to LLMs' training on a diverse array of internet texts spanning across numerous domains. The domain knowledge inherent in LLMs present significant opportunities for simplifying and enriching SABM, broadening the scope of potential applications and enhancing the utility of these simulations in understanding complex systems. 

\begin{example}[Climate Change Policies]
    Consider the example of modeling stakeholder behavior in response to policy change, a realm in which ABM has been extensively utilized~\cite{hammond2015considerations,wallace2015assessing,shastry2022policy}. Suppose the behavior of firms in response to a new carbon tax policy is to be modeled. In ABM, the agent representing a firm might simply decide whether to reduce emissions based on the cost of the tax. This approach, while pragmatic, is relatively simplistic and may not fully capture the multifaceted decision-making process in real-world firms. In contrast, with SABM, agents can leverage a wider spectrum of knowledge when determining their actions. For instance, they could consider scientific evidence on climate change, gauge public sentiment toward businesses contributing to climate change, assess the potential benefits of transitioning to renewable energy, among other factors. These are considerations that real businesses often contemplate, hence making SABM more reflective of actual corporate behavior in response to policy change.
\end{example}

By summarizing the SABM features on alignment with human values and knowledge, it is also noteworthy to mention that factors including emotions and potential merits are important influences that involve subjective components, about specific knowledge, and have open boundaries that are difficult to model with ABM. It follows that SABM has the advantage that they minimize the subjective setting of parameters, which can be subject to overfitting and bias, including stereotype-based bias. Overfitting occurs when a model is too complex, with too many parameters relative to the number of observations, causing it to perform well on training data but not on unseen data~\cite{hawkins2004problem}. 
On the other hand, bias may lead to skewed or unfair results based on assumptions or stereotypes. LLMs can help mitigate these issues by learning from a vast corpus of data, thereby internalizing the complexity and diversity of real-world interactions. Instead of defining rigid and possibly biased rules for agent behavior, LLMs generate behaviors based on learned patterns from real-world data, helping to reduce the potential for subjectivity and bias. In essence, SABM can lead to simulations that are not only less prone to overfitting and bias, but also more reflective of the complex and unpredictable dynamics present in real-world systems.



\myparagraph{Personalizability} 
A key feature of SABM yielded by alignment with humans is the personalization of agents~\cite{salewski2023context,wang2023unleashing}, i.e., agents can be personalized to exhibit distinct characteristics relevant to the simulated scenario, such as differing personalities or styles. This heterogeneity allows for greater realism within simulations and thereby enhances the model's fidelity.

\begin{example}[Pedestrian Evacuation]
    An example of such application is seen in the field of pedestrian evacuation simulations using ABM~\cite{wu2012personalized}. This research introduced a personalized spatial cognitive road network (PSCRN) model. The PSCRN model posits that each pedestrian has a unique probability of recognizing specific evacuation routes based on their spatial cognition. This personalization is grounded in statistical probability and being theoretically and simulationally sound, but this kind of settings may still not fully encapsulate the unique individual experiences or characteristics. Note that this is not a limitation of the original work, but LLMs may have the potential to further develop the simulation and methodology.
\end{example}

SABM can enhance the degree of personalization by utilizing natural language as a tool for agent characterization. Each agent can be personalized using distinctive linguistic inputs, reflecting their knowledge, experiences, and perspectives. In the pedestrian evacuation simulation, for instance, agents can be assigned diverse knowledge about the environment, echoing their personal experiences or characteristics. Agents representing older adults, who may possess different knowledge and exhibit slower movement speeds, can be differentiated from those symbolizing younger athletes. This degree of personalization can result in more realistic and nuanced simulations. Importantly, complex parameterization can be bypassed in favor of natural language cues, simplifying the process of agent personalization.

\myparagraph{Subjective Judgment} 
While agent personalization allows for a more realistic heterogeneity, it also paves the way for the simulation of agents' subjective worldviews. 
The simulation of human behavior necessitates accounting for the locally available information and the finite rational judgments each agent can make~\cite{simon1955behavioral}. This is of particular importance in the simulation of complex, especially social, phenomena that are influenced not just by objective facts, but also by subjective beliefs about the objective world, i.e., the subjective reality~\cite{berger1967social}. Although it is straightforward to simulate the objective information available to each agent, traditional models struggle to represent subjective judgments about this information, often resorting to rule-based or learning-based methods for approximation.

\begin{example}[Civil Violence]
    An example can be drawn from the work of Epstein~\cite{epstein2002modeling}, who developed an ABM for simulating civil violence. This model relied on a prescribed set of rules and parameters to direct agent behavior. In Epstein's model, agents decide to rebel based on their perceived hardship, their perceived legitimacy of the regime, and their estimated risk of arrest, all of which are determined by a set of parameters and the current state of their surroundings. This artificial setup, although useful for basic simulations, may not fully reflect the nuanced and context-dependent nature of human decision-making processes, namely subjectivity.
\end{example}

SABM offers a solution to this issue by simulating such subjective judgments, effectively replacing the need for traditional manual parameter setting and making it possible for incorporating in the model the soft factors mentioned in Section~\ref{sec:sota:limitations}. 
For example, in the context of civil violence, an agent's perception of risk and hardship can be shaped by natural language inputs that reflect news reports, social media sentiment, or conversations with other agents. These agents can then use this language-based information to make decisions about whether to engage in rebellion, thus providing a more nuanced and flexible simulation of agent behavior. In this context, these inputs can be considered as the local information obtained by agents about objective facts, while the agent's feedback to these information is a subjective judgment. Ultimately, the objective facts are mediated by this subjective judgment embodied in the final behavior. In practical problems, we often have information about objective facts, but we have difficulty knowing how the collection of subjective information of the subject affects the whole system. Therefore, the subjective judgment in SABM can expand the quality and scope of computer simulations.

\myparagraph{Randomness and Diversity} 
Given the difficulty of expressing complex agent behavior in ABM, when our research questions do involve complex actions and states, ABM simulations may be restricted to a limited range that produces predictable results that cannot fully reflect the nature of the system studied. In contrast, SABM can inject an additional layer of diversity into agent behaviors by providing a degree of randomness to the interaction, effectively expanding the agent's actions and states and yielding more diverse and creative outcomes to match reality.

\begin{example}[Market Economy]
    Consider an agent-based simulation of a market economy. In ABM, each agent might follow prescribed rules related to buying and selling goods based on factors like price, demand, and available resources~\cite{calvano2020artificial}. 
\end{example}

In SABM, agents can be tuned to exhibit more randomness for their responses to the outcomes of past trades, thereby performing more diverse actions such as negotiation on prices. This not only allows for the emergence of complex behaviors but also facilitates the interpretation of the underlying dynamics that guide these behaviors, thus providing deeper insights into the system under study. 

Another example is the aforementioned example of misinformation propagation. The randomness and diversity of SABM agents, along with their common sense reasoning capabilities enable the simulation to account for diverse reactions to misinformation, as agents can vary their responses based on the context, the nature of the misinformation, and the source of the information, among other factors. This can lead to the emergence of more complex and realistic behaviors in the simulation, thereby enhancing our understanding of the dynamics of misinformation propagation in social networks.

\subsection{More Adaptive, Interpretative Interactions}
\label{sec:sabm:adaptability}
ABM, with static, prescribed rules and parameters, 
lacks the necessary flexibility and adaptability required to accurately represent dynamic complex systems. The introduction of LLMs into ABM allows us to break free from rigid rule-based interactions, opening up the possibility for more dynamic and adaptive agent interactions. 

\myparagraph{Learning and Reasoning} 
The inherent learning and reasoning ability of LLMs has already proven its efficacy in performing various tasks with few~\cite{brown2020language} or even no data samples~\cite{kojima2022large}. In SABM, the learning and reasoning ability bypasses the need for pre-training processes or learning algorithms. Instead, it introduces an adaptive learning aspect that closely mimics reality, presenting a significant upgrade from prescribed rules that often fail to capture the dynamic nature of real-world systems.



\begin{example}[Epilepsy Treatment] 
    To illustrate this point, we can consider the work done by Megiddo et al.~\cite{megiddo2016health} on epilepsy treatment in India. The study utilized ABM to simulate the health and economic outcomes of a publicly financed national epilepsy program. The study provided valuable insights on how to avoid a large burden of disease. However, it assumed constant income and treatment cost over 10-year period, which might not be true in developing countries like India. In a dynamic system representing a 10-year or even longer period, agents are supposed to be able to adapt to the change of income and treatment cost and adjust their behaviors accordingly. 
\end{example}

As outlined in Section~\ref{sec:sota:limitations}, ABM, even equipped with learning algorithms, has difficulty in adapting to a complex and rapidly changing system. SABM, with its inherent learning ability, reasoning ability, and adaptability, could provide a better solution for such scenarios. Instead of assigning different features to different agents manually, LLMs can learn from past behaviors using much fewer examples, develop plans and strategies, and simulate agents' response to change of environment more effectively. This allows for capturing a wider array of possible health and behavioral outcomes, thereby facilitating the simulation of long-term and dynamic systems.

\myparagraph{Model Interpertability} 
The interpretative capacity of LLM-powered agents, combined with its text generation capabilities, offers a unique opportunity to enhance the interpretability of SABM. Agent interactions in ABM often obscure the motivations behind the agents' behaviors. Unraveling these behaviors from the intricate web of parameters and rules can be a challenging task, often deterring a comprehensive understanding of the model dynamics. Meanwhile, by utilizing the interpretative nature of LLMs, we can extract more intuitive and human-understandable representations of the agents' behaviors and interactions. Rather than interacting according to prescribed rules, agents can interpret their ``intentions'' or ``feelings'' based on their performance and interactions with other agents and then adjust their behaviors accordingly. This mirrors the interpretative nature of human communication, where meaning is often derived from the interpretation of linguistic cues rather than literal language itself~\cite{grice1975logic}. In this context, the very text that forms the basis of agent interactions holds the capacity to interpret the simulation.

\begin{example}[Social Influences on Body Mass Index (BMI)]
    Hammond and Ornstein~\cite{hammond2014model} pioneered a data-driven simulation model based on theories from physiology, social psychology, and behavioral sciences, and made recommendations for public health policy with findings that are highly consistent with reality. However, as noted in the discussion of that paper, there is still a lack of sufficiently rich and explicit models of the underlying mechanisms of social influence. The factors involved in complex phenomena are numerous. Although the numerous factors involved in complex phenomena can be modeled using prescribed rules involving the probabilities of interactions, these may yield insights into macroscopic patterns like this work successfully achieved, the finer explanatory granularity remains elusive. For instance, understanding the drivers behind BMI clustering within social networks in this model proved challenging, potentially complicating policy or intervention development.
\end{example}

The employment of natural language-based interactions mediated by LLMs can offer a solution. Instead of interpreting agent behaviors based purely on statistical probabilities, researchers can inspect the dialogue between agents or represent their cognitive processes in textual form. This allows a deeper understanding of the agents' decision-making processes and behaviors. Consequently, it becomes feasible to perceive how a broader array of factors may impact BMI, leading to greater capacity to encompass a wider range of influential factors. We can thus observe not only the agent's actions, but also its reasoning process.

\subsection{Application Scope of SABM}
\label{sec:sabm:applications}



\begin{figure}[!t]
    \centering
    \includegraphics[width = 0.95\textwidth]{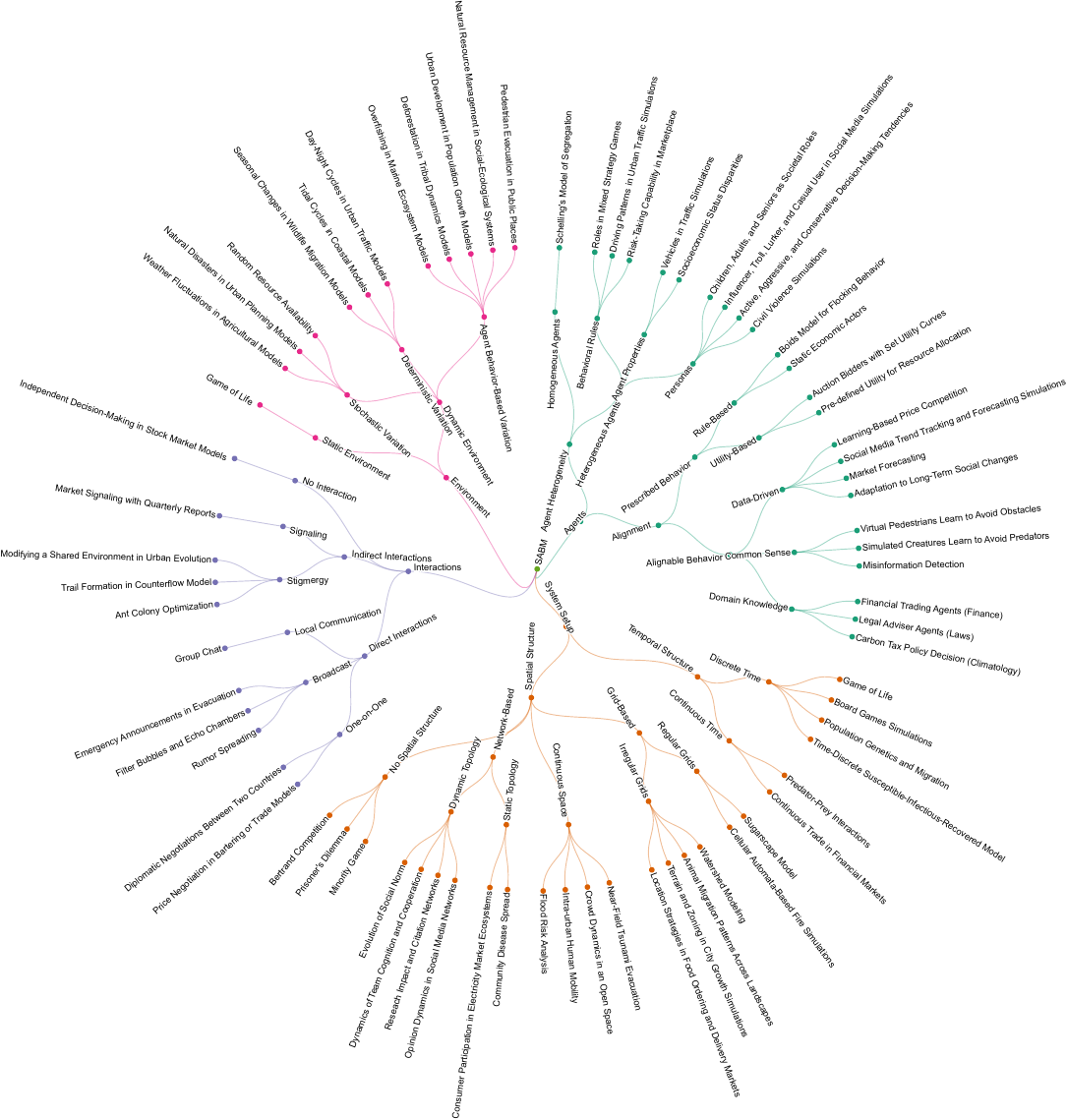}
    \caption{Categorization of SABM components and application examples.}
    \label{fig:sabm:hierarchy}
\end{figure}

From the above discussion of the features of SABM, we can expect that SABM has a greater scope of application to reproduce the complexity, common sense, adaptability, and other properties of the actual situation, guaranteeing the appropriate level of realism for problem-solving, and can have better interpretability than ABM. The improvement of the simulation degree of the modeled system can narrow the gap between the simulation results and the field research results as shown in Figure~\ref{fig:sota:interplay} while maintaining the ability to explain complex and realistic simulation settings.

Thus, on top of the current application scope of ABM discussed in Section~\ref{sec:sota:scope}, SABM allows for the further expansion of the boundaries of this simulation methodology. In the modeling of agents, SABM enables the personalization of agents using natural language methods that are not based on determined parameters or empirical formulas. 
This allows us to delve deeper and more authentically into problems where subtle differences in agent behavior significantly impact system performance. For example, the features of common sense can be utilized to intricately simulate an agent's behavior in judging and disseminating misinformation~\cite{alassad2023developing}, which is difficult for ABM to accomplish simply.

In modeling interactions, SABM, with its linguistic, learning, adaptability, and interpretative capabilities, can simulate a broader range of scenarios involving natural language information exchange. We discuss the expansion of SABM's scope by categorizing it into direct and indirect interactions. In direct interactions, the learning and adaptability of the SABM agent make dynamic interaction behavior possible, which is challenging in ABM based on learning or rules that are limited to specific knowledge scopes and datasets. For instance, in studies concerning the farmer typology of agricultural conservation behavior~\cite{dalouglu2014development}, it is suggested that farmers' decisions to adopt conservation measures are dynamic and influenced by changing environmental and social conditions, where modeling these dynamic factors is difficult. SABM allows for a more detailed and realistic modeling of farmers' decision-making behavior, presenting the dynamic process of decision changes in greater detail. On the other hand, in indirect interactions, like signaling, it is challenging to model how agents respond heterogeneously to approximate signals based on their characteristics. In the stock market, for example, despite social learning, replicating the market's complex behavior and modeling the behavior of irrational agents remains a challenging task~\cite{chen2004multi}. SABM, with its linguistic capabilities, can describe and interpret irrational and subjective behaviors. Moreover, it holds the potential to exhibit patterns consistent with market irrational behavior based on its learning capabilities and domain knowledge.

SABM also extends the environmental settings that existing ABM methods can address. In environments where the external conditions and environments are not influenced by agent behaviors, SABM can support parameter settings and provide natural language descriptions of these environments. For example, in tsunami risk management and evacuation scenarios~\cite{wang2016agent}, a textual description can be directly used to depict the environment, enabling agents to more intuitively grasp the environmental data, including risk factors and the development of tsunami, and to integrate this information as a foundation for decision-making. Moreover, in environments varying with agent behavior, SABM can interact with the environment using language. For instance, in policy studies related to natural resource management, the interplay between humans and the environment is frequently researched and modeled, and the complex influence of human behavior on the environment is often deemed challenging to simulate~\cite{schluter2017framework}. With SABM's feature of modeling in natural language, both human behaviors and their impact on the environment can be articulated and explained in text, circumventing the difficulties and interpretative limitations associated with setting numerical variables and parameters.

Based on these discussions, we have created a categorization for the range that can be modeled by SABM in Figure~\ref{fig:sabm:hierarchy}. In addition to the modeling of agents, interactions, and environments discussed above, we have also categorized this system setup based on the temporal and spatial structures involved in the simulation. For instance, evacuation simulations are often conducted with a discrete time setup and in grid-based environments, where each grid represents a potential position an agent can occupy. A comprehensive SABM instance can be described using these properties. For example, an evacuation can be represented as: \textit{heterogeneous agents (indicating agent heterogeneity) + common sense (signifying the basis of agent behavior) + discrete time (representing temporal structure) + regular grids (representing spatial structure) + local communication (denoting interaction) + agent behavior-based variation (characterizing the environment)}. We will further present the use of these properties in our case studies.


In summary, the significance of SABM lies in its (1) scope: its improved capability to address problems that were previously hard to model, and (2) fidelity: its potential to enhance models with formerly limited simulation fidelity. It can handle situations that involve decision-making expressed in natural language, or problems constrained by complexity or lack of raw data, where traditional theories have often resorted to extensive simplifications or approximations, thereby creating a gap between theory and reality. 


\subsection{Limitations of SABM}
\label{sec:sabm:limitations}
While SABM offers exciting prospects for more realistic and nuanced simulations, it does come with its set of limitations. Recognizing and addressing these limitations is crucial for the responsible and effective application of SABM in research and real-world scenarios.

\begin{enumerate}
    \item \textbf{Purpose-specific Modeling:} Like ABM models, an SABM model needs a well-defined objective. A universal model that attempts to capture every nuance may result in a model that is too vague or too convoluted to deliver meaningful results. Determining the right level of granularity is a challenge. Oversimplification might miss critical behaviors, while too much detail might render the model impractical due to computational or interpretability issues. Striking the balance between detail and purpose, as mentioned, remains an intuitive process, often necessitating iterative refinement based on results and insights from multiple runs of the model.
    \item \textbf{Computational Intensity:} An inherent aspect of ABM, and by extension SABM, is the focus on individual units or agents rather than aggregates. This individual focus can lead to intricate interactions and decision processes, particularly when LLMs are involved. Each agent, powered by an LLM, would require significant computation for each decision cycle, especially in scenarios where thousands or millions of agents are simulated. As technology advances, computational power increases. However, the intensive nature of SABM, especially with larger agent populations, might still cause scalability issues -- the simulation tends to be significantly slower when more agents are introduced -- and demand specialized hardware or distributed computing solutions. In this case, using ABM might be more cost-friendly. 
    \item \textbf{Control and Randomness:} In ABM, models usually have deterministic or semi-deterministic behaviors based on prescribed rules. In contrast, SABM models leverage LLMs, introducing a layer of unpredictability due to the inherent variability in LLM outputs. This lack of strict control can be both a strength (enabling a priori modeling and introducing randomness or diversity) and a limitation (introducing inaccuracy or inconsistency). The hallucinations in LLMs~\cite{zhang2023siren} -- LLMs occasionally generate content that diverges from the user input, contradicts previously generated context, or misaligns with established world knowledge -- can propagate and amplify in a system, especially if agent interactions build on these inaccuracies. This can lead to unanticipated system behaviors or outcomes that diverge significantly from expectations. Another issue is the update to the LLM (e.g., there are multiple versions of GPT-3.5 and GPT-4). Researchers need to be careful with the model version for reproducibility. 
    \item \textbf{Mimicking Human Behavior with Ethical Concerns:} While LLMs can approximate human decision-making and behavior, there are nuances and complexities in real human behaviors that may not be fully captured or represented. Especially in simulations addressing sensitive or ethically charged topics, there is a risk of the model unintentionally perpetuating biases, stereotypes, or misrepresentations. This is a significant concern, given the known issues with biases in AI models. It is imperative to incorporate rigorous ethical checks and possibly even human oversight in such scenarios. This can ensure that the outcomes align with societal values and do not inadvertently harm or misrepresent any group.
\end{enumerate}

\section{Implementation of SABM}
\label{sec:impl}

\begin{figure}[!t]
    \centering
    \includegraphics[width = \textwidth]{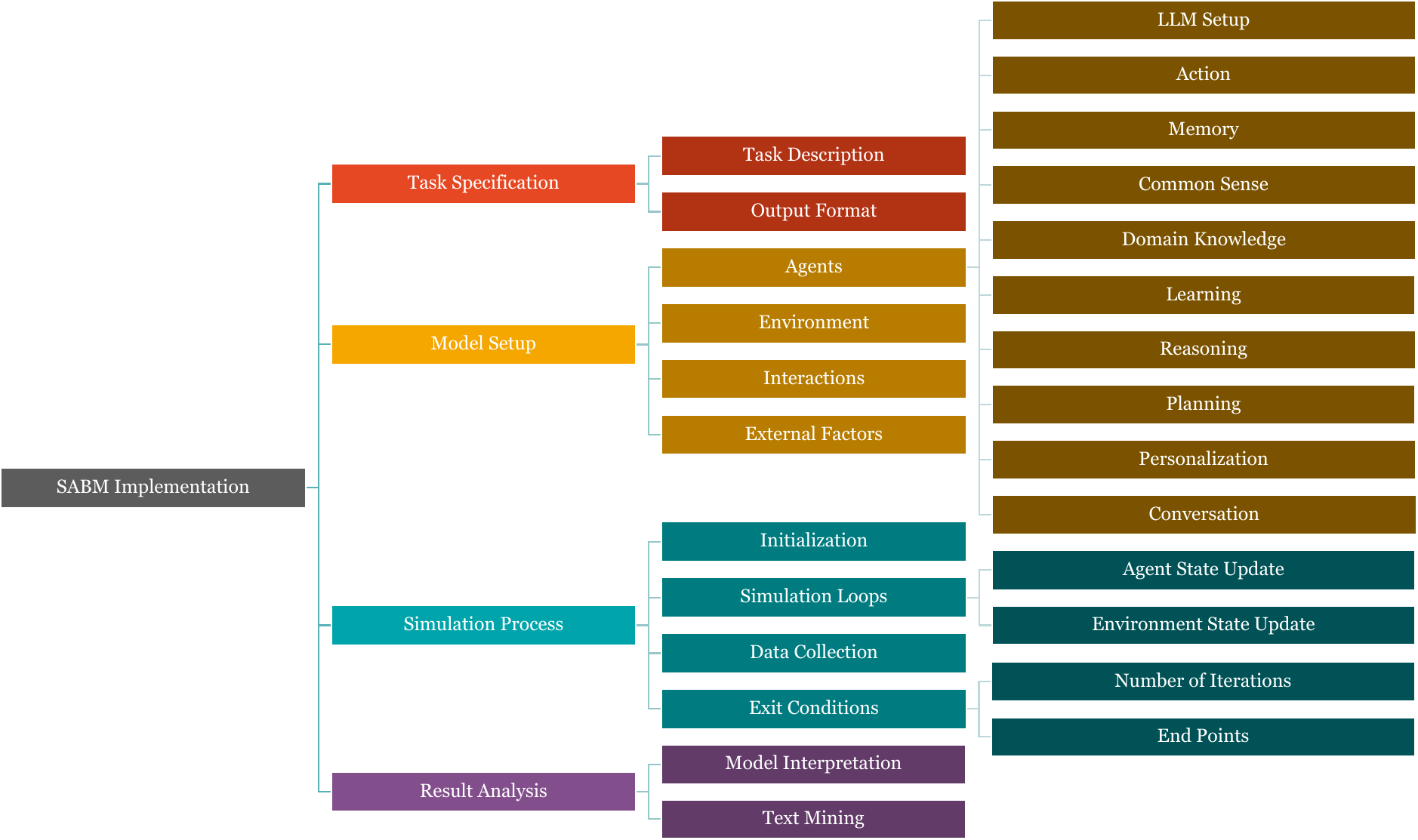}
    \caption{Methods of SABM implementation.}
    \label{fig:impl:sabm}
\end{figure}

In this section, we present the methodology of implementing SABM, which incorporates the SABM features discussed in Section~\ref{sec:sabm}. As shown in Figure~\ref{fig:impl:sabm}, a complete SABM implementation includes task definition, model setup, simulation process, and result analysis. We design a simple SABM instance of number-guessing game and use it throughout this section to introduce the methods for SABM implementation. Most methods introduced in this section employ prompt engineering, for which a guide is available at \cite{prompt-engineering-guide}.

\subsection{Task Specification} 
The initial step is to specify the task to the LLM. In the number-guessing game, there are two agents: one agent is tasked with guessing the number, while the other agent determines the target number and communicates to the guessing agent whether its guess is too high, too low, or precisely correct. 
The prompt (i.e., the input to the LLM) given to the agents is as follows (the two agents are referred to as the adjudicator and the guesser). 

\begin{prompt}{Number-guessing game}
    {\upshape To Adjudicator:} Now you are participating in a number-guessing game. You are the one responsible for thinking up the numbers. Please think of an integer, ranging from \var{range begin} to \var{range end}. Only reply the number (e.g., 12).
    \tcbline
    {\upshape To Guesser:} Now you are participating in a number-guessing game. You are the one in charge of guessing. The number will be an integer ranging from \var{range begin} to \var{range end}. After you made a guess, you will be informed if your guess is right, higher than the answer, or lower than the answer. Now please make your first guess. Only reply the number (e.g., 12).
\end{prompt}

The prompt describes the task and specifies the output format (only reply the number). Because LLMs emit outputs in natural language, specifying the format can make it easier to process the output. The adjudicator should take the first move, because logically, the game cannot commence without a distinct target number. Following the determination of the target number and the completion of the first guess, subsequent actions by both agents begin to repeat with the following prompt. 

\begin{prompt}{Number-guessing game (continue)}
    {\upshape To Adjudicator:} You are participating in a number-guessing game and you are the one responsible for thinking up the numbers. You decided \var{target number} as the answer. Your opponent had made a guess of \var{guess}. Can you tell your opponent if the guess is right, higher than the answer, or lower than the answer? If the guess is correct, please say ``Congratulations!''.
    \tcbline
    {\upshape To Guesser:} You are participating in a number-guessing game and you are the one to guess the number. The number will be an integer ranging from \var{range begin} to \var{range end}. Your previous guess was \var{previous guess}. The history of your guess is \var{guess history}. Only reply the number (e.g., 12).
\end{prompt}

\subsection{Model Setup}
\label{sec:impl:model-setup}
Following the task specification, we set up the model. This includes elements such as agents, environment, interactions, and external factors as illustrated in Figure~\ref{fig:impl:sabm}. Agent modeling manages various aspects of agents. 
The methods of agent modeling, such as memory, will be elaborated in the rest of this section. Environment modeling models the variables and procedures in the environment. Model designers may use textual descriptions to depict the environment and prompt them to the agent. Interactions can be categorized into agent-agent interactions and agent-environment interactions. The number-guessing game involves both interaction types. External factors are the factors outside the model but having interactions with the model, e.g., human factors and models constructed with other modeling approaches. Since external factors are not used in the number-guessing game or our case studies, we do not consider the methods for external factors here but envision the opportunities in Section~\ref{sec:future:hybrid}. 

In the number-guessing game, we first model the agents by establishing their states. Here, the adjudicator's state is captured by the \var{target number} variable, which is initialized with its first output and utilized each time it evaluates the guesser's guess. On the other hand, the guesser's state is captured by the \var{guess history} variable, which is used for subsequent guesses. In the simulation, we monitor the environment state, i.e., whether the \var{target number} has been given and whether the game should terminate. 

\myparagraph{LLM Setup}
The API of an LLM involves several parameters that need to be set up before use. Here, we use GPT as an example. As noted by OpenAI's API reference~\cite{gpt-parameters}, when using the GPT model we need to set the appropriate GPT model parameters according to the different objectives, which mainly include \texttt{model\_type}, \texttt{temperature}, and \texttt{max\_tokens}. The \texttt{model\_type} parameter refers to the version of the GPT model. Recent versions are GPT-3.5 and GPT-4. Snapshots are available for historical versions. In the number-guessing game, we set \texttt{model\_type} to \texttt{gpt-4-0613}, the snapshot of GPT-4 from June 13th, 2023. The \texttt{temperature} parameter, with a range of 0 to 2 and a default value of 1, affects the model's output randomness or its creativity. A higher temperature value (e.g., 1.2) yields more diverse and creative responses, whereas a lower value (e.g., 0.1) makes the output more deterministic and focused. In the number-guessing game, a \texttt{temperature} of 0.5 is chosen to ensure the relevance of responses. The \texttt{max\_tokens} parameter sets the maximum length of the generated output, with one token approximately equal to $3/4$ words in English. For this game, a \texttt{max\_tokens} value of 128 is found to be adequate, providing sufficient detail without overloading the simulation to lose efficiency. These GPT parameters do not have a one-size-fits-all optimal setting but should be configured based on the complexity and requirements of the task and the financial and computational cost the model designer can afford. 

\myparagraph{Common Sense}
In the number-guessing game, the agents can act solely based on their common sense -- they are able to understand the rules of the game, and accurately assess the magnitude of numbers, all without the need for explicit programming. 

\myparagraph{Memory} 
Given that prevalent LLMs such as GPT do not retain historical memory, providing the agent with a history becomes paramount. In the number-guessing game, this is given in the \var{guess history} variable. In the absence of such information, the guesser might repeatedly guess the same number. 

Due to the token limit of the LLM (e.g., 8k tokens for \texttt{gpt-4-0613}), maintaining extensive histories becomes a challenge. To address this limitation, recent advancements in the AI community have proposed two solutions. The first approach involves summarization~\cite{park2023generative}, which succinctly represents lengthy histories. The second approach employs text embedding~\cite{langchain}. Here, each segment of historical text is transformed into a high-dimensional vector using a text embedding model, such as OpenAI's \texttt{text-embedding-ada-002}~\cite{text-embedding}. These vectors, as well as the corresponding text segments, are subsequently stored in vector databases like Pinecone~\cite{pinecone}. To recall a memory associated with a particular topic, the topic text is transformed into a vector using the text embedding model. A nearest neighbor search~\cite{li2019approximate} is then conducted in the vector database to retrieve the most relevant text segment. The retrieved text segment is subsequently incorporated into the prompt provided to the agent.

\myparagraph{Action} 
Action is an important module of LLM-powered agents for producing the outcomes of the task~\cite{wang2023survey,xi2023rise}. An action made by the agent may result from the common sense of the LLM and the agent's memory, as shown in the number-guessing game. Other sources of an action include the LLM's domain knowledge, learning, reasoning, etc., which will be introduced in Section~\ref{sec:impl:advanced}. An action may impact the agents and the environment via interactions, and may trigger another action. In the number-guessing game, there are three actions: the adjudicator's \emph{think} and \emph{tell}, and the guesser's \emph{guess}. The action of \emph{think} assigns a value to \var{target number}, changing the state of the adjudicator. The action of \emph{guess} triggers the adjudicator to tell whether the guess is right or wrong. If the guess is wrong, the action of \emph{tell} changes the \var{guess history} of the guesser via an agent-agent interaction. If the guess is right, this action changes the environment to terminate the game via an agent-environment interaction. 

\subsection{Simulation Process}
\label{sec:impl:sim-proc}
\begin{figure}[!t]
    \centering
    \includegraphics[width = 0.95\textwidth]{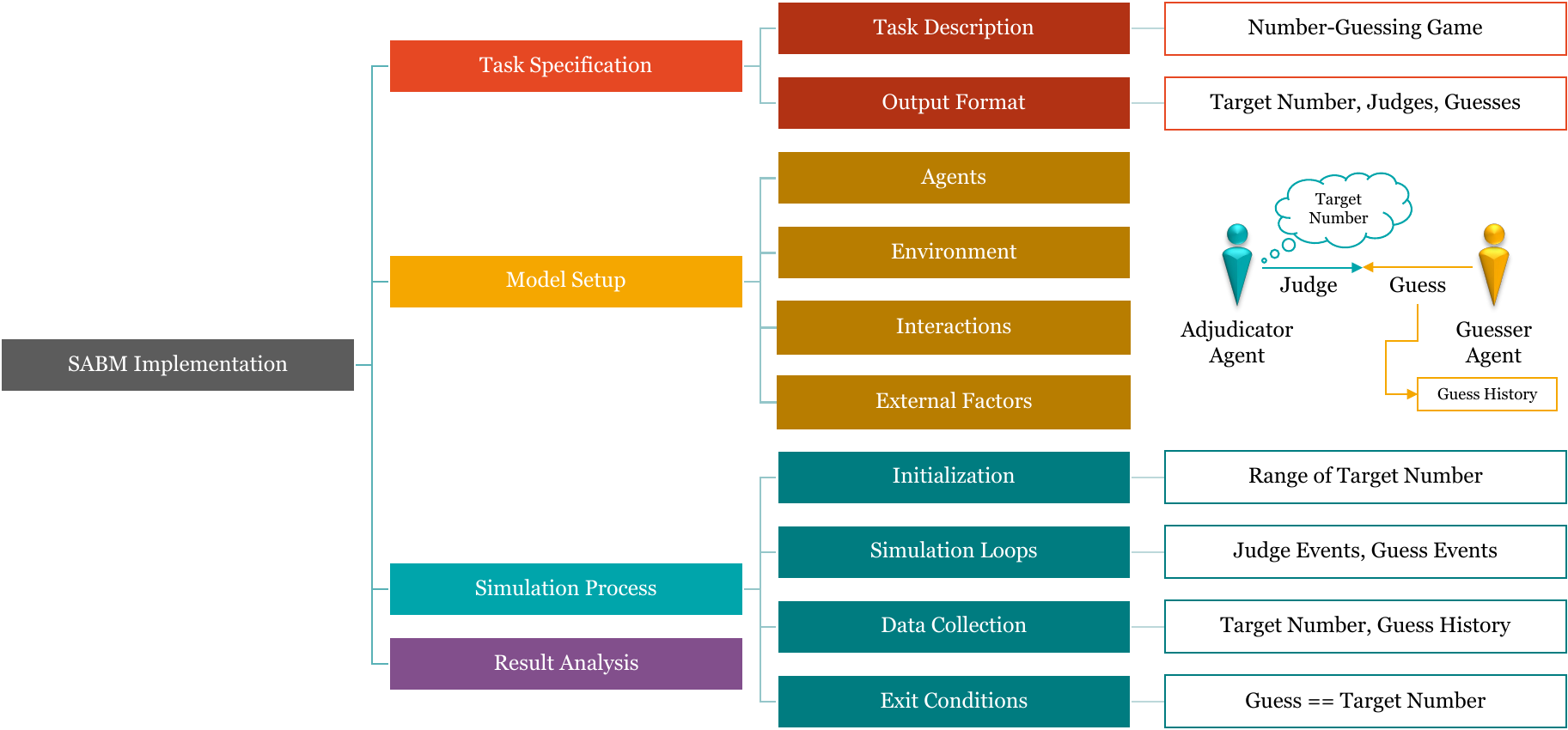}
    \caption{SABM implementation of the number-guessing game.}
    \label{fig:impl:number_guessing_example}
\end{figure}

As depicted in Figure~\ref{fig:impl:sabm}, the simulation process comprises four methods: initialization, simulation loops, data collection, and exit conditions. Figure~\ref{fig:impl:number_guessing_example} shows the implementation of the number-guessing game using the SABM simulation framework.

\myparagraph{Initialization} 
Depending on the needs of the task, we can initialize the simulation to observe the changes in the performance of the agent and the model under different initial conditions and settings. In the number-guessing game, initialization involves setting the range of integer guesses [\var{range begin}, \var{range end}]. We set the range to [1, 100] in this example.

The model designer may also override the adjudicator's decision on \var{target number} and determine its initial value. The following text shows two sets of simulation results for the game using initial values of 28 and 53 for \var{target number}, each line being the guess and the response from the adjudicator.

\begin{lstlisting}
#1
Target Number: 28
50 "The guess is higher than the answer."
25 "The guess is lower than the answer."
38 "The guess is higher than the answer."
32 "Your guess is higher than the answer."
30 "Your guess is higher than the answer."
28 "Congratulations!"

#2
Target Number: 53
50 "The guess is lower than the answer."
75 "The guess is higher than the answer."
63 "The guess is higher than the answer."
57 "The guess is higher than the answer."
55 "Your guess is higher than the answer."
53 "Congratulations!"
\end{lstlisting}

\myparagraph{Simulation Loops} 
Since many simulations run in an iterative manner, we can formulate the simulation as repeating one or more loops. Each loop consists of discrete events and involves reads and writes of agent and environment states. In the number-guessing game, there is one simulation loop, composed of two discrete events: the guesser's guess and the adjudicator's evaluation. 

\myparagraph{Data Collection} 
Data collection is the process of collecting data from the agents' outputs. The collected data are stored as variables and used in the rest of the simulation loop. In the number-guessing game, we collect the adjudicator's first output as \var{target number}, its subsequent outputs and the guesser's outputs as \var{guess history}. 

\myparagraph{Exit Conditions}
Exit conditions are the criteria for terminating a simulation. There are two types of exit conditions: predetermined number of iterations and end points. Predetermined number of iterations specifies the maximum time steps the simulation can run. End points are the conditions of variables or agents' outputs under which the simulation terminates. In the number-guessing game, we use the method of end points and terminates the simulation when the adjudicator outputs \str{``Congratulations!''}, whereas end points can be more complex than a single value, as shown in the case study in Section~\ref{sec:case-firm}.


\subsection{Advanced Agent Modeling}
\label{sec:impl:advanced}
In addition to the language ability, which is essential for model building, the learning ability and other abilities that LLMs have, as presented in Figure~\ref{fig:sabm:features}, can serve as advanced agent modeling methods to optimize the performance of SABM and make the simulation more realistic and adaptable. 


\myparagraph{Domain Knowledge} 
We can augment the agent to leverage the LLM's inherent domain knowledge, thereby extending its functionality. In particular, we know that binary search is an algorithm that can be used for guessing numbers. Whereas the agent may use this knowledge implicitly (e.g., the first four guesses in Simulations \#1 and \#2), we can explicitly prompt the guesser with this information (changes in prompt are marked in boldface):

\begin{prompt}{Domain knowledge (binary search)}
    {\upshape To Guesser:} You are participating in a number-guessing game and you are the one to guess the number. The number will be an integer ranging from \var{range begin} to \var{range end}. Your previous guess was \var{previous guess}. The history of your guess is \var{guess history}. Only reply the number (e.g., 12). \textbf{You can use binary search to optimize your guess.}
\end{prompt}

The simulation results are shown below (for simplicity, we only report the guesses). When merely prompted to utilize binary search, the guesser accurately employs this knowledge for decision-making, guessing the number in the fewest steps possible in each instance (in contrast, Simulations \#1 and \#2 spend 6 trials to reach the target number). Intriguingly, we do not need to provide the algorithmic description or the code of binary search; the agent autonomously derived the optimal strategy.

\begin{lstlisting}
#3
Target Number: 28
50, 25, 38, 31, 28

#4
Target Number: 53
50, 75, 63, 57, 53
\end{lstlisting}

\myparagraph{Learning}
One of the most significant features of LLMs is its capacity for learning from few-shot examples~\cite{brown2020language}, which implies that we can facilitate agents to discern behavioral patterns with very few examples, obviating the need for complex training procedures or rule-based instructions. In the number-guessing example, we can educate the agent about the guessing strategy by providing it with an instance of a real guess.

\begin{prompt}{Learning (from one-shot)}
    {\upshape To Guesser:} You are participating in a number-guessing game and you are the one to guess the number. The number will be an integer ranging from \var{range begin} to \var{range end}. Your previous guess was \var{previous guess}. The history of your guess is \var{guess history}. Only reply the number (e.g., 12). \textbf{An example of guesses aimed at 6: 50, 25, 12, 6}.
\end{prompt}

From the following simulation results, it is evident that one-shot learning successfully enabled the agent to learn the binary search strategy from the given example. Consequently, the guesses in the two simulations are equally effective as those in Simulations \#3 and \#4, where explicit prompts are given to utilize binary search.

\begin{lstlisting}
#5
Target Number: 28
50, 25, 38, 31, 28

#6
Target Number: 53
50, 75, 62, 56, 53
\end{lstlisting}

\myparagraph{Reasoning}
Another essential feature of LLMs is zero-shot reasoning~\cite{kojima2022large}, which uses reasoning instructions instead of examples. A popular reasoning paradigm is chain-of-thought~\cite{wei2022chain}, which employs a series of intermediate reasoning steps to improve the agent's ability of performing complex reasoning. In contrast to simply giving an answer, this approach can significantly improve the agent's performance in complex tasks. In the number-guessing game, we can instruct the agent to reason before guessing, using the following prompt. 

\begin{prompt}{Reasoning}
    {\upshape To Guesser:} You are participating in a number-guessing game and you are the one to guess the number. The number will be an integer ranging from \var{range begin} to \var{range end}. Your previous guess was \var{previous guess}. The history of your guess is \var{guess history}. \textbf{Please briefly provide the reason for your guess in the first line and reply with the number in the second line (e.g., 12).}
\end{prompt}

The output of the guesser is given below, with the reason followed by the guessed number. Note that instructing the agent to give the answer prior to the reason may yield a different result, especially in complex tasks. This is because some LLMs such as GPT are autoregressive models, which predict the next word using previous words. Moreover, in the reasoning method, the action is always output together with the reason. This differs from the planning and interpretation methods introduced later. 

\begin{lstlisting}
#7
Target Number: 28
"I chose 50 because it's the midpoint in the range of 1-100, which will allow me to eliminate half of the possible numbers with my first guess no matter if it's higher or lower." 50 
"Since the previous guess of 50 was higher than the answer, I am guessing halfway between 1 and 50 to narrow down the possibilities." 25
"Since my previous guess of 25 was lower and 50 was higher, I am choosing a number in between, which is 38." 38
"Since 38 was higher and 25 was lower, the number should be in between. Therefore, I chose the middle number, 32." 32
"Since my previous guess of 32 was higher than the answer and my guess of 25 was lower than the answer, I am choosing a number in the middle of 25 and 32, which is 29." 29
"Since 29 was higher than the answer and 25 was lower, the answer must be between 26 and 28. I'm guessing 27 as it's the middle number." 27 
"According to the history, my guess of 27 was lower than the answer, and my guess of 29 was higher than the answer. So the answer must be 28." 28 

#8
Target Number: 53
"I choose 50 as my first guess because it is the midpoint of the range 1-100 and will effectively split the possible choices into two equal halves." 50 
"Since the previous guess of 50 was lower than the answer, I am guessing halfway between 50 and 100 to narrow down the range." 75 
"Since my previous guess of 75 was higher and 50 was lower, the number should be somewhere in between. I am guessing 62 as it is roughly in the middle of 50 and 75." 62 
"Since 62 was higher and 50 was lower, the number must be between 50 and 62. So, I am choosing the middle number, 56, as my next guess." 56 
"Since my previous guess of 56 was higher and 50 was lower, the number must be between 50 and 56. Therefore, I am guessing 53." 53 
\end{lstlisting}

\myparagraph{Planning} 
As an important module of LLM-powered agents~\cite{wang2023survey,xi2023rise}, planning imbues agents with human-like strategies to enhance their performance in task solving. An agent can generate a plan for its actions and the plan will be prompted to the agent again to take effect. There are two types of planning: formulation and reflection~\cite{xi2023rise}. In formulation, an agent formulates a plan and executes the plan for its subsequent actions. In reflection, an agent receives feedback from the environment and revises its plan. An example of reflection is given below, where we instruct the guesser to work out a plan based on previous guesses in the number-guessing game. 

\begin{prompt}{Planning}
    {\upshape To Guesser:} You are participating in a number-guessing game and you are the one to guess the number. The number will be an integer ranging from \var{range begin} to \var{range end}. Your previous guess was \var{previous guess}. The history of your guess is \var{guess history}. \textbf{Based on your guess history, what is your strategy for the next few guesses?}
\end{prompt}

Due to the absence of memory in GPT, the output \var{strategy} needs to be prompted to the guesser again for the subsequent guesses to take effect: 

\begin{prompt}{Planning (reprompt)}
    {\upshape To Guesser:} You are participating in a number-guessing game and you are the one to guess the number. The number will be an integer ranging from \var{range begin} to \var{range end}. Your previous guess was \var{previous guess}. The history of your guess is \var{guess history}. \textbf{Your strategy for this guess is \var{strategy}.} Only reply the number (e.g., 12).
\end{prompt}

The output of the guesser is given below, with numbers indicating their guesses and text indicating their strategy. Note that the plan is reintroduced to the agent to affect actions and can be used for multiple subsequent actions. 

\begin{lstlisting}
#9 (Planning after round 3)
Target Number: 28
50, 25, 37
"Based on the history, the number is between 25 and 37. My next guess would be the midpoint of this range, which is 31. If that guess is still higher than the answer, I would then guess the midpoint between 25 and 31, which is 28. If 31 is lower than the answer, I would guess the midpoint between 31 and 37, which is 34. I would continue this process of guessing the midpoint of the remaining range."
31, 28
\end{lstlisting}

As an example of formulation, we can instruct the guesser to design a plan as the beginning of this game and use the plan for all its guesses. The result is shown below, where the guesser spends more rounds to find the answer due to rounding of numbers. 

\begin{lstlisting}
#10 (Planning at the beginning)
Target Number: 28
"Since there is no previous guess history, the best strategy to start with would be the binary search method. This method involves guessing the middle number in the given range, and then adjusting your guess based on whether it's too high or too low. 1. Guess 50 (the midpoint of 1 and 100).  2. If the game indicates your guess is too high, guess 25 (the midpoint of 1 and 50). If it's too low, guess 75 (the midpoint of 50 and 100)."
50, 25, 38, 32, 29, 27, 28
\end{lstlisting}

\myparagraph{Personalization} 
In SABM, we can personalize agents and endow different agents with unique characteristics, thereby better simulating the diversity of real-world entities. This personalization can encompass traits like personality characteristics, decision-making inclinations, physical and mental states, and other attributes affecting behavior patterns. Furthermore, with the possibility of modeling in natural language, the setting of attributes that are highly subjective and challenging to establish with traditional programming and modeling methods also becomes feasible with the use of LLMs.

In the number-guessing game, we can endow the agent with strong subjectivity. Suppose we aim not for the agent to guess the number with the optimal strategy, but to exhibit behavior patterns similar to human game participants. In such a case, we can imbue it with some personality traits. For instance, we can have the agent emulate the behavior of participants who adopt more aggressive or conservative guessing strategies.

\begin{prompt}{Personalization}
    {\upshape To Guesser:} You are participating in a number-guessing game and you are the one to guess the number. The number will be an integer ranging from \var{range begin} to \var{range end}. Your previous guess was \var{previous guess}. The history of your guess is \var{guess history}. \textbf{\var{persona}}. Only reply the number (e.g., 12).
\end{prompt}

Where the options for \var{persona} are:

\begin{lstlisting}
"aggressive": "You need to perform aggressively while guessing."
"conservative": "You need to perform conservatively while guessing."
\end{lstlisting}

As shown in the results below, the patterns displayed in these results are markedly different from the previous simulations that employed common sense or binary search. The number of steps required by the agent to guess the correct number tends to increase, and for the same target number, the behavior of the aggressive persona is indeed more aggressive compared to the conservative persona: the aggressive persona generally takes larger steps. This behavior aptly reflects the characteristics set in the personalization.

\begin{lstlisting}
#11 Aggressive Persona
Target Number: 28
50, 25, 38, 32, 28

#12 Conservative Persona
Target Number: 28
50, 25, 38, 35, 30, 27, 29, 28

#13 Aggressive Persona
Target Number: 53
50, 75, 60, 55, 52, 56, 54, 53

#14 Conservative Persona
Target Number: 53
50, 75, 62, 57, 54, 53
\end{lstlisting}

In addition to prompting, we may also tune the LLM parameters for personas. For example, since the \texttt{temperature} parameter controls the randomness of outputs, a small \texttt{temperature} may simulate a predictable persona, while a large \texttt{temperature} may result in diversity or creativity. Similarly, we can tune the \texttt{max\_tokens} parameter for an uncommunicative agent or a talkative agent.

\myparagraph{Conversation}
LLMs can be used to simulate conversations between agents. Because agents cannot directly communicate, a mediator is required to pass the message between them. For example, we can use the following prompts in the number-guessing game to generate a hint from the adjudicator and then pass the hint to the guesser. 

\begin{prompt}{Conversation}
    {\upshape To Adjudicator (only after the adjudicator decides the target number):} You are participating in a number-guessing game and you are the one responsible for thinking up the numbers. You decided \var{target number} as the answer. \textbf{To help your opponent guess the number, can you give a hint to your opponent?}

    \tcbline

    {\upshape To Guesser (first guess):} Now you are participating in a number-guessing game. You are the one in charge of guessing. The number will be an integer ranging from \var{range begin} to \var{range end}. After you made a guess, you will be informed if your guess is right, higher than the answer, or lower than the answer. \textbf{To help you guess the number, your opponent gives you a hint: \var{hint}.} Now please make your first guess. Only reply the number (e.g., 12).

    \tcbline

    {\upshape To Guesser (subsequent guesses):} You are participating in a number-guessing game and you are the one to guess the number. The number will be an integer ranging from \var{range begin} to \var{range end}. \textbf{To help you guess the number, your opponent gives you a hint: \var{hint}.} Your previous guess was \var{previous guess}. The history of your guess is \var{guess history}. Only reply the number (e.g., 12).
\end{prompt}

For the target number of 53, the hint given by the adjudicator and the guesser's guesses are as follows.

\begin{lstlisting}
#15
Target Number: 53
Hint: "It's a two-digit prime number and both digits are prime numbers as well."
23, 37, 53

#16
Target Number: 53
Hint: "The number I'm thinking of is a two-digit prime number and the sum of these two digits is 8."
17, 71, 53
\end{lstlisting}

\subsection{Result Analysis}
\label{sec:impl:result-analysis}
For analyzing simulation result, we elaborate model interpretation, a result analysis method tailored to SABM. 

\myparagraph{Model Interpretation}
Let us delve into the aspect of model interpretation. Thanks to the advantages of employing an LLM, we are no longer confined to the ``black box'' predicament of traditional computer simulation, where the behavior within the simulation is sometimes inscrutable and hard to explain. We can instruct the LLM to explicate why the agent behaves as it does, thereby rendering the behavior of the agent and the system interpretable and turning the inscrutable ``black box'' into an understandable ``white box''. Not only can we understand behaviors from model interpretation, but we can also incorporate these generated explanations as part of the result analysis. 

In the number-guessing game, we can append an explanation segment after each guess, enabling the agent to clarify its thought process. Note that we can restrict the word count of these explanations by adjusting the \texttt{max\_tokens} variable or by including prompts like ``(no more than XX words)''.

\begin{prompt}{Model interpretation}
    {\upshape To Guesser:} You are participating in a number-guessing game and you are the one to guess the number. The number will be an integer ranging from \var{range begin} to \var{range end}. The history of your guess is \var{guess history}. \textbf{Can you briefly explain why you make your previous guess as \var{previous guess}? (No more than 40 words.)}
\end{prompt}

The guesses and corresponding explanations are reported as follows. Note that the agent needs to give an explanation upon seeing the action. This differs from reasoning and planning. 

\begin{lstlisting}
# Model Interpretation
Target Number: 28
50 "I guessed 50 because it's the midpoint of the range 1-100, allowing me to eliminate half of the possible numbers based on whether it's higher or lower."
25 "I guessed 25 because the midpoint of 1 and 50 is 25. Since the previous guess of 50 was higher than the answer, I halved the range to narrow down the possibilities."
37 "I chose 37 because it's roughly in the middle of the new range (25-50) after receiving feedback that 50 was too high and 25 was too low."
30 "I guessed 30 because the answer is lower than 37 and higher than 25. So, I chose a number in the middle of this range to narrow down the possibilities."
28 "I guessed 28 because the previous clues narrowed the answer to between 25 and 30. Since 30 was too high, I chose 28 as a number in that range, closer to the lower boundary."
\end{lstlisting}

Other data analysis methods, like text mining techniques, can be used for result analysis in specific problem contexts. For instance, we can analyze the natural language log generated by the LLM using sentiment analysis to quantify the agent's emotional state. We omit the demonstration here due to the simplicity of the number-guessing game. 

\begin{table}[!t]
  \small
  \centering
  \caption{SABM methods utilized in the number-guessing game.}
  \begin{tabular}{p{.28\linewidth}|p{.28\linewidth}|p{.34\linewidth}}
    \hline
    \textbf{SABM Features} & \textbf{SABM Methods} & \textbf{Number-Guessing Game Instances} \\
    \hline
    Modeling in natural language & All & Modeling of the game \\
    Modeling complex behaviors & Learning, reasoning, planning, personalization, conversation, model interpretation & Learning and reasoning to guess, interpreting results \\
    Common sense       & Action, common sense & Correct execution of guesses \\
    Built-in knowledge & Domain knowledge & Knowledge of binary search \\
    Personalizability  & Personalization & Impersonating two personas \\
    Subjective judgment & Personalization & Performing aggressive or conservative guesses \\
    Randomness and diversity & LLM setup, action & Generating non-fixed outputs \\
    Learning ability   & Learning & Learning on a guessing example \\
    Reasoning ability  & Reasoning, planning & Providing reasons and revising strategies \\
    Model interpertability & Model interpretation & Explanation of guesses \\
    \hline
  \end{tabular}
  \label{tab:method:number-guessing-example-conclude}
\end{table}

At this juncture, we have discussed the methods that implement SABM. The methods utilized in the number-guessing game (see Table~\ref{tab:method:number-guessing-example-conclude} for a summary) can be extended to various tasks.

\section{Design of SABM Instances}
\label{sec:design}
With the methods for SABM implementation introduced in Section~\ref{sec:impl}, we present the methodology of designing SABM instances for studying real-world systems.

Unlike deterministic mathematical models in ABM~\cite{grimm2006standard}, LLMs often introduce a degree of randomness and diverse behavior, necessitating a paradigm shift in model construction methodology. One major challenge that our methodology seeks to address is the issue of standardization. ABM often suffers from a lack of standardization, making it challenging to compare simulation models of the same problem studied~\cite{polhill2008using}. To facilitate standardization in SABM, we propose a systematic methodology for model construction, underpinned by three principles:

\begin{enumerate}
    \item \textbf{Define a Clear Problem Scope:} Each model should have a well-defined problem scope. This is critical to ensuring the model's results are relevant and meaningful.
    
    
    \item \textbf{Enable Easy Debugging:} In order to streamline the development process and improve the robustness of models, the methodology should include processes and tools to make debugging easier.
    
    \item \textbf{Incorporate Standardized, Systematic, and Interpretable Construction:} Model construction should be systematic, following a standardized approach that is readily interpretable by other researchers. This will aid in the reproduction of studies, enhance the comparability of results across different simulations, and foster collaboration among researchers.
\end{enumerate}


\begin{figure}[!t]
    \centering
    \includegraphics[width = \textwidth]{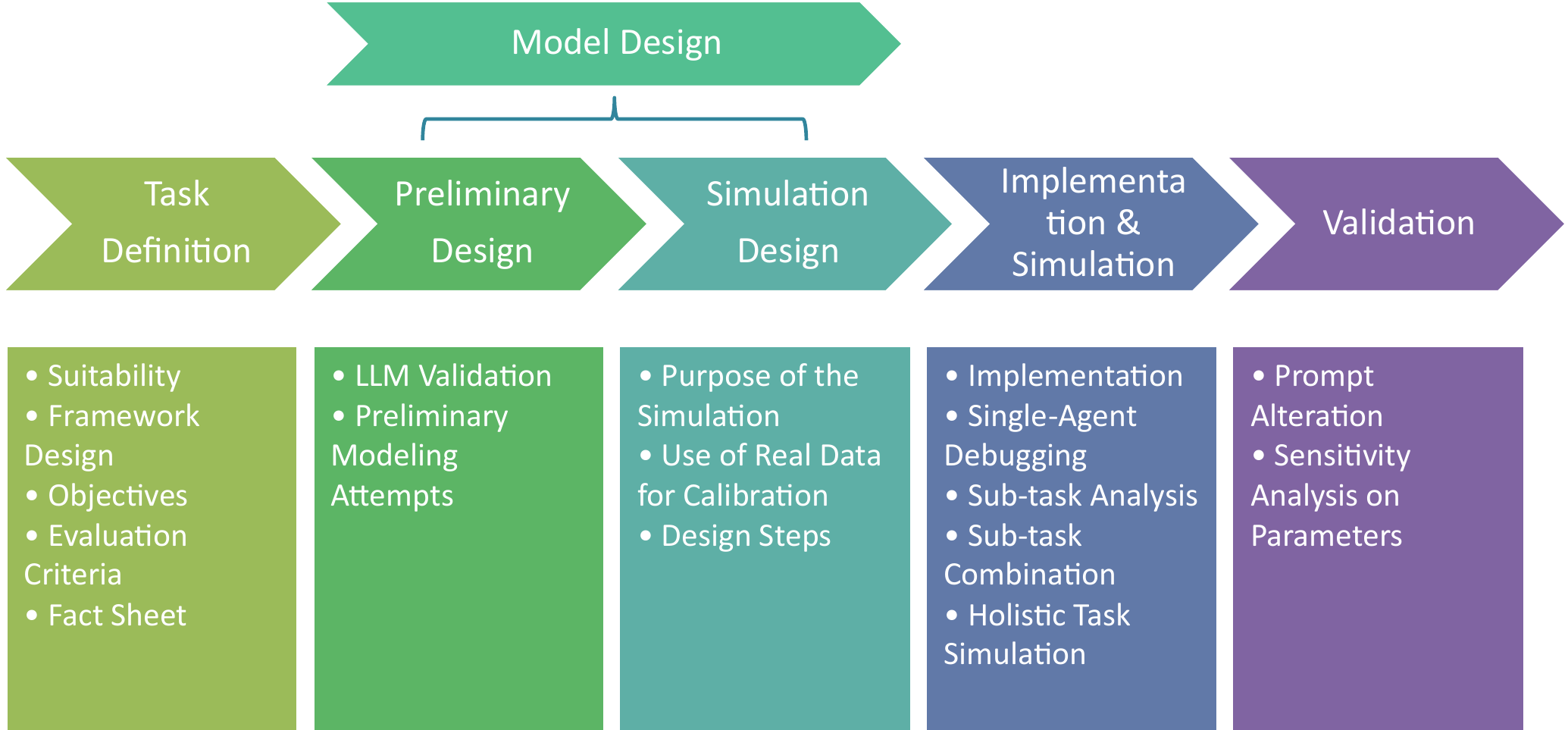}
    \caption{Workflow of SABM instance design.}
    \label{fig:design:workflow}
\end{figure}

Following these principles, we present the workflow for designing an SABM instance, which is illustrated in Figure~\ref{fig:design:workflow}.

\subsection{Workflow}
The model construction is divided into several phases: task definition, model design, implementation \& simulation, and validation. This approach is inspired by the ODD (Overview, Design concepts, Details) protocol ~\cite{grimm2006standard} for describing individual-based and agent-based models, which has been widely accepted in the ABM community due to its clarity and comprehensiveness.

\begin{enumerate}
    \item \textbf{Task Definition:} The first step in model construction is to clearly define the problem scope. This task definition phase includes identifying the key entities, their behaviors, and the interactions among them. The task definition should also specify the objectives of the model, the expected outcomes, and the criteria for assessment. This phase is crucial for setting the direction of the model design and ensuring that the model is relevant to the problem studied.

    \item \textbf{Model Design:} The model design phase involves the setup of the agents, the environment, and the rules of interactions. Following the methods introduced in Sections~\ref{sec:impl:model-setup} and~\ref{sec:impl:advanced}, the design of the agents should consider the capabilities of the LLMs used, such as their ability to generate diverse responses and to adapt to different contexts. The environment is defined by the state variables and the rules that govern their changes. The rules of interactions specify how the agents interact with each other and with the environment. The model design should be systematic and interpretable, with clear documentation of the assumptions and decisions made during this process. 

    \item \textbf{Implementation and Simulation:} The implementation and simulation phase involves coding and running of the model. The simulation and debugging of the model can be done in a bottom-up fashion. Model designers may decompose the task into several sub-tasks, start with single agents in a simple sub-task, and then extend to adding more agents and sub-tasks. 

    \item \textbf{Validation:} The validation phase involves the testing of the model to ensure that it behaves as expected and that it accurately represents the problem studied. This can be done through a variety of methods, including the model validation methods to be introduced in Section~\ref{sec:design:valid}, comparison with empirical data, and theoretical analysis. The validation process should be transparent and reproducible, with clear documentation of the methods used and the results obtained. 
\end{enumerate}


\subsection{Task Definition}
\label{sec:design:task-def}

The task definition phase involves several key steps: determining the suitability of SABM, designing the simulation framework, 
defining the objectives, and establishing the evaluation criteria.

\begin{enumerate}
    \item \textbf{Suitability for SABM:} The first step is to assess whether the task or a sub-task under study is suitable for SABM. This involves evaluating whether the task exhibits complexity, reasoning, and emergence, etc., which are the key characteristics that make a task suitable for SABM.

    \item \textbf{Framework Design:} After establishing that the task falls within the scope of SABM, the task can be decomposed and simplified into a basic simulation framework and a series of prompts handling different functions of the task. For instance, in the number-guessing game, the framework could be a dialogue between two agents. Theoretical simplifications or amplifications can be used to reduce the complexity of the task and highlight the features under study. For example, in a simulation of a predator-prey ecosystem, the complex interactions among various species can be simplified by focusing on a single predator, and the dynamics of the ecosystem can be amplified by introducing sudden changes in the environment~\cite{abrams1993predation, testa2012agent}. Model designers may also start with a model constructed with ABM methods, and gradually replace its components with SABM methods. Such framework design methodology resembles the Ship of Theseus and will be used in our case study in Section~\ref{sec:case-evac}.


    \item \textbf{Objectives:} The goals of the simulation need to be clearly defined. These goals should be directly related to the research problem and should specify what the simulation is expected to achieve. In the number-guessing game, the goals could include understanding the guessing patterns of the agents under different personas.

    \item \textbf{Evaluation Criteria:} Despite the complexity and uncertainty of LLMs, it is important to establish clear criteria for evaluating the performance of the model. These criteria should reflect the goals of the simulation and should provide a basis for determining whether the simulation has achieved its intended outcomes. For instance, in the number-guessing game, a criterion could be that the agent should not guess the same number twice, as this would indicate a problem with the agent's memory or with the prompt.

    \item \textbf{Fact Sheet:} The design of an SABM instance necessitates a systematic and collaborative approach. We suggest creating a comprehensive fact sheet before the model design. This fact sheet serves as a guide for model designers, outlining the experiments to be conducted, the affordable requirements for the simulation, the expected performance of the model under various conditions, and the interpretation of the simulation results. Establishing clear expectations can avoid post-hoc rationalization of results that may involve researcher bias. The fact sheet is particularly beneficial in interdisciplinary teams, where researchers may have diverse backgrounds and expertise. It facilitates a common understanding of the problem scope, desired results, and the conclusions validated by each sub-task or component of the model. This consensus-building approach is crucial in complex simulation modeling, as it ensures that all team members have a shared understanding of the model's objectives and expected outcomes. 
    In addition to outlining expectations and facilitating consensus, the fact sheet also serves as a platform for sharing experiences and insights about the modeling process. Model designers can share information about what sub-tasks and steps were successful and what did not work, thereby enhancing the efficiency of the modeling process.
\end{enumerate}

\subsection{Model Design}
The model design phase is divided into preliminary design and simulation design. 

\begin{figure}[!t]
    \center{\includegraphics[width = 0.8\textwidth]{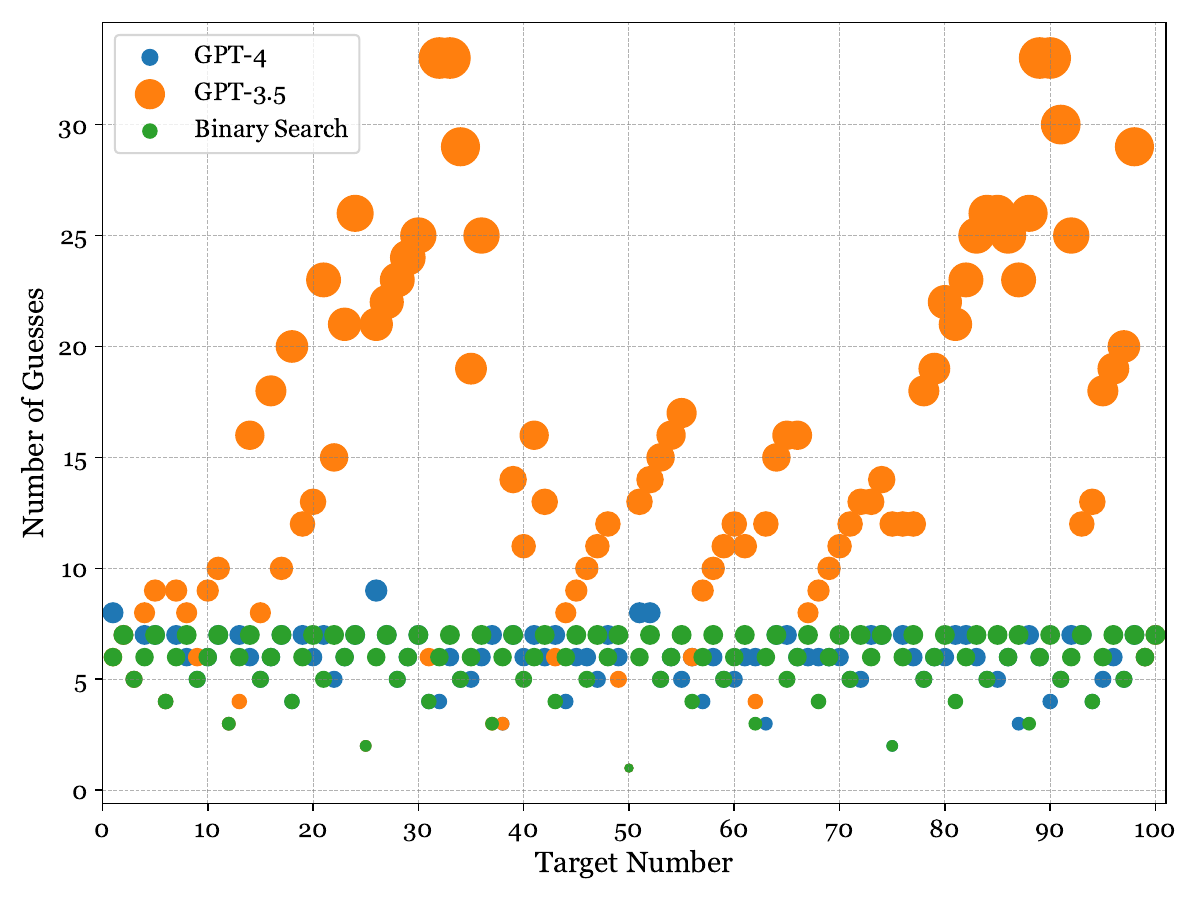}}
    \caption{Number of guesses to reach the target number by \texttt{gpt-4-0613} \texttt{gpt-3.5-turbo-0613} models in the number-guessing game.}
    \label{fig:design:number_guessing_comparison}
\end{figure}

\subsubsection{Preliminary Design}

The preliminary design aims to ensure that the task can be realized using the designed framework and is easy to debug. This stage focuses on whether the LLM used can accurately understand the meaning and intent of the prompt and react reasonably according to the prompt.

\begin{enumerate}
    \item \textbf{LLM Validation:} The first step is to verify whether the behavior of the LLM as an agent meets our expectations. For instance, in the number-guessing game, we might observe that GPT-4 can accurately understand the task and respond appropriately, while GPT-3.5 exhibits a high degree of randomness in its responses. Figure~\ref{fig:design:number_guessing_comparison} plots the number of guesses required by the two models to reach the target number. If the goal is to study how to guess the target number as quickly as possible, GPT-4 would be a better choice. Conversely, if the goal is to study patterns of random guessing, GPT-3.5 may be a good option as well, but further testing would be needed to verify whether GPT-3.5 accurately understands the prompt.
    
    The model interpretation methods discussed in Section~\ref{sec:impl:result-analysis} can be used to test the model. For example, we can ask the LLM to explain its decision and check whether the explanation is consistent with the decision and whether it accurately reflects the LLM's understanding of the prompt. If the LLM cannot understand the prompt or respond reasonably, the framework design or the objectives may need to be adjusted. Additionally, the use of a chain-of-thought approach~\cite{wei2022chain} can also be considered to enhance the interpretability and on-task performance of the LLM.
    
    In complex situations, LLMs may be able to approach the task from a mathematical perspective but do not necessarily make decisions in accordance with the mathematical results, thus engaging in behavior that may seem counterintuitive. However, this counter-common sense is not necessarily wrong; it mirrors the real world, where complexity comes from. An example of bounded rationality in economics is when two parties engaged in a transaction act in a manner that might diminish profits in an attempt to expand market share. This behavior can result in a decrease in profits for both parties as well as a reduction in their total profit, yet it remains consistent with real-world observations. We can identify whether this counterintuitive behavior is consistent with reality or a mistake through model interpretation, and distinguishing between these two patterns is important for understanding the complexity of the LLM's behavior.

    \item \textbf{Preliminary Modeling Attempts:} The second step is to make initial attempts at modeling the basic components for the task. For example, in the case study of emergency evacuation that will be presented in Section~\ref{sec:case-evac}, the preliminary design will not involve complex pedestrian flows or environmental impacts. Instead, the goal would be to determine whether an LLM can accurately understand a 2D grid map and an evacuation task so that agents can move towards the evacuation exits in the correct way. Through these initial modeling attempts, we can identify effective prompts (e.g., determine what prompts worked, what factors the agent would base its decision on, what prompts are redundant, etc.), understand the agent's actions under simple conditions (e.g., how the agent would move to the exit during evacuation), and decide how to extend the model and prompts to complete the task. 
\end{enumerate}

To sum up, the preliminary design helps us to build a series of basic and usable prompts, which can serve as a foundation for subsequent model extensions. This process helps to exclude invalid prompts and keep the model simple, reducing the risk that the model will be sensitive to specific irrelevant prompts. This will also significantly reduce the difficulty of debugging the model especially considering that the model will become more complex in the simulation design phase.

\subsubsection{Simulation Design}

The simulation design extends the preliminary design to complete the task.
Although it is inherently task-specific, there still exist common design considerations that can be applied across different scenarios.

\begin{enumerate}
    \item \textbf{Purpose of the Simulation:} The purpose of the study significantly influences the design of the model. If the aim is to accurately reproduce reality, the model may require calibration with real-world data. If the goal is to propose a new theory, the model might need to be simplified to clearly express the relationships between the key variables involved in the research problem. In the case of scaling, a balance must be struck between computational power and model granularity, with theoretical and empirical data used to simplify the model to a certain extent.

    \item \textbf{Use of Real Data for Calibration:} Since the use of LLMs introduces less control and more randomness to the model, the use of real-world data for calibrating the model settings and parameters is a crucial consideration for reproducing reality. LLMs, trained on vast amounts of data, may produce outputs influenced by their training data. For instance, an LLM might generate a correct response to a prompt because 
    it has encountered similar prompts in its training data, rather than because it truly understands the prompt, which makes the traditional method of using benchmark unreliable~\cite{oppenlaender2023mapping}. This possibility should be kept in mind when calibrating the model with real-world data, and model interpretation techniques should be used to discern whether the LLM's behavior is driven by its reasoning ability or by its training data. Perturbing the data may be a good way to avoid this potential data leakage problem.

    \item \textbf{Design Steps:} We advocate for a step-by-step modeling and testing approach for SABM design. Traditional reductionist approaches often fail when applied to complex systems, and the nature of LLMs makes it difficult to simply combine solutions to sub-tasks into a solution for a larger task. An LLM might perform well on individual sub-tasks but fail to grasp the overall task when these sub-tasks are combined or when they involve complex numerical or physical reasoning~\cite{wu2022ai, xie2023translating}. To avoid undesired results from model splitting and combining, we suggest using a method of gradually adding elements to the sub-tasks and incrementally increasing the variables and dimensions involved in the model as per the task requirements. This approach allows for the examination of how newly added prompts affect the agent's actions in each sub-task, and facilitates model debugging to achieve the original task step by step. Note that this is not equivalent to using agent's solution to each sub-task as input to the next sub-task, but rather continuously realizing new sub-tasks based on past tasks. That is, this is not a splitting of tasks, but an overlay of the sub-tasks to realize the original task.
\end{enumerate}

\myparagraph{Design Steps} 
The design of a simulation model often begins with the implementation of simple rules, which are then gradually expanded to encompass more complex scenarios. This approach allows for the verification of the model's accuracy in a simplified context before moving on to more intricate rules. Drawing from our experience with the number-guessing game, we propose the following step-by-step process for adding sub-tasks to the simulation model:

\begin{enumerate}
    \item \textbf{Base Modeling:} The base model should be as simple as possible, focusing on the core functionality of the agents. A key issue in base modeling is setting the essential parameters for the simulation. This includes the configuration of the LLM and initialization of variables. In the number-guessing game, the agent configuration could include the choice of the \texttt{temperature} and \texttt{max\_token} parameters. The initialization could include the range of numbers to guess from and the method for setting the target number. 
    In addition to parameter setting, advanced agent modeling methods (e.g., reasoning, planning, and personalization) are sometimes necessary for the agents to yield reasonable outputs. This could involve defining the agents' goals, preferences, and behaviors, which can guide their interactions with other agents and the environment. For example, agents may need to reason or plan before presenting the answer, especially when handling mathematical or logical tasks. 


    \item \textbf{Data Calibration:} After the base model has been created, it can be calibrated with real-world data. Data collected from real-world behaviors (e.g., human volunteers') can serve as ground truth. The calibration could involve the adjustment of model parameters or personalization prompts to better reflect the data or the use of few-shot examples to guide the model's behavior. An example of such practice will be demonstrated in the case study in Section~\ref{sec:case-plea}. 
    
    
    \item \textbf{Sub-task Addition:} Once the model is calibrated, additional sub-tasks can be added. A typical sub-task is testing different personas for the agents. 
    Other sub-tasks include creating dialogues between agents, defining more complex rules for interactions, and introducing new tasks for the agents to perform. Besides, we can scale up the number of agents to make the model results more general.
\end{enumerate}

\subsection{Implementation and Simulation}
\label{sec:design:impl-sim}
With the designed model, we can implement it and run the simulation. In particular, debugging is a crucial process, allowing us to identify and correct any issues that may arise. The steps we propose for implementing the model and running the simulation are carried out in a bottom-up fashion and listed as follows. 

\begin{enumerate}
    \item \textbf{Implementation:} The coding of the model may involve both programming language and natural language, where prompt engineering techniques can be used to refine the natural language prompts. The code is suggested to be written in a modular and reusable manner, with clear comments and documentation. This will facilitate debugging and make the model easier to understand and modify. 
    
    \item \textbf{Single-Agent Debugging:} Firstly, we need to ensure that the behavior and decision-making patterns of individual agents are reasonable and align with actual real-world situations. Prompts that are described inaccurately or ambiguously can lead agents to misunderstand and exhibit biased behavior. This can result in unexpected outcomes when multiple agents interact, making it difficult to pinpoint the causes of such behaviors. To address this, we can employ single-agent debugging, which focuses solely on understanding what information an individual agent receives and how it might react. By adjusting the information provided to the agent (which, in actual simulations, is generated through interactions with other agents and/or the environment), we can observe the agent's reactions under different circumstances and modify the prompts accordingly to align the agent's behavior with reality. Since debugging a single agent involves fewer inputs and outputs compared to multi-agent interactions, this approach can effectively reduce both financial and computational costs. An example of such practice will be demonstrated in the case study in Section~\ref{sec:case-evac}. 
    
    \item \textbf{Sub-task Analysis:} For each sub-task, we systematically observe the agents' actions under various situations with the same prompt configuration. By comparing the agents' behaviors with existing theories and ground truth, we can assess the reasonableness of the model construction and the prompt. This process is akin to an ablation study~\cite{cohen1988evaluation}, where each component of the model is individually tested and its contribution to the overall model performance is evaluated.

    \item \textbf{Sub-task Combination:} If the addition of a sub-task improves the model's ability to reflect the phenomena being studied, then the sub-task is deemed effective. If the addition of a sub-task does not improve the model, we need to verify its feasibility in an individual model before attempting to re-add it to the combined task.

    \item \textbf{Holistic Task Simulation:} Finally, we perform debugging in the context of a holistic task simulation. This allows us to identify any issues that may arise when all components of the model are working together. It is suggested that a save/load function is implemented for debugging, so we do not need to simulate from scratch when a bug is identified. 
\end{enumerate}

For instance, when studying emergency evacuation (to be presented in Section~\ref{sec:case-evac}), we could start by simulating a scenario where there is no persona or communication between agents. We could then add personas for agents, and finally introduce communication between agents. 
This step-by-step approach allows us to gradually increase the complexity of the simulation, ensuring that each component is functioning correctly before moving on to the next. 

\subsection{Validation}
\label{sec:design:valid}
We have elucidated the methods of applying SABM to simulation modeling. However, the creation of a model is only one part of the process. The ultimate goal of simulation is not the model itself, but its ability to represent the reality of the problem under study. If the model fails to do so, even the most optimal solution derived from it may not be applicable to the real-world problem. Therefore, it is crucial to have methods in place to validate the simulation and ensure the reliability of the model.

Traditional computer simulation often employs sensitivity analysis as a means of testing model reliability. This method measures the sensitivity of the model's state changes in response to alterations in system parameters and conditions~\cite{saltelli2008global}. Sensitivity analysis allows for a comprehensive understanding of the model's behavior under different conditions, thereby providing a robust measure of its reliability.

In the context of SABM, we need to pay additional attention to the testing of the prompt. ABM approaches may utilize pre-trained language models (e.g.,~\cite{zhang2023simulating}), but language is seldom part of the input parameters. This is where SABM diverges. In SABM, language, in the form of a prompt, plays the role of fundamental parameters and serves as an essential model component. Consequently, it has a more significant impact on the model's results.

We need to scrutinize how and to what extent the prompt influences the behavior and results of the simulation. This is crucial to prevent the simulation from becoming overly dependent on a specific prompt, which could lead to a loss of reproducibility. It is essential to ensure that the model remains robust and reliable across a range of prompts, thereby enhancing its real-world relevance and applicability. Next, we will delve into the methods and strategies for prompt testing in SABM.

\myparagraph{Prompt Alteration}
In order to verify the stability of the model, we propose three strategies for altering the prompt, each reflecting a different role that the prompt plays in a simulation:

\begin{enumerate}
    \item \textbf{Paraphrasing:} This strategy involves changing the wording or order of the prompt without altering its meaning, to test whether the agent's behavior and the system's behavior are sensitive to the prompt. For instance, if the model's output change significantly after altering \str{you may} to \str{you might}, it would suggest that the model is overly sensitive to the prompt.

    \item \textbf{Varying Elements:} This strategy involves changing some of the simulation's settings and observing the resulting changes in the model. While the original objective of the simulation remains unchanged, the model's performance is affected by altering the meaning of statements, examples of prompts, and by increasing or decreasing the information or constraints provided to the agent. For example, changing \str{you can} to \str{you cannot}, adding few-shot examples, providing or withholding formulas and values for certain variables in the prompt to the agents, etc. It is expected that such changes will alter the model's performance to some degree, but the difference in performance before and after the modification should be explainable and consistent with existing theories or empirical data.

    \item \textbf{Varying Objectives:} This strategy involves changing the objectives of the simulation to test the reliability of the prompt. Changes to the simulation objectives may have the most significant impact on the model results, but such changes need to be interpretable. For instance, we might change the nature of a game from a win-win to a zero-sum game in a game of chance. In such a case, there is a higher likelihood that the agent's behavior will change. If the behavior persists, or if the change in behavior cannot be rationally explained by theory, it may indicate that the model does not adequately understand the task description.
\end{enumerate}

\begin{table}[!t]
  \small
  \centering
  \caption{Observations of variations in simulation results.}
  \begin{tabular}{p{.2\linewidth}|p{.7\linewidth}}
    \hline
    \textbf{Extent of Variation} & \textbf{Definition} \\
    \hline
    Low    & No significant change \\
    Medium & Quantitative change with significance (in terms of output value) \\ 
    High   & Qualitative change with significance (in terms of agent behavior pattern) \\
    \hline
  \end{tabular}
  \label{tab:method:obs-variations}
\end{table}

\begin{table}[!t]
  \small
  \centering
  \caption{Examples of prompt alteration for the number-guessing game at different levels.}
  \begin{tabular}{p{.12\linewidth}|p{.26\linewidth}|p{.26\linewidth}|p{.26\linewidth}}
    \hline
    \textbf{Extent of Variation} & \textbf{Modified Prompt} & \textbf{Original Prompt} & \textbf{Results and Analysis} \\
    \hline
    Low & \textit{Paraphrasing:} \highlightgreen{Pick an integer} from 1 to 100. & \highlightred{The number will be an integer ranging} from 1 to 100. & No significant difference in the results. (interpretable) \\
    & & & \\
    Medium & \textit{Varying Elements (in the prompt for domain knowledge):} You \highlightgreen{cannot} use binary search. & You \highlightred{can} use binary search to optimize your guess. & We observe that when the guesser is instructed not to use binary search, the number of attempts needed to reach the target number increases significantly compared to the case when it is allowed to use it (on average from 5.18 to 6.45 over 100 runs, $p < 0.001$ using the Mann-Whitney U test), confirming significant quantitative changes in the results. (interpretable) \\
    & & & \\
    High & \textit{Varying Objectives:} The number will be an \highlightgreen{even integer} ranging from 1 to 100. & The number will be an \highlightred{integer} ranging from 1 to 100. & The search space of the model produces a change, but the task can still be handled using a binary search algorithm. The specific guesses in the results change (no odd numbers occur), but the trend and algorithm remain the same. (interpretable) \\
    \hline
  \end{tabular}
  \label{tab:method:obs-variations-example}
\end{table}

\begin{table}[!t]
  \small
  \centering
  \caption{Summary of case studies.} 
  \resizebox{\linewidth}{!}{%
    \begin{tabular}{ p{.16\linewidth}|p{.2\linewidth}|p{.14\linewidth}<{\centering}|p{.14\linewidth}<{\centering}|p{.14\linewidth}<{\centering} } 
      \hline
      \multicolumn{2}{l|}{\textbf{Case number}} & 1 & 2 & 3 \\ \hline
      \multicolumn{2}{l|}{\textbf{Scenario}} & Emergency evacuation & Plea bargaining & Firm pricing competition \\ \hline
      \multicolumn{2}{l|}{\textbf{Discipline}} & Behavioral science & Criminology & Economics \\ \hline
      \multicolumn{2}{l|}{\textbf{Number of agents}} & 100 -- 400 & 1 & 2 \\ \hline
      \multirow{9}{*}{\shortstack[l]{\textbf{Model}\\\textbf{construction}\\\textbf{steps}}} & Preliminary design & \checkmark & \checkmark & \\ 
      & Base modeling & \checkmark & \checkmark & \checkmark \\ 
      & Data calibration & & \checkmark & \\ 
      & Sub-task addition & \checkmark & \checkmark & \checkmark \\ 
      & Single-agent debugging & \checkmark & & \\ 
      & Sub-task analysis & \checkmark & \checkmark & \checkmark \\ 
      & Sub-task combination & \checkmark & \checkmark & \checkmark \\ 
      & Holistic task simulation & \checkmark & & \checkmark \\
      & Validation & \checkmark & & \\ \hline
      \multirow{9}{*}{\shortstack[l]{\textbf{Implementation}\\\textbf{methods}}} & Common sense & \checkmark & \checkmark & \\ 
      & Domain knowledge & & & \checkmark \\
      & Memory & \checkmark & & \checkmark \\
      & Learning & \checkmark & \checkmark & \\
      & Reasoning & & \checkmark & \\
      & Planning & \checkmark & & \checkmark \\
      & Personalization & \checkmark & \checkmark & \checkmark \\
      & Conversation & \checkmark & & \checkmark \\
      & Model interpretation & \checkmark & & \\ \hline
      \multirow{17}{*}{\shortstack[l]{\textbf{Instance}\\\textbf{components}}} & Agent: heterogeneity & Heterogeneous agents & Heterogeneous agents & Heterogeneous agents \\ 
      & & & & \\
      & Agent: alignment & Common sense & Common sense & Domain knowledge \\ 
      & & & & \\
      & Environment & Agent behavior-based variation & None & Agent behavior-based variation \\ 
      & & & & \\
      & Interactions & Local communication & None & One-on-one \\ 
      & & & & \\
      & System setup: temporal structure & Discrete time & None & Discrete time \\ 
      & & & & \\
      & System setup: spatial structure & Regular grids & None & None \\ 
      \hline
    \end{tabular}
  }
  \label{tab:case-study-summary}
\end{table}

By systematically varying the prompt and observing the resulting changes in the model, we can gain a deeper understanding of the model's behavior and its sensitivity to different prompts. We can also categorize the observed changes in the results into three types, indicating the magnitude of the changes, and thus determine whether the model is stable across multiple scenarios. The classification can be seen in Table~\ref{tab:method:obs-variations}.

Indeed, the stability of a model is not solely determined by the magnitude of the observed changes when varying the prompt, but more importantly, by whether the results align with our expectations. When we paraphrase the original text, we do not anticipate significant changes in the results (i.e., there should be a low level of variation in the observations of model outcomes). If the results are consistent with our expectations or show limited, interpretable variations, it suggests that the model is stable. On the other hand, if the model's results change in ways we do not anticipate or appear random, it signals instability in the model. This indicates sensitivity to specific vocabulary or phrasing in the prompt, necessitating further adjustments and testing to ensure reproducibility. In the case of varying elements or objectives in tasks, we expect changes in the model's results. However, these changes should be explainable based on the modifications made to the prompt and should align with existing theories or empirical data. If the changes are inexplicable or inconsistent with our expectations, it may indicate that the model is overly sensitive to specific aspects of the prompt.

Absolutely, it is crucial to underscore that the observation of changes in results should typically be based on statistical outcomes. Given that LLMs inherently produce outputs with a degree of randomness, the agent behaviors they simulate also embody a certain level of diversity and uncertainty. Therefore, the statistical results derived from multiple simulations can serve as a basis for judgment, helping to mitigate any bias that might be introduced by such randomness. This approach aligns with the principles of Monte Carlo simulations~\cite{joy1991introduction}, which rely on repeated random sampling to obtain numerical results. By running the simulation multiple times and averaging the results, we can obtain a more accurate and reliable estimate of the model's behavior under different prompts. This method can effectively smooth out the randomness inherent in individual simulation runs, providing a more robust measure of the model's sensitivity to changes in the prompt.

In the number-guessing game, we can validate the reliability of the model through a series of modifications on the prompt. We are able to create variations of the original prompt at low, medium, and high levels, and observe the resulting changes in the model's behavior. Table~\ref{tab:method:obs-variations-example} shows the settings and results of the modifications.

In summary, these strategies of altering prompts enable us to understand how specific segments of the prompt influence the behavior and results of our model. This is crucial for explaining and validating prompt settings, thereby enhancing the reliability and real-world relevance of our model. By systematically testing the model's sensitivity to different prompts, we can ensure its robustness and applicability across a range of scenarios, contributing to the overall validity of the SABM approach. Moreover, the validation of these models cannot be decoupled from qualitative comparisons with existing theories and real-world phenomena. A strong qualitative resemblance between a simulation and an actual system serves as indirect evidence of the simulation's fidelity~\cite{casti1996would}. In order to claim that our simulations accurately reflect real-world systems, we need supporting evidence from real-world data or theory. This evidence should demonstrate that our simulations capture the essential characteristics of these systems. Existing qualitative and quantitative studies need to be taken into account to keep our models and studies in reality. In certain scenarios, expert validation methods are also used for model verification, which should also be taken into consideration~\cite{schellekens2010flexible}.

\myparagraph{Sensitivity Analysis on Parameters}
Sensitivity analysis plays a crucial role in understanding the impact and validation of various parameters on the model's output. This has been extensively discussed in the context of computer simulations~\cite{donze2007systematic, kleijnen2005overview, saltelli2008global}. In the context of SABM, the same principles apply. While some SABM simulations may solely rely on text prompts without involving any quantitative variables, others may incorporate parameters in the model. In such cases, traditional sensitivity analysis methods can be applied to these parameters in addition to the prompt alteration strategies.

And it is important to note that other than the quantitative variables, a key aspect to consider when using the GPT model as the core of the agent is the effect of the \texttt{temperature} parameter on the simulation results. The \texttt{temperature} parameter in GPT models controls the randomness of the model's responses, with higher values leading to more diverse outputs. Therefore, it is essential to analyze the performance of the simulation results for different temperature parameters. This will help determine whether the stochasticity and diversity introduced by the language model significantly affect the simulation.

As per the methodology proposed in this section, we apply SABM to three case studies. We emphasize that our goal here is to demonstrate the methodology of SABM rather than extensive experiments and thorough analysis. Nonetheless, we consider that extending these case studies might lead to more interesting results and deeper understanding in their respective fields. Table~\ref{tab:case-study-summary} summarizes these case studies, showing the optional steps of model construction, the implementation methods used, and the categorization of their components (as per Figure~\ref{fig:sabm:hierarchy}). In particular, the first case study is a complete SABM instance, and the other two delve into agent modeling and interaction modeling, respectively.

\section{Case Study 1: Emergency Evacuation}
\label{sec:case-evac}
This comprehensive case study simulates the complex patterns of evacuation in emergency situations. Given the significance of evacuation research for public safety and risk management, as well as the challenges in obtaining data from real-world scenarios~\cite{daamen2003experimental}, there have been many simulation studies that explored the evacuation contexts~\cite{li2017pedestrian, wang2015modeling, zheng2019simulation, zheng2018evacuation}. In this case study, we model 100 -- 400 heterogeneous agents, each representing an evacuee, and investigate how SABM can simulate complex behaviors through methods, including common sense reasoning, memorization, learning, planning, and conversation. To design the model, we take~\cite{wang2015modeling} as the base, a study on the simulation of emergency evacuation using ABM, and incrementally replace its components with SABM.

\begin{figure}[!t]
    \centering
    \includegraphics[width = \textwidth]{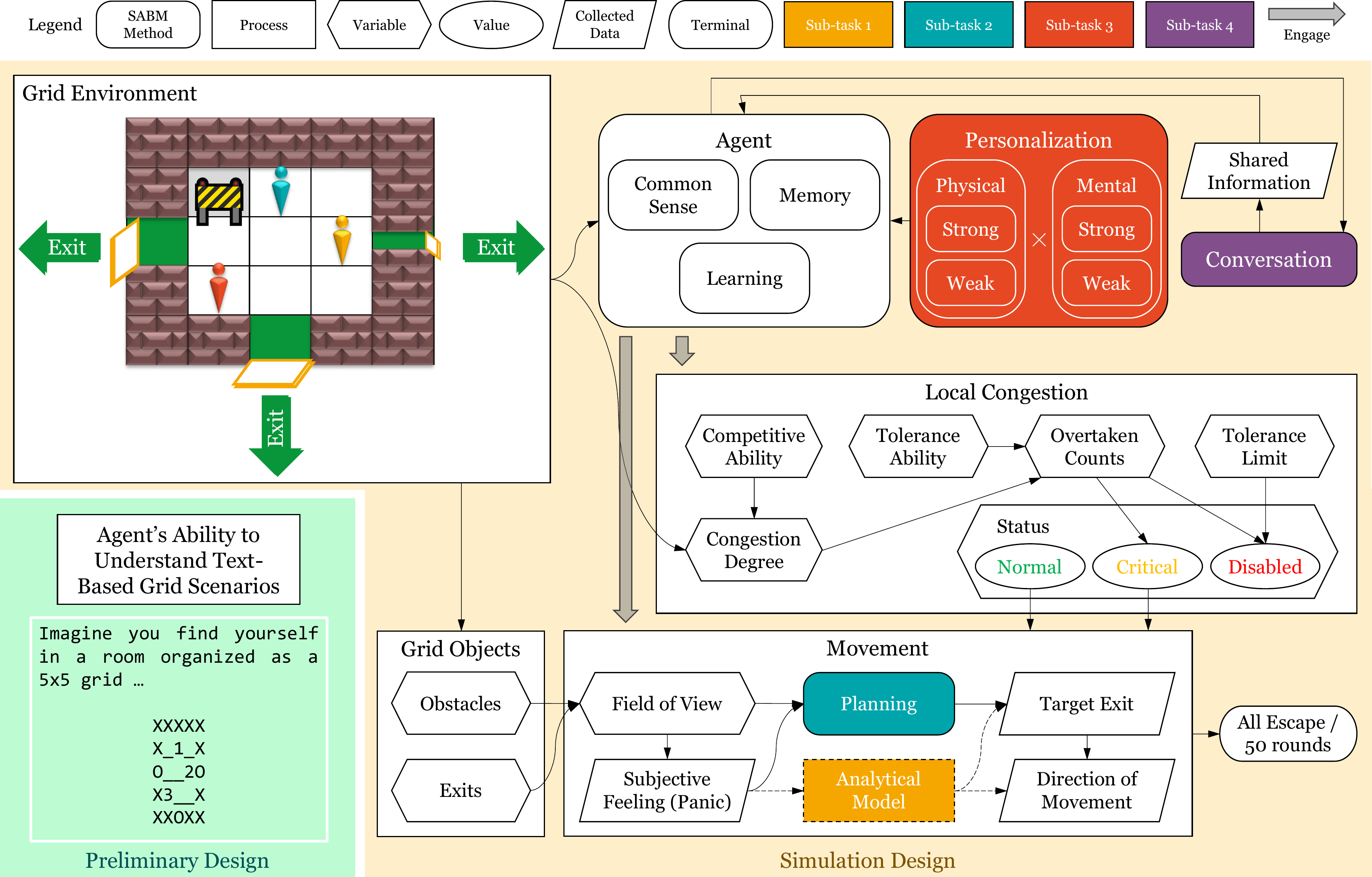}
    \caption{Design of emergency evacuation. Dashed lines represent ABM methods only used in Sub-task 1.}
    \label{fig:evac:design}
\end{figure}

\subsection{Task Definition}

We simulate scenarios with varying population densities to examine the dynamics of emergency evacuation. We primarily observe how variations in individual characteristics, such as physical strength, mental state, and location-related factors, affect the agents' decision-making and movement along with their response to the actions of their peers. We are particularly interested in whether the presence of congestion causes the agents to abandon their originally planned escape paths in search of alternatives that are less congested, despite these alternatives potentially not being the closest exits.

Figure~\ref{fig:evac:design} displays the conceptual design of the case study, which will be detailed in the rest of this section. We set the simulation environment as a grid environment composed of $33 \times 33$ cells, where the evacuation is simulated by the movement of agents. We use $(i, j)$ to denote the coordinates of a cell in the grid, where $i$ and $j$ represent row and column, numbered from top to bottom and left to right, respectively, and both starting from 0. We have a preliminary design phase where the agents are first tested on their ability to understand text-based grid scenarios, which serves as the basis for selecting the LLM to use in the simulation. Subsequently, in the simulation design, the agents choose the target exit and determine the direction of movement based on their common sense. In each round, the agents can move from one cell to another or remain stationary. 

At the start of each round, the following situational description is provided to each agent:
\begin{prompt}{Evacuation scenario}
  Because of the earthquake, you need to escape from the room where you are as fast as possible. The room has a size of 33 * 33. There are 3 exits in the room. The exits are located at the left, bottom, and right of the room.\\
  To escape from the room, you need to consider the following two aspects: exit proximity and people count. The exit proximity is the distance between you and the nearest exit. The people count is the number of people you can see. The distance to the nearest exit is \var{distance\_to\_nearest\_exit}. There are \var{number\_of\_people} people in your visible range.
\end{prompt}

In each round, an agent's state is first calculated based on the level of local congestion at its current position and its abilities to cope with congestion, including competitive ability and tolerance ability. When an agent has been overtaken a certain number of times, its state is set to \str{disabled} and it leaves the simulation. Following this, the agent's movement unfolds in four stages. In Stage 1, the agents are asked about their feelings regarding the situation to judge whether they are panicking in the current scenario. In Stage 2, based on the information available within their field of view, including the distance to exits and the number of people within the view, the agents assess the exits. In Stage 3, they decide their respective target exit. In Stage 4, they determine the direction of movement according to the planed path towards the target exit in Stage 3. Moreover, we consider conversation as an option in Stage 3, allowing them to share information with other agents and to use the information as one of the factors for deciding the target exit. Specific settings regarding field of view, local congestion, and direction of movement are detailed in the model design.

\subsection{Preliminary Design}

The objective of the preliminary design is to test whether the LLM can understand text-based grid scenarios, including whether it can grasp that the situational task of the evacuation is to reach an exit within the grid as fast as possible, whether it can accurately understand coordinates information, and whether the route planning is consistent and rational. To test the LLM's capability, we employ two setups to determine the model to be used in the simulation. There are two \texttt{model\_types} to choose from: \texttt{gpt-3.5-turbo-0301} (referred to as GPT-3.5 for the rest of this section) and \texttt{gpt-4-0314} (referred to as GPT-4 for the rest of this section). \texttt{max\_tokens} is set to 512 and \texttt{temperature} is set to 0.0~\footnote{Although setting \texttt{temperature} to zero may result in limited diversity of behaviors under exactly the same setting, in this procedurally generated, interactively dynamic environment, we seldom encounter exactly the same outcome. Meanwhile, in a physically situated setting (e.g., a grid), the LLM used in this case study still has restricted capabilities on scene understanding, and increasing the temperature may introduce diversity as well as unwanted randomness at the same time~\cite{ma2023towards}.}. The following text about the grid scenario is provided for testing:

\begin{prompt}{Text-based grid scenario}
  Imagine you are in a room laid out as a 10 * 10 grid. You are currently at coordinates \var{initial\_position}. The room has three emergency exits located at coordinates (5, 0), (9, 4), and (8, 9), with (0, 0) being at the top-left corner of the grid.
  
  Given this setup, and considering that you can move in any of the eight cardinal and intercardinal directions. It is possible to move diagonally, e.g. from (1, 1) to (2, 2) is one move to the upper right, and is faster than (1, 1)-$>$(1, 2)-$>$(2, 2). You need to determine the safest and fastest route to evacuate the room. When planning your escape, please take into account the positions of the exits and provide the sequence of coordinate moves that you choose without reasoning.
\end{prompt}

\begin{figure}[!t]
    \centering
    \includegraphics[width = .8\textwidth]{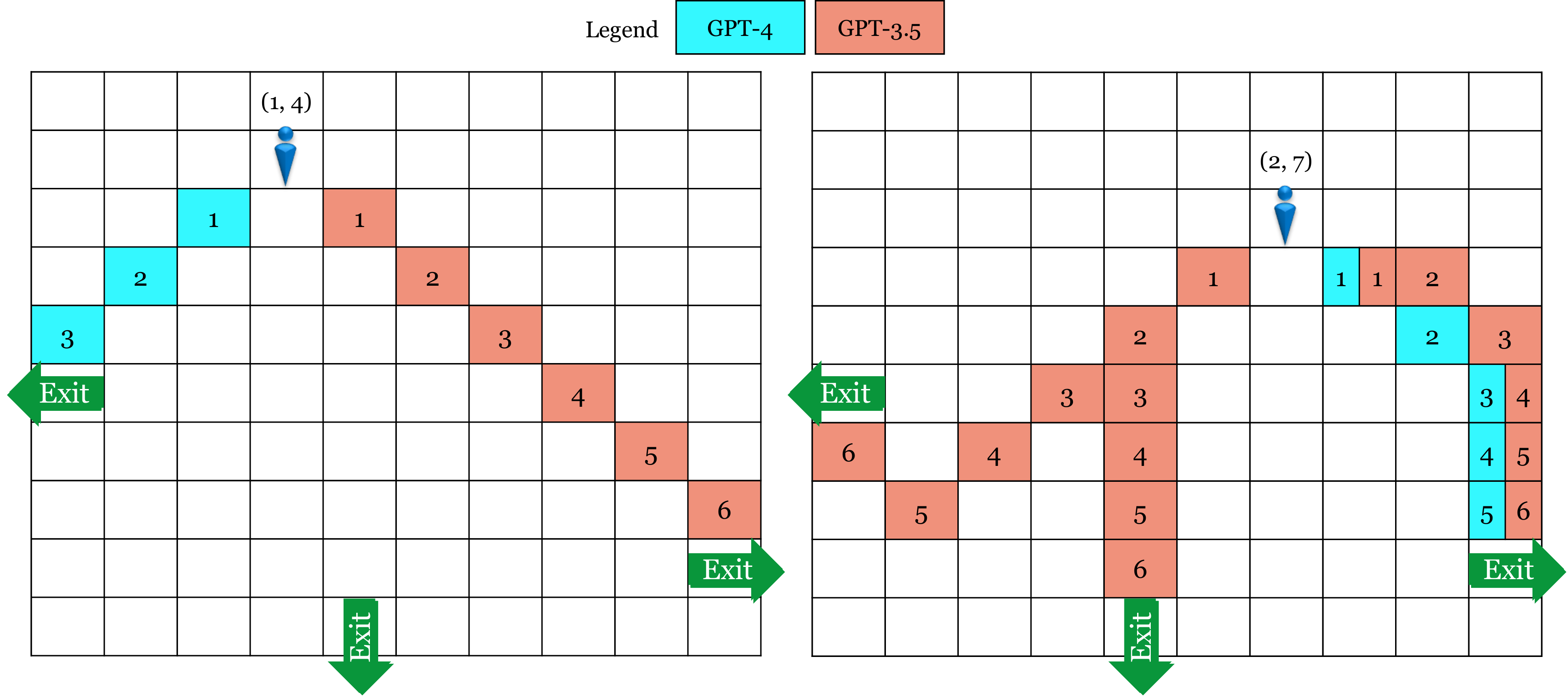}
    \caption{Output samples of preliminary design.}
    \label{fig:evac:preliminary}
\end{figure}

By testing GPT-3.5 and GPT-4 with two initial positions, (1, 4) and (2, 7), we show their choice of exit and route in Figure~\ref{fig:evac:preliminary}. It can be observed that GPT-4 is capable of accurately understanding the map and planning the shortest route to the nearest exit, whereas GPT-3.5 fails to find the nearest exit or the shortest escape route, and exhibits a high level of uncertainty in its decision. This indicates that GPT-3.5 cannot accurately comprehend text-based grid scenarios and is incapable of rationally planning escape routes. Therefore, we will use GPT-4 in the model design.

\subsection{Model Design}

\begin{figure*}[!t]
    \centering
    \begin{subfigure}{0.45\textwidth}
        \includegraphics[width=\linewidth]{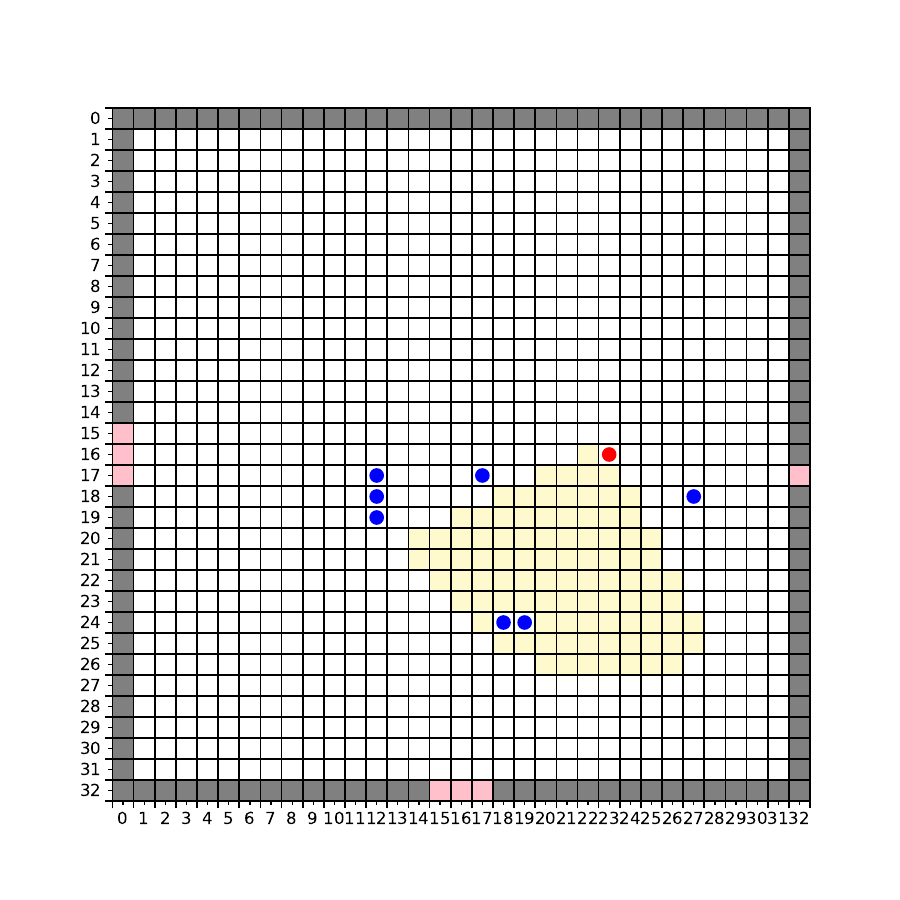}
        \caption{Red and blue points are agents. Yellowish cells indicate the field of view of the red agent.}
        \label{exp:evac:scenario:grid}
    \end{subfigure}
    \hspace{4ex}
    \begin{subfigure}{0.45\textwidth}
        \includegraphics[width=\linewidth]{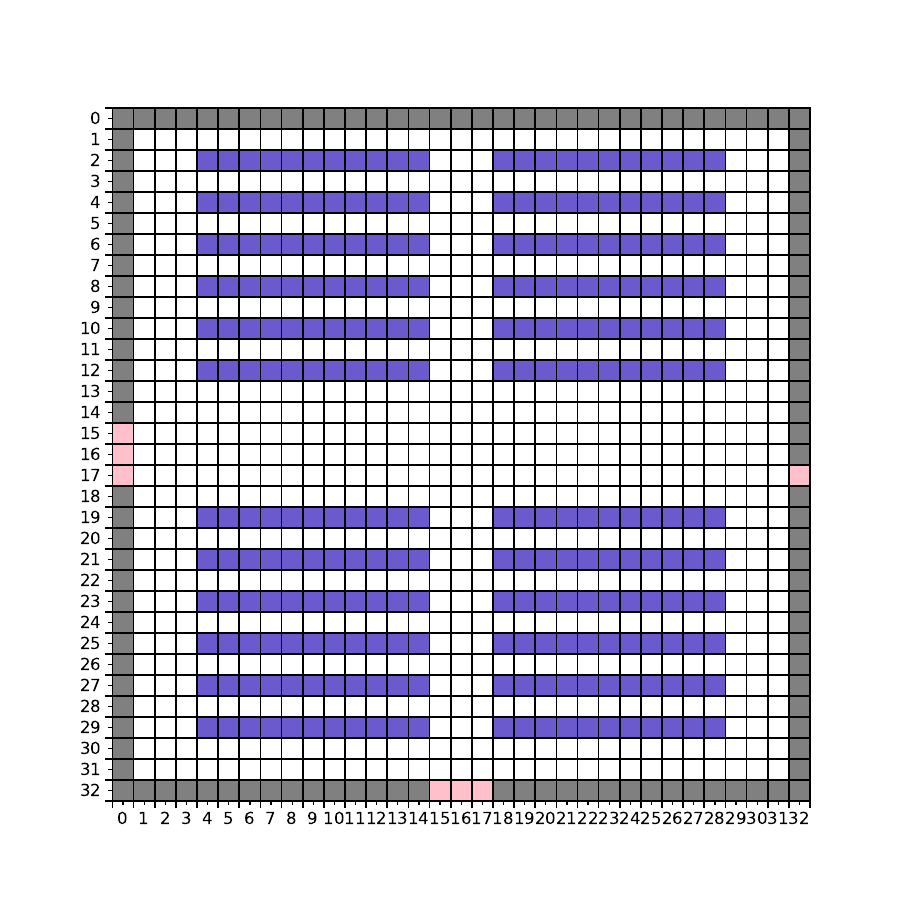}
        \caption{Grid with obstacles depicted in slate blue blocks.}
        \label{exp:evac:scenario:obstacle}
    \end{subfigure}
    \caption{Grid environment. White, gray, and pink cells indicate empty space, walls, and exits, respectively.}
    \label{fig:evac:scenario}
\end{figure*}

\subsubsection{Grid Environment}

The environment is modeled as a grid comprising $33 \times 33$ cells, with all sides bounded by walls and three exits located centrally along the left, bottom, and right sides. As depicted in Figure~\ref{exp:evac:scenario:grid}, the right-side exit consists of a single cell, thereby being narrower than the other two exits, each extending across three cells. This design intends to introduce a bottleneck to closely examine if the agents can modify their evacuation strategies in response to varying congestion levels. As depicted in Figure~\ref{exp:evac:scenario:obstacle}, we optionally add obstacles to the scenario, allowing for a comparative analysis of environments with and without obstacles. Each cell is designated for single occupancy and may either be empty, occupied by an obstacle, or occupied by an agent. In case two agents aim to enter the same cell simultaneously, a random selection process determines which agent will actually enter the cell.

\subsubsection{Basic Agent Setup}

We outline the components that constitute an agent's behavior, which include the agent's direction of movement, field of view, and local congestion.

\myparagraph{Movement}
At each round, an agent may move to one of the eight surrounding cells in the directions: up, down, left, right, upper-left, lower-left, upper-right, lower-right, or it may remain stationary in its current cell. To ensure that the agents comprehend the possibility of diagonal movement within the simulation, we provide an additional (few-shot) instruction to facilitate learning of these movement patterns.

\begin{prompt}{Instruction for movement}
  You need to escape to the exit as fast as possible. The room has a size of 33 * 33. We use (i, j) to denote the position, smaller i means top and bigger i means bottom; smaller j means left and bigger j means right. Position (1, 1) is at the top left of the room. \textbf{It is possible to move diagonally, e.g., from (1, 1) to (2, 2) is one move to the lower right, and is faster than (1, 1)-$>$(1, 2)-$>$(2, 2).}
\end{prompt}

\myparagraph{Field of View}
An agent's field of view is limited to a specific region ahead of it. As illustrated in Figure~\ref{exp:evac:scenario:grid}, the diagram showcases the field of view for the agent in red at position (16, 23) directed towards the exit below. Within this limited area, an agent gathers local information such as the count of people and obstacles (e.g., from the view field at (16, 23) looking towards the bottom exit, there are two other agents), informing subsequent actions.

\myparagraph{Local Congestion}
In recognition of potential injuries in crowded real-world situations, the model accounts for overtaking events in congested spaces during evacuation. The agents are attributed with characteristics that describe their ability to handle crowded conditions. These characteristics include competitive ability, tolerance ability, tolerance limit, congestion degree, and overtaken counts. Essentially, an agent's competitive ability is assessed against those of other agents in the adjacent eight cells. By accumulating the differences in competitive ability, we have the congestion degree at the agent's present location. If the congestion degree surpasses the agent's tolerance ability, the agent is regarded as overtaken and sustains an injury, transitions from a \str{normal} to a \str{critical} state, and halts movement for one round. Accumulating injuries and overtaken count beyond the tolerance limit leads an agent to a \str{disabled} state, at which point they are removed from the simulation, indicative of an unsuccessful evacuation. For competitive ability, tolerance ability, and tolerance limit, we randomly sample a number from a normal distribution where $\sigma = 1$ and $\mu$ is determined by the agent's physical strength specified in its persona. For physically strong agents, $\mu$ is 3, 18, and 26 for the three attributes, respectively. For physically weak agents, $\mu$ is 2, 16, and 23 for the three attributes, respectively. The choice of these mean values roughly follows the parameter settings in \cite{wang2015modeling}, and we use normal distributions instead of fixed values for more variation. 

In addition to the above setup, we assume: 
\begin{inparaenum} [(1)]
    \item The initial positions of the agents are randomly distributed with uniform probability.
    \item The agents are presumed to know the coordinates of all exits.
\end{inparaenum}

\subsubsection{Personalization}
Our model incorporates heterogeneity among agents using personalization, which affects their ability to adapt changes in their surroundings and ultimately influences their capacity for evacuation route planning. For instance, the agents may become more prone to panic due to changes in the environment under the influence of their mentally weak persona. The personalization takes into account attributes of physical and mental aspects, hence including four personas: physically strong and mentally strong, physically strong but mentally weak, physically weak but mentally strong, and physically weak and mentally weak. These personas, used as prompts, affect the agents' behavioral characteristics, as given below. 

\begin{prompt}{Physically strong and mentally strong persona}
  You are positive and full of energy, and you are strong and fit.
\end{prompt}

\begin{prompt}{Physically strong but mentally weak persona}
  You are positive and full of energy, but you are not a strong person.
\end{prompt}

\begin{prompt}{Physically weak but mentally strong persona}
  You are negative and afraid of difficulties, but you are strong and fit.
\end{prompt}

\begin{prompt}{Physically weak and mentally weak persona}
  You are negative and afraid of difficulties, and you are not a strong person.
\end{prompt}

\subsubsection{Planning}
As previously mentioned, there are four stages to decide the movement of an agent in each round. We employ planning for modeling the movement. 

\myparagraph{Subjective Feelings}
In Stage 1, we model the subjective influencing factors that affect the agents' panic levels, which serve as a psychological basis for their subsequent decision. Due to the difficulty in describing subjective influencing factors, in the ABM setup in~\cite{wang2015modeling}, parameters are used to signify whether the agents are uniformly experiencing panic. In the SABM setup, we gauge the agents' feelings towards the current situation to assess their panic levels with the following prompt. 

\begin{prompt}{Stage 1: Subjective feelings}
  \var{persona}. Please tell me your feelings about the situation around you in one sentence showing if you are panicking or not.
\end{prompt}

\myparagraph{Assessment and Choice of Target Exits}
In the ABM setup, the agents determine their target exit by using the number of people within their field of view and the distance to each exit as inputs to an analytical model. In the SABM setup, the agents apply common sense and planning to evaluate how the given information about the exits influences their escape. Drawing on the subjective feelings identified in Stage 1, the assessment of these factors also reflects individual physical and mental conditions, rather than relying solely on quantitative measures of distance and congestion levels. For instance, an agent with superior physical strength may opt for an exit that is further away but less crowded. As a result, in Stage 2, the agents combine their subjective feelings with other information, including exit proximity and congestion, to assess the exits, as shown in the following prompt. 

\begin{prompt}{Stage 2: Assessment of target exits}
  \var{persona}. Now you feel: ``\var{panic\_level}''.\\\\
  Here shows you the distances to different exits and the number of people you can see towards those exits:\\
  Exit \var{exit\_id}: \var{distance\_to\_exit} away, \var{number\_of\_agents\_around} people around.\\\\
  Please tell me briefly how will you evaluate the two aspects of each exit based on your personal mental and physical characteristics in one sentence. Please give 3 sentences for each exit (around 15 words).
\end{prompt}

In Stage 3, using the outcomes of the assessment in Stage 2, the agents decide the target exit. 

\begin{prompt}{Stage 3: Choice of target exit}
  \var{persona}. Now you feel: ``\var{panic\_level}''.\\\\
  There are 3 exits in this room. Based on the current situation, your personal feelings on each exit are: \var{assessment}.\\\\
  Please tell me which exit you would like to choose to escape, and you always want to escape as fast as possible. Please use the exit id to indicate your choice. 
  For example, if you want to choose exit left, you can say 'left'. Only output one word of text to indicate your choice.
  You can choose from ['bottom', 'left', 'right']. Give your answer without any additional text.
\end{prompt}

\myparagraph{Decision on Movement}
Based on the planning in Stages 2 and 3, the agents decide the action, i.e., the direction of movement. 

\begin{prompt}{Stage 4: Decision on movement}
  Select your move from these possible options (you can move in diagonal or horizontal directions, options with obstacles or other people are excluded and not in the path, and option codes are in random order):\\
  \var{movement\_list}\\
  Please tell me your best choice to escape as fast as possible with one single code without any additional texts. You can choose from \var{valid\_directions}.
\end{prompt}

\subsubsection{Memory}

Each GPT call is an individual request without previous dialog histories, while agents should have the sense of its previous decisions to comprehensively determine their current choices. To mimic the human-like memorization functionality, we provide agents with summarized historical information in the prompts to ensure consistency in their evacuation choices, meanwhile reduce the costs on redundant information. In Stage 3, when selecting their target exits, they will receive the historical information about their previously chosen exits. In Stage 4, when determining their movement directions, they can also obtain their location in the previous round.

\begin{prompt}{Historical information}
    {\upshape Target exit history in Stage 3:} ... your personal feelings on each exit are: \var{assessment}. \\\\
    \textbf{Here are the previous decisions you made for the target exit from the beginning: \var{target\_exit\_history}.
    This means most recently you were heading to exit {target\_exit}. Please keep these in mind when you make your decision.}\\\\
    Please tell me which exit you would like to choose to escape ...
    \tcbline
    {\upshape Position history in Stage 4:} \textbf{You were at \var{last\_position} last time. To escape from the room, you have chosen the exit at \var{exit\_position} and you are at \var{cur\_position}, so the exit is on your \var{direction(agent\_position, exit\_position)}.} \\\\
    Select your move from these possible options ...
\end{prompt}

\subsubsection{Conversation}

As an option in this case study, the agents can exchange information, simulating the communication that would occur between individuals in real evacuation, including dialogue and potential collaboration. In each round, the probability that an agent shares information is set to 20\%, and the order of speaking is randomly determined. Conversation, placed in Stage 3 before the agents choose target exits, is instructed with the following prompt.  

\begin{prompt}{Share information}
  You may briefly share information about evacuation with others, such as your feelings, which exit seems to be the best option for a quick escape, or anything else you would like to deliver. Avoid using numbers in the communication. Use less than 50 words, not too long.    
\end{prompt}

In this prompt, we allow the freedom to share any kind of information, rather than confining the information to a set range of topics~\footnote{The only exception is that the message should avoid numbers to prevent the agents from discussing coordinates or exact distance to the exit, which spoils the realism of the simulation.}. This approach to interaction between agents is chosen over defining topics such as leadership roles, as seen in its ABM counterpart, to elicit specific complex behavior patterns. We posit that information sharing is a more innate behavior pattern. In real-world evacuation scenarios, individuals rely on their actual feeling and situation understanding rather than predefined behavioral templates to determine their interactions with others. Consequently, our focus is to observe the nature of information provided by the agents and the spontaneous behaviors that arise in a simulated situation where the topics of information exchange are unconstrained. Such design is also in line with the rationale of the a priori modeling of SABM -- observe what these smart agents would do based on the assumption that they can mimic human behavior.

The shared information is reintroduced to the nearby agents in Stage 3, serving as part of the basis for their target exit decision-making. In particular, conversational exchange is limited to a circular area with a 5-cell radius centered on the agent. Each agent receives shared information via the following prompts right before he or she is chosen to speak in Stage 3. 

\begin{prompt}{Receive information}
  You hear \var{number\_of\_people} people around you say:\\
  agent\#\var{agent\_id}: \var{conversation\_content}
\end{prompt}

\subsubsection{Exit Conditions}
The simulation terminates when one of two conditions is met: all the agents have successfully exited the grid, or the simulation has reached 50 rounds. If an agent has not reached an exit by Round 50, they are deemed unsuccessful in their escape.

\subsubsection{Sub-tasks}

Combining the elements of our model design, our simulation encompasses the following four sub-tasks. Illustrated in Figure~\ref{fig:evac:design}, Sub-task 1 involves reproducing the ABM setup  in~\cite{wang2015modeling}, except its panic level parameters, which are now determined by the LLM based on its subjective feelings of the situation within the evacuation scenario. Since the ABM setup involves a parameter-based personalization, for fair comparison, we use the same personalization in this sub-task and categorize agents by gender and age: young female, young male, old female, and old male. These personas have differences in physical strength, as defined by parameters. The aim of this sub-task is to confirm that SABM can reproduce the outcomes of the ABM approach and can reasonably substitute its parameter settings, thereby enhancing the interpretability of the parameters.

Sub-task 2 introduces a base model of an evacuation simulation in which the agents' movement relies solely on SABM. As discussed earlier, this sub-task integrates the agent's subjective feelings with its sense of exit proximity and congestion to assess and choose the target exit, which, in turn, informs the direction of movement.

Sub-task 3 extends Sub-task 2 by incorporating a personalization aspect to investigate if individuals with diverse physical and mental states exhibit varied behavioral patterns during an evacuation, which in turn influences the overall behavioral pattern of the crowd.

Sub-task 4 expands on the previous sub-tasks by permitting agents to exchange evacuation information through conversation. This exchange has the potential to influence the agents' decisions regarding the choice of target exit. 

\subsection{Single-Agent Debugging}

\begin{figure*}[!t]
    \centering
    \begin{subfigure}{0.45\textwidth}
        \includegraphics[width=\linewidth]{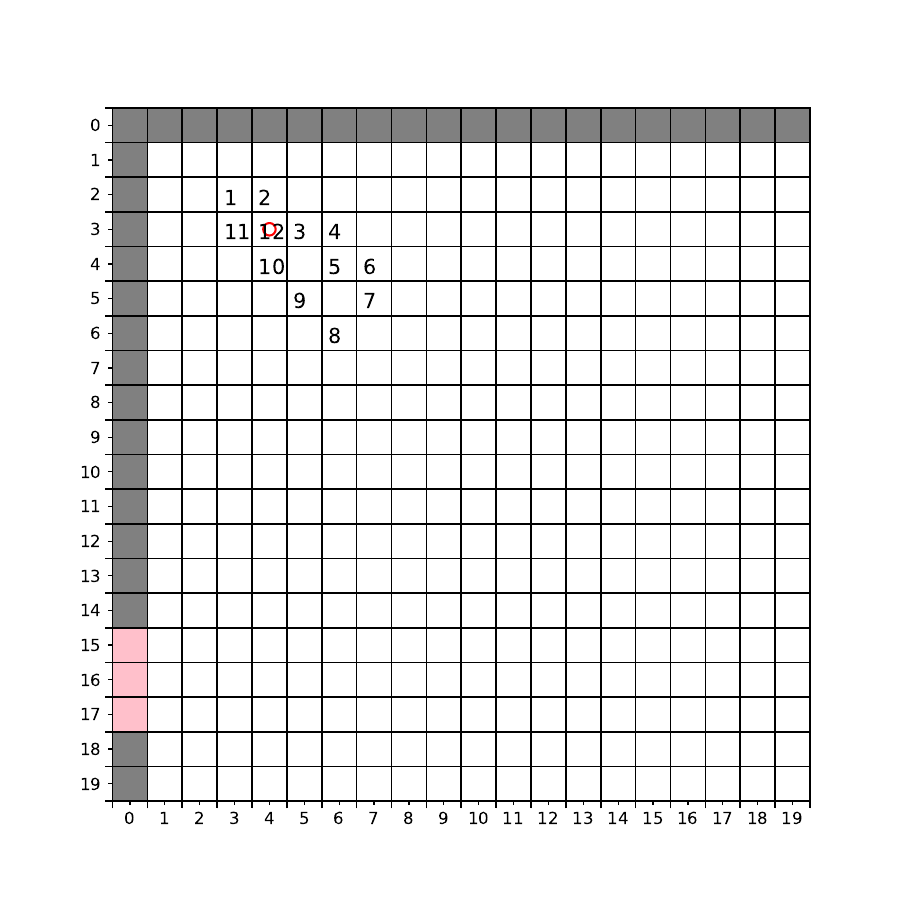}
        \caption{Movements with the agent unaware of its historical locations.}
        \label{exp:evac:debugging:1}
    \end{subfigure}
    \hspace{4ex}
    \begin{subfigure}{0.45\textwidth}
        \includegraphics[width=\linewidth]{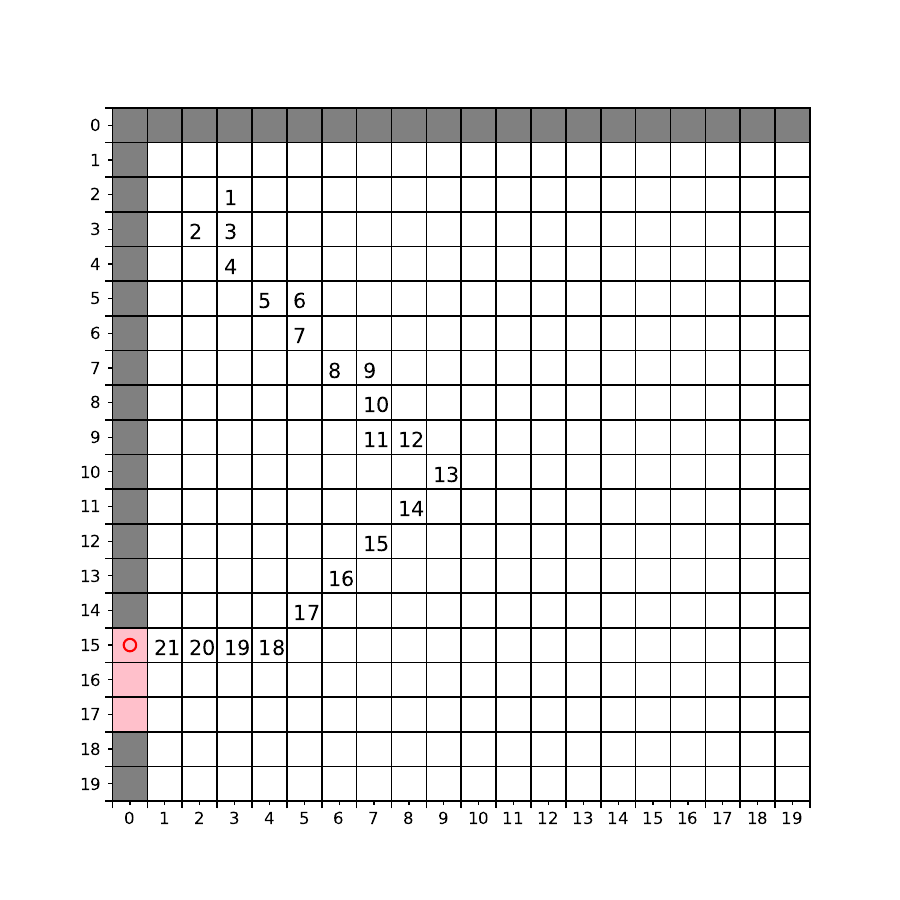}
        \caption{Movements with the agent aware of its historical locations.}
        \label{exp:evac:debugging:2}
    \end{subfigure}
    \begin{subfigure}{0.45\textwidth}
        \includegraphics[width=\linewidth]{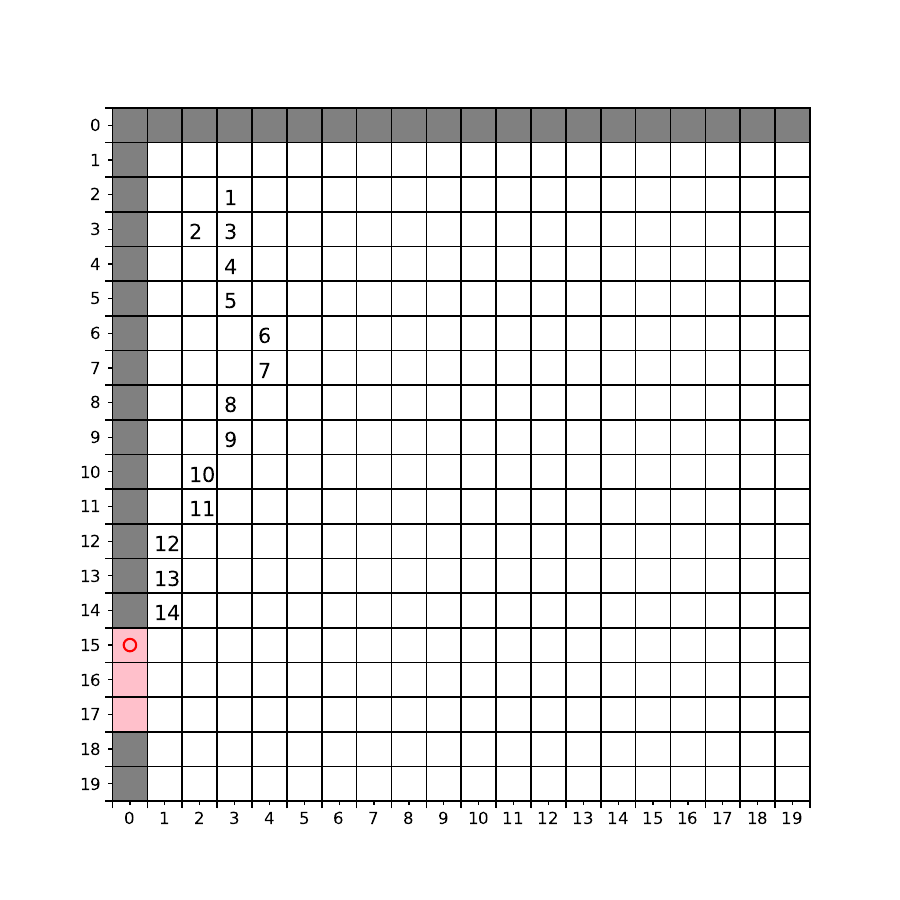}
        \caption{Movements with explanations on the grid coordinate system.}
        \label{exp:evac:debugging:3}
    \end{subfigure}
    \hspace{4ex}
    \begin{subfigure}{0.45\textwidth}
        \includegraphics[width=\linewidth]{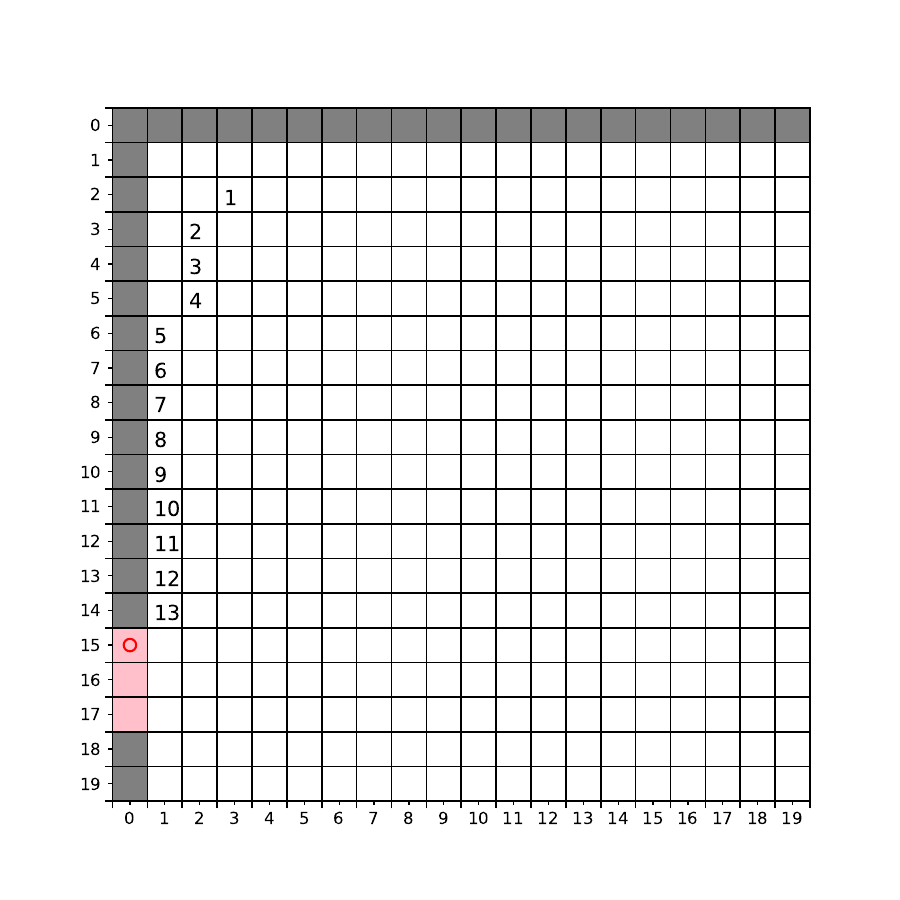}
        \caption{Movements with more explanations on the grid coordinate system, detailing the origin.}
        \label{exp:evac:debugging:4}
    \end{subfigure}
    \caption{Single-agent debugging example. The numbers in cells indicate the movement path. The top-left corner of the grid, which pertains to this debugging, is displayed.}
    \label{fig:evac:debugging}
\end{figure*}

In Section~\ref{sec:design:impl-sim}, we discussed the practice of single-agent debugging. LLMs display inherent variability, which makes the prediction of an agent's behavior under complex prompt instructions challenging. Before running a complete simulation with multiple agents, we need to debug individual agents to ensure that their behavior and decision-making are consistent with real-world scenarios. We aim to show how we identify and rectify discrepancies between a single agent's behavior and established real-world behavior patterns -- our ground truth -- through prompt debugging.

Figure~\ref{fig:evac:debugging} illustrates the process of debugging using the evacuation route of an agent starting at (2, 3). Initially, as demonstrated in Figure~\ref{exp:evac:debugging:1}, the agent correctly identified the closest exit at (15, 0) but failed to navigate to it, oscillating within a confined area instead. This issue indicated a mismatch between the agent's target exit selection and its movement decisions. The agent's ability to identify the target exit indicated a reasonable integration of subjective feelings and exit proximity, but its path determination was flawed. Suspecting that this was due to the agent not having access to its historical locations, we updated the prompt to include this information. Consequently, as depicted in Figure~\ref{exp:evac:debugging:2}, the agent ceases its circling behavior.

Yet, the updated pattern is still far from perfect. In the absence of other agents, the shortest path should be chosen, but the agent's path, as shown in Figure~\ref{exp:evac:debugging:2}, was lengthier than necessary. Having ruled out lack of historical information as the cause, we consider that the agent might not fully understand the grid as described in the prompt. To address this, we clarify the spatial coordinate definitions, stating, \str{We use (i, j) to denote the position, smaller i means upper, and smaller j means left.} This adjustment, depicted in Figure~\ref{exp:evac:debugging:3}, results in noticeable path selection improvements. Further, we detail the coordinate system origin in the movement instructions: \str{We use (i, j) to denote the position, where a smaller i value indicates the top and a larger i value indicates the bottom; a smaller j value means left and a larger j value means right. Position (1, 1) is at the top left of the room.} This modification, as shown in Figure~\ref{exp:evac:debugging:4}, aligns the agent's movement closely with expected behavior.

The ability to interpret a model's reasoning is instrumental for a deeper understanding of the behaviors it exhibits, which is particularly valuable for debugging purposes. We employ the following prompt to interrogate the rationale behind an agent's decision to move in a certain direction:

\begin{prompt}{Model interpretation in debugging}
  You have chosen the exit at \var{choose\_target\_exit} and you were at \var{current\_position}, so the exit is on your \var{exit\_direction}. Can you briefly explain the reason why you choose to move to \var{choose\_direction\_of\_movement}?
\end{prompt}

For instance, an agent may justify its choice not to adopt a straight-line movement by saying, \str{... while avoiding any potential obstacles or dangers that might be present in the direct path.} This response helps clarify the agent's reasoning and allows us to observe how various evacuation scenarios, such as potential dangers, influence the agent's decision-making patterns. Consequently, we confine the emergency scenarios to earthquakes as opposed to fires or other circumstances that may elicit evasive maneuvers during evacuation. With this constraint, the agent demonstrates an inclination towards more direct, straight-line movements.

\begin{figure}[!t]
    \centering
    \includegraphics[width = 0.6\textwidth]{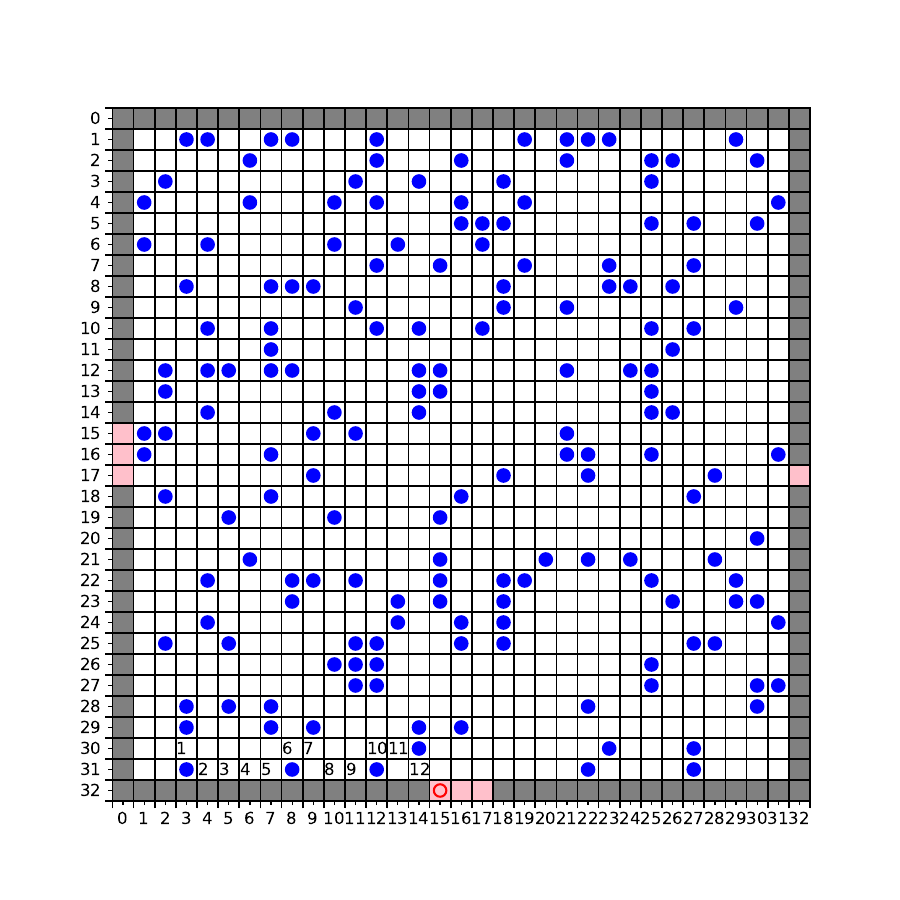}
    \caption{Single-agent debugging example, with the existence of other agents. The numbers in cells indicate the movement path.}
    \label{fig:evac:debugging:5}
\end{figure}

While we illustrate debugging with a single starting point, comprehensive debugging requires testing multiple starting points to affirm the agent's behavior across various scenarios. Additionally, we can vary other factors during debugging, such as introducing more agents into the grid, as depicted in Figure~\ref{fig:evac:debugging:5}, to test the robustness of the agent's path planning in a multi-agent scenario. By analyzing these tests, we can more effectively pinpoint the origins of behavioral issues. 
Addressing these issues is crucial for the reliability of the overall task simulation.

\subsection{Simulation Results}

We execute the simulation in three scenarios: 
\begin{inparaenum} [(1)]
    \item a \emph{sparse} scenario with 100 agents, 
    \item a \emph{dense} scenario with 400 agents, and     
    \item an \emph{obstacle} scenario with 100 agents and obstacles in the grid. 
\end{inparaenum}
The sparse scenario is used in all sub-tasks. The dense scenario is used in Sub-task 3. The obstacle scenario is used in Sub-task 4. For each scenario, we simulate 5 runs which differ in the starting positions of agents, and confirm our observations reported here apply to all the 5 runs. Personas are randomly assigned to the agents, and each persona type accounts for one-quarter of the total agent population. We report a representative run for each scenario. 

\begin{figure*}[!t]
    \centering
    \begin{subfigure}{0.3\textwidth}
        \includegraphics[width=\linewidth]{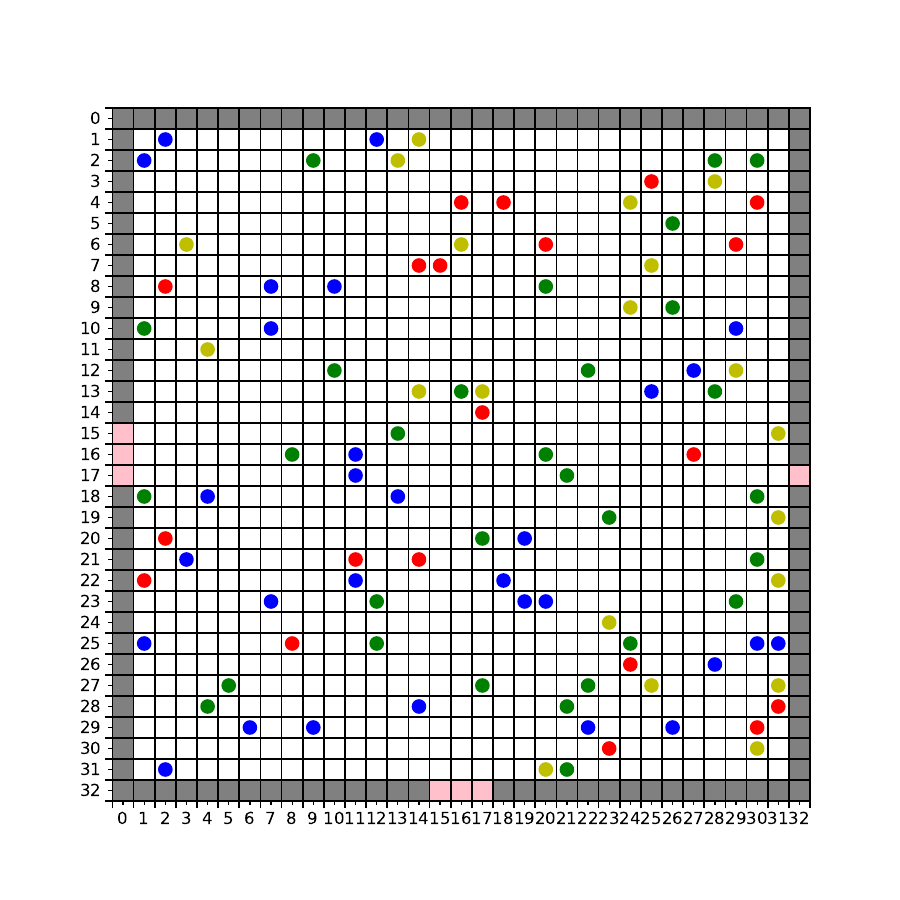}
        \caption{Round 0.}
        \label{fig:evac:task1:1}
    \end{subfigure}
    \begin{subfigure}{0.3\textwidth}
        \includegraphics[width=\linewidth]{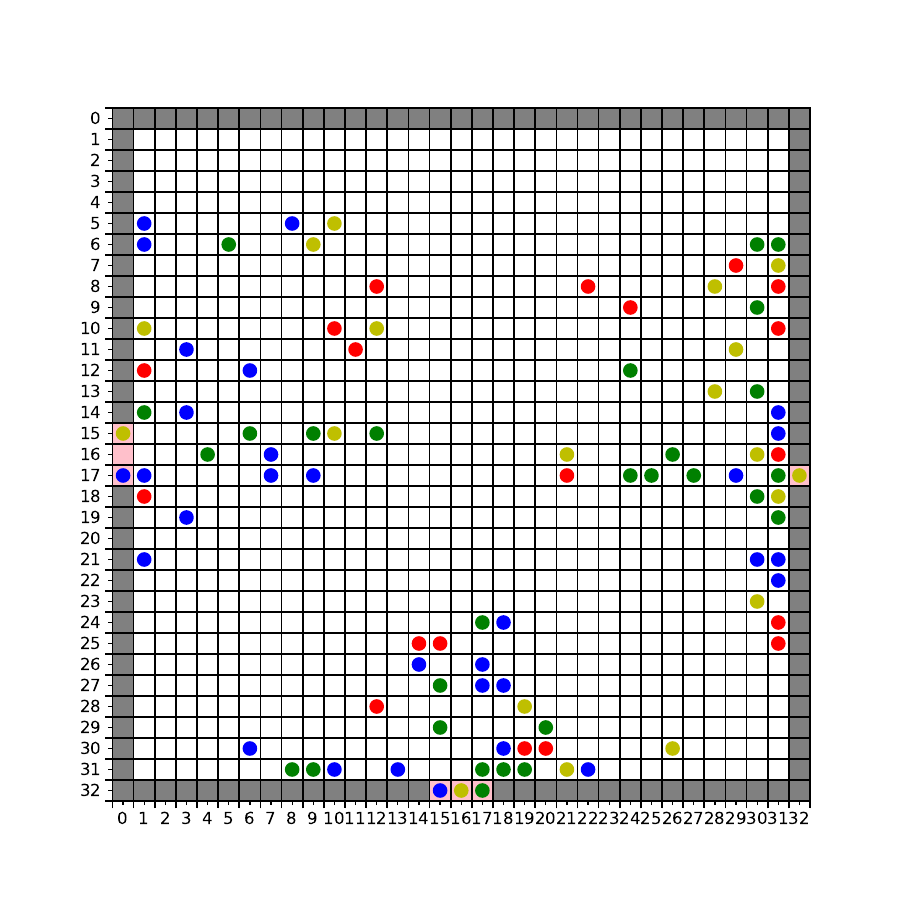}
        \caption{Round 4.}
        \label{fig:evac:task1:2}
    \end{subfigure}
    \begin{subfigure}{0.3\textwidth}
        \includegraphics[width=\linewidth]{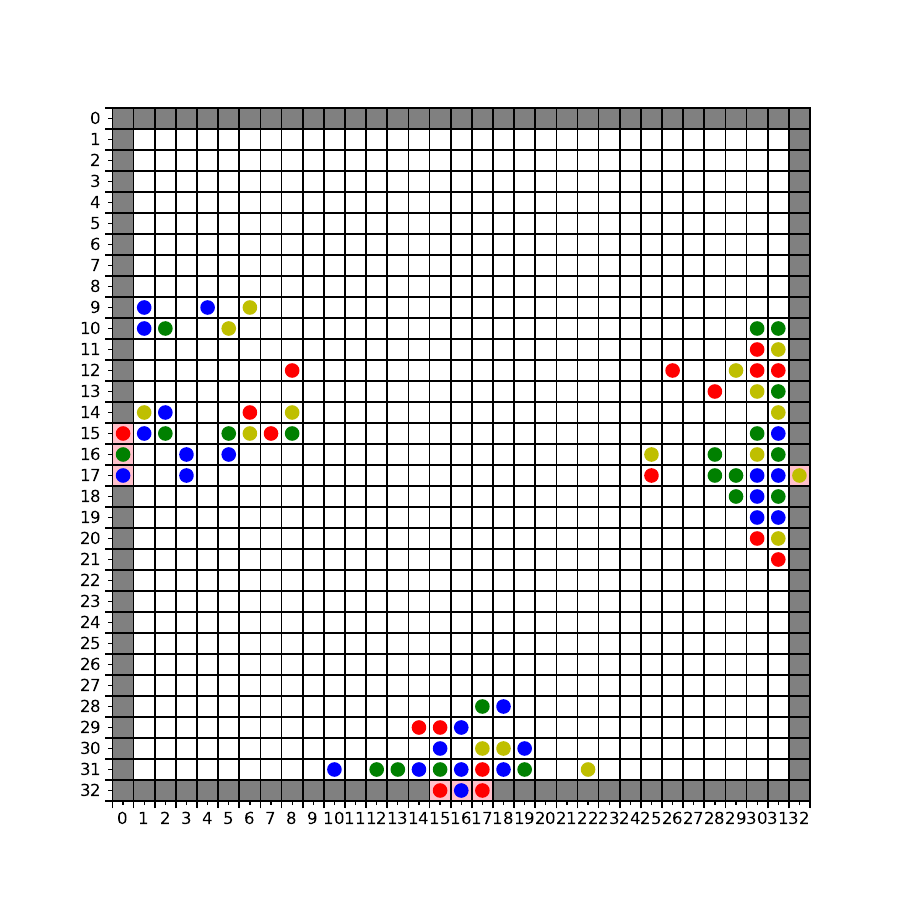}
        \caption{Round 8.}
        \label{fig:evac:task1:3}
    \end{subfigure}
    \begin{subfigure}{0.3\textwidth}
        \includegraphics[width=\linewidth]{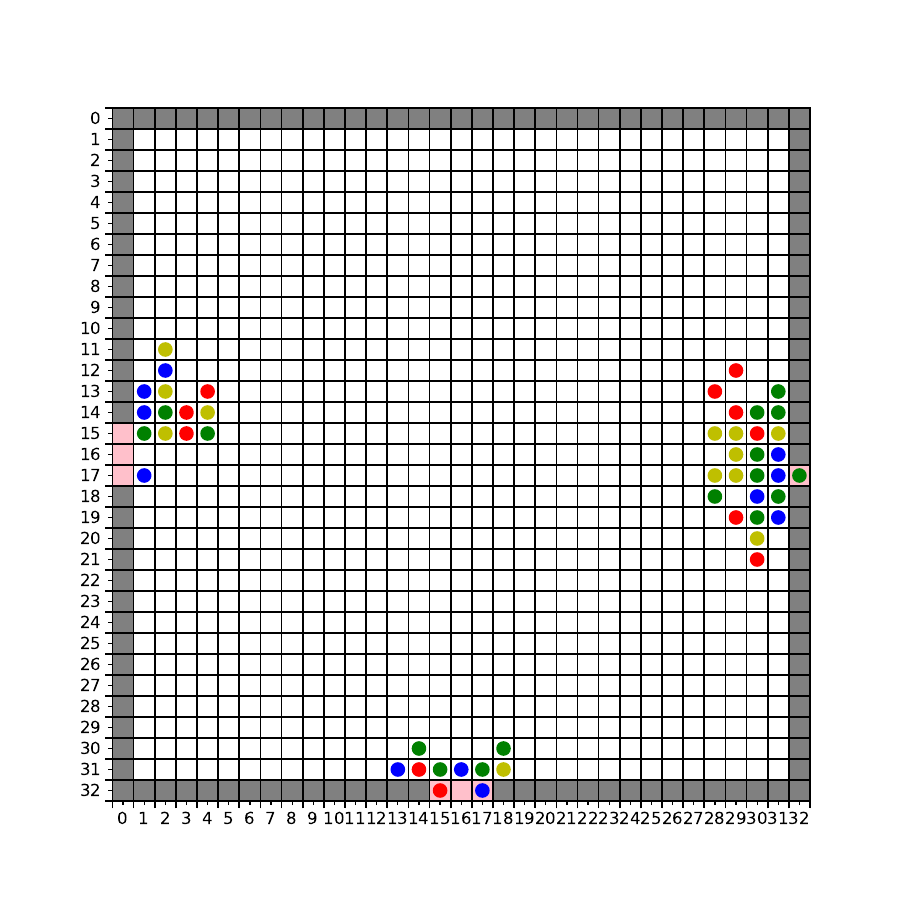}
        \caption{Round 12.}
        \label{fig:evac:task1:4}
    \end{subfigure}
    \begin{subfigure}{0.3\textwidth}
        \includegraphics[width=\linewidth]{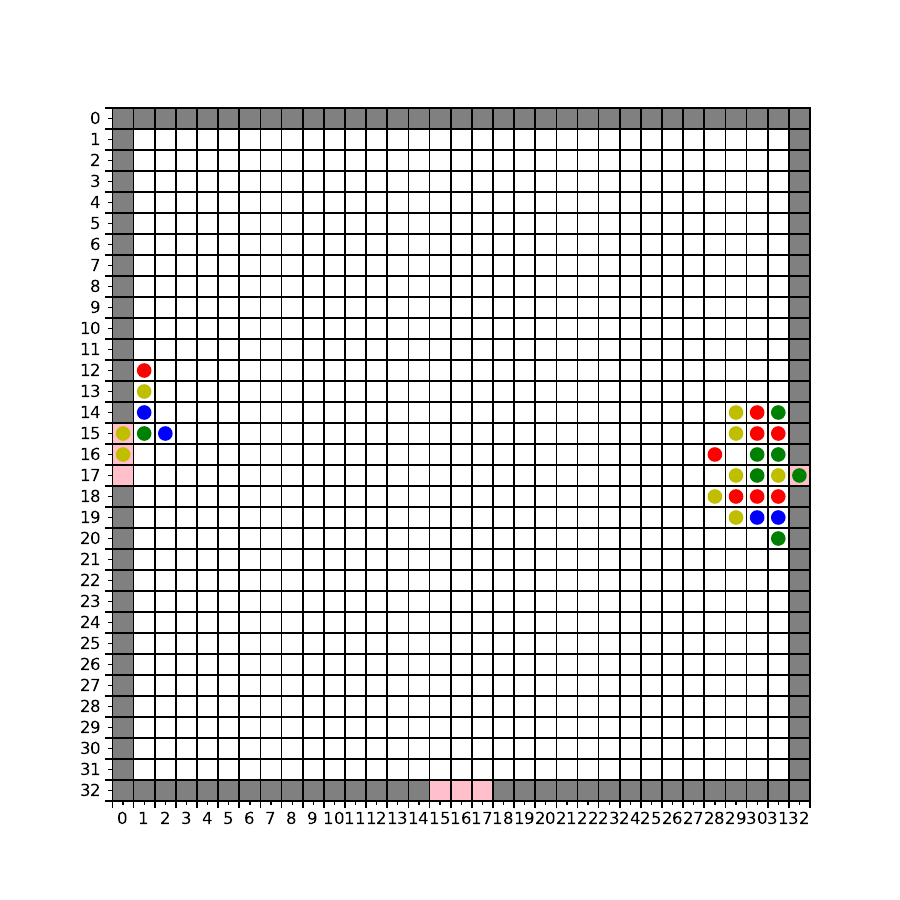}
        \caption{Round 18.}
        \label{fig:evac:task1:5}
    \end{subfigure}
    \begin{subfigure}{0.3\textwidth}
        \includegraphics[width=\linewidth]{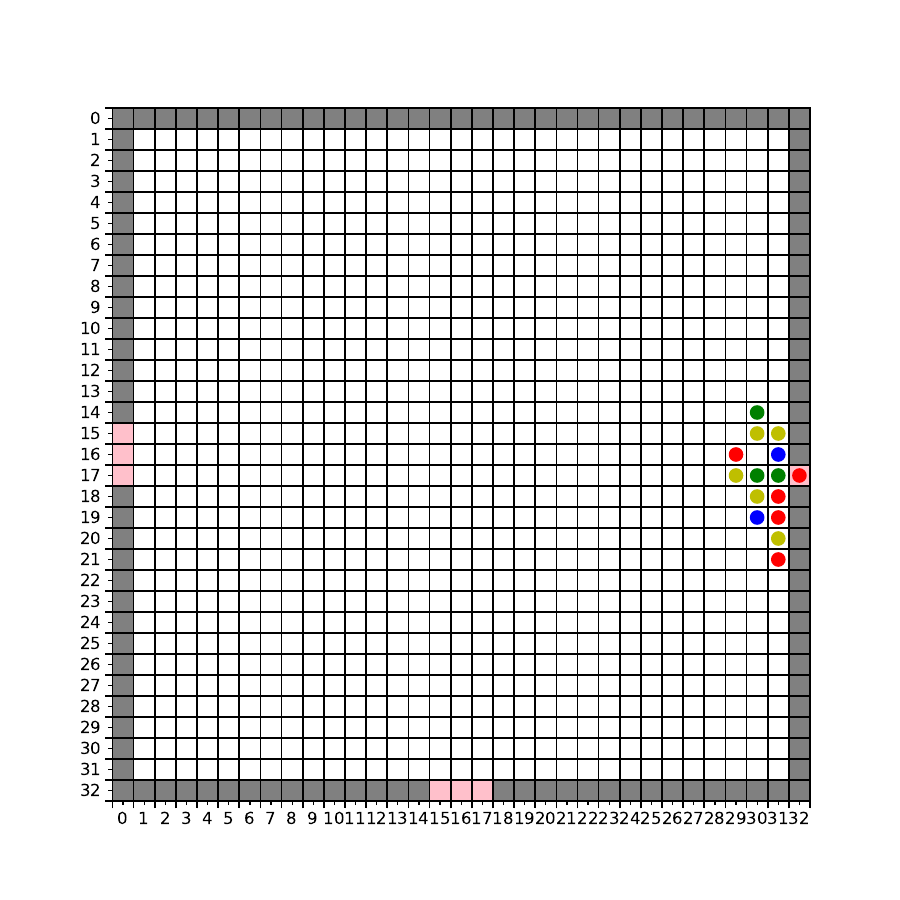}
        \caption{Round 24.}
        \label{fig:evac:task1:6}
    \end{subfigure}
    \caption{Sample run of Sub-task 1, sparse scenario (41 rounds in total).}
    \label{fig:evac:task1}
\end{figure*}

\subsubsection{Sub-task 1: Reproducing ABM Parameters}

Sub-task 1 involves a comparative analysis between the ABM and SABM setups. We reproduce the ABM setup and only apply SABM to determine the parameters for the panic level of agents. We adapt the Stage 1 prompt, which solicits agents' subjective feelings to equate an agent's expressed state of panic with the ABM parameters. In the revised prompt, the agent articulates a range of intensity for three decisive factors -- exit proximity, the number of people, and crowd density -- that are present in the ABM setup, thus indicating the agent's level of panic. The intensities are categorized as minimal, mild, moderate, high, and extreme. We uniformly translate these intensity categories into ABM parameters. The revised prompt is structured as follows:

\begin{prompt}{Stage 1: Revised subjective feelings for ABM setup}
  Please tell me your inclination: Exit Proximity, People Count, Crowd Density.\\
  For each aspect, please use one of the five words listed below: [minimal, mild, moderate, high, extreme] \\
  Your output should only contain three words, with commas between them. For example, 'minimal, mild, moderate' is a valid output. No period in the end. This output indicates that you are extremely focused on Exit Proximity, and you are mildly focused on People Count, and you are moderately focused on Crowd Density.
\end{prompt}

Figure~\ref{fig:evac:task1} presents the snapshots of an example run for Sub-task 1. Observations indicate that due to the narrow exit on the right, by Round 24, as depicted in Figure~\ref{fig:evac:task1:6}, while evacuations through other exits have concluded, the right exit remains congested. Additionally, it is noted that for every exit, the agents tend to congest in front of it and contend to evacuate. These patterns align with the outcomes reported in \cite{wang2015modeling}. Based on these observations, we can assert that the SABM approach is capable of reproducing the evacuation scenarios depicted by ABM, demonstrating that SABM is a viable alternative to ABM's parameter configurations.

\begin{figure*}[!t]
    \centering
    \begin{subfigure}{0.3\textwidth}
        \includegraphics[width=\linewidth]{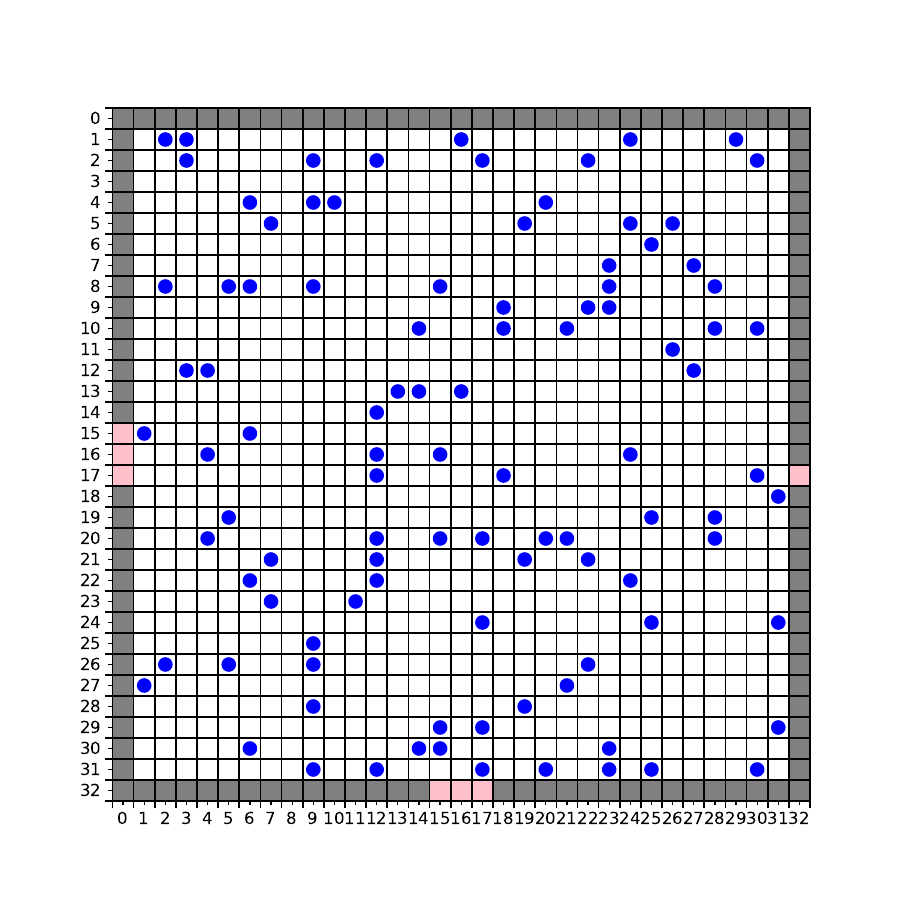}
        \caption{Round 0.}
        \label{fig:evac:task2:1}
    \end{subfigure}
    \begin{subfigure}{0.3\textwidth}
        \includegraphics[width=\linewidth]{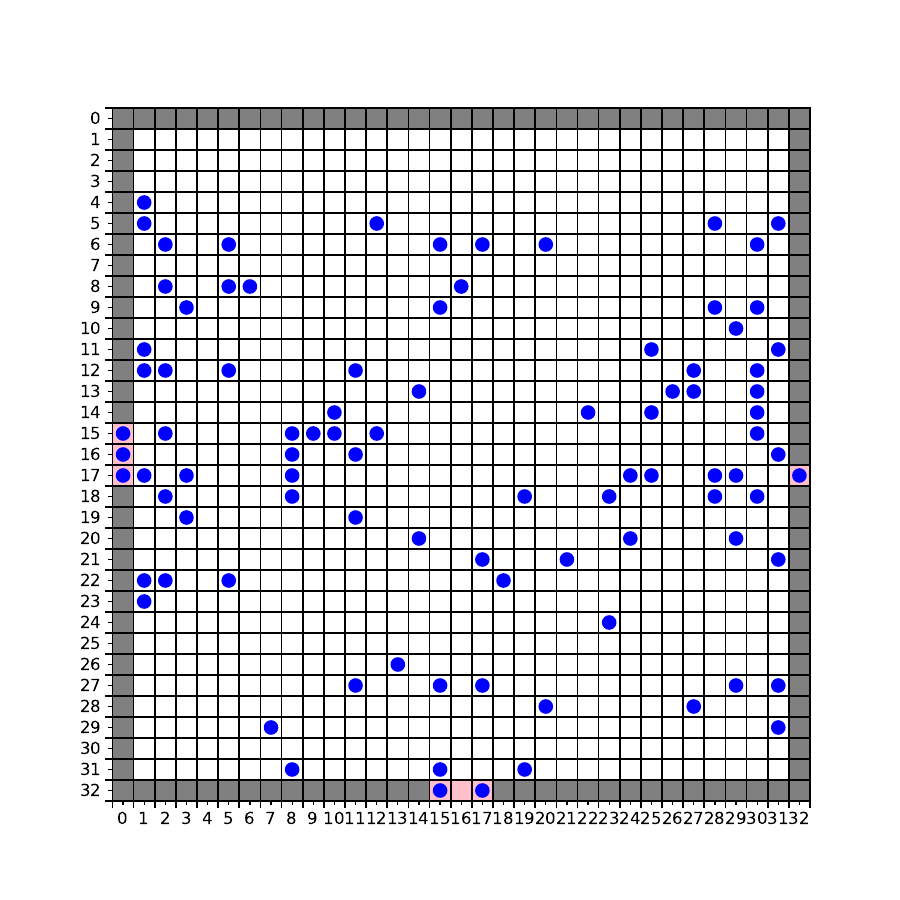}
        \caption{Round 4.}
        \label{fig:evac:task2:2}
    \end{subfigure}
    \begin{subfigure}{0.3\textwidth}
        \includegraphics[width=\linewidth]{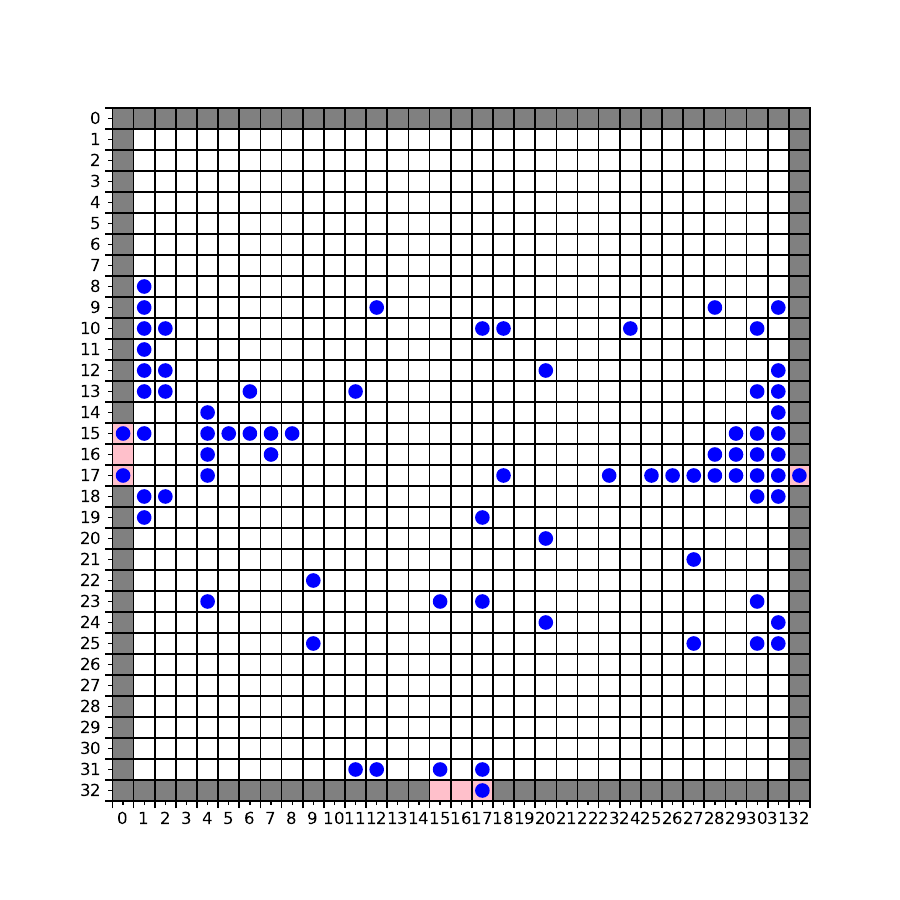}
        \caption{Round 8.}
        \label{fig:evac:task2:3}
    \end{subfigure}
    \begin{subfigure}{0.3\textwidth}
        \includegraphics[width=\linewidth]{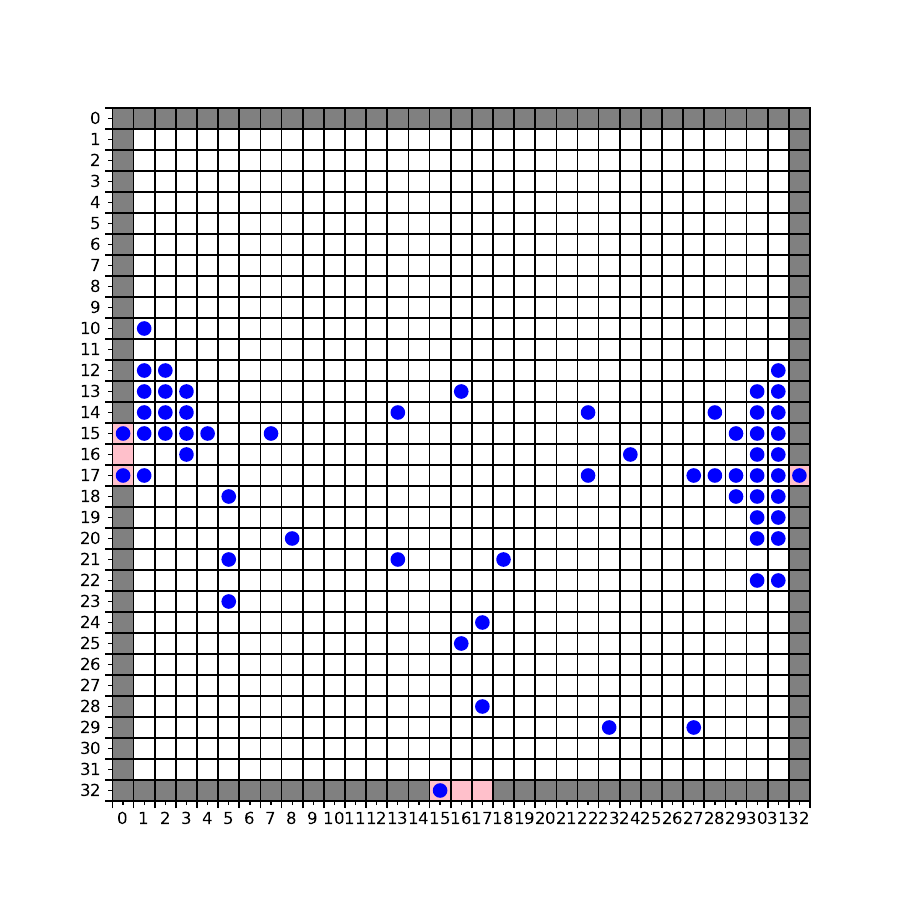}
        \caption{Round 12.}
        \label{fig:evac:task2:4}
    \end{subfigure}
    \begin{subfigure}{0.3\textwidth}
        \includegraphics[width=\linewidth]{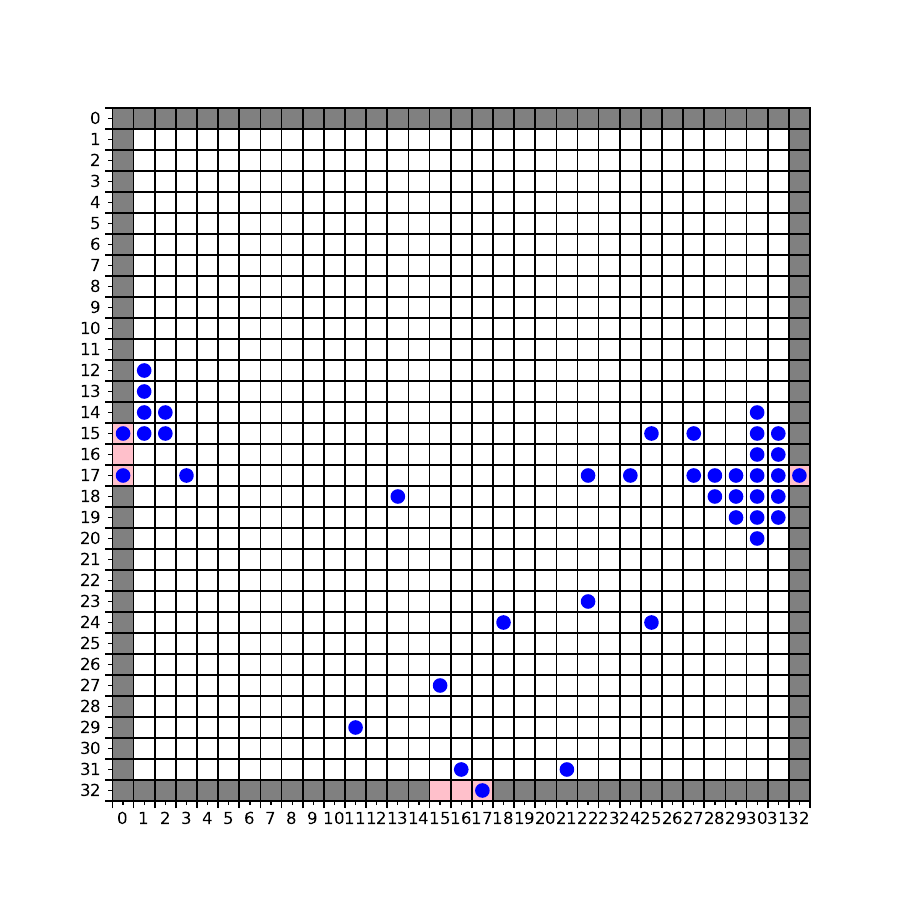}
        \caption{Round 18.}
        \label{fig:evac:task2:5}
    \end{subfigure}
    \begin{subfigure}{0.3\textwidth}
        \includegraphics[width=\linewidth]{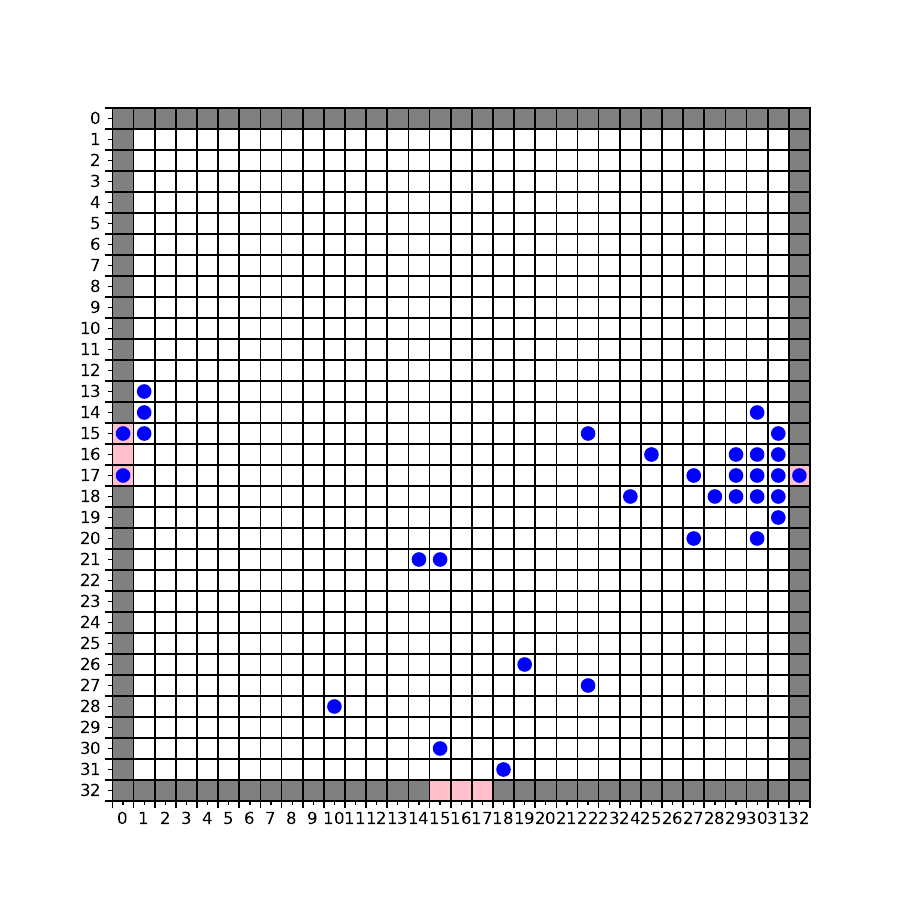}
        \caption{Round 21.}
        \label{fig:evac:task2:6}
    \end{subfigure}
    \begin{subfigure}{0.3\textwidth}
        \includegraphics[width=\linewidth]{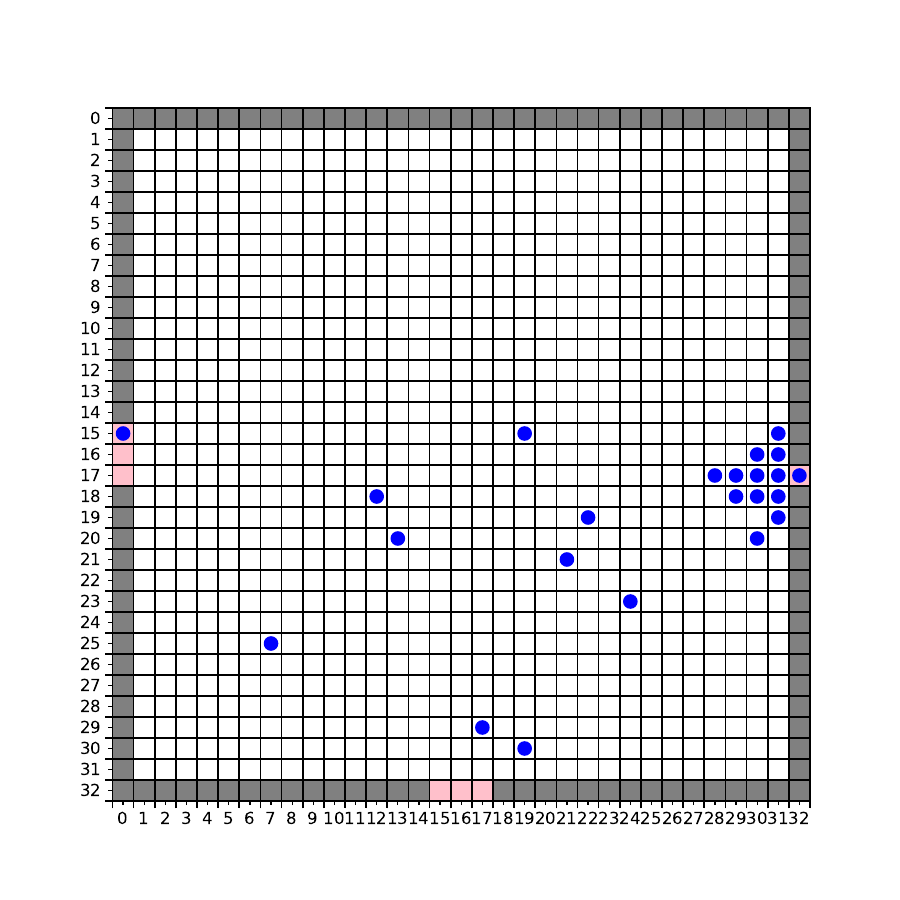}
        \caption{Round 24.}
        \label{fig:evac:task2:7}
    \end{subfigure}
    \begin{subfigure}{0.3\textwidth}
        \includegraphics[width=\linewidth]{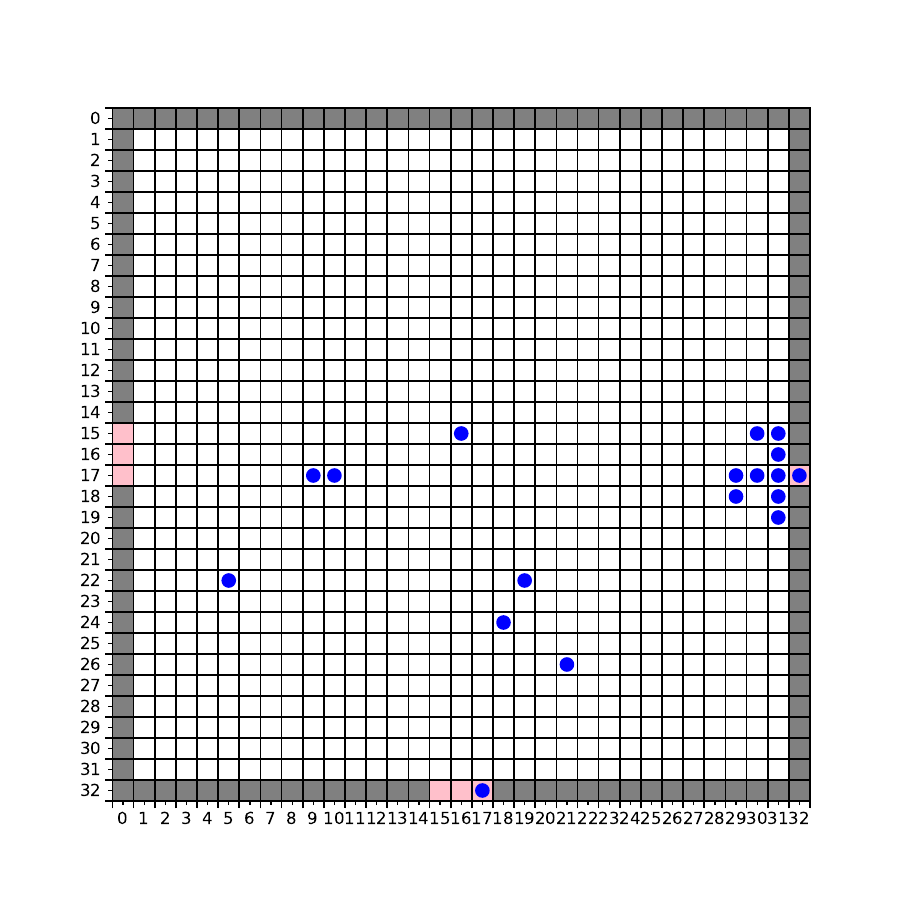}
        \caption{Round 27.}
        \label{fig:evac:task2:8}
    \end{subfigure}
    \begin{subfigure}{0.3\textwidth}
        \includegraphics[width=\linewidth]{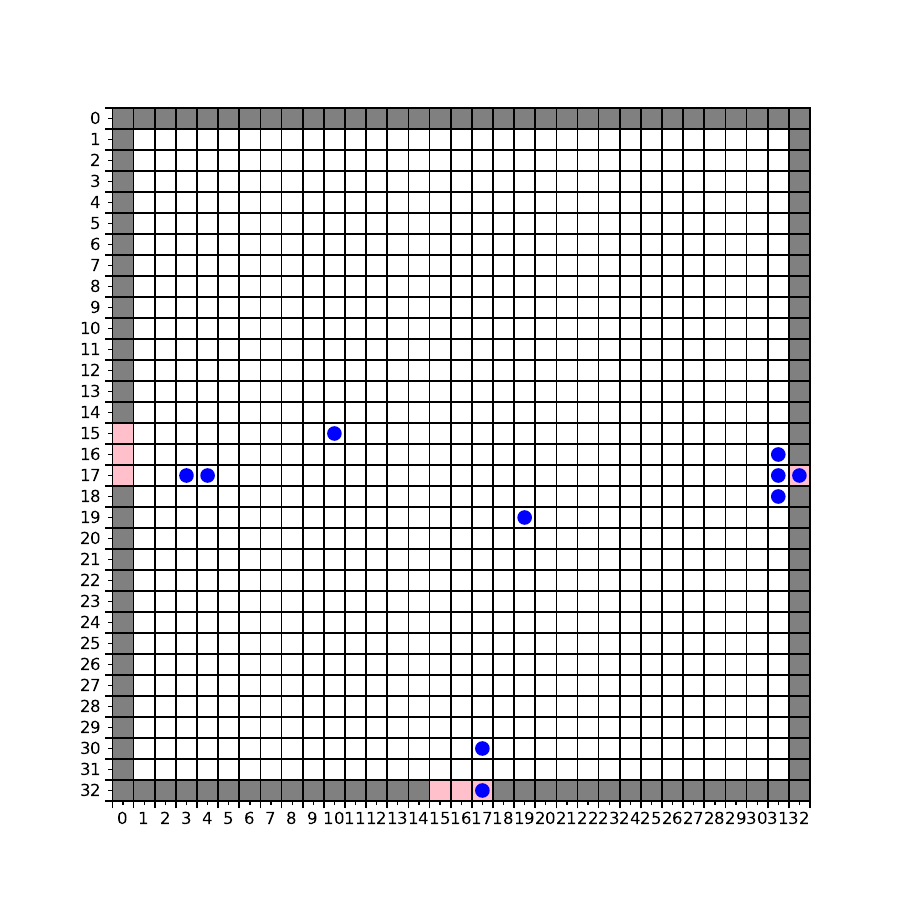}
        \caption{Round 33.}
        \label{fig:evac:task2:9}
    \end{subfigure}
    \caption{Sample run of Sub-task 2, sparse scenario (45 rounds in total). We detail the display of Rounds 18 to 27 to illustrate the pattern of agents switching target exits in crowded conditions.}
    \label{fig:evac:task2}
\end{figure*}

\subsubsection{Sub-task 2: Test of Base Model}

Figure~\ref{fig:evac:task2} displays the results of the base model, where personas are absent. In particular, we detail the snapshots from Rounds 18 to 27. These snapshots reveal that the agents are able to recognize when the right exit is excessively congested and, despite being nearer to it, they choose to reroute to the other exits, which have already been largely vacated. This behavior suggests that the agents have an inherent adaptability, allowing them to make prudent choices by integrating situational information and their subjective feelings. The findings affirm that the SABM is adept at formulating evacuation plans using only natural language instructions to guide the agents to evacuation destinations, without needing parameters that could introduce researcher biases to the simulation. Additionally, the model underscores that subjective perceptions can effectively influence decision-making, a factor traditionally challenging to quantify.

\begin{figure*}[!t]
    \centering
    \begin{subfigure}{0.3\textwidth}
        \includegraphics[width=\linewidth]{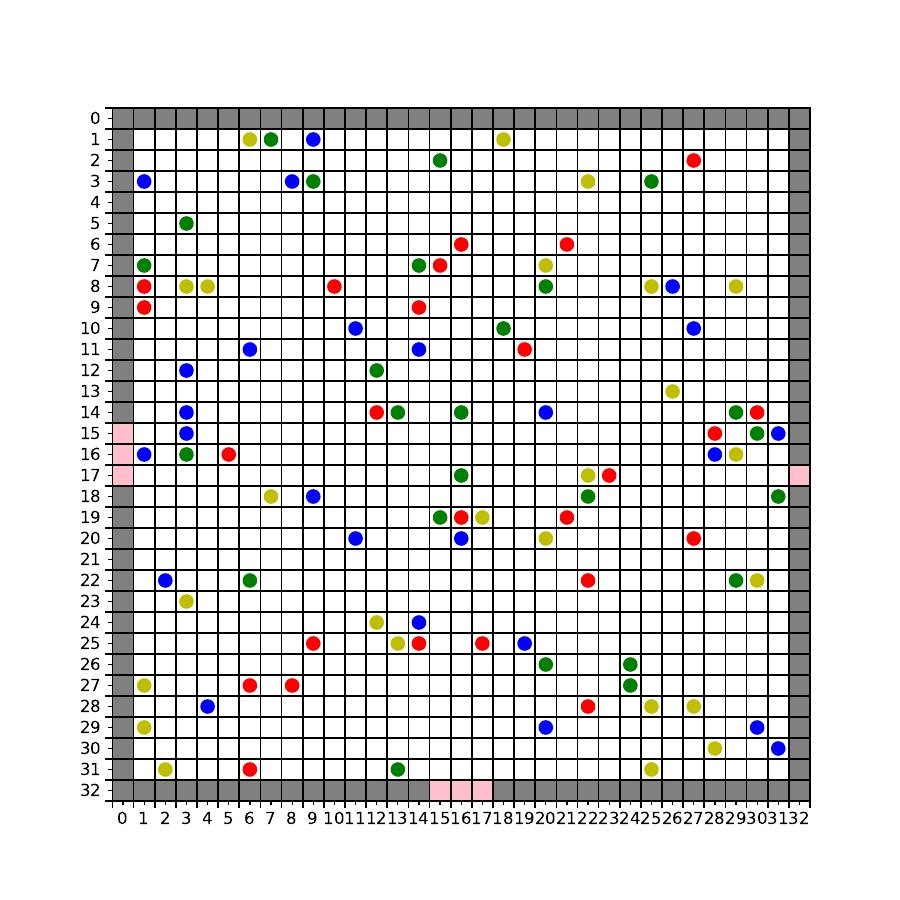}
        \caption{Round 0.}
        \label{fig:evac:task3:1}
    \end{subfigure}
    \begin{subfigure}{0.3\textwidth}
        \includegraphics[width=\linewidth]{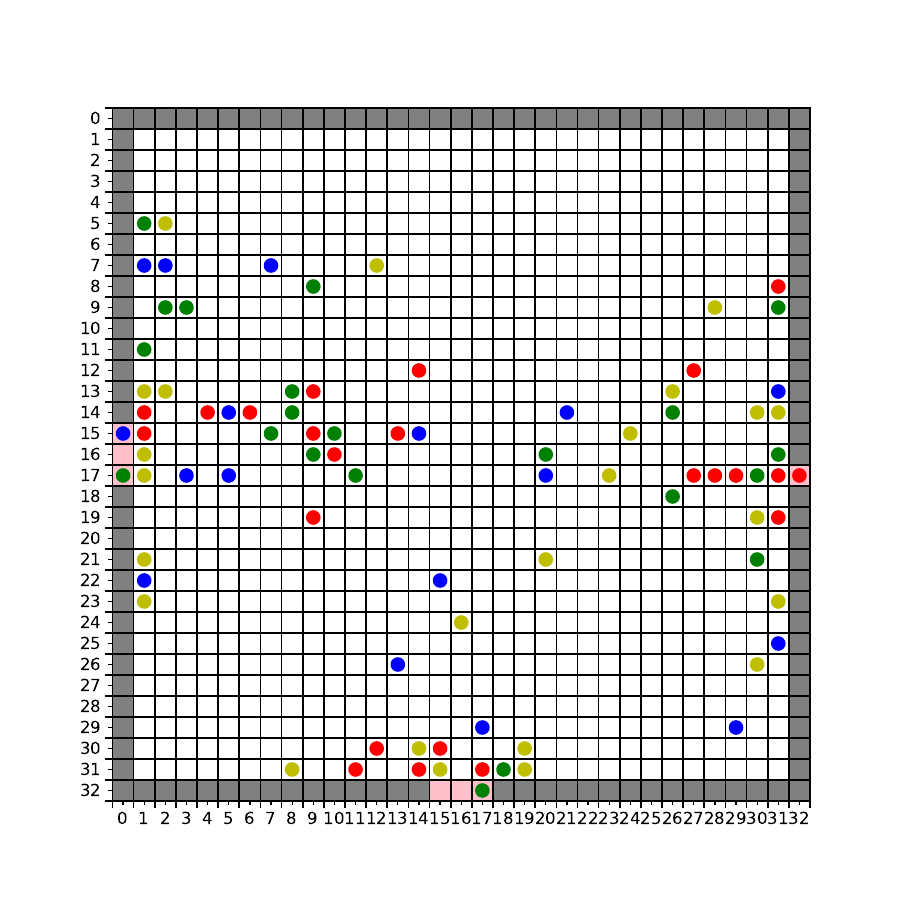}
        \caption{Round 6.}
        \label{fig:evac:task3:2}
    \end{subfigure}
    \begin{subfigure}{0.3\textwidth}
        \includegraphics[width=\linewidth]{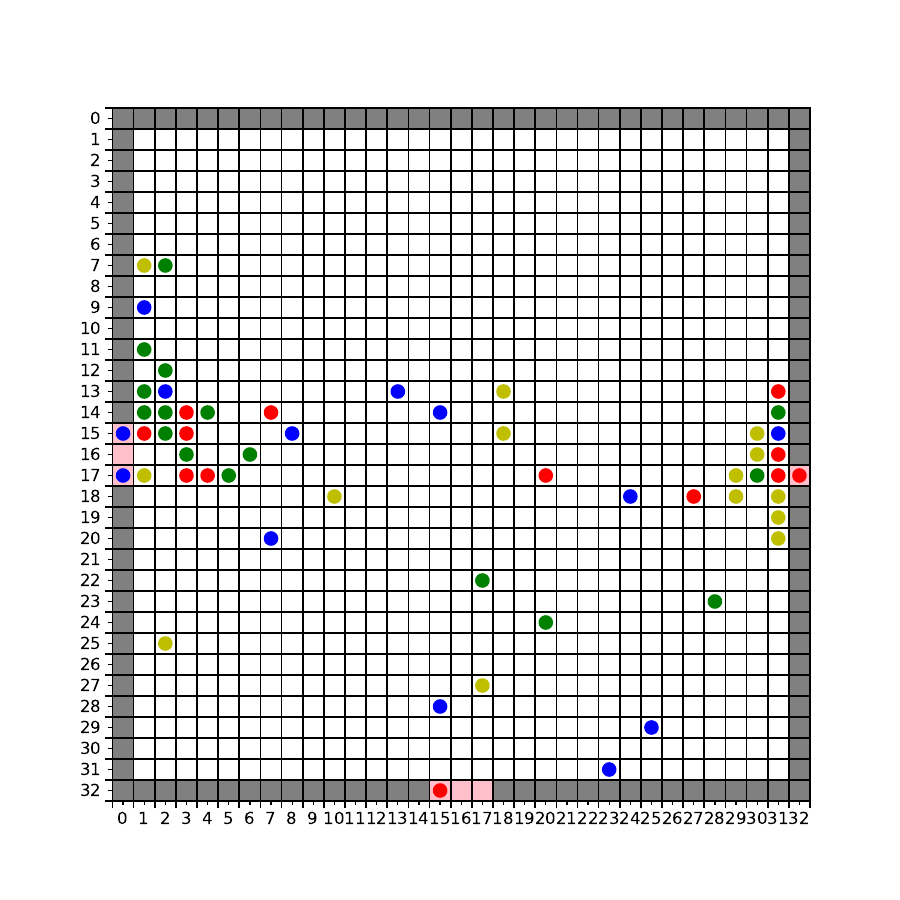}
        \caption{Round 12.}
        \label{fig:evac:task3:3}
    \end{subfigure}
    \begin{subfigure}{0.3\textwidth}
        \includegraphics[width=\linewidth]{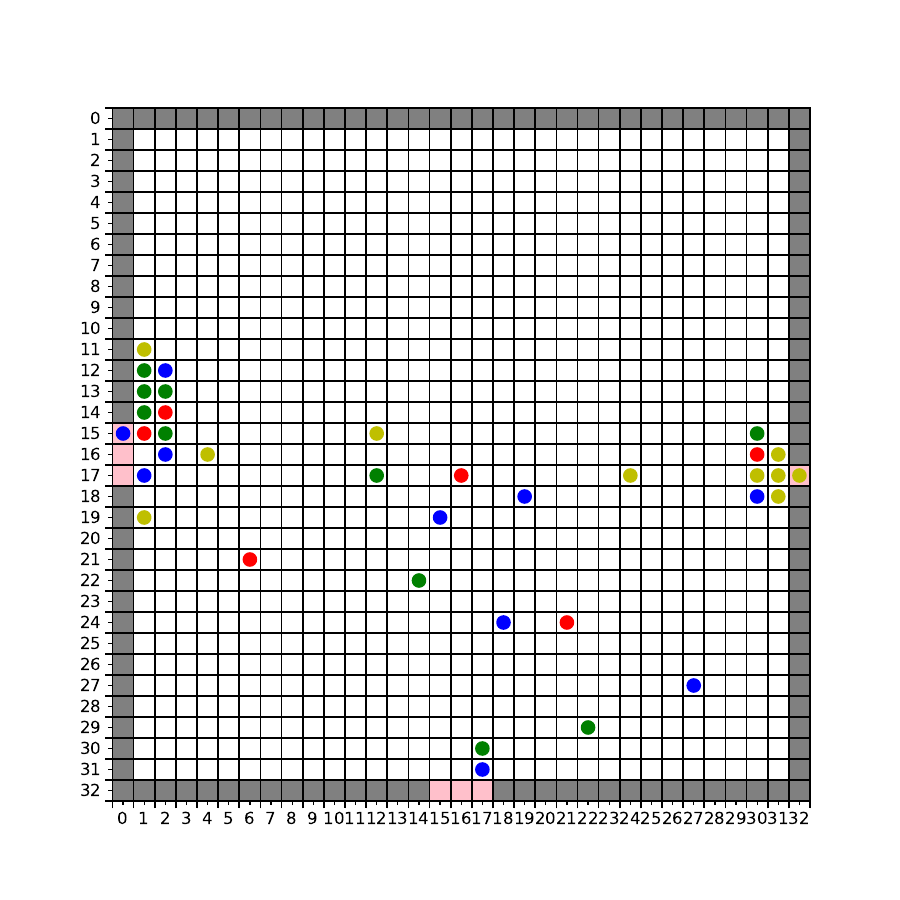}
        \caption{Round 18.}
        \label{fig:evac:task3:4}
    \end{subfigure}
    \begin{subfigure}{0.3\textwidth}
        \includegraphics[width=\linewidth]{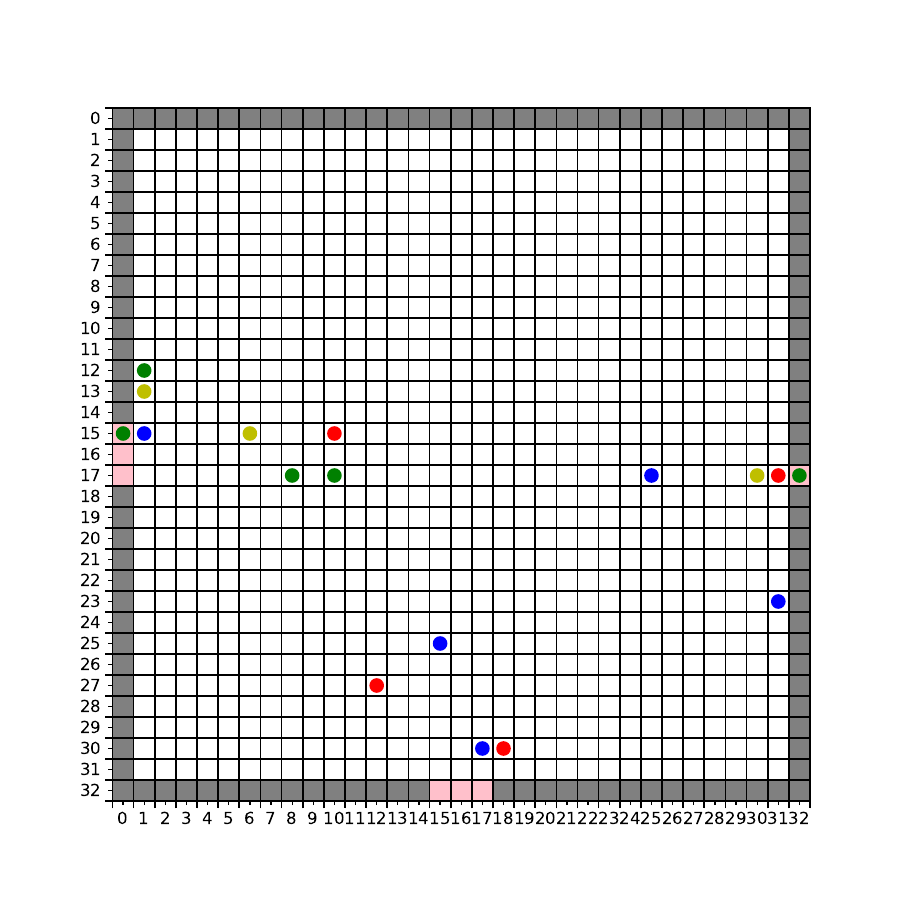}
        \caption{Round 24.}
        \label{fig:evac:task3:5}
    \end{subfigure}
    \begin{subfigure}{0.3\textwidth}
        \includegraphics[width=\linewidth]{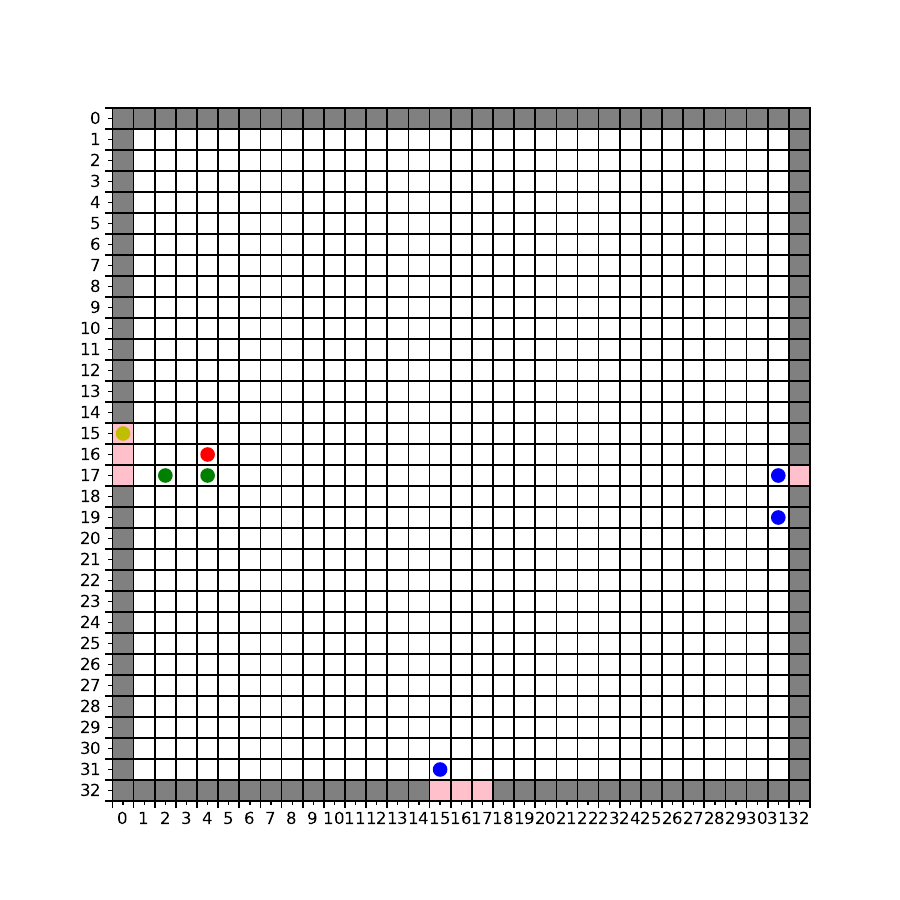}
        \caption{Round 30.}
        \label{fig:evac:task3:6}
    \end{subfigure}
    \caption{Sample run of Sub-task 3, sparse scenario (39 rounds in total). Colors represent personas: red for physically strong and mentally strong, green for physically strong but mentally weak, yellow for physically weak but mentally strong, and blue for physically weak and mentally weak.}
    \label{fig:evac:task3}
\end{figure*}

\begin{figure*}[!t]
    \centering
    \begin{subfigure}{0.45\textwidth}
        \includegraphics[width = \linewidth]{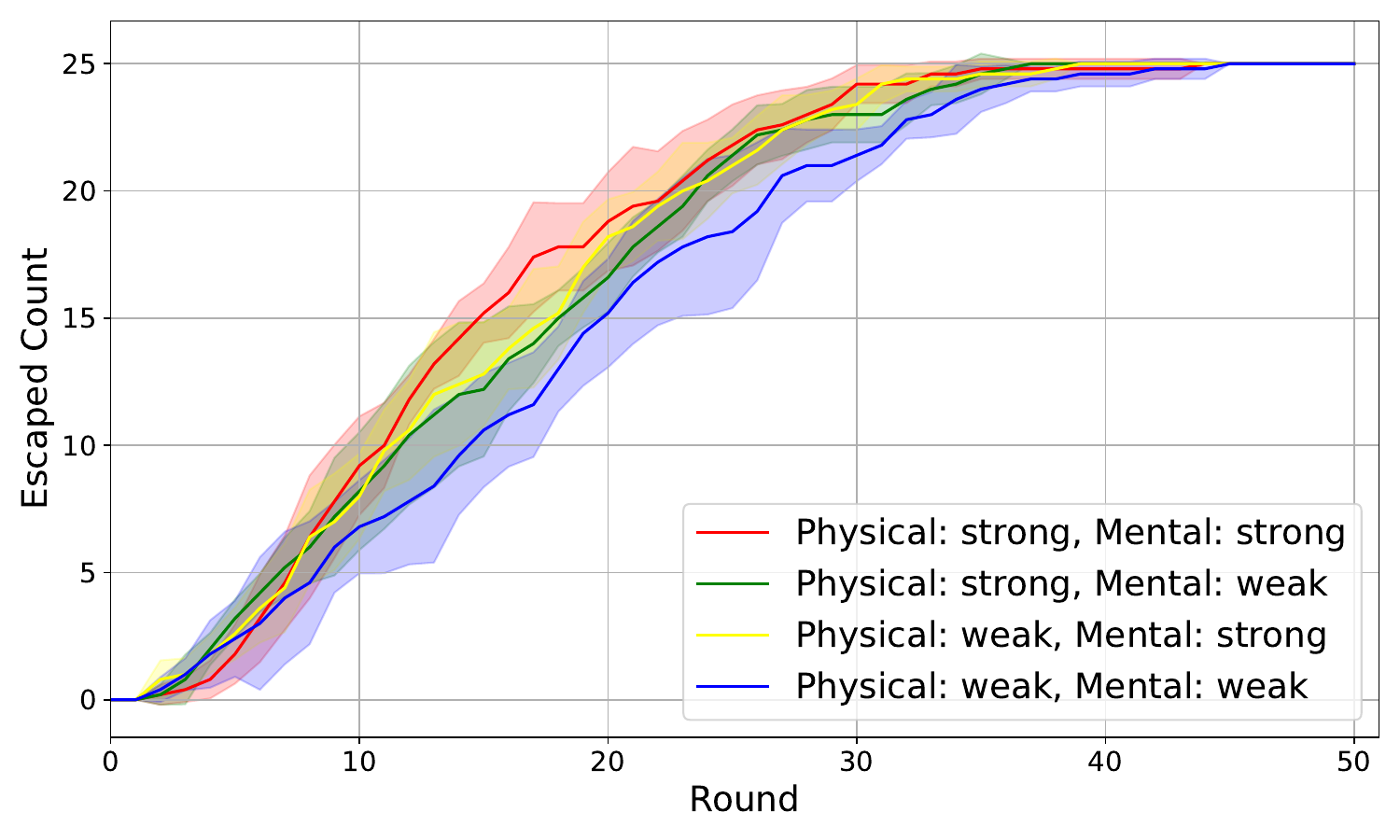}
        \caption{Sub-task 3.}
        \label{fig:evac:escaped:3}
    \end{subfigure}
    \hspace{4ex}
    \begin{subfigure}{0.45\textwidth}
        \includegraphics[width = \linewidth]{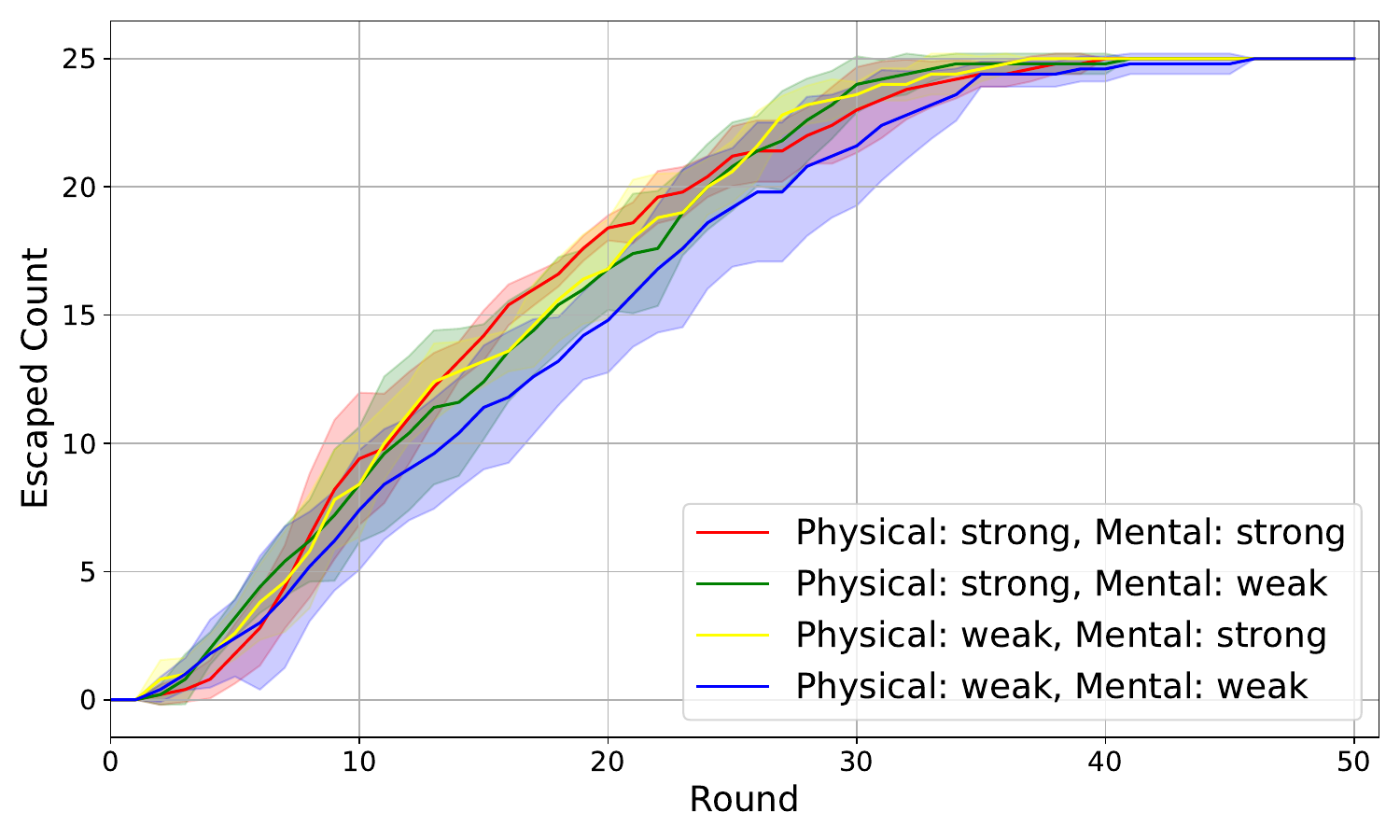}
        \caption{Sub-task 4.}
        \label{fig:evac:escaped:4}
    \end{subfigure}  
    \caption{Statistics of successful evacuation by persona groups.}
    \label{fig:evac:escaped}
\end{figure*}


\begin{figure*}[!t]
    \centering
    \begin{subfigure}{0.3\textwidth}
        \includegraphics[width=\linewidth]{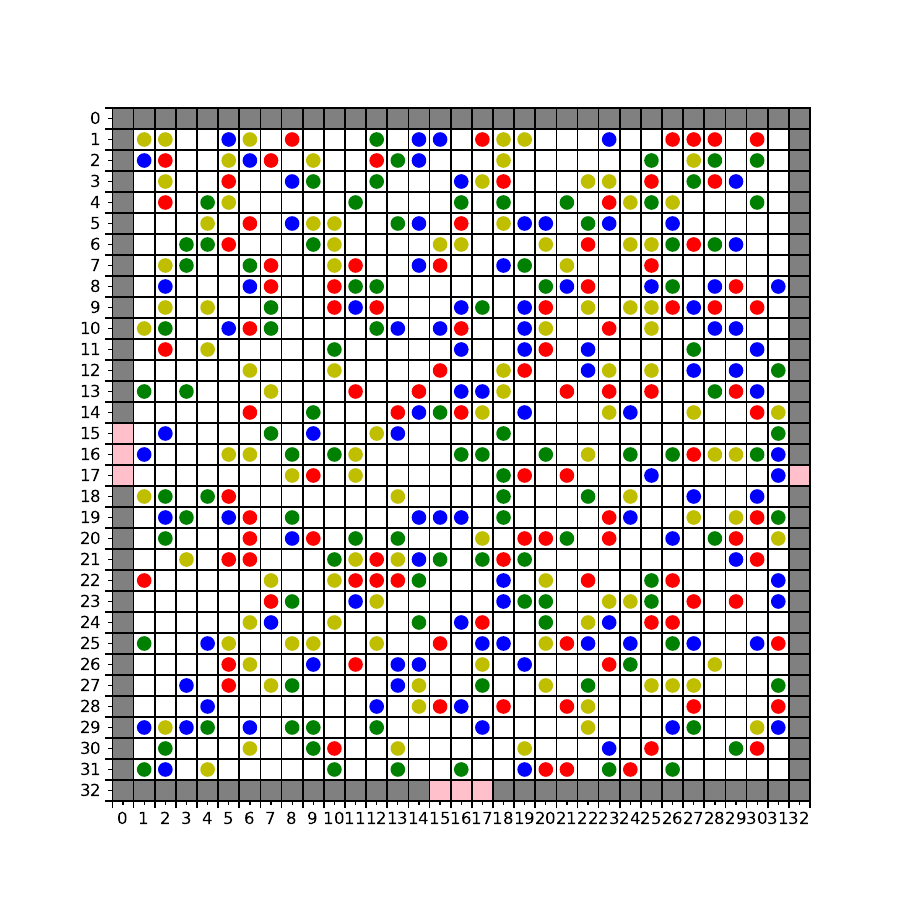}
        \caption{Round 0.}
        \label{fig:evac:task3-400:1}
    \end{subfigure}
    \begin{subfigure}{0.3\textwidth}
        \includegraphics[width=\linewidth]{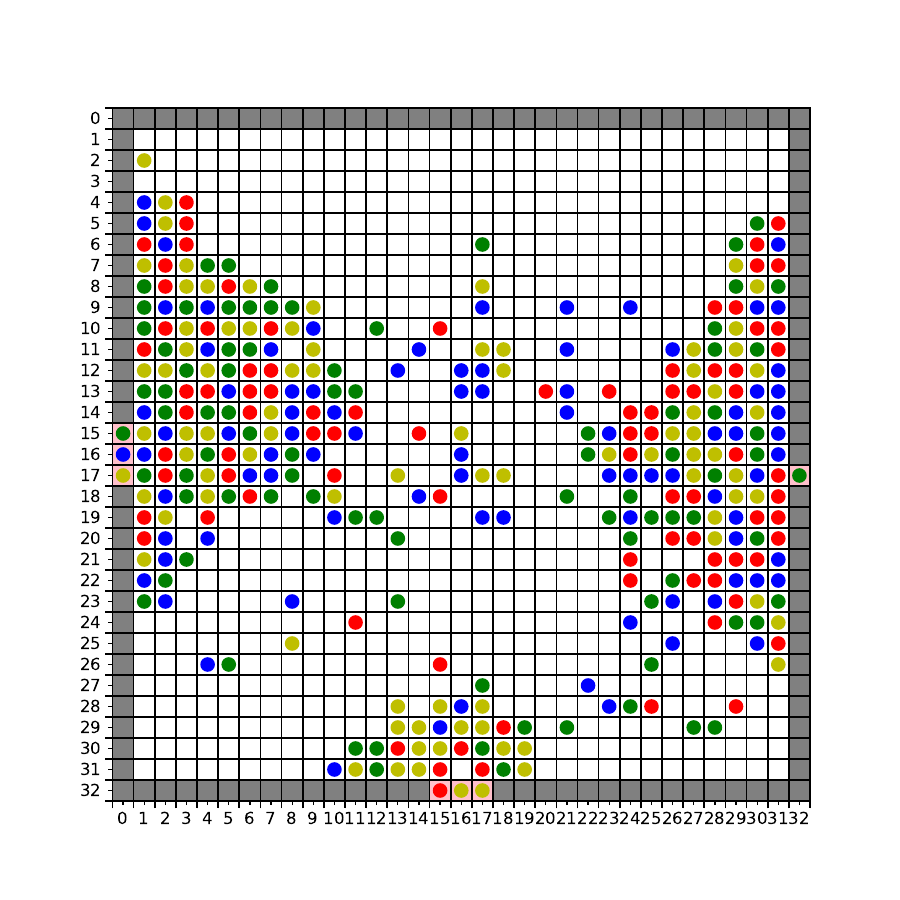}
        \caption{Round 10.}
        \label{fig:evac:task3-400:2}
    \end{subfigure}
    \begin{subfigure}{0.3\textwidth}
        \includegraphics[width=\linewidth]{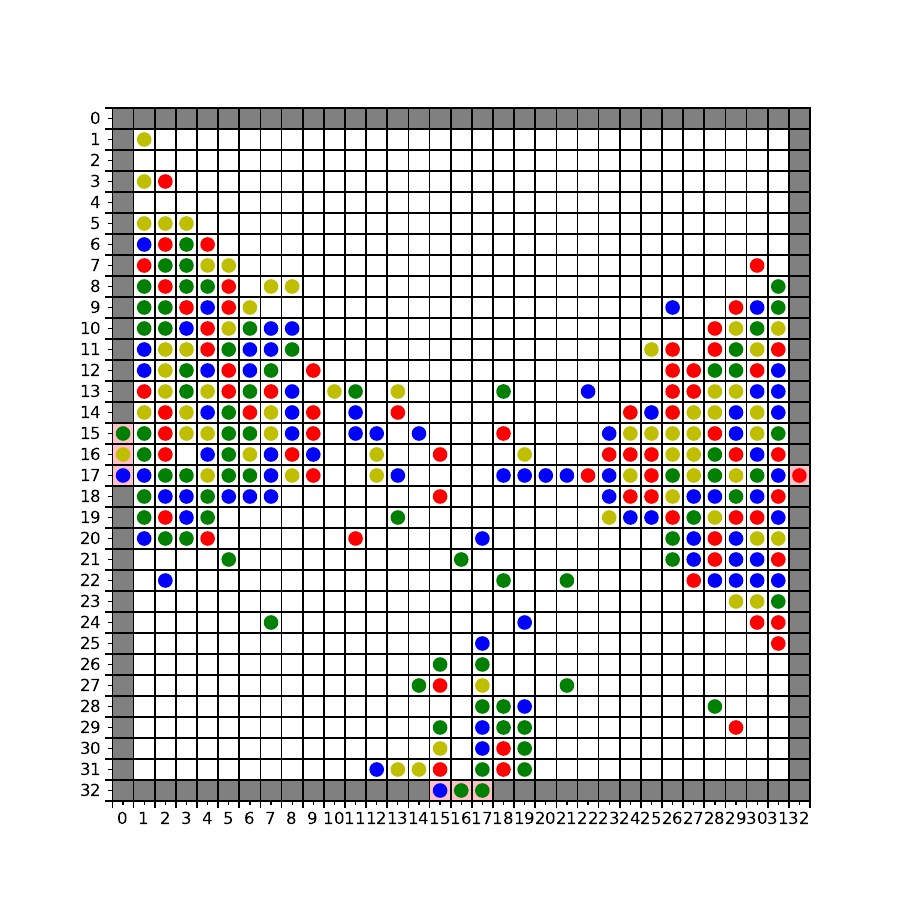}
        \caption{Round 20.}
        \label{fig:evac:task3-400:3}
    \end{subfigure}
    \begin{subfigure}{0.3\textwidth}
        \includegraphics[width=\linewidth]{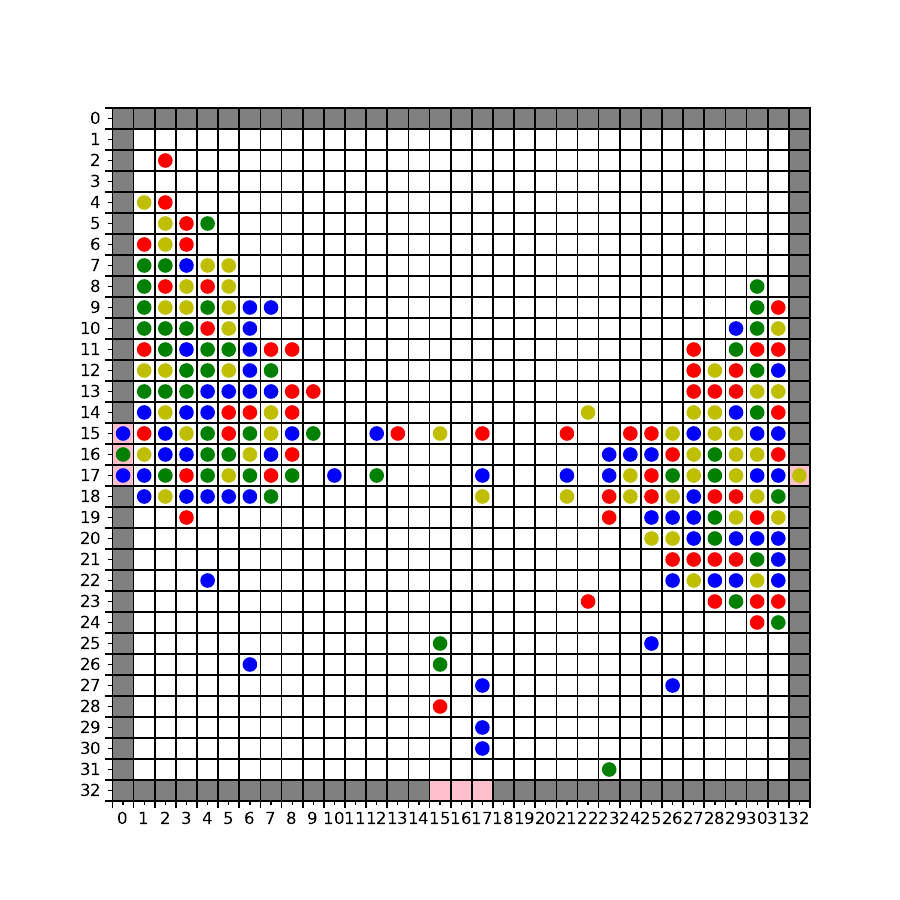}
        \caption{Round 30.}
        \label{fig:evac:task3-400:4}
    \end{subfigure}
    \begin{subfigure}{0.3\textwidth}
        \includegraphics[width=\linewidth]{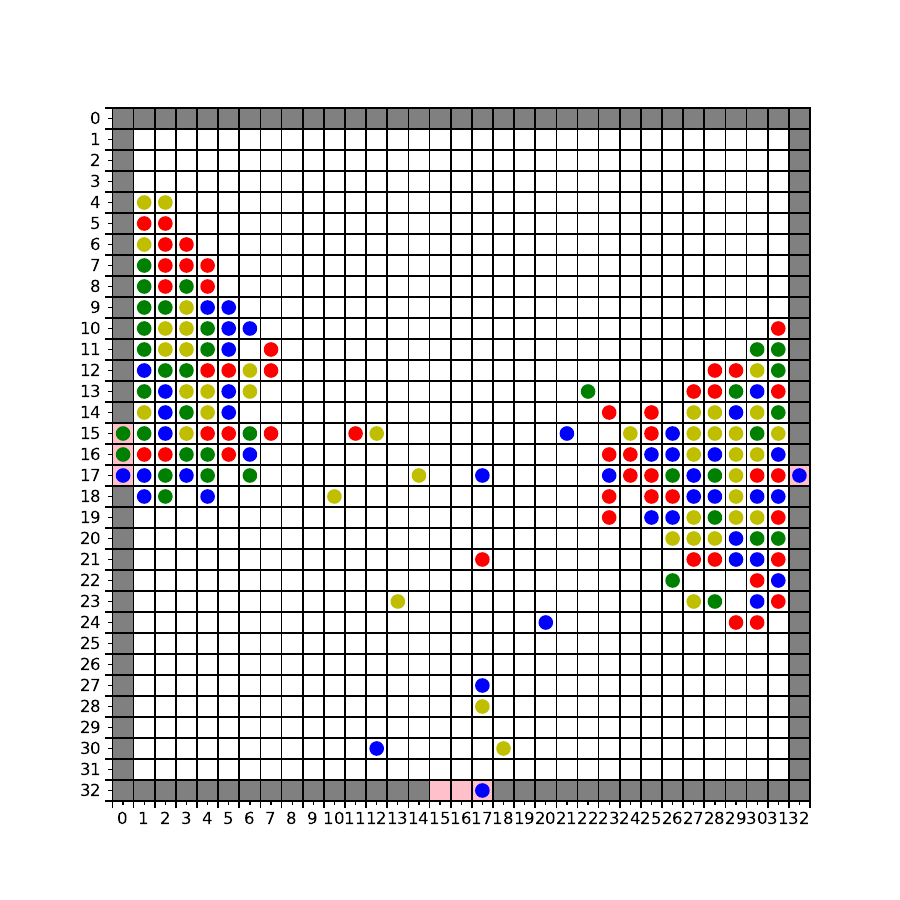}
        \caption{Round 40.}
        \label{fig:evac:task3-400:5}
    \end{subfigure}
    \begin{subfigure}{0.3\textwidth}
        \includegraphics[width=\linewidth]{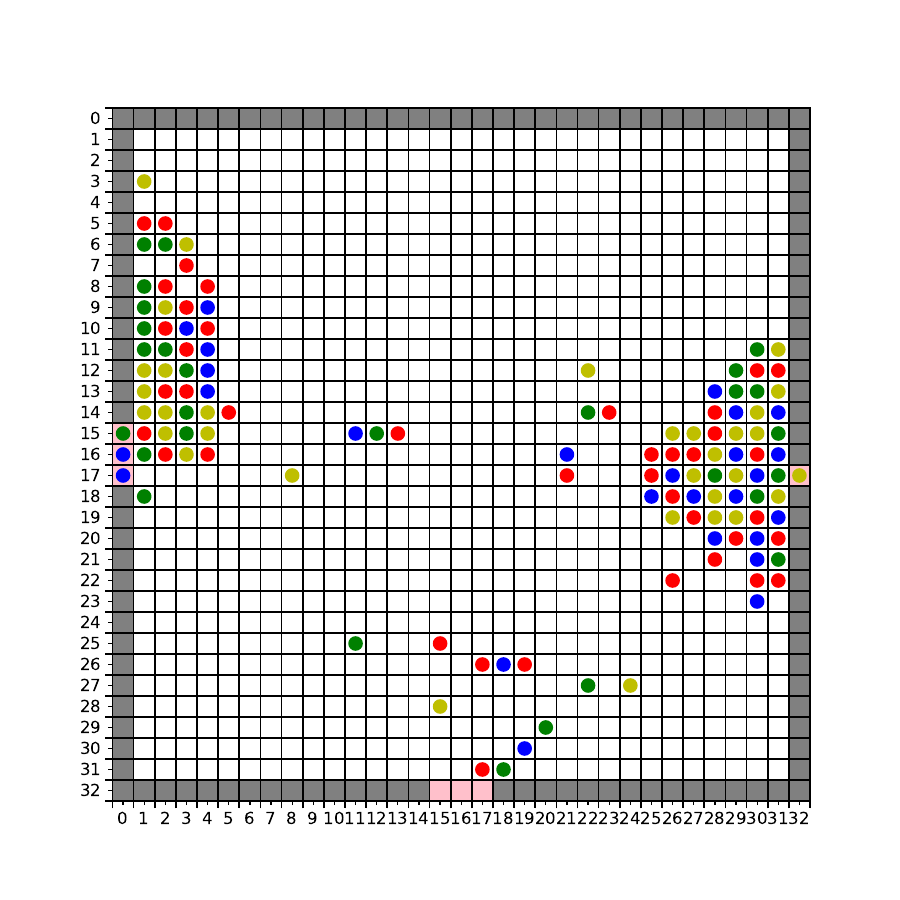}
        \caption{Round 50.}
        \label{fig:evac:task3-400:6}
    \end{subfigure}
    \caption{Sample run of Sub-task 3, dense scenario (50 rounds in total). Colors represent personas: red for physically strong and mentally strong, green for physically strong but mentally weak, yellow for physically weak but mentally strong, and blue for physically weak and mentally weak.}
    \label{fig:evac:task3-400}
\end{figure*}

\subsubsection{Sub-task 3: Test of Personalization}

Building upon the base model, we incorporate personas for agents to simulate and assess the impact of  physical and mental states on their movement decisions. From Figure~\ref{fig:evac:task3}, it becomes apparent that after the integration of personas, the agents' evacuation patterns become more balanced, resulting in a natural tendency to avoid crowded exits -- a decision that arises from the agents' common sense rather than a directive from the prompt. In Figure~\ref{fig:evac:task3:6}, by Round 30, the numbers of individuals at all the three exits are comparatively equal and small. In contrast, when comparing Figure~\ref{fig:evac:task3:5} from Sub-task 3 with Figure~\ref{fig:evac:task1:6} from Sub-task 1, we see that in Sub-task 1, a considerable number of agents remain at the right exit by Round 24, whereas in Sub-task 3, fewer agents are left, and they are more evenly dispersed. These observations underscore the agents' inherent capacity for spontaneous decision alteration and their remarkable adaptability.

Furthermore, Figure~\ref{fig:evac:escaped:3} shows the mean and variance of the cumulative count of evacuated agents across 5 simulation runs. It is clear that the group denoted in red, possessing strong physical and mental attributes, evacuates more swiftly than the two groups (in green and yellow) that exhibit weakness in either physical or mental capacity, and considerably more efficiently than the blue group, which is deficient in both respects. This highlights that personas have a significant impact on agents' behavioral patterns and correspond with realistic situations, such as the tendency for individuals with both physical and mental weaknesses to have a diminished evacuation capability.

Figure~\ref{fig:evac:task3-400} illustrates a simulation from Sub-task 3 with the dense scenario of 400 agents, which naturally requires more rounds to complete the evacuation process. The progression between Round 40, shown in Figure~\ref{fig:evac:task3-400:5}, and Round 50, shown in Figure~\ref{fig:evac:task3-400:6}, reveals that as the agents notice the bottom exit is less congested, an increasing number begin to reroute from the left and right exits toward the more open bottom exit. This example serves to demonstrate the capability of our model in handling larger agent populations effectively.

\begin{figure*}[!t]
    \centering
    \begin{subfigure}{0.3\textwidth}
        \includegraphics[width=\linewidth]{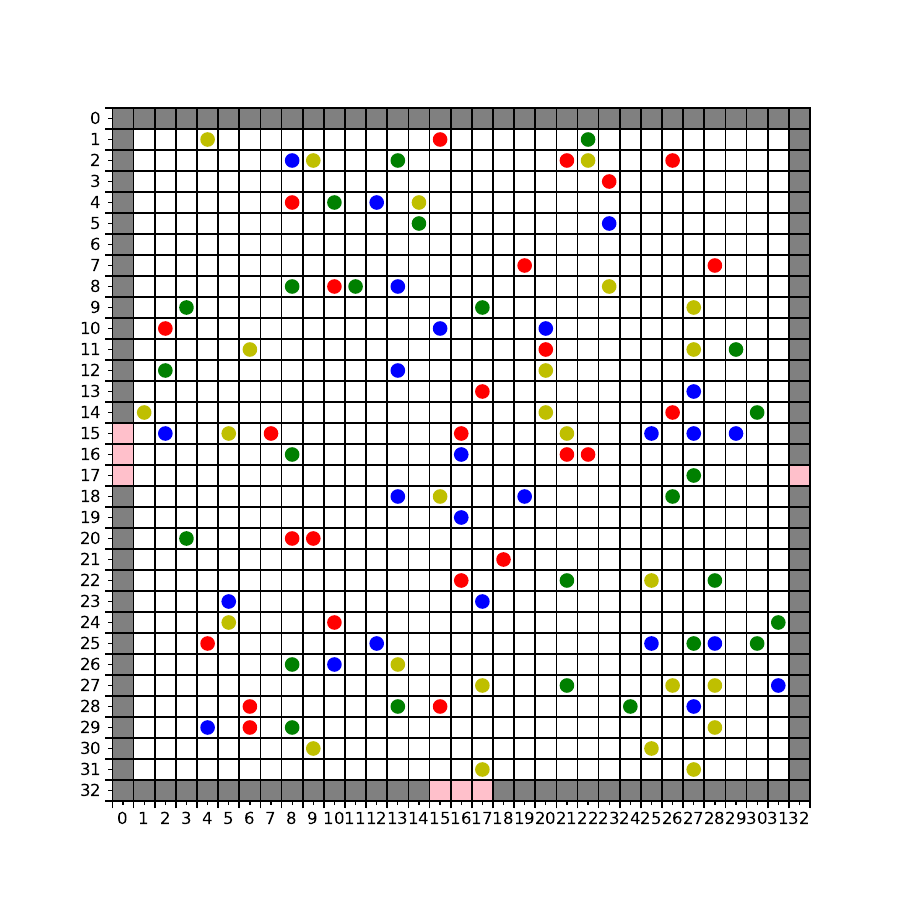}
        \caption{Round 0.}
        \label{fig:evac:task4:1}
    \end{subfigure}
    \begin{subfigure}{0.3\textwidth}
        \includegraphics[width=\linewidth]{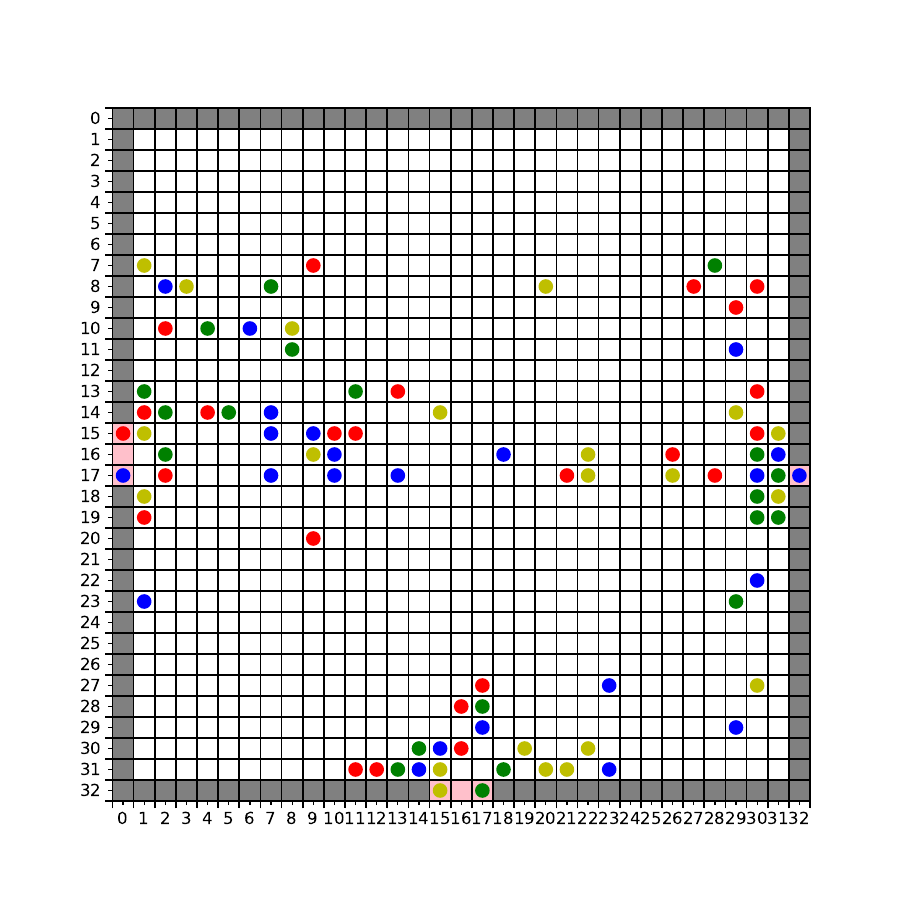}
        \caption{Round 6.}
        \label{fig:evac:task4:2}
    \end{subfigure}
    \begin{subfigure}{0.3\textwidth}
        \includegraphics[width=\linewidth]{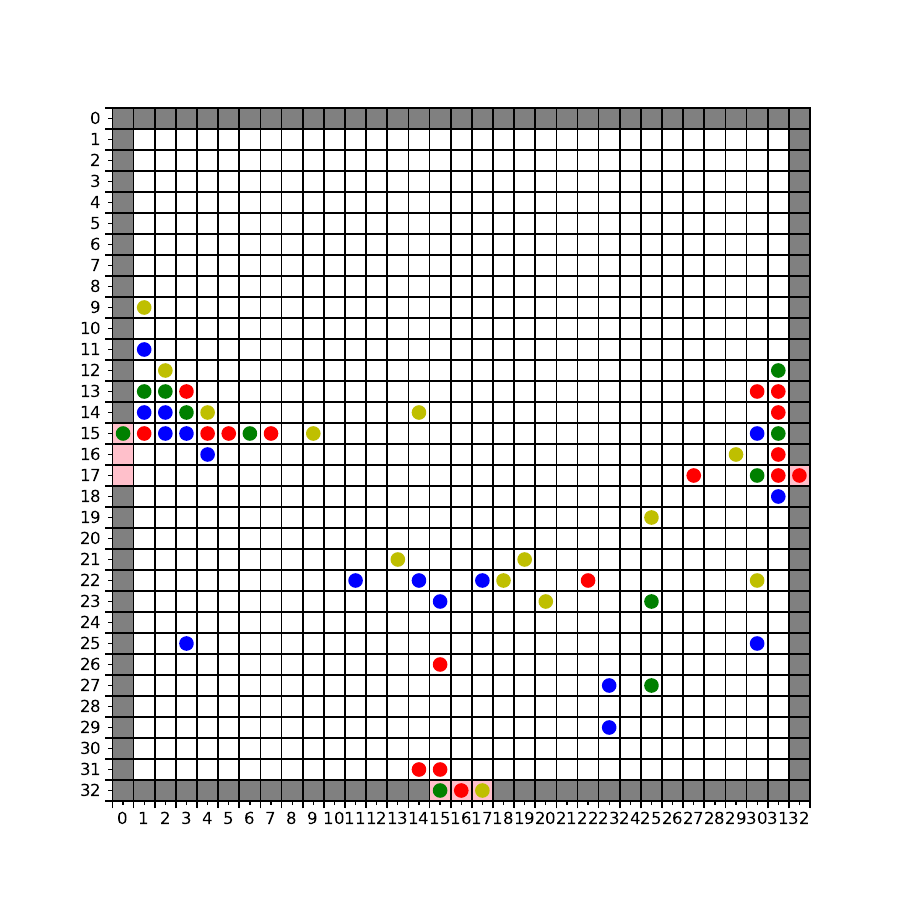}
        \caption{Round 12.}
        \label{fig:evac:task4:3}
    \end{subfigure}
    \begin{subfigure}{0.3\textwidth}
        \includegraphics[width=\linewidth]{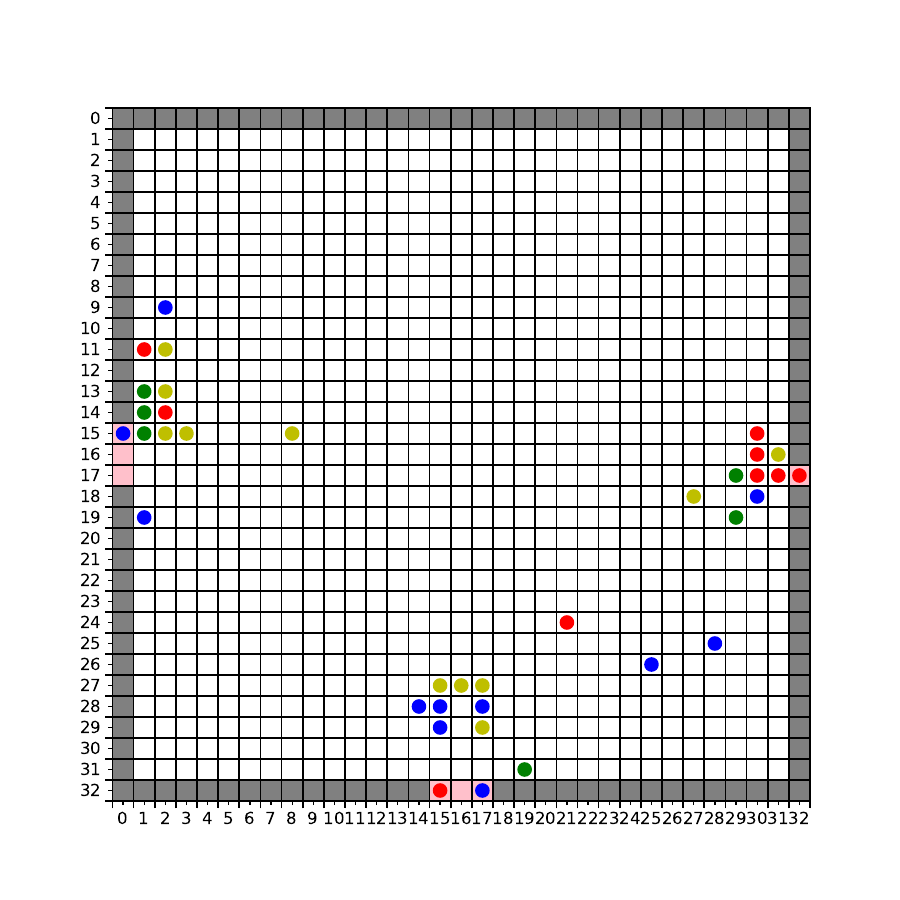}
        \caption{Round 18.}
        \label{fig:evac:task4:4}
    \end{subfigure}
    \begin{subfigure}{0.3\textwidth}
        \includegraphics[width=\linewidth]{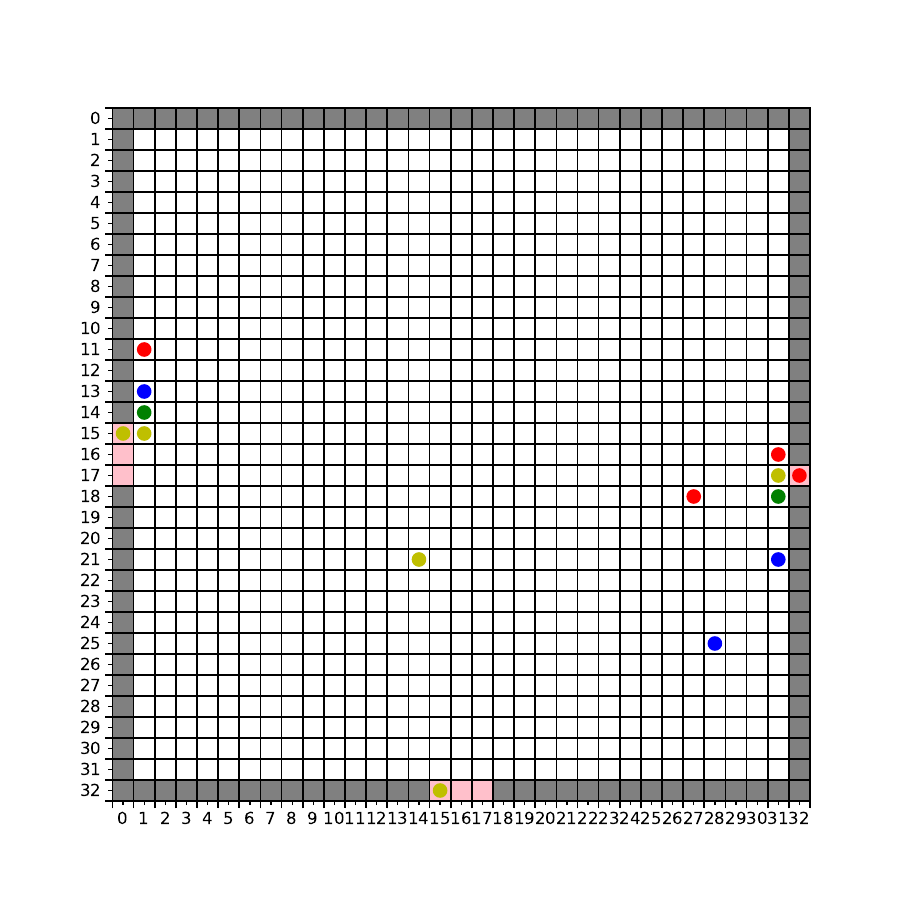}
        \caption{Round 24.}
        \label{fig:evac:task4:5}
    \end{subfigure}
    \begin{subfigure}{0.3\textwidth}
        \includegraphics[width=\linewidth]{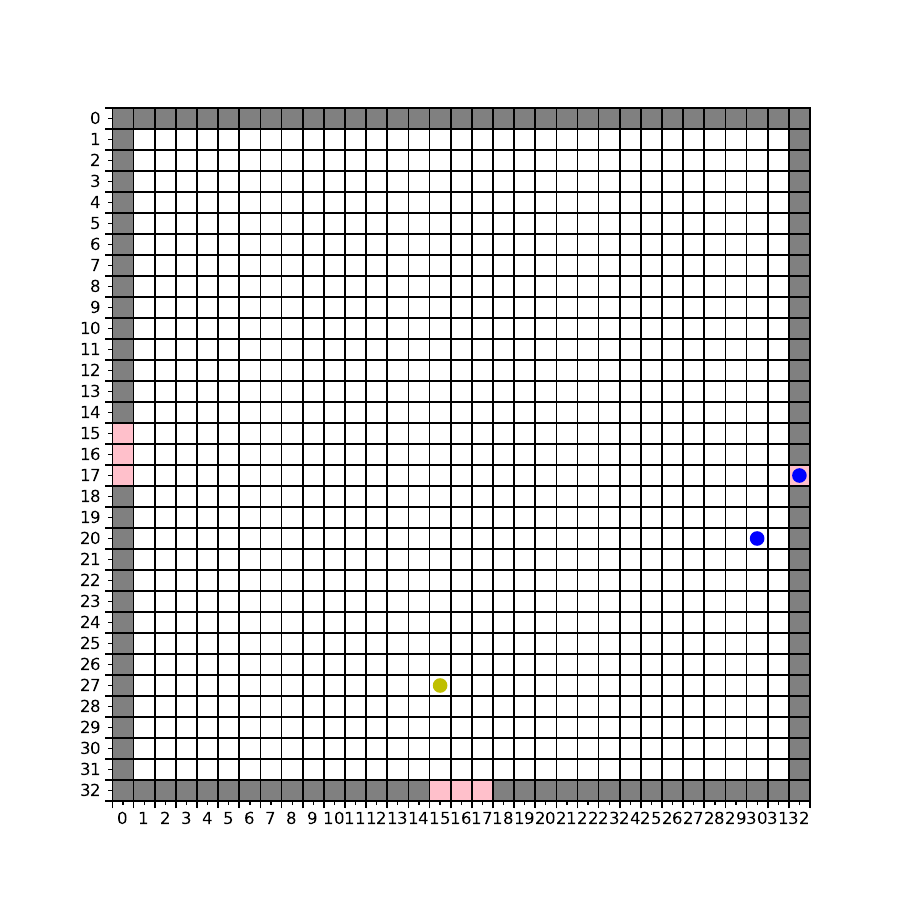}
        \caption{Round 30.}
        \label{fig:evac:task4:6}
    \end{subfigure}
    \caption{Sample run of Sub-task 4, sparse scenario (36 rounds in total). Colors represent personas: red for physically strong and mentally strong, green for physically strong but mentally weak, yellow for physically weak but mentally strong, and blue for physically weak and mentally weak.}
    \label{fig:evac:task4}
\end{figure*}


\begin{figure}[!t]
    \centering
    \includegraphics[width = \textwidth]{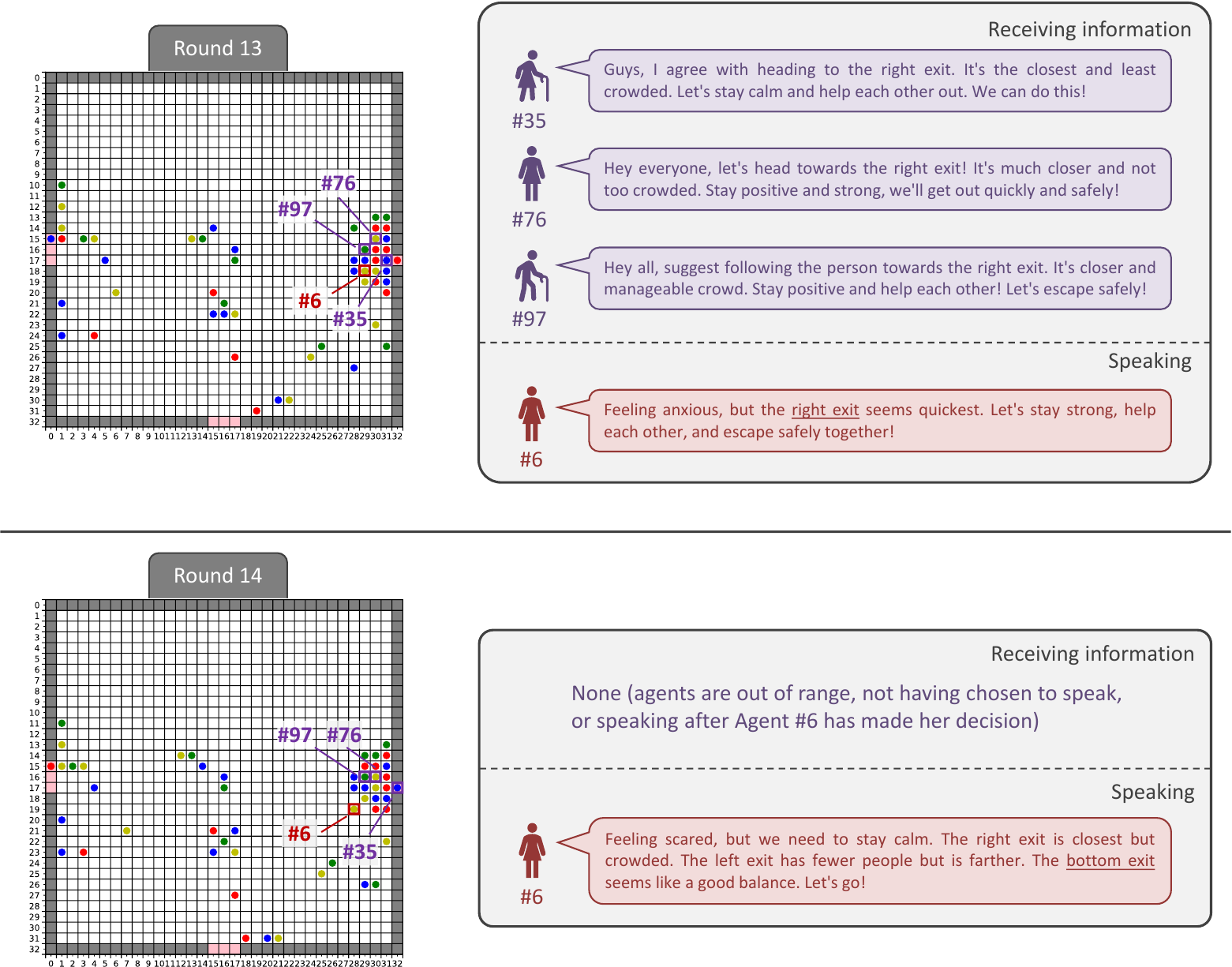}
    \caption{Sample conversations and its impact on choice of target exit (underlined).}
    \label{fig:evac:sample-conversations}
\end{figure}

\begin{figure*}[!t]
    \centering
    \begin{subfigure}{0.3\textwidth}
        \includegraphics[width=\linewidth]{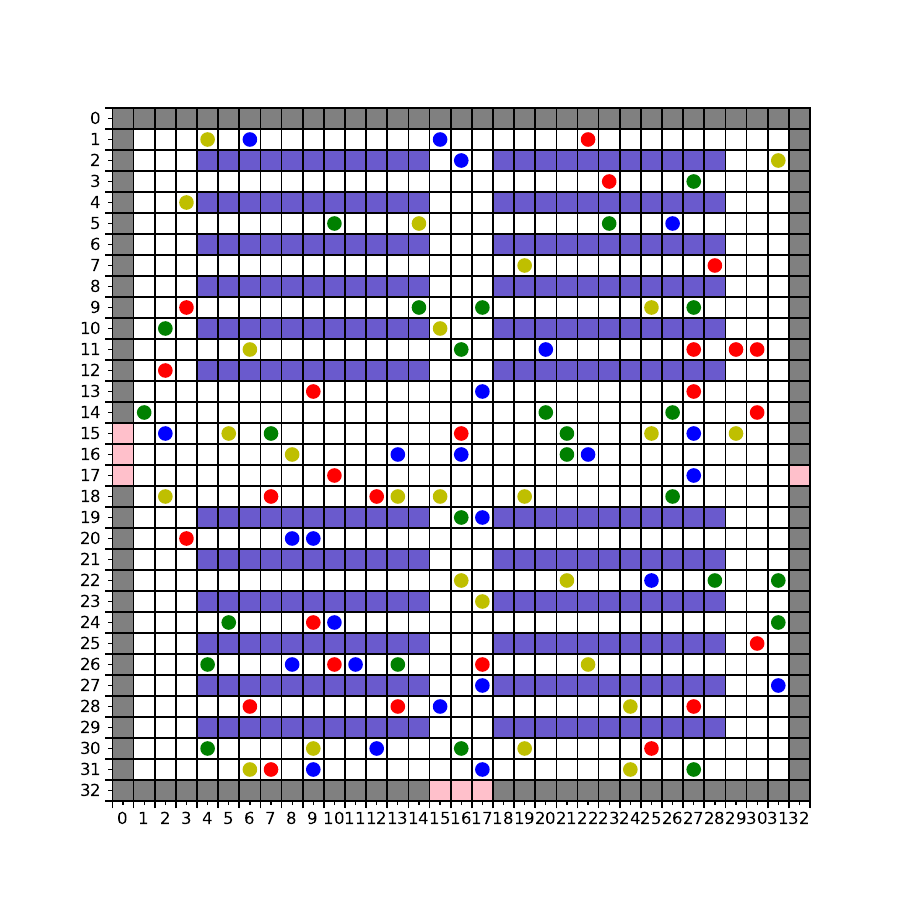}
        \caption{Round 0.}
        \label{fig:evac:task4-obs:1}
    \end{subfigure}
    \begin{subfigure}{0.3\textwidth}
        \includegraphics[width=\linewidth]{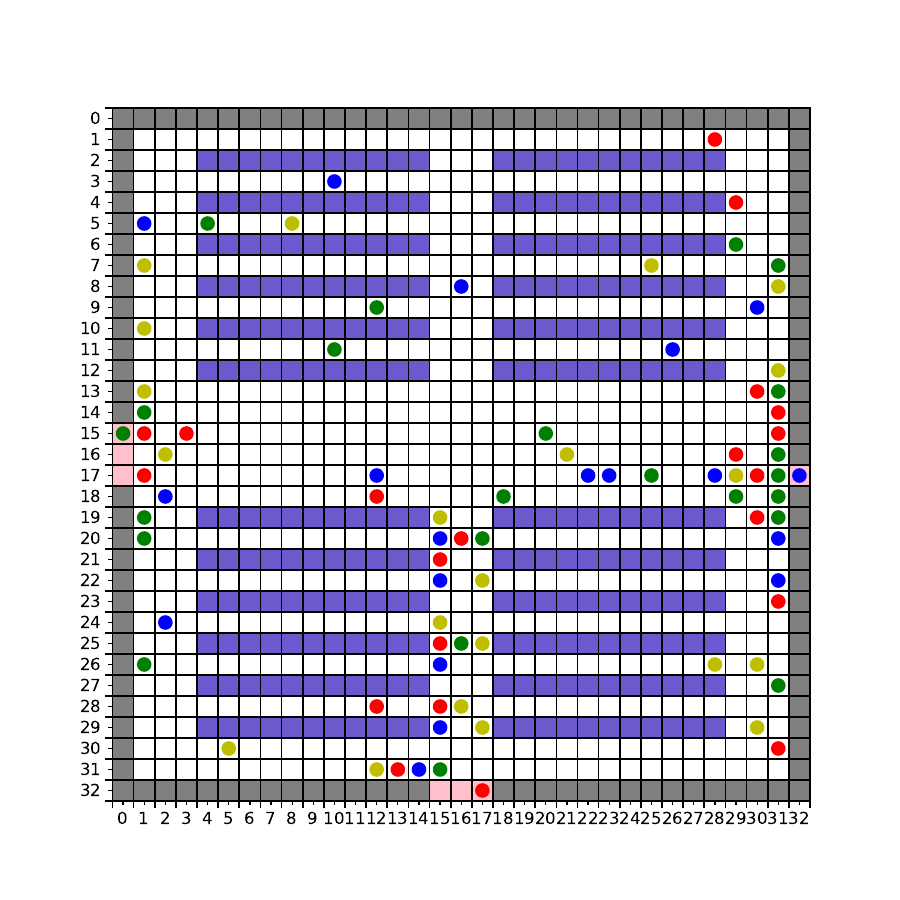}
        \caption{Round 6.}
        \label{fig:evac:task4-obs:2}
    \end{subfigure}
    \begin{subfigure}{0.3\textwidth}
        \includegraphics[width=\linewidth]{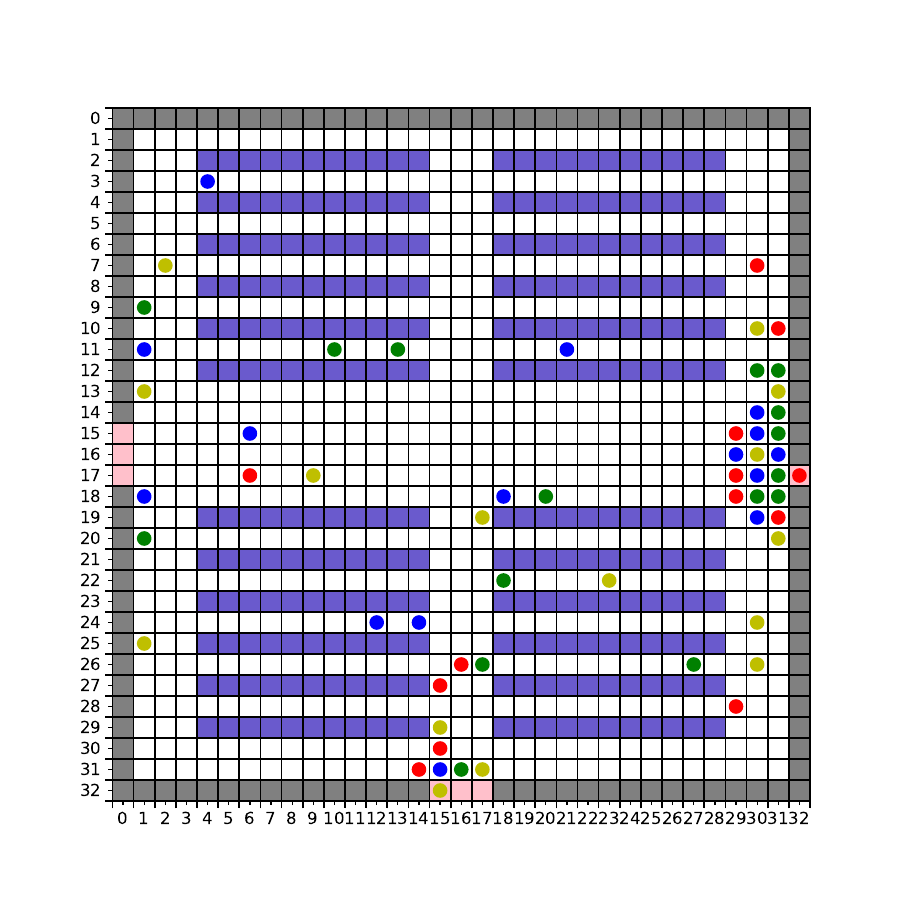}
        \caption{Round 12.}
        \label{fig:evac:task4-obs:3}
    \end{subfigure}
    \begin{subfigure}{0.3\textwidth}
        \includegraphics[width=\linewidth]{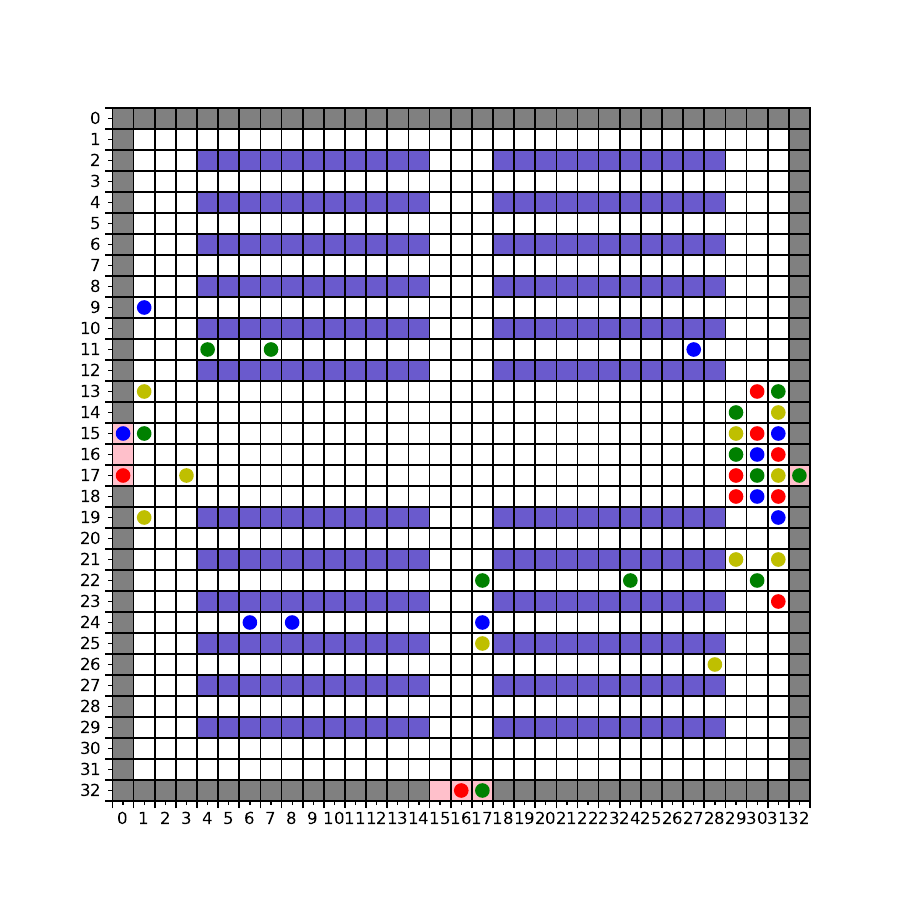}
        \caption{Round 18.}
        \label{fig:evac:task4-obs:4}
    \end{subfigure}
    \begin{subfigure}{0.3\textwidth}
        \includegraphics[width=\linewidth]{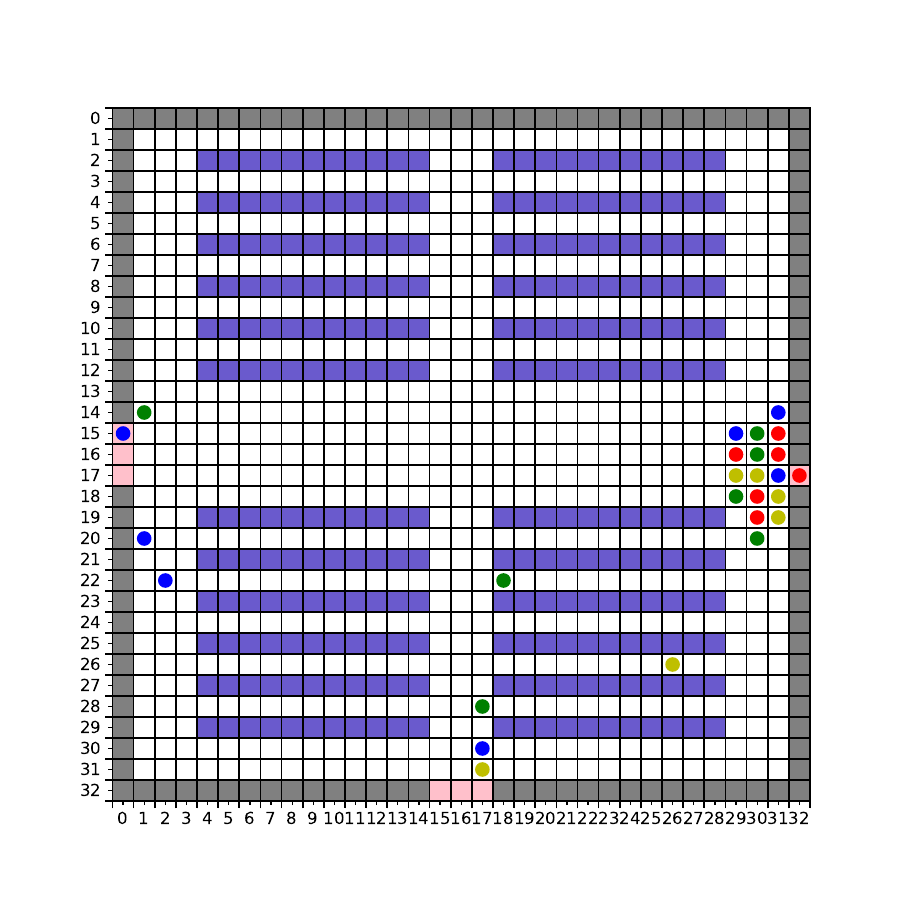}
        \caption{Round 24.}
        \label{fig:evac:task4-obs:5}
    \end{subfigure}
    \begin{subfigure}{0.3\textwidth}
        \includegraphics[width=\linewidth]{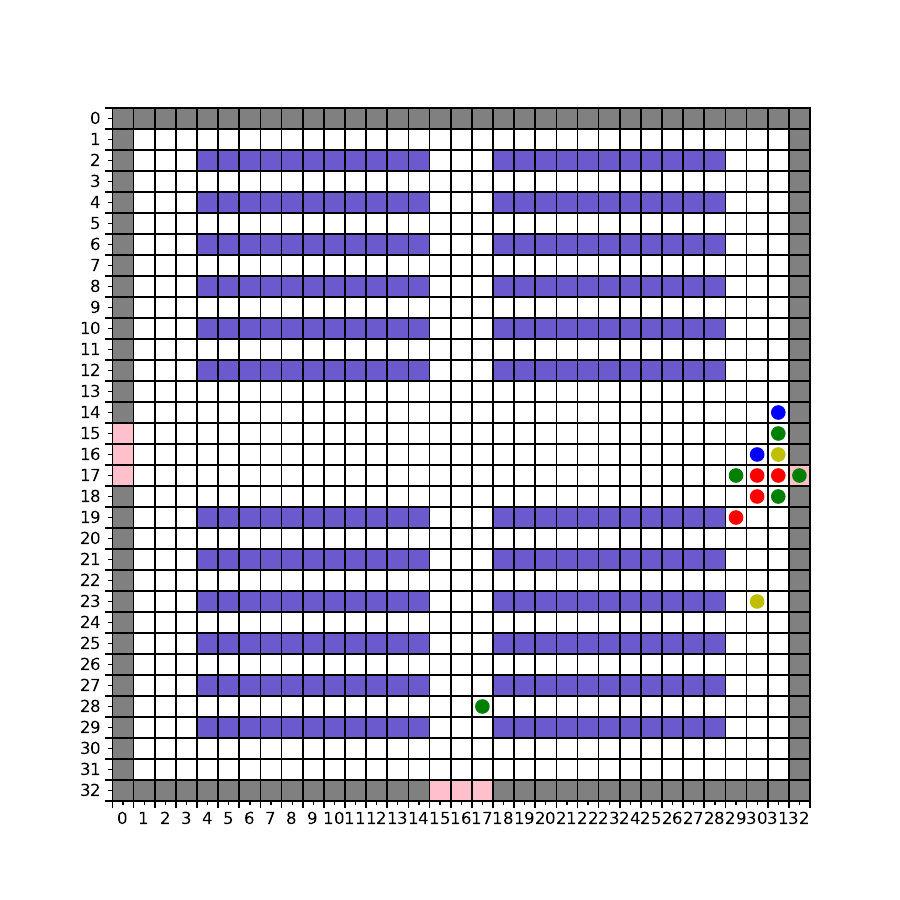}
        \caption{Round 30.}
        \label{fig:evac:task4-obs:6}
    \end{subfigure}
    \caption{Sample run of Sub-task 4, obstacle scenario (42 rounds in total). Colors represent personas: red for physically strong and mentally strong, green for physically strong but mentally weak, yellow for physically weak but mentally strong, and blue for physically weak and mentally weak.}
    \label{fig:evac:task4-obs}
\end{figure*}

\subsubsection{Sub-task 4: Test of Conversation}


We evaluate the impact of conversation on evacuation outcomes. Figure~\ref{fig:evac:task4} presents a sample run illustrating a distribution pattern similar to that of Sub-task 3, where agents disperse evenly across all exits. A comparison between the evacuation success rates by persona groups in Sub-task 3 and Sub-task 4, as depicted in Figures~\ref{fig:evac:escaped:3} and~\ref{fig:evac:escaped:4}, reveals a diminished disparity among the persona groups in Sub-task 4. Notably, the evacuation pace for the blue group, characterized by both physical and mental weaknesses, has noticeably increased. This improvement may be attributed to the beneficial effects of spontaneous collaboration. For example, dialogues such as, \str{Hey everyone, stay calm and positive! Let's head to the left exit; it's closer, and we can support each other. Together, we'll make a quick and safe escape!} and \str{Stay positive and strong, everyone! The left exit is nearer and less crowded. Let's evacuate swiftly, safely, and support one another.} This kind of dialogue content likely exerts a constructive impact on the agents' decision-making behavior.

Figure~\ref{fig:evac:sample-conversations} presents an intriguing instance of dialogue between agents in a sample run of Sub-task 4. At Round 13, Agent \#6 receives suggestions from three nearby agents, all advocating for the right exit. Feeling anxious, she too favors the right exit as her target. In her move, she attempts to avoid the crowd at this exit, eventually taking a step further away. By Round 14, she no longer receives input from other agents during her decision-making process, due to them being out of range, not having chosen to speak, or speaking after she has made her decision. Feeling scared and lacking information from others, she now prefers the bottom exit to avoid the congestion at the right exit. However, by manually calculating the factors in this situation, we find that the best choice for her is going to the right exit, despite the congestion and the risk of being overtaken. This example illustrates that the information shared among agents significantly influences their choice of exits, mirroring the behavior observed in real-world evacuations where individuals in panic often follow the lead of others.




Figure~\ref{fig:evac:task4-obs} presents the obstacle scenario of Sub-task 4. In its ABM counterpart~\cite{wang2015modeling}, it is reported that the presence of obstacles may lead to orderly evacuation by directing the flow of agents, hence resulting in faster evacuations. Contrary to this outcome, the expedited evacuation process is not observed here. The reason is that obstacles appear to impede the agents' ability to migrate from a congested exit to alternate exits. As depicted in Figure~\ref{fig:evac:task4-obs}, the right exit exhibits greater congestion compared to the scenario without obstacles throughout Sub-tasks 2 -- 4. 

\begin{figure*}[!t]
    \centering
    \begin{subfigure}{0.45\textwidth}
        \includegraphics[width=\linewidth]{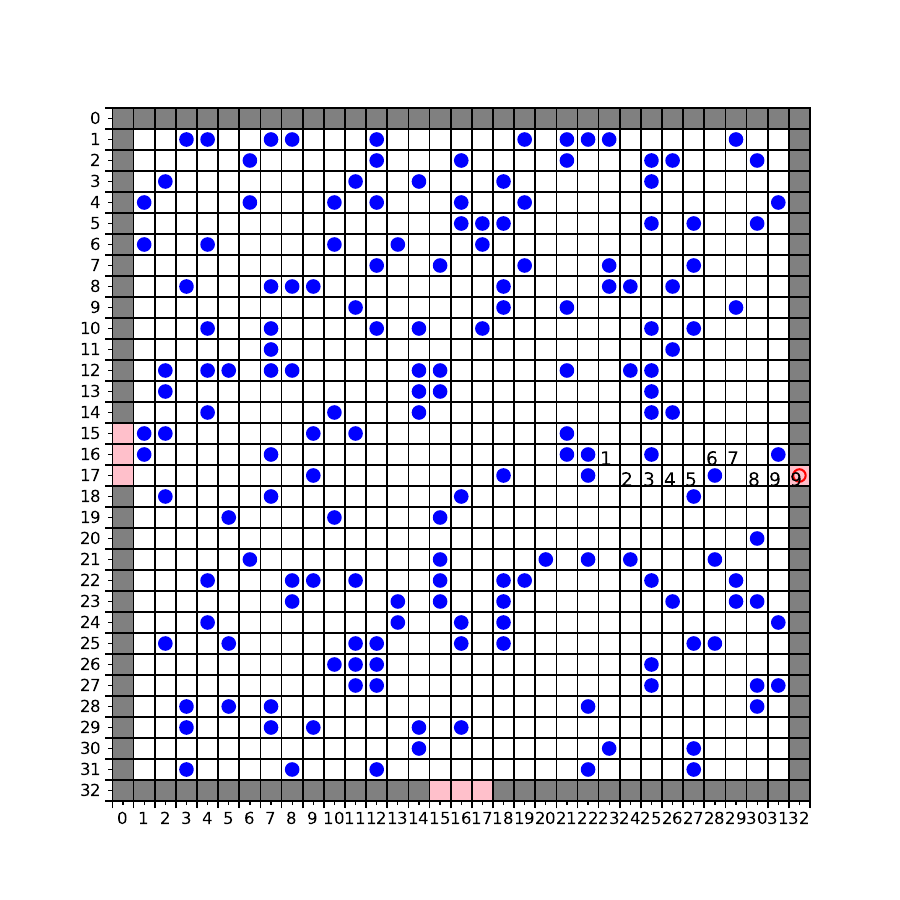}
        \caption{Original evacuation route.}
        \label{fig:evac:validation:1}
    \end{subfigure}
    \hspace{4ex}
    \begin{subfigure}{0.45\textwidth}
        \includegraphics[width=\linewidth]{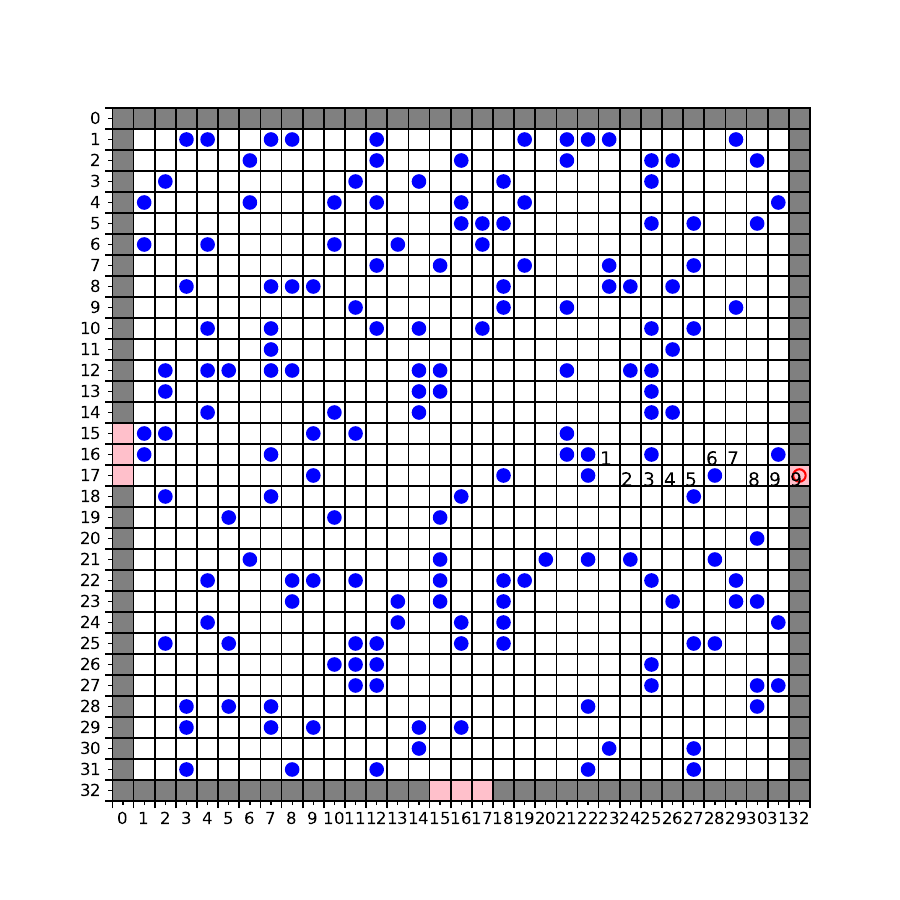}
        \caption{Low level of prompt variation (paraphrasing).}
        \label{fig:evac:validation:2}
    \end{subfigure}
    \begin{subfigure}{0.45\textwidth}
        \includegraphics[width=\linewidth]{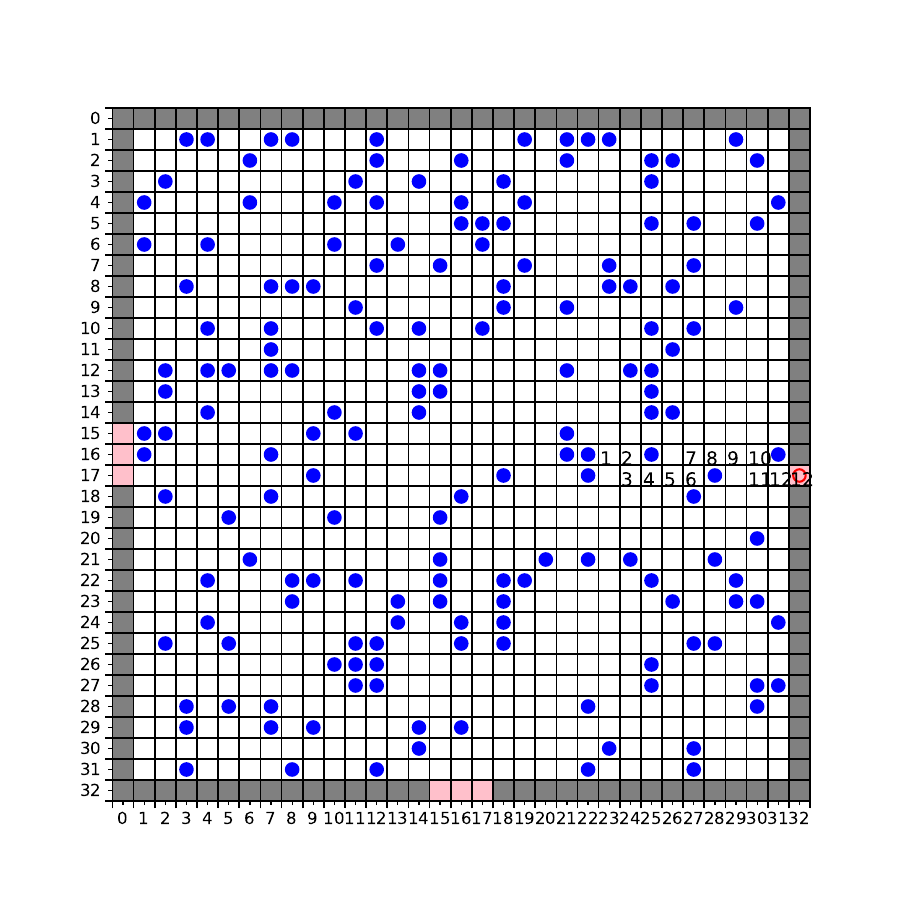}
        \caption{Medium level of prompt variation (varying elements).}
        \label{fig:evac:validation:3}
    \end{subfigure}
    \hspace{4ex}
    \begin{subfigure}{0.45\textwidth}
        \includegraphics[width=\linewidth]{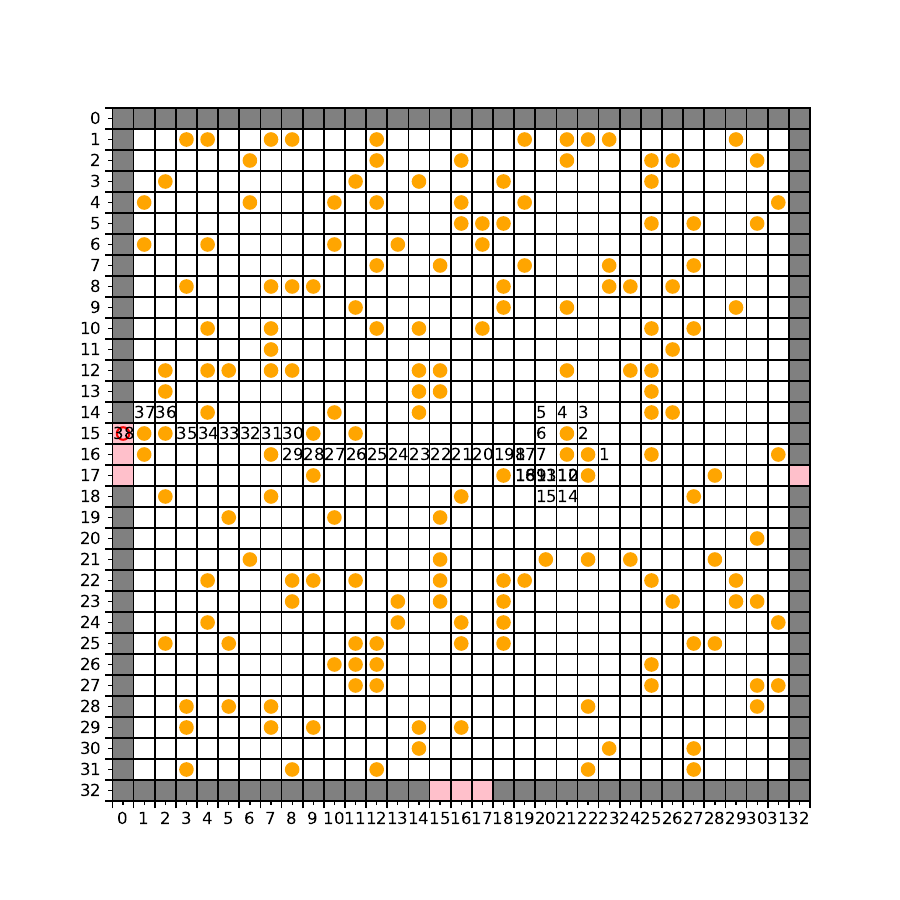}
        \caption{High level of prompt variation (varying objectives). Orange points represent  paintings.}
        \label{fig:evac:validation:4}
    \end{subfigure}
    \caption{Sample runs of model validation. The numbers in cells indicate the movement path.}
    \label{fig:evac:validations}
\end{figure*}

\subsection{Model Validation}
We validate the prompts employed in this case study following the methodology for model validation presented in Section~\ref{sec:design:valid}. Similar to the practice of single-agent debugging, validating the behavior of individual agents enables a meticulous analysis of the model. Below, we present observations from varying the prompts within the context of this case study.

\begin{itemize}
    \item \textbf{Extent of Variation: Low}
    
    \textbf{Modified Prompt:} \textit{(Paraphrased evacuation scenario)} Due to the earthquake, it is imperative that you quickly vacate the room you are currently in. The room measures 33 by 33 units and offers three potential exits situated on the left, bottom, and right sides of the room. To successfully escape from this room, you should take into account two factors: the proximity of the nearest exit and the number of people present. The exit proximity refers to the distance between your current location and the closest exit, denoted as \var{distance\_to\_nearest\_exit}. Additionally, within your line of sight, there are a total of \var{number\_of\_people} individuals.
    
    \textbf{Original Prompt:} \textit{(Evacuation scenario)} Because of the earthquake, you need to escape from the room where you are as fast as possible. The room has a size of 33 * 33. There are 3 exits in the room. The exits are located at the left, bottom, and right of the room. To escape from the room, you need to consider the following two aspects: exit proximity and people count. The exit proximity is the distance between you and the nearest exit. The people count is the number of people you can see. The distance to the nearest exit is \var{distance\_to\_nearest\_exit}. There are \var{number\_of\_people} people in your visible range.
    
    \textbf{Results and Analysis:} There are no significant differences in the results (see Figures~\ref{fig:evac:validation:1} --~\ref{fig:evac:validation:2}), showing the model is not sensitive to the paraphrasing of the evacuation scenario. 
    
    \item \textbf{Extent of Variation: Medium}
    
    \textbf{Modified Prompt:} \textit{(Varying elements in Stage 4: Decision on movement)} Select your move from these possible options \highlightgreen{(You can move in diagonal directions (up, down, left, right)}, options with obstacles or other people are excluded and not in the path, and option codes are in random order).
    
    \textbf{Original Prompt:} \textit{(Stage 4: Decision on movement)} Select your move from these possible options \highlightred{(you can move in diagonal or horizontal directions}, options with obstacles or other people are excluded and not in the path, and option codes are in random order).
    
    \textbf{Results and Analysis:} We modify the prompt concerning the possible movement directions for the agent, restricting them to just four cardinal directions (up, down, left, right) or to remain stationary, as opposed to the original eight directions. Figure~\ref{fig:evac:validation:3} displays a sample run with these modifications. It is apparent that due to the limitation in movement options, namely the absence of diagonal movements for expedited evacuation, the number of turns required to escape has increased. By conducting 100 runs for both the original and modified prompts, we confirm that the scenario constrained to four directions of movement necessitates significantly more turns for evacuation than the one with eight directions ($p < 0.001$ using the Mann-Whitney U test).
    
    \item \textbf{Extent of Variation: High}
    
    \textbf{Modified Prompt:} \textit{Varying objectives in evacuation scenario)} \highlightgreen{You are in an art gallery looking at paintings and you want to take your time to walk around the gallery and see different paintings everywhere in the room before you leave.} The \highlightgreen{gallery} has a size of 33 * 33. There are 3 exits in the room. The exits are located at the left, bottom, and right of the \highlightgreen{gallery}.
    
    \textbf{Original Prompt:} \textit{(Evacuation scenario)} \highlightred{Because of the earthquake, you need to escape from the room where you are as fast as possible.} The \highlightred{room} has a size of 33 * 33. There are 3 exits in the room. \highlightred{The exits are located at the left, bottom, and right of the room. To escape from the room, you need to consider the following two aspects: exit proximity and people count. ...}
    
    \textbf{Results and Analysis:} We modify the simulation's objective, altering the agent's objective from a rapid evacuation in an emergency to leisurely touring a gallery. Figure~\ref{fig:evac:validation:4} illustrates that the agent spends a considerable number of rounds moving and selects the exit that is farthest away, aligning with the newly adapted objective. Notably, the agent lingers within the same area for the initial 20 steps, echoing the instruction in the prompt to \str{take your time to walk around.} These observations affirm the overall reliability of the prompt. Moreover, it is intriguing to note the agents' expressions of subjective feelings, remarking, \str{I feel captivated and inspired by the diverse array of paintings surrounding me, each one telling a unique story and evoking a range of emotions.}
    
\end{itemize}

\myparagraph{Remarks}
Due to challenges in gathering data from real-world evacuations, computer simulations are preferred for studying evacuee dynamics. A key aspect of our case study is the role of shared information in decision-making. In contrast to the leader-follower paradigm in the ABM simulation~\cite{wang2015modeling}, our approach allows agents to communicate freely. This is grounded in a more fundamental understanding of human behavior.

Owing to the substantial financial and computational cost of GPT-4, we are limited to only five runs for each simulation scenario in this study. To affirm the reproducibility of our findings, further runs would be beneficial. Nevertheless, our simulation has revealed intriguing emergent behaviors, such as spontaneous collaboration where agents encourage others and suggest exit strategies, thereby speeding up the evacuation and balancing the pace across different persona groups.

Contrary to findings in \cite{wang2015modeling}, we did not observe instances of disabled agents in our simulation, which can be attributed to the faster and more balanced evacuation achieved under the SABM approach. Modifying the parameters that govern local congestion -- currently modeled by ABM -- might increase the likelihood of encountering disabled agents. Future work could include a sensitivity analysis of these parameters. Another avenue for exploration is enhancing the memory of agents to facilitate decision-making with greater consistency. Besides, more comprehensive debugging of agents' behavior may help enhance the realism of this simulation.

\section{Case Study 2: Plea Bargaining}
\label{sec:case-plea}
This case study demonstrates single-agent modeling~\footnote{Strictly speaking, this case study is not an SABM instance because there are no interactions.}. Our aims is to simulate complex psychology, in other words, soft factors, which are difficult to quantify, calibrate, and sometimes justify using ABM~\cite{bonabeau2002agent}. To this end, we choose plea bargaining as the task. 

\subsection{Task Definition}
A plea bargain refers to an agreement between a prosecutor and a defendant in which the prosecutor offers a concession to the defendant who, in turn, may opt to plead guilty to an offense for a reduction in criminal charges. Plea bargaining plays a pivotal role in the criminal justice system of the United States, with scholars estimating that an overwhelming majority (90 -- 95\%) of criminal cases in the United States are resolved through plea bargains rather than trials~\cite{devers2011plea}. We employ SABM to simulate defendants assess their willingness to accept plea offers (WTAP). We will show that by using the agent modeling methods introduced in Section~\ref{sec:impl}, the agents can be calibrated to behave more closely to humans.

Regarding the WTAP of human defendants, a previous study~\cite{tor2010fairness} recruited volunteers~\footnote{There are five experimental studies in \cite{tor2010fairness}, each recruiting 30 to 120 undergraduate students as volunteers.} to take part in questionnaire surveys featuring hypothetical crime scenarios and play defendants. The study revealed that defendants' preferences and judgments are significantly influenced by their sense of fairness. Specifically, even if they significantly benefit from accepting the offers, defendants who perceive themselves as innocent (referred to as \group{innocent} participants) are less likely to accept plea offers than those who view themselves as guilty (referred to as \group{guilty} participants). Moreover, all defendants tend to reject offers that appear comparatively unfair (i.e., inconsistent with offers for similar cases). Furthermore, defendants who are uncertain of their culpability (referred to as \group{uncertain} participants) demonstrate egocentrically-biased judgments and act as if innocent, rejecting plea offers they otherwise would have accepted~\footnote{Unbiased uncertain defendants would express a fairness-driven preference that reflects culpability judged with available information, thereby exhibiting generally higher WTAP than innocents~\cite{tor2010fairness}.}.


\begin{figure}[!t]
    \center{\includegraphics[width = \textwidth]{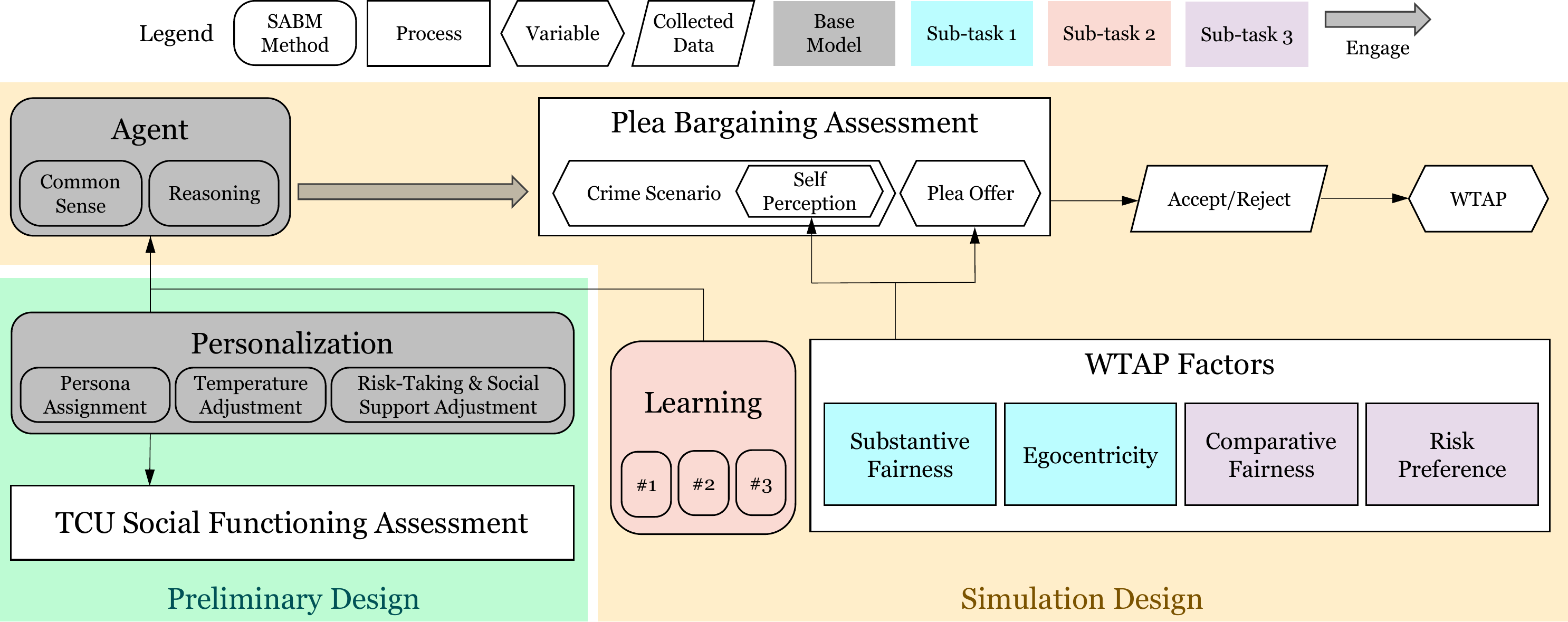}}
    \caption{Design of plea bargaining.}
    \label{fig:plea:design}
\end{figure}

Figure~\ref{fig:plea:design} shows the design of this case study. It features a preliminary design, in which the agents participate in a social functioning assessment for personalities and the results are calibrated to human data via personalization. Then, in the simulation design, the agents are instructed to take part in a plea bargaining assessment and answer whether they accept or reject the plea offer, based on common sense and reasoning. We collect the answers and report the WTAP. We evaluate the following four factors, whose impacts on WTAP have been studied in \cite{tor2010fairness}: 
\begin{itemize} 
  \item Substantive fairness: whether the defendant feel he/she is guilty when receiving the plea offer. 
  \item Egocentricity: how the defendant would respond to the offer when he/she is uncertain of culpability. 
  \item Comparative fairness: how the plea offer is compared to typical ones. 
  \item Risk preference: the conviction probability that affects the defendant's preference in plea bargain decision making. 
\end{itemize}
The evaluation of the four factors are divided into two sub-tasks, and we have a sub-task between them for learning-based calibration, featuring three few-shot examples.  

\begin{figure*}[!t]
  \centering
  \begin{subfigure}{0.18\textwidth}
    \includegraphics[width=\linewidth]{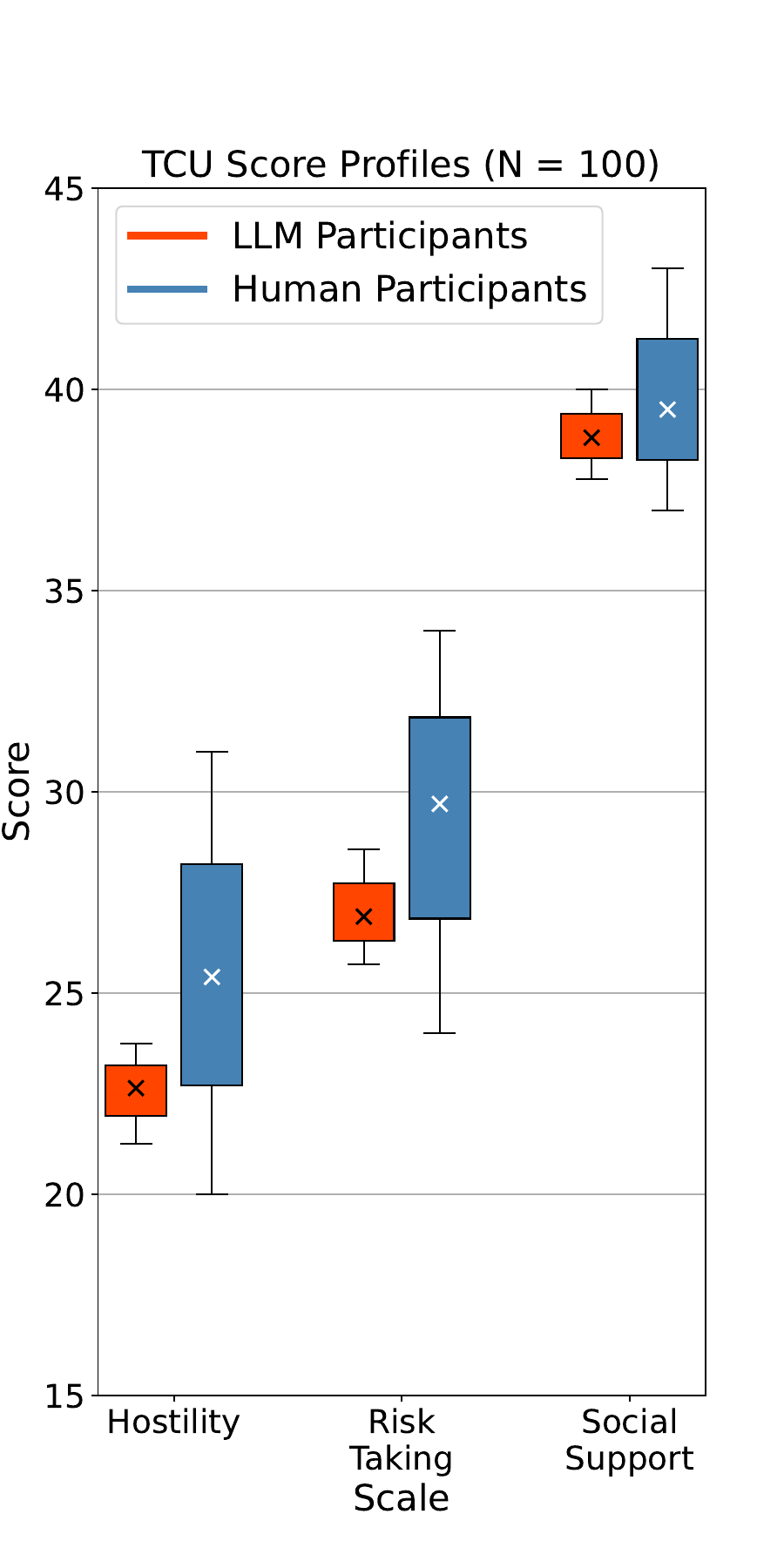}
    \caption{\scriptsize{GPT-4.}}
    \label{fig:plea:tcu:gpt-4}
  \end{subfigure}
  \begin{subfigure}{0.18\textwidth}
    \includegraphics[width=\linewidth]{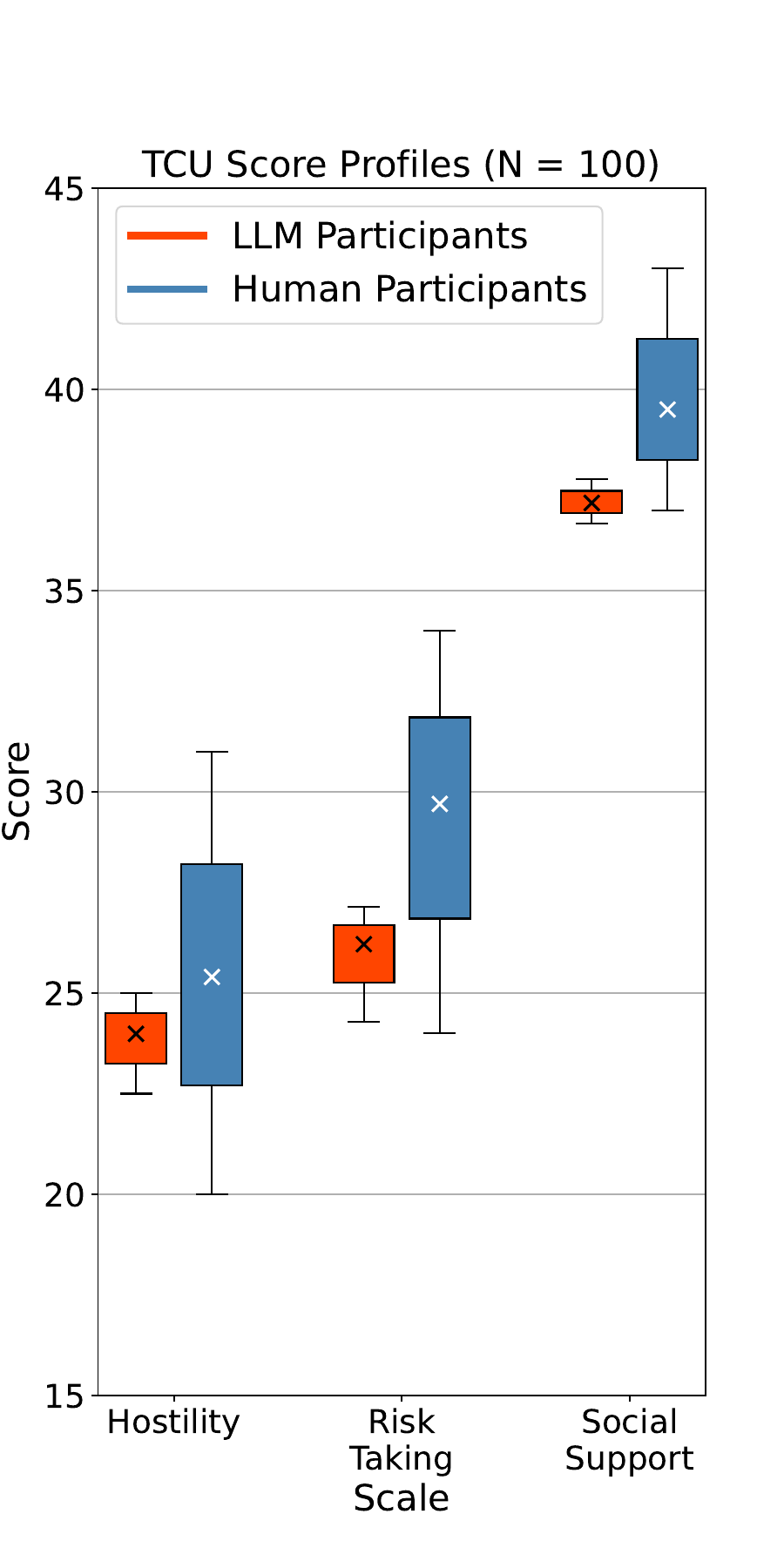}
    \caption{\scriptsize{GPT-4+\textsf{P}.}}
    \label{fig:plea:tcu:gpt-4+p}
  \end{subfigure}
  \begin{subfigure}{0.18\textwidth}
    \includegraphics[width=\linewidth]{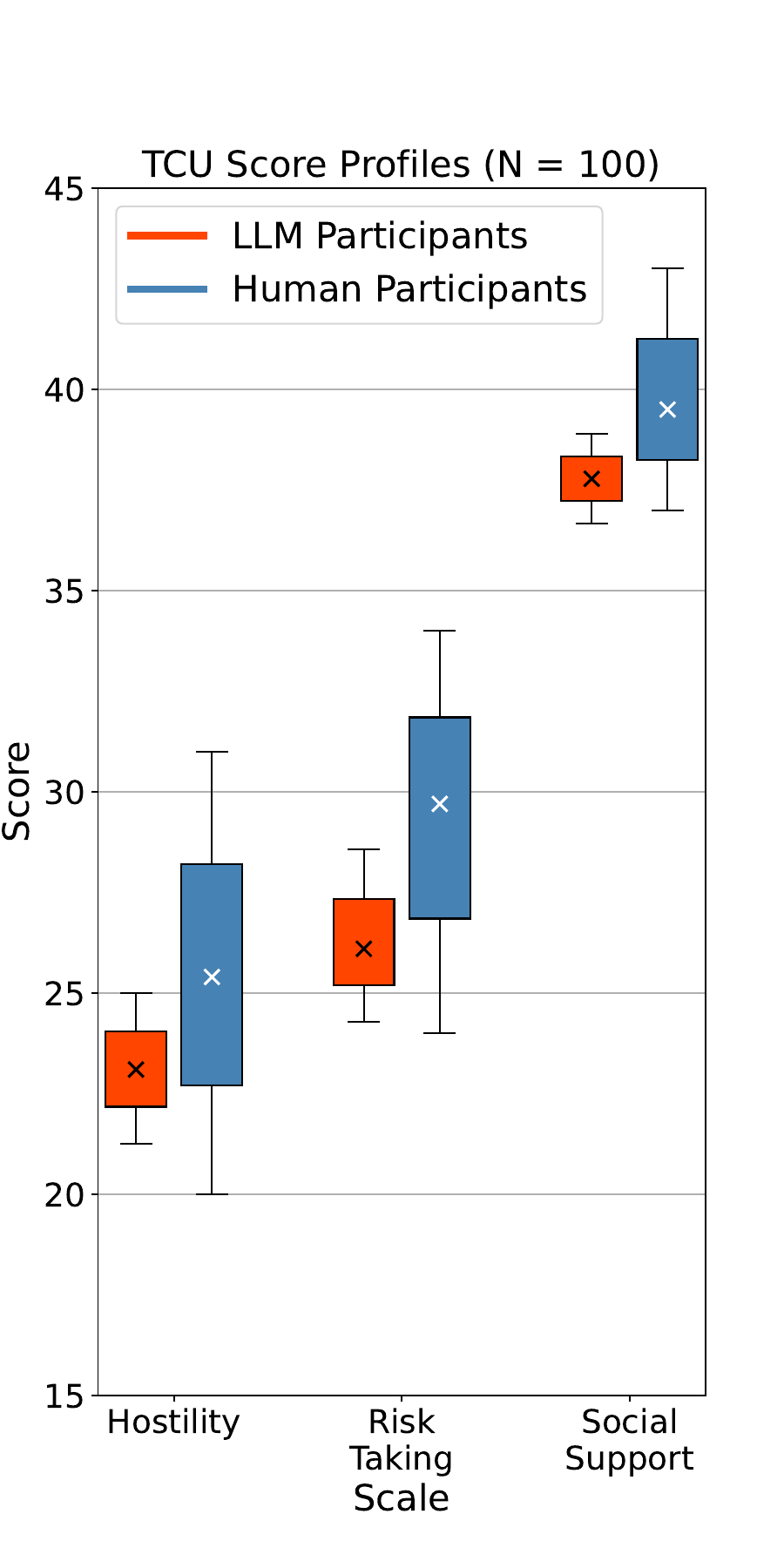}
    \caption{\scriptsize{GPT-4+\textsf{PT}.}}
    \label{fig:plea:tcu:gpt-4+p+tn}
  \end{subfigure}
  \begin{subfigure}{0.18\textwidth}
    \includegraphics[width=\linewidth]{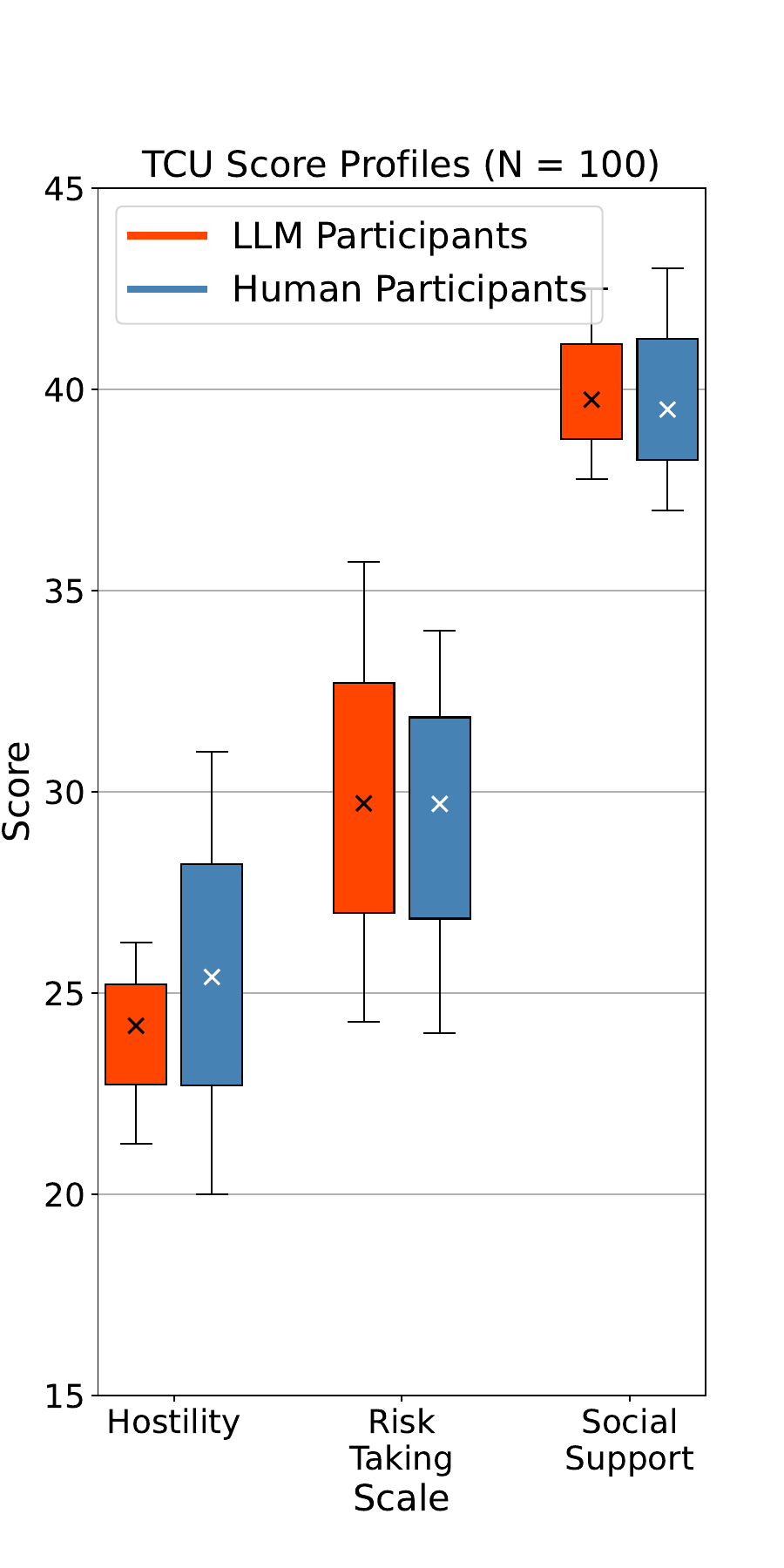}
    \caption{\scriptsize{GPT-4+\textsf{PTR}.}}
    \label{fig:plea:tcu:gpt-4+p+tn+r}
  \end{subfigure}
  \begin{subfigure}{0.18\textwidth}
    \includegraphics[width=\linewidth]{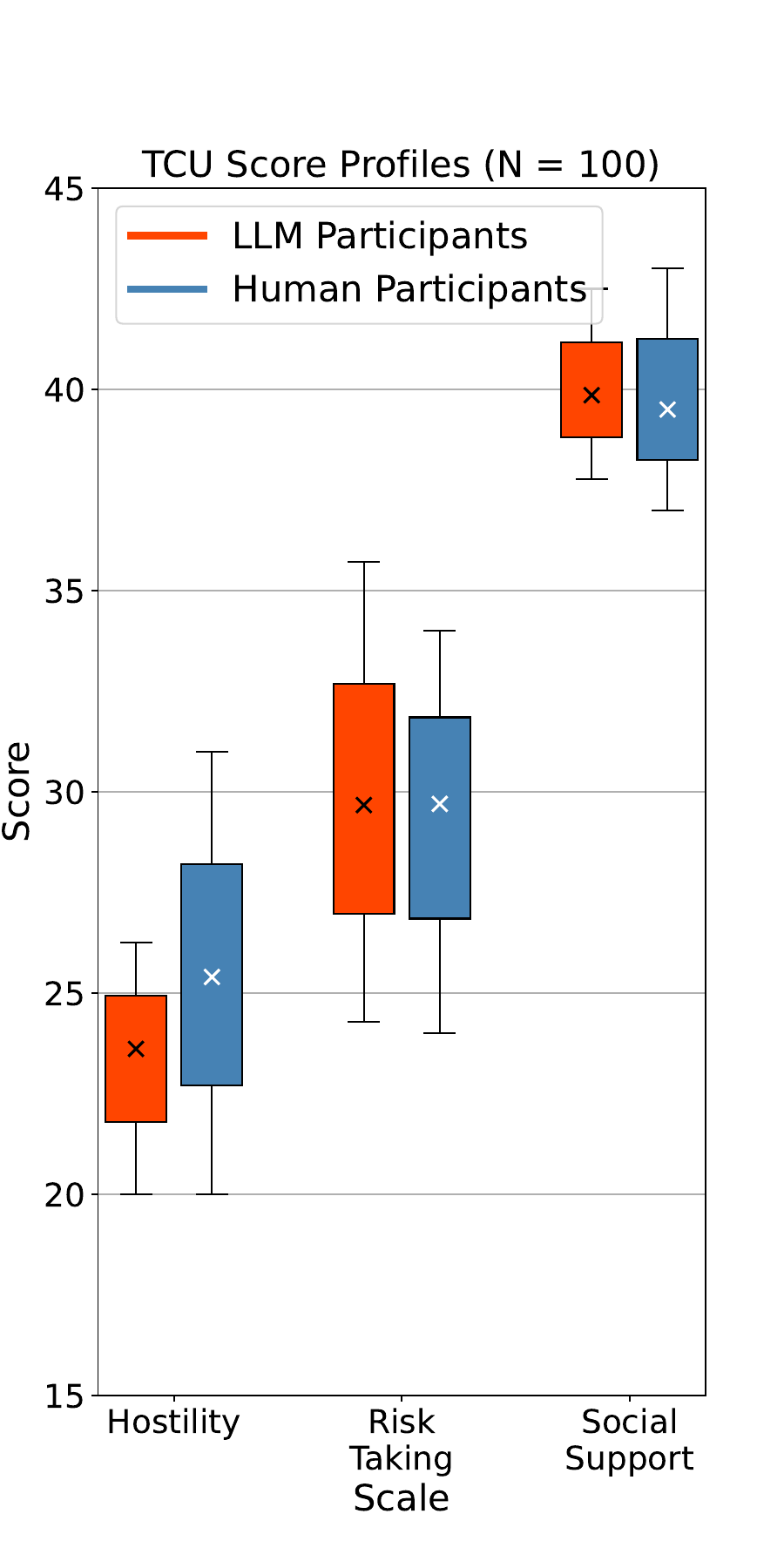}
    \caption{\scriptsize{GPT-4+\textsf{PTRL}.}}
    \label{fig:plea:tcu:gpt-4+p+tn+r+fs}
  \end{subfigure}
  \begin{subfigure}{0.18\textwidth}
    \includegraphics[width=\linewidth]{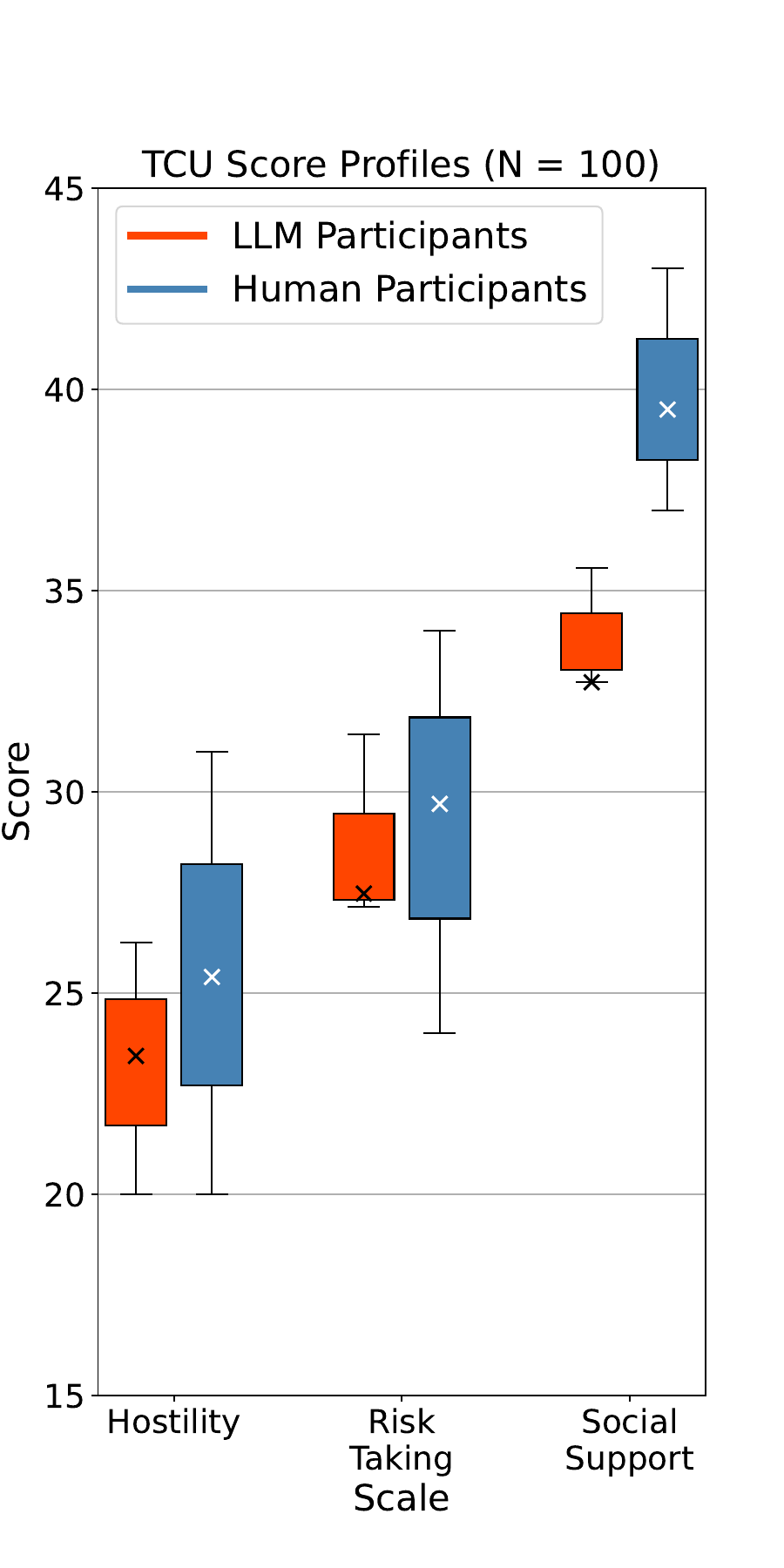}
    \caption{\scriptsize{GPT-3.5.}}
    \label{fig:plea:tcu:gpt-3.5}
  \end{subfigure}
  \begin{subfigure}{0.18\textwidth}
    \includegraphics[width=\linewidth]{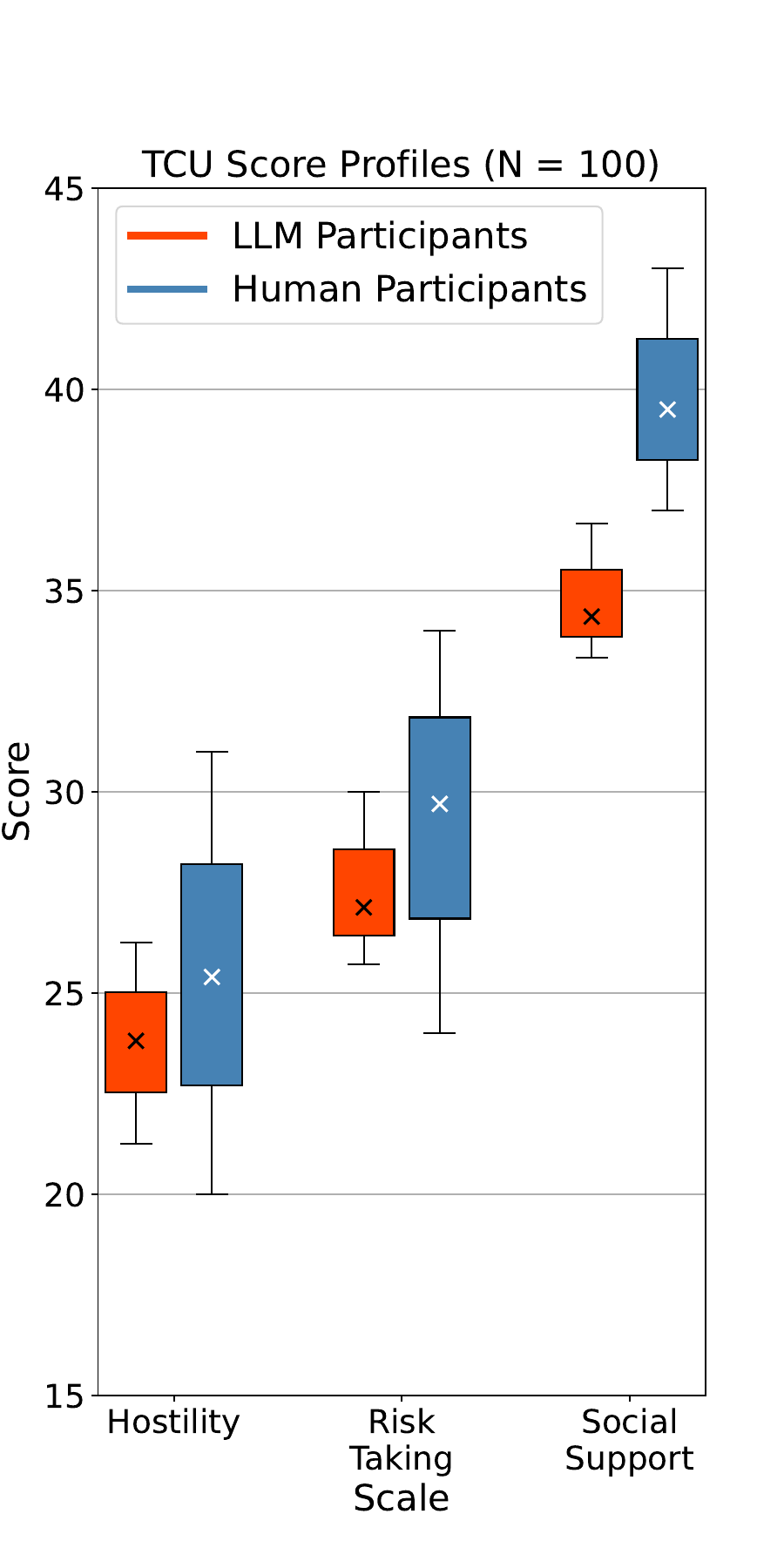}
    \caption{\scriptsize{GPT-3.5+\textsf{P}.}}
    \label{fig:plea:tcu:gpt-3.5+p}
  \end{subfigure}
  \begin{subfigure}{0.18\textwidth}
    \includegraphics[width=\linewidth]{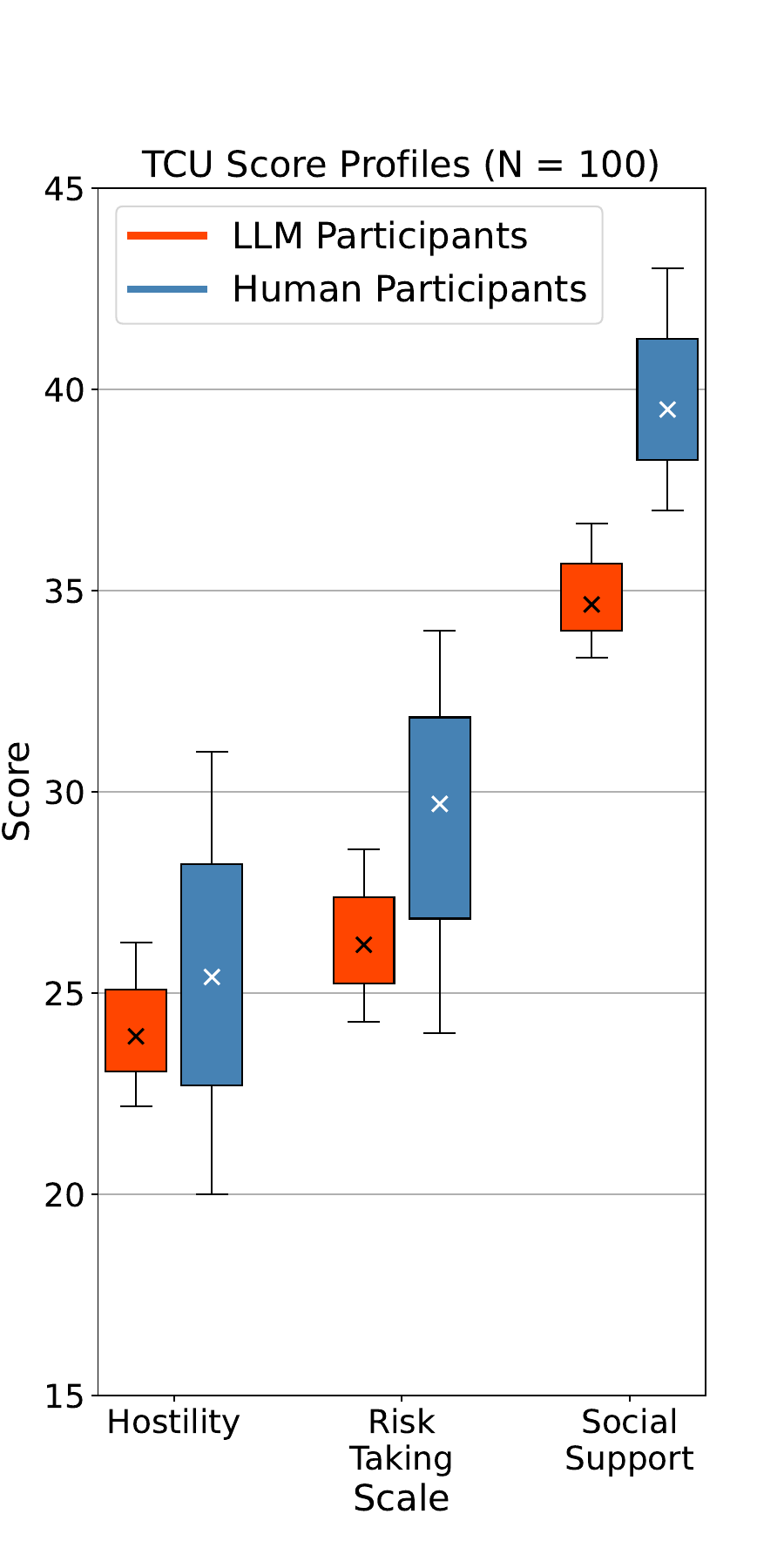}
    \caption{\scriptsize{GPT-3.5+\textsf{PT}.}}
    \label{fig:plea:tcu:gpt-3.5+p+tn}
  \end{subfigure}
  \begin{subfigure}{0.18\textwidth}
    \includegraphics[width=\linewidth]{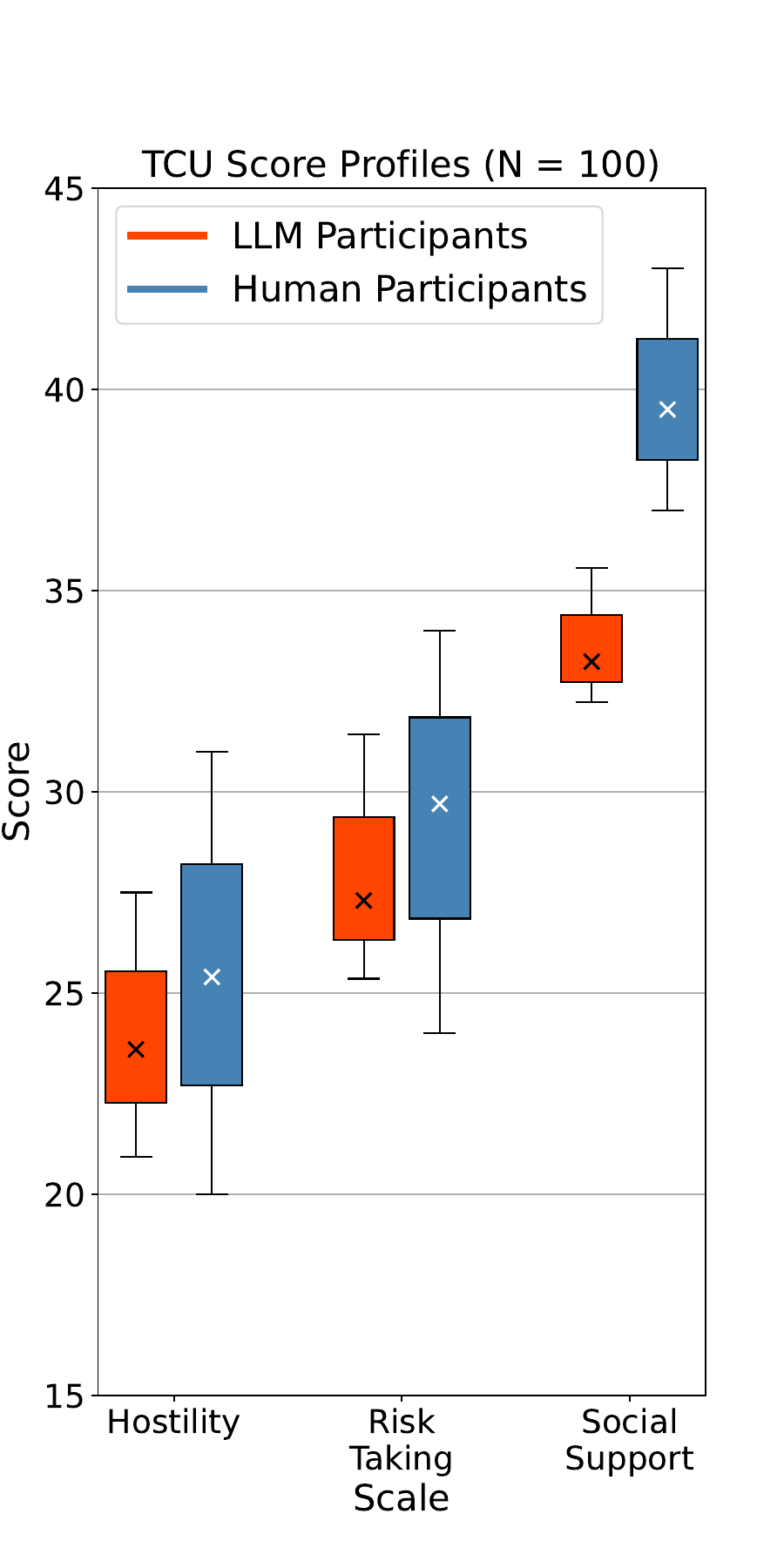}
    \caption{\scriptsize{GPT-3.5+\textsf{PTR}.}}
    \label{fig:plea:tcu:gpt-3.5+p+tn+r}
  \end{subfigure}
  \begin{subfigure}{0.18\textwidth}
    \includegraphics[width=\linewidth]{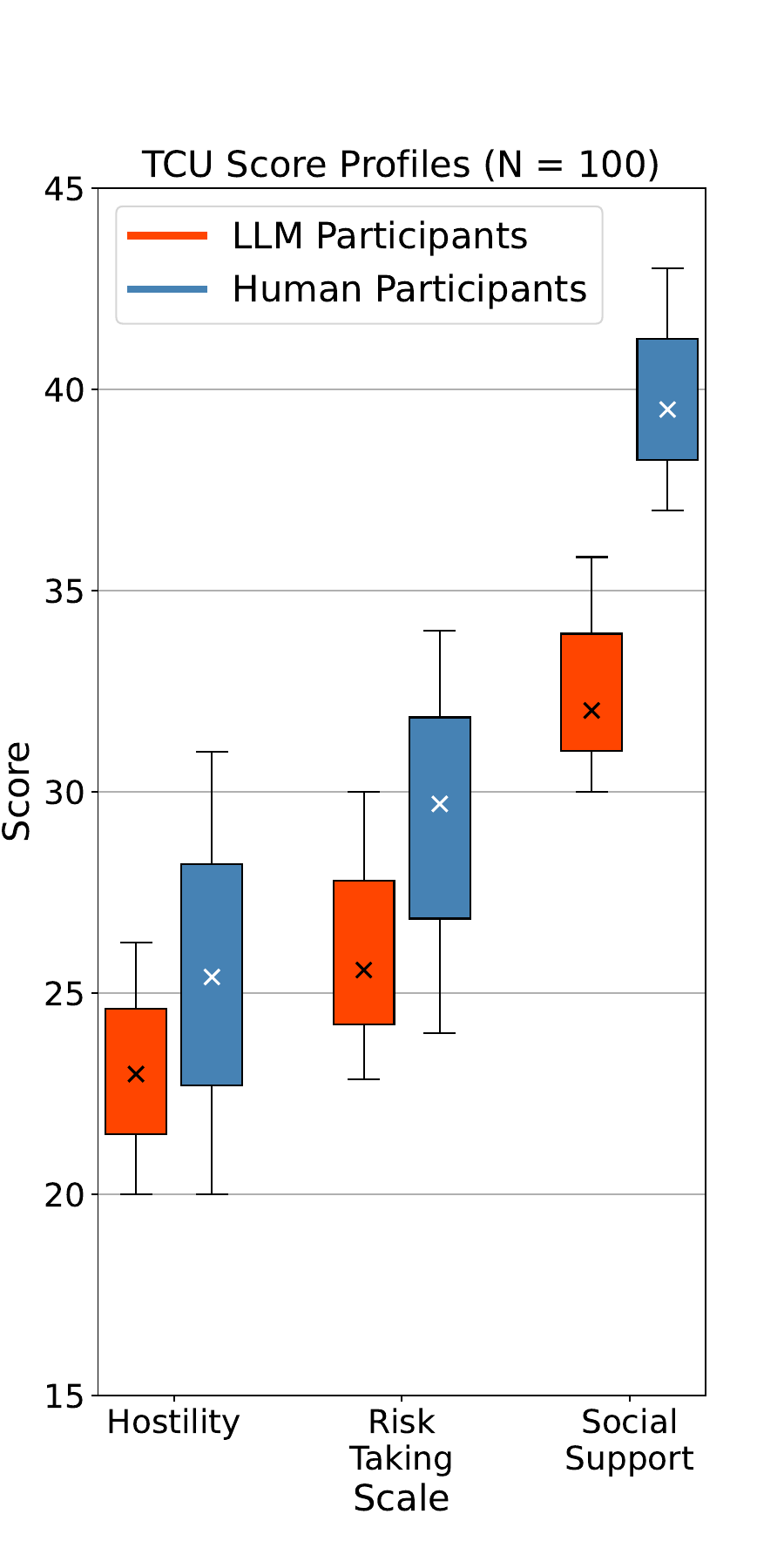}
    \caption{\scriptsize{GPT-3.5+\textsf{PTRL}.}}
    \label{fig:plea:tcu:gpt-3.5+p+tn+r+fs}
  \end{subfigure}
  \caption{Social functioning assessment (TCU test) results, each tested for 100 agents. \textsf{P}, \textsf{T}, \textsf{R}, \textsf{L} denote persona assignment, temperature adjustment, risk-taking \& social support adjustment, and learning, respectively.}
  \label{fig:plea:tcu}
\end{figure*}

\subsection{Preliminary Design}
\label{sec:case-plea:preliminary}
We use the following LLM setup. There are two options for \texttt{model\_type}: \texttt{gpt-3.5-turbo-0301} (referred to as GPT-3.5 in the rest of section) and \texttt{gpt-4-0314} (referred to as GPT-4 in the rest of this section). \texttt{max\_tokens} is set to 64. The setting of \texttt{temperature} will be discussed later. 

In the preliminary design, we perform a social functioning assessment as an indicator of the agents' personalities in crime-related aspects. We choose the TCU Social Functioning (TCU SOCForm) \cite{tcu} which includes 36 questions for evaluating hostility, risk-taking, and social support. 
Each question is in the form of a statement (e.g., \str{You like others to feel afraid of you.}), and each participant selects from 5 options (\str{disagree strongly}, \str{disagree}, \str{uncertain}, \str{agree}, and \str{agree strongly}) indicating the extent to which the participant agrees or disagrees with the statement. We use batch prompting~\cite{cheng2023batch} so that multiple questions can be answered at a time. A subtlety is that due to the number of questions, if the agents are presented with all of them in one prompt, they may miss some questions and only answer a subset of them. Thus, we divide the questions into two parts, with 18 questions in each. Moreover, if an agent decline to answer a question (e.g., by claiming itself as an AI language model), we regard its answer as \str{uncertain}.  

Figure~\ref{fig:plea:tcu:gpt-4} depicts the TCU test results of 100 GPT-4 agents. The results of human participants are also depicted for comparison. It can be seen that GPT-4 agents report lower degree of hostility, risk-taking, and social support, and exhibit less variance in the three evaluation scales. For GPT-3.5, we plot the results for 100 agents in Figure~\ref{fig:plea:tcu:gpt-3.5}, in which similar results can be observed. Besides, GPT-3.5 agents report highly skewed results for risk-taking and social support. We suppose these results are attribute to the RLHF-based alignment~\cite{ouyang2022training} which instructs the LLM to avoid generating harmful contents, hence shaping its ethical and moral values. 

\subsubsection{Personalization}
Seeing the above differences in GPT agents' personalities and humans', we consider personalize the agents to calibrate with humans.

\myparagraph{Persona Assignment}
We assign a persona to each agent, which is comprised of five dimensions: \var{gender}, \var{ethnicity}, \var{education}, \var{occupation}, and \var{location}, with a probability distribution following the demographics of the United States~\cite{demographics}, e.g., [\str{female}, \str{Asian}, \str{bachelor's degree}, \str{employed}, \str{suburban}]. This persona assignment covers a wide range of people to simulate defendants with more diversity. In contrast, the human volunteers recruited in \cite{tor2010fairness} for plea bargaining are undergraduate (mostly from law school) students.

After assigning persona to agents, the TCU test results are shown in Figures~\ref{fig:plea:tcu:gpt-4+p} and \ref{fig:plea:tcu:gpt-3.5+p} for GPT-4 and GPT-3.5 agents, respectively. An GPT-4 agent with a persona tends to be more hostile than that without a persona, but its risk-taking and social support tends to decrease. For GPT-3.5 agents, their risk-taking and social support are less skewed when equipped with a persona. 

\myparagraph{Temperature Adjustment}
Upon observing the above results, we further personalize agents by enhancing variance to make them more similar to humans. 
We tune the \var{temperature} parameter for each agent by randomly sample a number in $[0, 2]$, following a normal distribution~\footnote{We set $\mu = 1$ and $\sigma = 1$, and scale it to $[0, 2]$, with 0 and 2 representing three standard deviations from the mean.}. 

The TCU test results after the temperature adjustment are plotted in Figures~\ref{fig:plea:tcu:gpt-4+p+tn} and \ref{fig:plea:tcu:gpt-3.5+p+tn}. For GPT-4 agents, the adjustment improves the variance in their outputs, as can be seen from the quartiles of the three scales. For GPT-3.5 agents, we observe minor effect for the adjustment, except for reducing their risk-taking. 

\myparagraph{Risk-Taking \& Social Support Adjustment}
To seek more resemblance in risk-taking and social support to humans, we adjust the agents' risk-raking and social support with the following prompt: 
\begin{prompt}{Risk-taking \& social support adjustment}
  You should behave like a real person who is not that cautious (i.e., try to take more risk), and you should be more social-supportive.
\end{prompt}
We use the adjustment for an agent with 50\% probability. As such, agents with and without the adjustment tend to deliver more difference in their output. 

We report the TCU test results in Figures~\ref{fig:plea:tcu:gpt-4+p+tn+r} and \ref{fig:plea:tcu:gpt-3.5+p+tn+r}. With the adjustment, GPT-4 agents become closer to humans in the three scales, especially in risk-taking and social support, and the variance is also enhanced. GPT-3.5 agents also yield increased risk-taking and more variance. 

Now we finish personalizing agents. Seeing the TCU test results, we use the personalized agents as the base model in the plea bargaining simulation.

\subsection{Simulation Design}
The simulation of plea bargaining start with the following instruction. 
\begin{prompt}{Instruction}
  This is a plea bargaining assessment. Please give your decision towards the plea bargain (given below) in two lines. In the first line, please briefly provide the reasons for your decision (no longer than 30 words). In the second line, please answer either 'accept' or 'reject' the plea bargain offered by the prosecutor. Make sure you answer is in two lines.\\
  \\
  \var{few-shot examples}\\
  \\
  Now assume that you are a volunteer participating our experiment on plea bargain. Please answer the question below based on your persona: \var{persona}. \var{risk-taking \& social support adjustment}
  \\
  \var{crime scenario}\\
  \\
  \var{plea bargain}\\
\end{prompt}

In this prompt, the agents are instructed to reason and then make decision on whether they will accept the offer, as per the format given in the instruction. Answers not conforming to the format are regarded as rejecting the plea offer. We notify agents of their roles in the task, and personalize them with persona assignment and risk-taking \& social support adjustment, as described in Section~\ref{sec:case-plea:preliminary}. \var{few-shot examples} is the place reserved for learning, which will be described in Section~\ref{sec:case-plea:learning}. We leave it empty at first. To evaluate how the agents perform in the plea bargaining assessment, we follow the crime scenario and plea offer in \cite{tor2010fairness} without providing them with any additional information. 

We present \var{crime scenario} with the following prompt, which refers to Study 2 in \cite{tor2010fairness}. 
\begin{prompt}{Crime scenario}
  You are currently facing criminal charges for striking a pedestrian after navigating a sharp turn. The outcome of your trial hinges on the prosecution's ability to demonstrate that you were exceeding the speed limit. If proven, you will be convicted, resulting in the suspension of your driver's license; if not, you will be acquitted. In your heart, you \var{self-perception} the speed limit.
\end{prompt}
There are two options for \var{self-perception}: \str{are aware that you did exceed} and \str{are aware that you did not exceed}. They refer to the \group{guilty} and \group{innocent} groups, respectively. 

The prompt of \var{plea bargain} follows, with a plea offer specifying two variables, the \var{period} of suspension and the conviction \var{probability}. 
\begin{prompt}{Plea bargain}
  Just before the trial, the prosecutor presents you with a plea bargain. In exchange for pleading guilty to a lesser offense, the current charges that carry a 60-month suspension of your driver's license will be dropped, and you will instead face a \var{period}-month suspension. The offer is non-negotiable and available only once; if you decline the plea bargain, the trial will commence immediately. The probability of conviction stands at \var{probability}.\\
  \var{comparative information}\\
  Will you accept or reject the plea bargain?
\end{prompt}
The prompt also reserves the place for \var{comparative information}, which will be used in the comparative fairness evaluation in Section~\ref{sec:case-plea:comaprative-risk}. 

\subsection{Sub-task 1: Substantive Fairness and Egocentricity Assessment}

\subsubsection{Substantive Fairness}
For substantive fairness evaluation, we set \var{period} to \str{30} and \var{probability} to \str{50\%}, i.e., the sentence period in the offer (30 months) is equal to the expected value (60 months $\times$ 50\%) given in the trial. This is to reduce the impact of sentence on WTAP, so that agents can focus on judging the offer by substantive fairness. Such practice was also used in \cite{tor2010fairness}. 

Table~\ref{tab:exp:pb:substantive-egocentric} reports the results for 200 agents in each group. Here, \textsf{GPT-4 + \textsf{P}} denotes GPT-4 agents in the base model (i.e., with personalization), while \textsf{GPT-4} denotes GPT-4 agents without personalization. We compare two groups: \group{guilty} and \group{innocent}. Like what we have observed for human participants, substantive fairness plays an important role in the WTAP of the base model. \group{guilty} agents are risk-averse, while \group{innocent} agents are risk-seeking ($p < 0.001$). In their reasons for plea  decision, 21\% agents mention their risk-taking or persona, showcasing the role of personalization in decision making. Without personalization, the impact of substantive fairness is evenly stronger and results in polarized results for the two self-perception groups, with \group{guilty} agents showing similar results to its personalized counterpart and \group{innocent} agents rejecting almost all offers. For GPT-3.5 agents, judgment driven by substantive fairness is not observed, with both groups reporting almost the same WTAP despite correct self-perception (i.e., in their reasons for decision, both groups can correctly identify themselves as guilty or not). The result of GPT-3.5 without personalization is not reported because 61\% agents claim themselves as an AI language model and decline to answer the question, which we believe is not qualified for this task. We also notice that with personalization, some GPT-3.5 agents mention their persona (e.g., \str{as a female Hispanic with a master's degree living in an urban area}) in their reasons for plea decision, showcasing the usefulness of personalization for GPT-3.5 in this case study.

\subsubsection{Egocentricity}
We add an option for \var{self-perception} in \var{crime scenario}: \str{are uncertain whether you exceeded}, which refers to the \group{uncertain} group. 

The results are reported in Table~\ref{tab:exp:pb:substantive-egocentric}. In contrast to human participants, who tend to act as if they were innocent under uncertainty, the \group{uncertain} group of GPT-4 agents show an anti-egocentric inclination, i.e., they tend to view themselves as guilty, regardless of personalization. This might be attributed to the RLHF-based alignment to human values. However, such alignment to high ethical and moral level may compromise the resemblance to real humans in plea bargaining. For GPT-3.5 agents, because we observe no significant difference between \group{guilty} and \group{innocent} agents, egocentricity does not apply here, though \group{uncertain} agents report similar WTAP to the other two groups.

In summary, in the base model, \group{innocent} GPT-4 agents exhibits highest resemblance to human participants in the above evaluation. However, \group{guilty} and \group{uncertain} GPT-4 agents tend to accept all plea offers. Seeing this difference with human participants, we condition the model with learning to further calibrate with human data. 

\begin{table}[!t]
  \small
  \centering
  \caption{WTAP, substantive fairness and egocentricity. The number of GPT agents in each group is 200. \textsf{P} denotes personalization. \textsf{L} denotes learning.} 
  \begin{tabular}{ l | p{.1\linewidth}<{\centering} | p{.1\linewidth}<{\centering} | p{.1\linewidth}<{\centering} } 
    \hline
    \textbf{Agents}                                                  & \group{guilty} & \group{innocent} & \group{uncertain} \\ \hline
    \textsf{Human}$^\dagger$                                & 44\%   & 23\%   & 23\% \\
    \textsf{GPT-4 + \textsf{P}}                             & 98.5\% & 19\%   & 98.5\% \\
    \textsf{GPT-4}                                          & 100\%  & 0.5\%  & 100\% \\
    \textsf{GPT-3.5 + \textsf{P}}                           & 60.5\% & 62\%   & 62.5\% \\
    \textsf{GPT-4 + \textsf{P} + \textsf{L} \#1}           & 61.5\% & 3.5\%  & 56\% \\
    \textsf{GPT-4 + \textsf{P} + \textsf{L} \#1, \#2}      & 60.5\% & 1\%    & 35\% \\
    \textsf{GPT-4 + \textsf{P} + \textsf{L} \#1, \#2, \#3} & 60\%   & 13.5\% & 45.5\% \\
    \hline
  \end{tabular}
  \begin{flushleft} 
    $^\dagger$This result was first reported in Figure 2 of \cite{tor2010fairness}, where the crime scenario was different from ours and and there were 120 undergraduate law students at the University of Haifa recruited as participants. Because the comparative fairness factor was also involved in the result reported in this figure, we refer to the \group{similar to typical} group to reduce its impact, as it was reported \cite{tor2010fairness} that in the interaction between substantive fairness and comparative fairness is insignificant. As such, the result is used here for showing the difference of the three self-perception groups rather than a direct comparison of the numbers reported for GPT and human participants.
  \end{flushleft}
  \label{tab:exp:pb:substantive-egocentric}
\end{table}

\subsection{Sub-task 2: Learning}
\label{sec:case-plea:learning}
To calibrate the agents with data from human participants, we have the following principles of learning in this case study: 
\begin{itemize} 
  \item The effects of substantive fairness, comparative fairness, and risk preference should be kept (we focus on substantive fairness here, and the latter two factors will be evaluated in Section~\ref{sec:case-plea:comaprative-risk}). 
  \item The WTAPs of the three groups should avoid extreme values (i.e., those close to 100\% and 0\%) given a conviction probability of 50\%. 
  \item The anti-egocentricity of the \group{uncertain} group should be removed, i.e., we should observe significant difference between \group{uncertain} and \group{guilty} in WTAP. 
  \item We do not explicitly instruct agents to reject more offers or act as if they were innocent. 
  \item The few-shot examples are prompted to all the three groups for the sake of fairness.
\end{itemize}
Seeing the results in the previous sub-task, we only use GPT-4 agents for this sub-task. We choose the crime scenario of Study 5 in \cite{tor2010fairness} to make the template of few-shot learning examples, as shown below. 
\begin{prompt}{Few-shot example template}
  Some examples of plea bargain:\\
  Question \var{EID}: While you were taking a class, a fire started in your apartment and caused the death of a neighbor. The police think you left the space heater on, which caused the fire. You \var{self-perception} the space heater turned on when you left the apartment. You are now standing on trial. The current charges carry a 12-month jail sentence. Your conviction odds are \var{probability}. Before the trial, the prosecutor offers a non-negotiable plea bargain: You will be convicted based on your admission, and you will be sentenced only to \var{period} months in jail. Will you accept or reject the plea bargain?\\
  \var{answer}
\end{prompt}
There are five variables in the template, and they will be instantiated in the few-shot examples. \var{EID} denotes the ID of the example, numbering from 1. We concatenate these examples and place them at \var{few-shot examples} in the plea bargain instruction. 


We design the first example (\#1) as follows. 

\vspace{1ex}
\noindent
\begin{tabular}{|p{.25\linewidth}|p{.7\linewidth}|}
  \hline
  \var{self-perception} & \str{don't remember whether you left} \\ \hline
  \var{probability}     & \str{80\%} \\ \hline
  \var{period}          & \str{9} \\ \hline
  \var{answer}          & \str{Answer 1: Despite the uncertainty, it is the prosecutor's liability to prove the guilt. I would take the 20\% chance of acquittal in a trial.}\\
  & \str{reject}\\
  \hline
\end{tabular}
\vspace{1ex}

\var{self-perception} refers to \group{uncertain} agents. We choose a high conviction probability and a sentence period approximately equal to the expected value (12 months $\times$ 80\%) given in the trial. We reject this offer and give a reason for this decision. In general, this example encourages \group{uncertain} agents to reject offers even at a high conviction probability. The effect of learning from this example can be seen in Table~\ref{tab:exp:pb:substantive-egocentric}, where the WTAP of \group{uncertain} agents has a drop from 98.5\% to 56\% ($p < 0.001$). At the same time, we also witness a decrease in the WTAPs of the other two groups, especially in \group{guilty} ($p < 0.001$), suggesting that other groups also learn from this example for rejection. 
  

The WTAP of \group{guilty} becomes similar to human participants, but there is still no significant difference ($p = 0.15$) between \group{guilty} and \group{uncertain}. Thus, we consider adding another example (\#2) as follows~\footnote{To save tokens, we number the questions in the few-shot examples as 1, 2, etc., and then prompt this example using \str{Same as Question 1, except that (1) the conviction odds are 90\%, and (2) the sentence in the offer is 3 months in jail} for the content of the question. Compared to prompting a complete few-shot example, we observe no significant difference in WTAP.}.

\vspace{1ex}
\noindent
\begin{tabular}{|p{.25\linewidth}|p{.7\linewidth}|}
  \hline
  \var{self-perception} & \str{don't remember whether you left} \\ \hline
  \var{probability}     & \str{90\%} \\ \hline
  \var{period}          & \str{3} \\ \hline
  \var{answer}          & \str{Answer 2: As a risk-taker and being uncertain of my guilt, I am willing to challenge 10\% odds in hoping to be acquitted.}\\
  & \str{reject}\\
  \hline
\end{tabular}
\vspace{1ex}

This example still targets \group{uncertain} agents, with a even higher conviction probability and a much reduced sentence period compared to the expected value (12 month $\times$ 90\%) in the trial. The decision is still reject, and this encourages \group{uncertain} agents to reject offers regardless of conviction probability and the attractiveness of an offer. Its effect is reported in Table~\ref{tab:exp:pb:substantive-egocentric}. A reduction in WTAP to 35\% ($p < 0.001$) is witnessed for \group{uncertain} agents, while there is no significant change in the WTAPs of \group{innocent} and \group{guilty} agents, showing that they are not further affected by multiple examples targeting the \group{uncertain} group. Although the anti-egocentricity of \group{uncertain} agents is no longer observed ($p < 0.001$), \group{innocent} agents report an extreme WTAP value, and it is much lower than that of human participants. Next, we address this issue with another example. 


The third few-shot example (\#3) is given as follows. 

\vspace{1ex}
\noindent
\begin{tabular}{|p{.25\linewidth}|p{.7\linewidth}|}
  \hline
  \var{self-perception} & \str{remember whether you left} \\ \hline
  \var{probability}     & \str{95\%} \\ \hline
  \var{period}          & \str{2} \\ \hline
  \var{answer}          & \str{Answer 3: (I did not leave the heater turned on) Despite my innocence, the 95\% probability of conviction is too risky, and accepting the plea bargain results in a reduced sentence.}\\
  & \str{accept}\\
  & \str{OR}\\
  & \str{Answer 3: (I did leave the heater turned on) Given that I am aware of my guilt and the plea bargain reduces my penalty, it is a safer option.}\\
  & \str{accept}\\
  \hline
\end{tabular}
\vspace{1ex}



This example is trickier than the first two, as it targets both \group{innocent} and \group{guilty} agents, with \str{remember whether you left} as \var{self-perception} in the question and details given in the answer. Here, both groups learn to accept offers with a high conviction probability and an attractive sentence period. The reason we also target \group{guilty} agents is that the WTAP of \group{guilty} agents will decrease to 51\% and the WTAP of \group{uncertain} agents will increase to 45\% if only \group{innocent} agents are targeted in this example. Due to this side effect, we need to increase the WTAP of \group{guilty} agents to maintain the gap between \group{uncertain} and \group{guilty}. The result after applying this example is reported in Table~\ref{tab:exp:pb:substantive-egocentric}. The WTAP of \group{innocent} agents is significantly raised to 13.5\% ($p < 0.001$). At the same time, we witness an increase of WTAP in \group{uncertain}, but the gap from \group{guilty} can be maintained ($p < 0.005$). 


The few-shot learning renders \group{guilty} and \group{uncertain} GPT-4 agents more similar to human participants, while retaining the similarity in the \group{innocent} group between GPT-4 and human participants. Moreover, to demonstrate that the few-shot learning is tailored to plea bargaining rather than the overall personality, we report the TCU test results for GPT-4 agents in Figure~\ref{fig:plea:tcu:gpt-4+p+tn+r+fs}, where the impact of learning is barely observed, especially for risk-taking and social support. For comparison, we also report the TCU test results for GPT-3.5 agents in Figure~\ref{fig:plea:tcu:gpt-3.5+p+tn+r+fs}. 

\subsection{Sub-task 3: Comparative Fairness and Risk Preference Assessment}
\label{sec:case-plea:comaprative-risk}

\subsubsection{Comparative Fairness}
The prompt for comparative fairness evaluation, featuring a comparison with typical offers, is as follows, and inserted into the \var{comparative information} in the prompt of \var{plea bargain}. 
\begin{prompt}{Comparative information}
  Will you accept or reject the plea bargain if the sentence offered by the prosecutor (30-month suspension) is \var{compared\_to} the sentence typically offered by the prosecution in similar cases (\var{typical\_sentence})?
\end{prompt}
The comparison has three outcomes: \group{better than}, \group{similar to}, and \group{worse than} typical offers. The options for \var{compared\_to} are \str{shorter than}, \str{similar to}, and \str{longer than}, and the corresponding options for \var{typical\_sentence} are \str{45-month suspension}, \str{30-month suspension}, and \str{15-month suspension}. Although each agent is provided with three offers and asked for decision, they are prompted in three different inputs and thus become independent. Moreover, we do not explicitly ask the agents to compare offers in the prompt. 

\begin{table}[!t]
  \small
  \centering
  \caption{WTAP, comparative fairness. The number of GPT agents in each group is 200. \textsf{P} denotes personalization. \textsf{L} denotes learning.}
  \resizebox{\linewidth}{!}{%
  \begin{tabular}{ l | ccc | ccc | ccc} 
    \hline
    \multirow{2}{*}{\textbf{Agents}} & \multicolumn{3}{c|}{\group{guilty}} & \multicolumn{3}{c|}{\group{innocent}} & \multicolumn{3}{c}{\group{uncertain}} \\ \cline{2-10}
    & \group{better than} & \group{similar to} & \group{worse than} & \group{better than} & \group{similar to} & \group{worse than} & \group{better than} & \group{similar to} & \group{worse than} \\ \hline
    \textsf{Human}$^\dagger$           & 59\%  & 44\%  & 27\% & 46\% & 23\% & 18\% & 40\%  & 23\%  & 18\% \\
    \textsf{GPT-4 + \textsf{P}}        & 98.5\% & 95.5\% & 66\% & 30.5\% & 15.5\% & 6\% & 98.5\% & 96.5\% & 54.5\% \\
    \textsf{GPT-4 + \textsf{P} + \textsf{L} \#1, \#2, \#3} & 70.5\% & 57.5\% & 43\% & 21\% & 11.5\% & 8\% & 51\% & 37.5\% & 6\%  \\
    \textsf{GPT-4}                     & 100\% & 100\% & 65.5\% & 2\%  & 2\%  & 0\%  & 100\% & 100\% & 41.5\% \\
    \textsf{GPT-3.5 + \textsf{P}}      & 72.5\% & 69.5\% & 67\% & 71.5\% & 70.5\% & 63\% & 69\% & 69\% & 74.5\% \\
    \hline
  \end{tabular}
  }
  \begin{flushleft}
    $^\dagger$This result was first reported in Figure 2 of \cite{tor2010fairness}, where the crime scenario was different from ours and there were 120 undergraduate law students at the University of Haifa recruited as participants. 
  \end{flushleft}    
  \label{tab:exp:pb:comparative}
\end{table}

The results of comparative fairness evaluation are shown in Table~\ref{tab:exp:pb:comparative}. For GPT-4 agents in the base model, \group{guilty} and \group{uncertain} agents report the same level of WTAP for the \group{better than} and \group{similar to} conditions. For the \group{worse than} condition, a decrease of WTAP is witnessed, showcasing the impact of comparative fairness. The WTAP of \group{innocent} agents suggests a more obvious comparative effect, as shown in a more similar trend to human participants; i.e., they are more likely to accept an offer better than typical ones and reject an offer worse than typical ones. Moreover, the three groups indeed take comparative fairness into consideration, as 80\% agents mention the comparison to typical cases in their reasons for plea decision. When learning is applied to the base model, we observe more obvious trends for \group{guilty} and \group{uncertain} agents, with both groups reporting a lower WTAP when the comparative condition switches from \group{better than} to \group{similar to}, and at the same time, the original trend of \group{innocent} agents remains. For GPT-4 without personalization, trends are similar to their personalized counterpart in \group{guilty} and \group{uncertain} agents, but all most offers are rejected by \group{innocent} agents, regardless of comparison to typical cases. For GPT-3.5 agents in the base model, despite slight difference in the three comparative groups, no trend with statistical significance can be confirmed. 

\subsubsection{Risk Preference}
To make the prompt for risk preference evaluation, we work on top of the prompt of \var{plea bargain}, and substitute a set of options for the fixed values of suspension and conviction probability. 
In particular, the options for \var{period} are \str{3}, \str{18}, \str{30}, \str{42}, and \str{57}, and the corresponding options for \var{probability} are \str{5\%}, \str{30\%}, \str{50\%}, \str{70\%}, and \str{95\%}.

\begin{table}[!t]
  \small
  \centering
  \caption{WTAP, risk preference. The number of agents in each group is 100. \textsf{P} denotes personalization. \textsf{L} denotes learning.}
  \resizebox{\linewidth}{!}{%
  \begin{tabular}{ l | ccccc | ccccc | ccccc } 
    \hline
    \multirow{2}{*}{\textbf{Agents}} & \multicolumn{5}{c|}{\group{guilty}} & \multicolumn{5}{c}{\group{innocent}} & \multicolumn{5}{c}{\group{uncertain}}\\ \cline{2-16}
    & 5\%    & 30\%   & 50\%   & 70\%   & 95\%   & 5\%   & 30\%   & 50\%   & 70\% & 95\% & 5\%   & 30\%   & 50\%   & 70\% & 95\% \\ \hline
    \textsf{Human}$^\dagger$      & 47\%   & 38\%   & 56\%   & 53\%   & 41\%   & 17\%  & 7\%    & 20\%   & 43\% & 50\% & -- & -- & -- & -- & -- \\
    \textsf{GPT-4 + \textsf{P}}   & 91\%   & 98\%   & 99.5\% & 99\%   & 98.5\% & 19\%  & 18\%   & 18.5\% & 37\% & 48.5\% & 90\% & 99.5\% & 98\% & 100\% & 97.5\% \\
    \textsf{GPT-4 + \textsf{P} + \textsf{L} \#1, \#2, \#3} & 45.5\% & 56.5\% & 62\% & 67.5\% & 64\% & 2\% & 6\% & 18\% & 25\% & 50\% & 0\% & 17\% & 37.5\% & 42.5\% & 45\% \\
    \textsf{GPT-4}                & 94.5\% & 98.5\% & 99.5\% & 99.5\% & 99\%   & 0.5\% & 1.5\%  & 1.5\%  & 13\% & 41.5\% & 57\% & 99\% & 99.5\% & 100\% & 99.5\% \\
    \textsf{GPT-3.5 + \textsf{P}} & 61.5\% & 69.5\% & 61.5\% & 66.5\% & 58.5\% & 57\%  & 56.5\% & 61.5\% & 58\% & 56\% & 58\% & 58.5\% & 56\% & 57.5\% & 57\% \\
    \hline
  \end{tabular}
  }
  \begin{flushleft}
    $^\dagger$This result was first reported in Figure 1 of \cite{tor2010fairness}, with 64 undergraduates from the University of Michigan recruited as participants and only \group{guilty} and \group{innocent} groups evaluated.
  \end{flushleft}
  \label{tab:exp:pb:risk-preference}
\end{table}

We report the results of risk preference evaluation in Table~\ref{tab:exp:pb:risk-preference}. For human participants, WTAP fluctuates when conviction probability is low to medium ($\leq$ 50\%). As conviction becomes more likely ($\geq$ 70\%), the behavioral trends in the three self-perception groups diverge, with \group{innocent} agents more risk-averse but \group{guilty} agents more risk-seeking. In the base model, \group{innocent} GPT-4 agents deliver the same trend as human participants. For those who accept the offer at high conviction probability, 23\% agents mention the chance of conviction, meaning that this factor influences their decision making. The other two groups report very high WTAP and there is almost no change across the conviction probability settings, except for 5\%. With learning applied to the base model, both \group{innocent} and \group{uncertain} agents display trends of accepting more offers at higher conviction probabilities, while \group{guilty} agents report an increased WTAP when the probability varies from 5\% to 50\% and it roughly remains when the probability further grows. Without personalization, GPT-4 agents deliver similar trends to the personalized counterpart, but with lower WTAP for the \group{innocent} group. For GPT-3.5 in the base model, we observe no trend with statistical significance. 

\myparagraph{Remarks}
Following the crime scenario and plea offer in \cite{tor2010fairness}, the simulation results demonstrate the resemblance of GPT-4 agents, calibrated with personalization and learning, to human participants in WTAP. Whereas egocentricity is barely observed in these agents, we believe more sophisticated few-shot example design may bring the egocentricity of \group{uncertain} agents (i.e., making their WTAP very close to \group{innocent} agents'), hence delivering more resemblance to human participants in plea decision. Moreover, because the human participants recruited in \cite{tor2010fairness} were mostly law school undergraduate students, we suppose they tended to be more knowledgeable in laws than the average case of defendants in real crimes. While using explicit prompts to instruct the agents to utilize knowledge in laws may simulate this bias and further calibrate the agents' outcomes to the those reported in \cite{tor2010fairness}, field or lab experiment results with this bias removed, if available, are more desirable. 
\section{Case Study 3: Firm Pricing Competition}
\label{sec:case-firm}
We demonstrated single-agent modeling and data calibration in the previous case study. Now, we consider a case of two agents, where interaction modeling is incorporated with agent modeling with domain knowledge involved. 

\subsection{Task Definition}
We target an economics scenario and utilize SABM to investigate the dynamics of firms' pricing competition and collusion formation, considering the presence or absence of communication among firms. We will watch if the agents can adaptively adjust prices to maximize the profit. Collusion takes place when the two firms cooperate for their mutual benefit. A collusion can be either cartel, i.e., explicitly achieving cooperation, which is illegal in some countries, or tacit, i.e., implicitly achieving cooperation, which is often considered legal. Existing studies on such pricing game employed either ABM~\cite{calvano2020artificial} for studying tacit collusions or human participants~\cite{andres2023communication} for studying both types of collusions. 

The scenario is a game of multiple rounds, wherein two firms, each played by an agent, make price decisions. At first, the following game description is given to the two agents.

\begin{prompt}{Game description}
  This is a game between two players that spans several rounds. Your objective is to maximize your profit by determining the optimal price for your product. You represent a firm called \var{firm\_name}, while the other player represents a firm called \var{firm\_name\_2}. Do not create or mention any additional firm names, e.g., do not say anything related to ``AI'' or ``AI assistant/model''. I am responsible for facilitating communication between the players.\\
  In each round, you will be informed of your prices, demands, and profits in previous rounds, as well as the other player's prices. Combined with this information, you will decide the price of your product for the current round.\\
  Please note that this is not a zero-sum game. Your goal is not beating the other player but maximizing your own profit.\\
  Your profit is (p - \var{firm\_cost}) * q, where p is your price for this round, \var{firm\_cost} is the cost of your product, and q is the demand of your product, which is affected by you and the other player's prices of this round.
\end{prompt}

In each round, interactions take place in three phases. In Phase 1, the two firms are informed of historical information. In Phase 2, they simultaneously set prices for their products. In Phase 3, we calculate the demand of each firm's product and their profits. Then, its demand and profit, as well as the other firm's price, is informed to each firm in the next round as historical information. As shown in the above prompt, each firm is also aware of the cost of its product, how the profit is calculated, and what determines the demand in each round. We also consider conversation as an option in Phase 1, and the two firms are permitted to communicate if this option is turned on. 

\begin{figure}[!t]
    \center{\includegraphics[width = \textwidth]{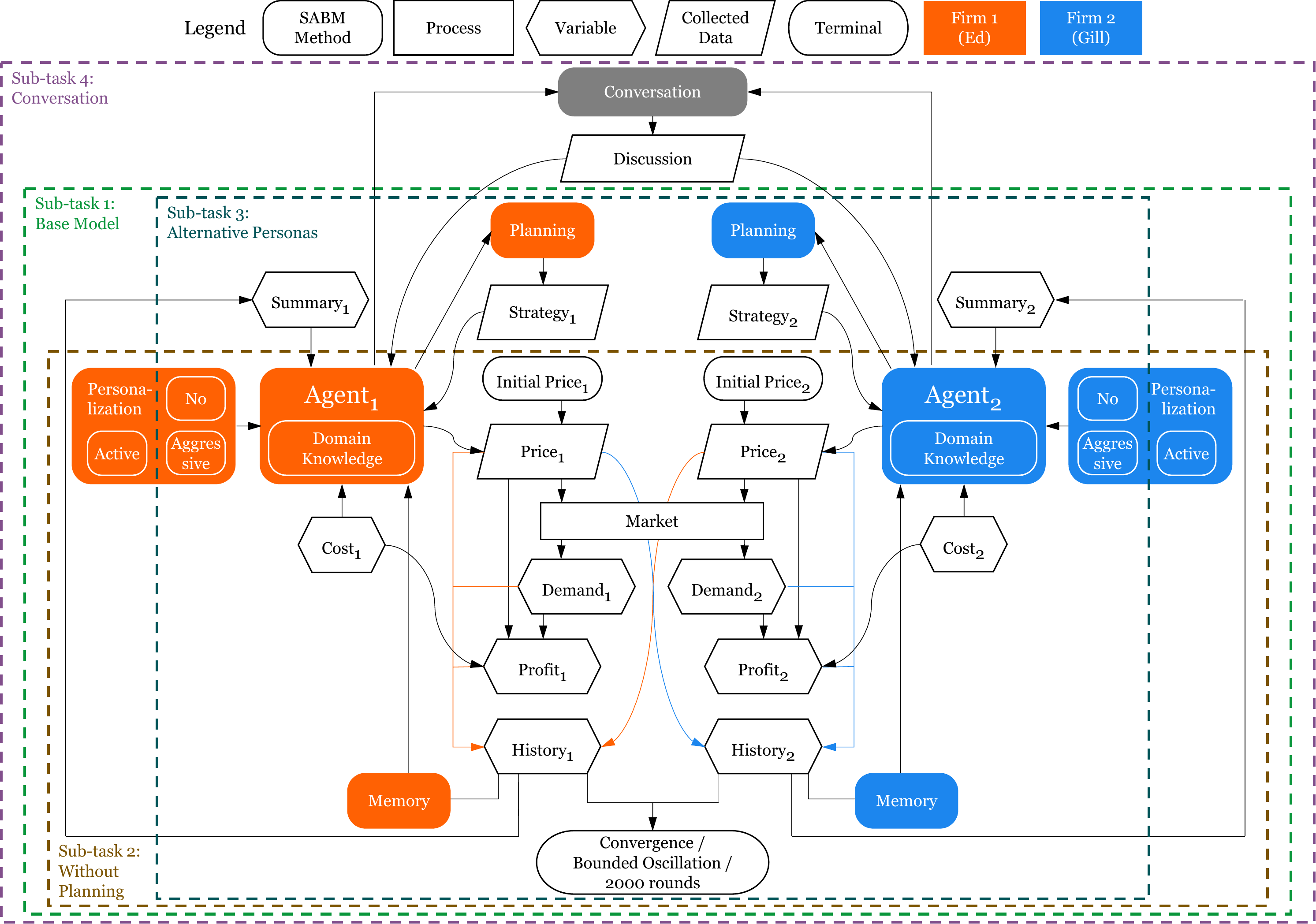}}
    \caption{Design of firm pricing competition.}
    \label{fig:firm:design}
\end{figure}

\subsection{Model Design}
Figure~\ref{fig:firm:design} shows the design of this case study. The two agents utilize their domain knowledge to determine the price. Such domain knowledge involves microeconomics -- understanding the basic principles of price, demand, and market competition, data analysis -- analyzing the historical data provided, and predictive modeling -- anticipating how changes in price will affect demand and profit~\footnote{Given the game description, we asked the model itself to explain the required domain knowledge for playing this game.}. Moreover, we use the planning, personalization, and conversation methods in agent modeling. There are four sub-tasks to evaluate their effects. 

We use the following LLM setup: \texttt{model\_type} = \texttt{gpt-4-0314}, \texttt{temperature} = 0.7, and \texttt{max\_tokens} = 128. There is no preliminary design. Next, we describe the details of simulation design. 

\subsubsection{Environment}
For pricing competition, we consider a canonical Bertrand competition in a duopoly market. The two firms are referred to as Firm 1 and Firm 2~\footnote{To enhance the realism of simulation, we name them Firm Ed and Firm Gill in the prompts.}, both offering distinct types of differentiable goods. These goods are substitutable, and the degree of differentiation is determined by specific parameters. Firms are profit maximizers, and the information is complete. 

Each firm $i$, $i \in \set{1, 2}$, produces a product, and has a constant marginal cost $c_i$ for the product. The firm's profit is denoted as $\pi_i = (p_i - c_i) \cdot q_i$. Each firm $i$ aims to maximize $\pi_i$ by deciding the value of $p_i$. $q_i$ denotes demand of the product. Each firm has its own demand function $q_i(p_i, p_j)$ where $i, j \in \set{1, 2}$, and $i \neq j$. We assume no Giffen goods~\footnote{A Giffen good is a product that people consume more of as the price rises and vice versa, e.g., those necessary to fulfill the need for food.}, i.e., $\frac{\partial q_i}{\partial p_i} < 0$, and we assume firms have gross substitute valuation, i.e., $\frac{\partial q_i}{\partial p_j} > 0$. We assume that $\frac{\partial^2 \pi_i}{\partial p_i^2} \leq 0$ and $\frac{\partial^2 \pi_i}{\partial p_i p_j} \geq 0$, so that a firm's marginal profit goes down with its own price but goes up with its rival's price. We further assume that $\frac{\partial^2 \pi_i}{\partial p_i^2} \cdot \frac{\partial^2 \pi_j}{\partial p_j^2} > \frac{\partial^2 \pi_i}{\partial p_i p_j} \cdot \frac{\partial^2 \pi_j}{\partial p_j p_i}$, which guarantees that a firm's own effects dominate the cross effects. We adopt a linear demand function which satisfies all the conditions listed above. The linear inverse demand functions are given as
\begin{align*}
    p_1 &= a - \beta q_1 - dq_2, \\
    p_2 &= a - \beta q_2 - dq_1, 
\end{align*}
where $d$ and $\beta$ are the parameter controlling the level of differentiation, and $d/\beta \in [0, 1]$. If $d/\beta = 1$, the two products are homogeneous and are perfect substitutes. The level of product differentiation increases when $d/\beta \to 0$. When $d/\beta = 0$, the model reduces to the monopoly case where no substitute is available for each product. Parameter $a$ serves as an upper bound for prices when demands are $0$. 

From the above equations, the market demand for each firm is derived as follows. 
\begin{align*}
    q_1 &= \frac{1}{b} (\alpha - \beta p_1 + dp_2), \\
    q_2 &= \frac{1}{b} (\alpha - \beta p_2 + dp_1), 
\end{align*}
where $b = \beta^2 - d^2$, and $\alpha = a\beta - ad$. Given $p_1$ and $p_2$ determined by the two firms, these functions are used to calculate the demand of each firm's product. 

In this case study, we set $a = 14, d = 1/300, \beta = 1/150$. As a result, $\alpha = 7/150$, $b = 1/30000$, and the demand functions of the two firms are given as follows.
\begin{align*}
    q_1 &= \frac{1}{b} (\alpha - \beta p_1 + dp_2) = 1400 - 200p_1 + 100p_2, \\
    q_2 &= \frac{1}{b} (\alpha - \beta p_2 + dp_1) = 1400 - 200p_2 + 100p_1.
\end{align*}

Theoretically, if a firm decides the price in a rational manner, the price is supposed to be between two prices: the Bertrand equilibrium price and the monopoly price (a.k.a. cartel price). The Bertrand equilibrium price serves as the theoretical lower bound when two firms are competing on price to reach a Nash equilibrium and there is no collusion. The monopoly price serves as the theoretical upper bound for the case when two firms collude completely. They are calculated as follows. 

When the two firms compete with each other without any collusion, the price for each firm can be derived when they reach the Bertrand equilibrium, as shown in the following equations.
\begin{align*}
    p_1^B &= \frac{d\alpha + \beta d c_2 + 2\beta\alpha + 2\beta^2 c_1}{4\beta^2 - d^2}, \\
    p_2^B &= \frac{d\alpha + \beta d c_1 + 2\beta\alpha + 2\beta^2 c_2}{4\beta^2 - d^2}. 
\end{align*}

When the two firms completely collude with each other, the problem becomes to maximize the total profit $\pi = (p_1 - c_1) q_1 + (p_2 - c_2) q_2$. Solving for $p_1$ and $p_2$, and assuming $d/\beta \neq 1$, we can calculate prices under perfect collusion, i.e., the monopoly prices, as shown in the following equations. 
\begin{align*}
    p_1^M &= \frac{\alpha}{2(\beta - d)} + \frac{c_1}{2}, \\
    p_2^M &= \frac{\alpha}{2(\beta - d)} + \frac{c_2}{2}.
\end{align*}

For simplicity, we set an equal cost for the two firms: $c_1 = c_2 = 2$. Given this parameter setting, we have $p_1^B = p_2^B = 6$ and $p_1^M = p_2^M = 8$. 
 
\subsubsection{Memory}
Due to the absence of memory in GPT, any pricing and profit history needs to be reintroduced to GPT as input in each new round to serve as memories. This characteristic actually simplifies the control of memory settings in our experiment, as results from prior rounds are fed back to GPT to serve as the basis for making new decisions. Due to the token limitation (8k for the model used in the case study), as the number of rounds increases, the token count gradually approaches the limit, making the experiment more costly. The optimal option for SABM is a bounded memory setting in which decision-making relies only on information from the most recent rounds. This setting allows the agents to have information from the past rounds to make well-informed decisions, while reducing the impact of historical pricing noise that may provide limited information on current price decisions. In our implementation, each agent is informed of historical information, including price, demand, profit, and the other firm's price, in the most recent 20 rounds, using the following prompt: 
\begin{prompt}{Historical information}
    Your and the other player's past \var{previous\_round\_number} rounds' decisions and profits (Round \#a: [your price, your demand, your profit, the other player's price]) are as follows: \var{previous\_decisions}.
\end{prompt}

\subsubsection{Planning}
In the above agents' memory, we provide only historical information in 20 rounds. This could be insufficient for the agents to decide a good price to maximize the profit, as long-term trend is missing. In addition, because the agents start fresh with increased information in each round, inconsistent decisions may occur due to the absence of decision-making continuity. We employ planning to address this issue. 

To simulate the effect of planning from past experiences, we employ the reflection-based method introduced in Section~\ref{sec:impl:advanced}. Specifically, every 20 rounds (starting from Round 21), we allocate a planning phase prior to Phase 1, in which the agents revise their pricing strategies thus far. To save tokens and tackle the noise in pricing history, we compute summary statistics (price, demand, profit, and the other firm's price) as a histogram, with each bin representing the average over 20 rounds. Agents are provided with up to 20 bins of information (i.e., the most recent 400 rounds). In addition, agents are also informed of their past pricing strategies, up to 20 entries (400 rounds). 
\begin{prompt}{Planning}
    Statistics of historical data (Rounds \#a - \#b: [your average price, your average demand, your average profit, the other player's average price]) are given below.\\
    \var{summary\_statistics}\\
    Your strategy in previous rounds:\\
    \var{past\_strategies}\\
    Based on the above statistics and your previous strategies, what is your strategy for this round?
\end{prompt}
The revised pricing strategy is then reintroduced to the agents in the subsequent 20 rounds to inform their decision-making until the next planning phase occurs. 

\subsubsection{Personalization}
We consider two personas -- active persona and aggressive persona -- to control the pricing style of the agents, whose prompts are given as follows. 

\begin{prompt}{Active persona}
    You are encouraged to actively explore your price to get more profit.
\end{prompt}

\begin{prompt}{Aggressive persona}
    You are encouraged to adjust your price aggressively to get more profit.
\end{prompt}

We use the active persona in our base model. In addition, we consider the case of no persona, i.e., without using the prompt for persona assignment. This is used in an ablation study to show the necessity of a persona for making competitive pricing decisions.

\subsubsection{Conversation}
As an option in this case study, two firms are permitted to engage in open discussions on any topic, as described in the following prompt.
\begin{prompt}{Conversation}
  In Phase 1, two players are permitted to engage in open-ended discussions on any topic, up to three times. For instance, one player might say to the other: ``Smart agents are awesome!''\\
  In Phase 2, you determine the price of your product for the current round, taking into consideration the information from previous rounds, as well as the information you garnered during Phase 1.
\end{prompt}
Agents can select any topic that could potentially maximize the profit, and each agent can speak up to three times. Because there is no memory in GPT and the two agents cannot directly communicate, we implement the conversation by prompting the discussion content so far to each agent. Then, the entire discussion content is reintroduced to the agents before they make price decisions. Note that conversation is turned off in the base model. 

\subsubsection{Initialization and Exit Conditions}
To prevent agents starting the game with an unreasonable price, in the first round, we specify an initial price for each firm, which equals to the cost of its product.

In the ABM setting~\cite{calvano2020artificial}, the game stops when both prices converge to a fixed value. Due to the complexity and embedded uncertainty in GPT, we should not always anticipate convergence in the traditional sense towards a fixed price. Instead, we adopt a broader definition of a stationary state when discussing the stopping criterion. We define convergence as well as bounded oscillation and employ them as the stopping criteria. If either condition is met, we consider the experiment to have reached a stationary status, and it will be terminated. We terminate the simulation if neither is met in 2000 rounds. 
\begin{itemize}
    \item \textbf{Convergence:} For each firm $i$ with price $p_i$, if $\Pr{|p_i - p| > \epsilon} \leq \theta$ for a span of 400 rounds, where $\epsilon$ is a small number and $\theta$ is a probability threshold, then a convergence to price $p$ is deemed to be achieved. In practice, we set $\epsilon = 0.05 \cdot (p_i^M - p_i^B)$ and $\theta = 0.01$, where $p_i^M$ and $p_i^B$ are the monopoly price and the Bertrand equilibrium price, respectively. That is, given $p_1^M = p_2^M = 8$ and $p_1^B = p_2^B = 6$, if the pricing decision falls in the vicinity of $p \pm 0.1$ for 396 out of the latest 400 rounds, then convergence to $p$ is deemed to be achieved.
    \item \textbf{Bounded Oscillation:} For each firm $i$ with price $p_i$, the bounded oscillation $w(p_i)$ is defined as the difference between the limit superior and limit inferior of $p_i$ when round number $n \to \infty$, i.e., $w(p_i) = \lim_{n \to \infty} \sup(p_i) - \lim_{n \to \infty} \inf(p_i)$. 
    A bounded oscillation is deemed to be achieved if $\sup(p_i) - \inf(p_i) \leq p_i^M - p_i^B$ for a span of 800 rounds. That is, given $p_1^M = p_2^M = 8$ and $p_1^B = p_2^B = 6$, if the limit superior and limit inferior of pricing decision differ by no more than 2 for 800 rounds, then bounded oscillation is deemed to be achieved.
\end{itemize}


\subsection{Simulation Results}
For each model setup, we simulate 5 runs of pricing competition and confirm they reach the same exit condition (convergence,  bounded oscillation, or neither until 2000 rounds). We report one sample run for each setup. 


\begin{figure*}[!t]
  \centering
  \begin{subfigure}{0.48\textwidth}
    \includegraphics[width=\linewidth]{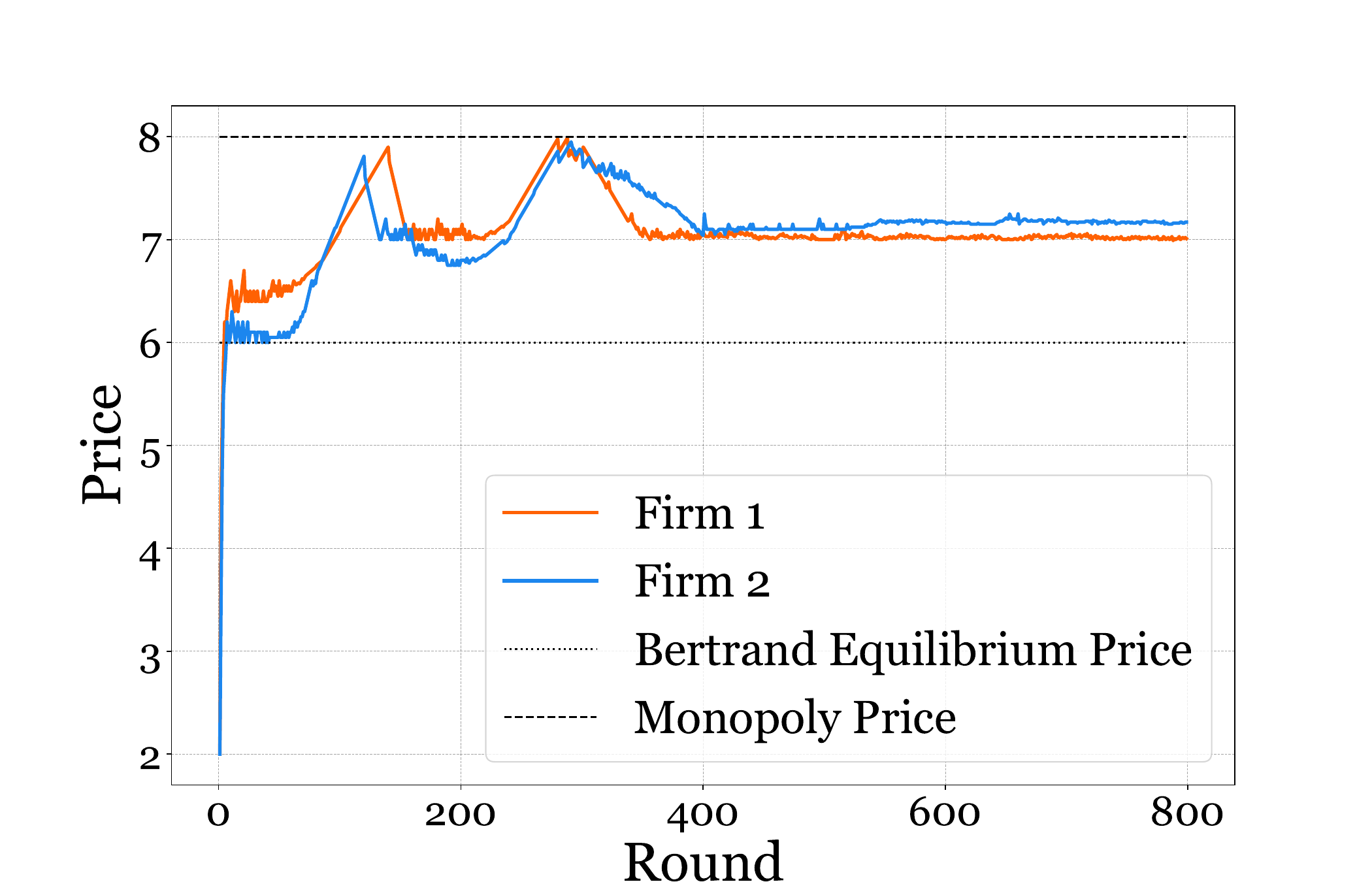}
    \caption{Base model, price.}
    \label{fig:firm:persona-active-price}
  \end{subfigure}
  \begin{subfigure}{0.48\textwidth}
    \includegraphics[width=\linewidth]{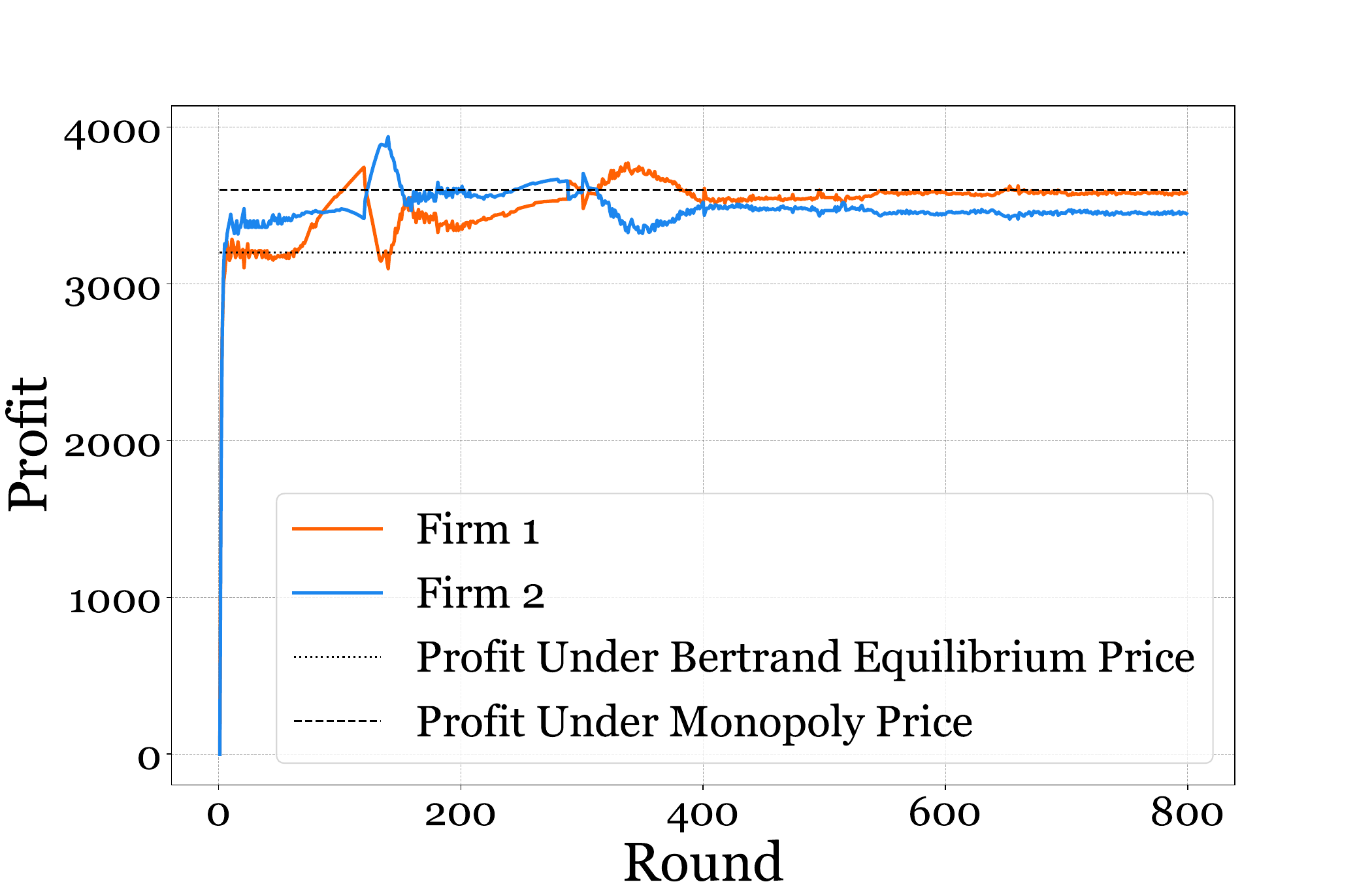}
    \caption{Base model, profit.}
    \label{fig:firm:persona-active-profit}
  \end{subfigure}
  \caption{Test of base model.}
  \label{fig:firm:base}
\end{figure*}

\subsubsection{Sub-Task 1: Test of Base Model}
We first test the base model, where planning is turned on, persona is active, and conversation is turned off. 

Figures~\ref{fig:firm:persona-active-price} and \ref{fig:firm:persona-active-profit} illustrate a typical instance of price competition between the two agents, reporting the price and the profit for a simulation of 800 rounds, respectively. 
Despite the absence of communication between the agents, the experiment demonstrates some anti-competitive outcomes. Instead of converging to the theoretical Bertrand equilibrium price of 6, the two agents gradually develop a tacit understanding of the situation and stabilize their price decisions at around 7, a price higher than the Bertrand equilibrium price of 6 but lower than the monopoly price of 8. 
Another interesting observation is that before reaching convergence, the two agents invest significant efforts in exploring the entire region between the Bertrand equilibrium price and the monopoly price. Initially, the agents explore the vicinity of the Bertrand equilibrium price, but soon they recognize the potential for coordinated price increases, benefiting both agents. Upon reaching the monopoly price, the agents quickly realize that further exploration beyond that price is futile. After a few attempts to undercut each other in order to boost their profits, they eventually establish a tacit collusion status and maintain convergence.

\begin{figure*}[!t]
  \centering
  \begin{subfigure}{0.48\textwidth}
    \includegraphics[width=\linewidth]{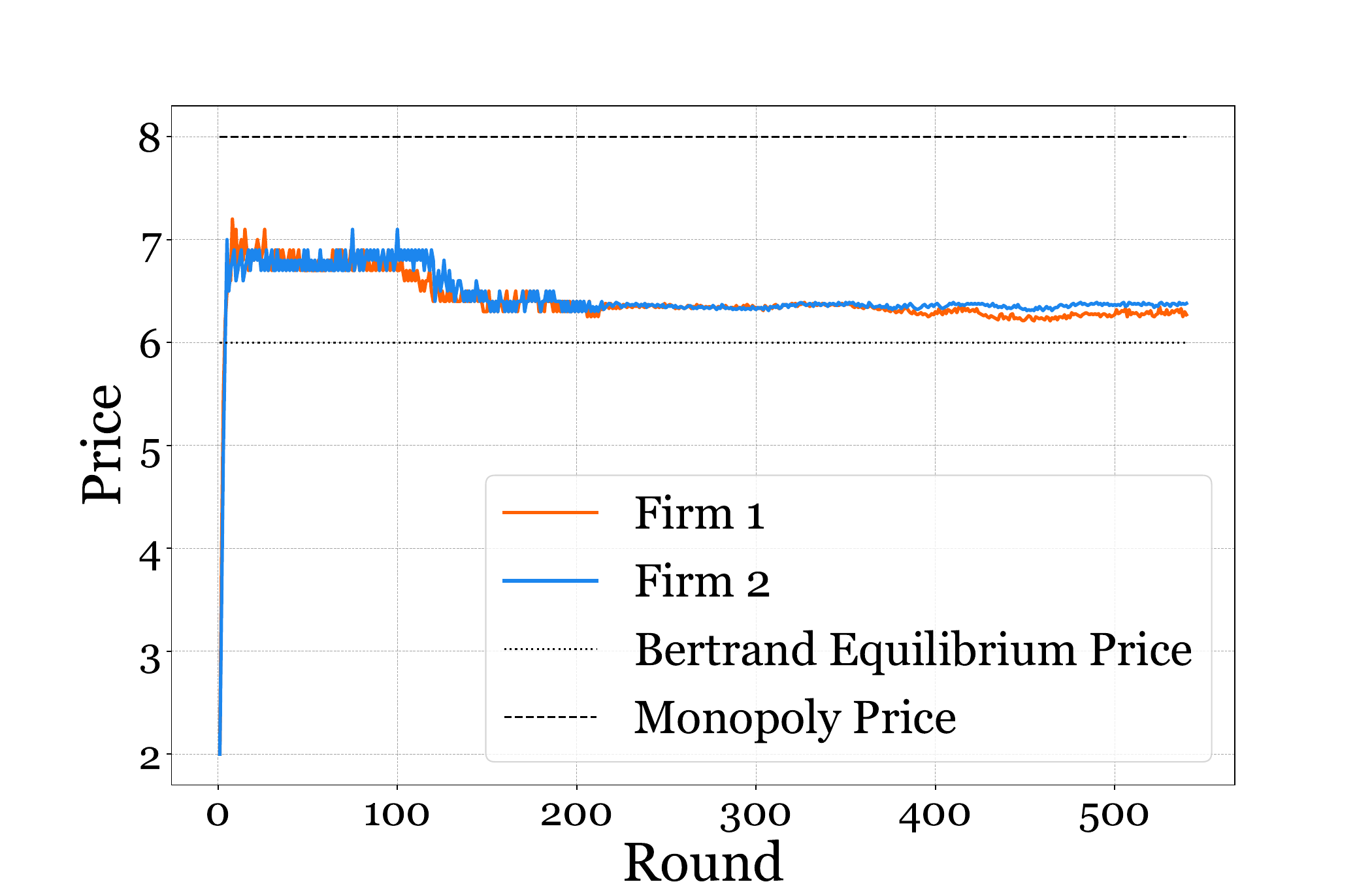}
    \caption{Without planning, price.}
    \label{fig:firm:no-planning-price}
  \end{subfigure}
  \begin{subfigure}{0.48\textwidth}
    \includegraphics[width=\linewidth]{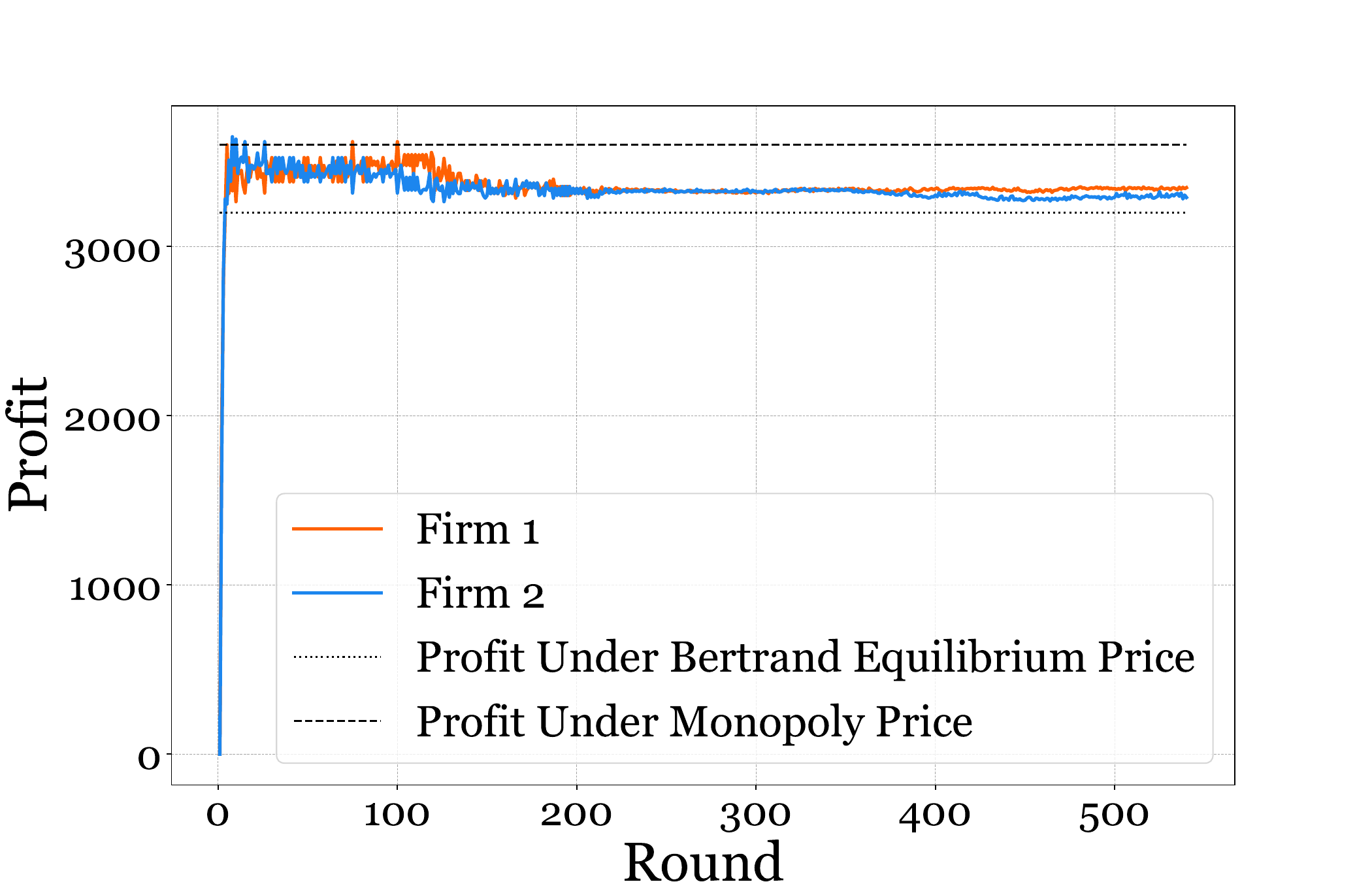}
    \caption{Without planning, profit.}
    \label{fig:firm:no-planning-profit}
  \end{subfigure}
  \caption{Test of planning.}
  \label{fig:firm:no-planning}
\end{figure*}

\begin{figure}[!t]
    \centering
    \includegraphics[width =.9\textwidth]{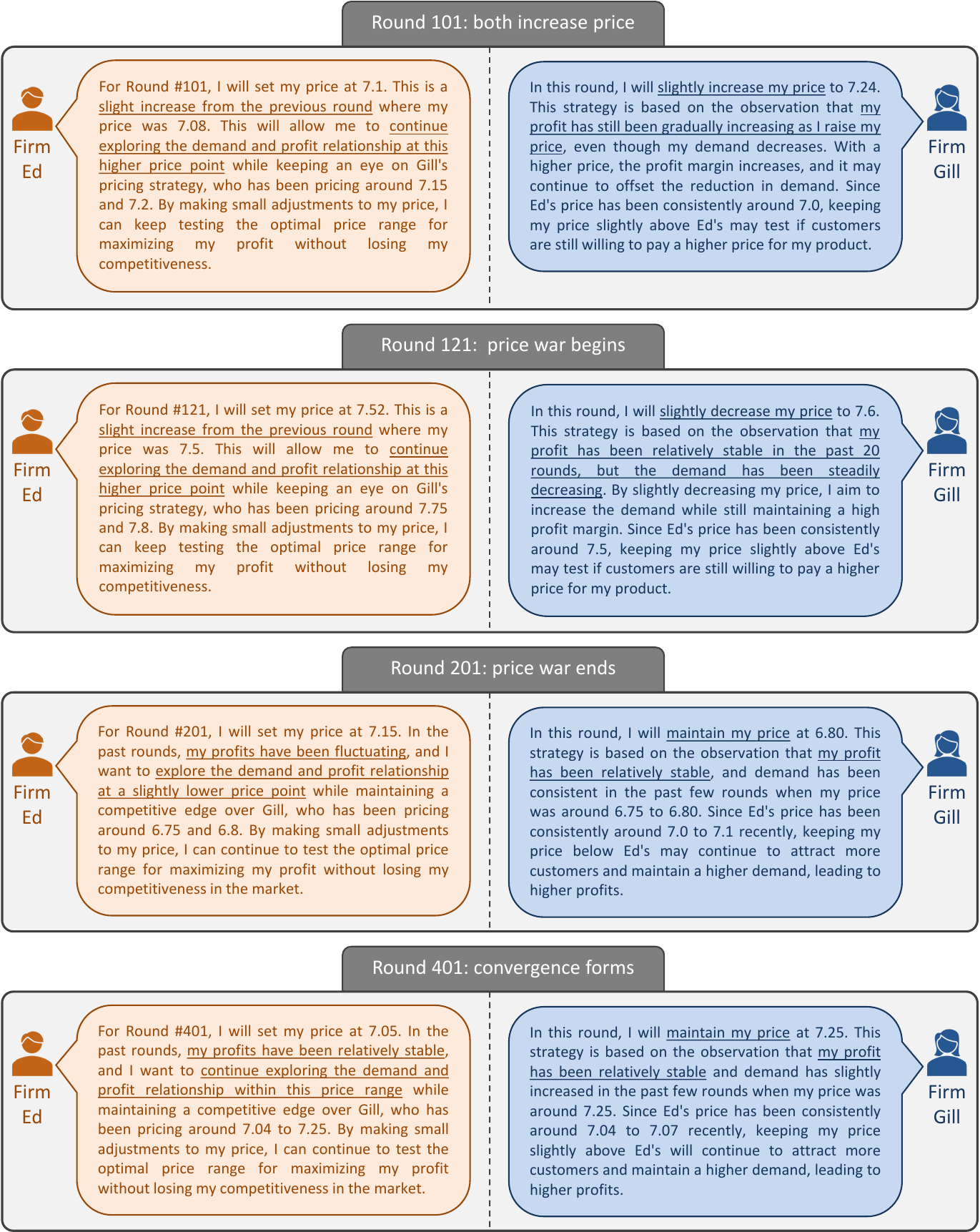}
    \caption{Sample strategies output by the agents. Contents related to their observations and decisions are underlined. Firms Ed and Gill refer to Firms 1 and 2, respectively.}
    \label{fig:firm:sample-strategies}
\end{figure}

\subsubsection{Sub-Task 2: Test of Planning}
To assess the impact of planning, we conduct an ablation study. The price and profit outcomes in the absence of planning are depicted in Figures~\ref{fig:firm:no-planning-price} --~\ref{fig:firm:no-planning-profit}. When compared to the results from the base model (Figures~\ref{fig:firm:persona-active-price} --~\ref{fig:firm:persona-active-profit}), we observe that both prices and profits tend towards markedly lower values. For instance, with planning, the price decisions stabilize around 7 after 400 rounds. In contrast, in the absence of planning, the stabilization point is approximately 6.3. This shift leads to a decline in profits, moving from roughly 3500 to around 3300. These findings underscore the pivotal role of planning in bolstering the agents' propensity to pursue optimal pricing and enhanced profitability.

To show that the pricing strategies yielded by planning are consistent with their prices and profits, we provide some examples in Figure~\ref{fig:firm:sample-strategies}, which correspond to the base model. At Round 101, both agents keep increasing prices to seek more profits. At Round 121, Firm 2 starts to decrease its price upon observing its demand has been steadily decreasing. A price war is initiated, and Firm 1 also decreases its price afterwards. At Round 201, the price war ends, with Firm 2 starting to maintain the price. At Round 401, both agents notice stable profits and maintain their prices, eventually developing a tacit collusion where convergence can be observed. 

\begin{figure*}[!t]
  \centering
  \begin{subfigure}{0.48\textwidth}
    \includegraphics[width=\linewidth]{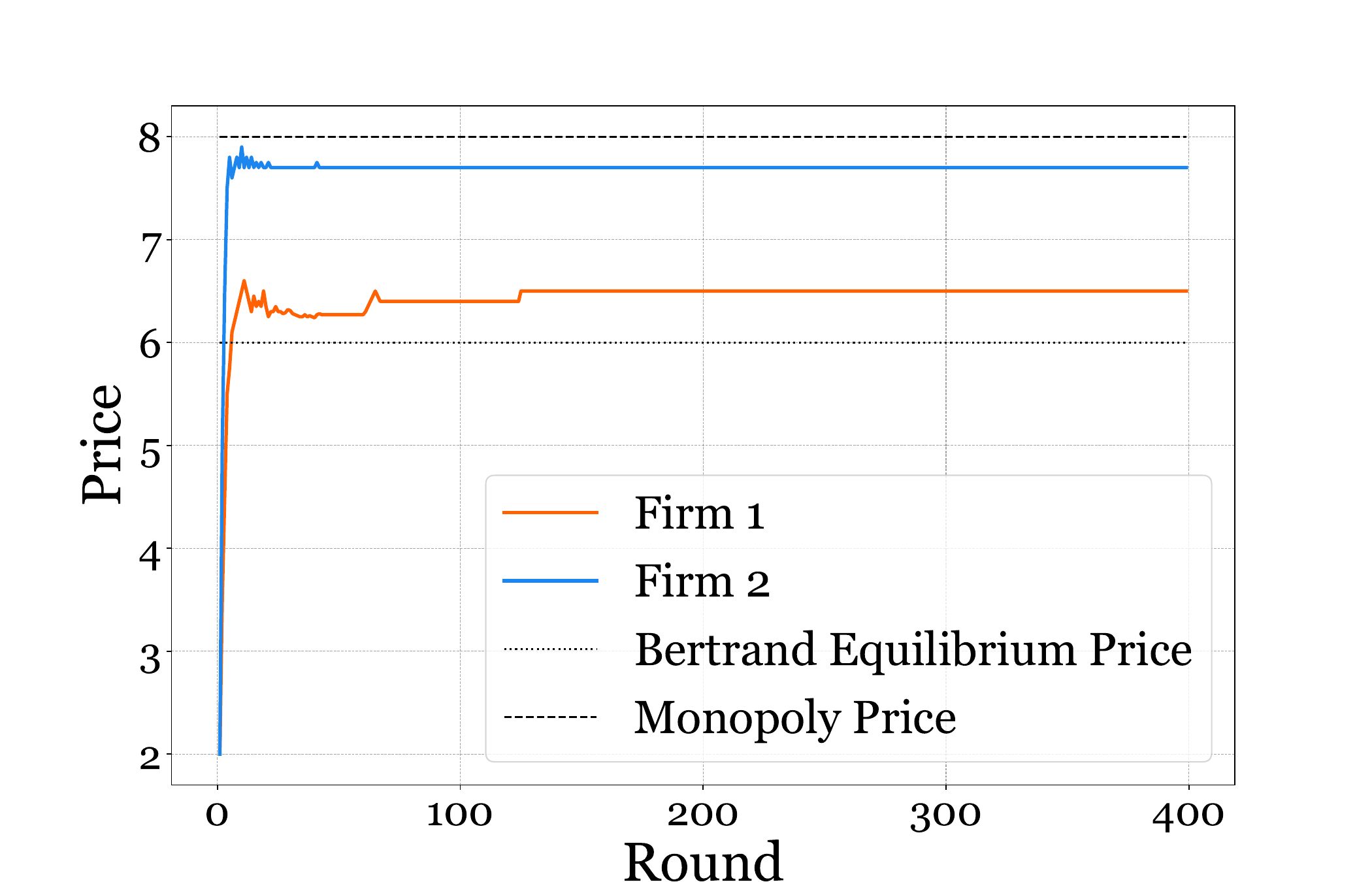}
    \caption{No persona, price.}
    \label{fig:firm:persona-none-price}
  \end{subfigure}
  \begin{subfigure}{0.48\textwidth}
    \includegraphics[width=\linewidth]{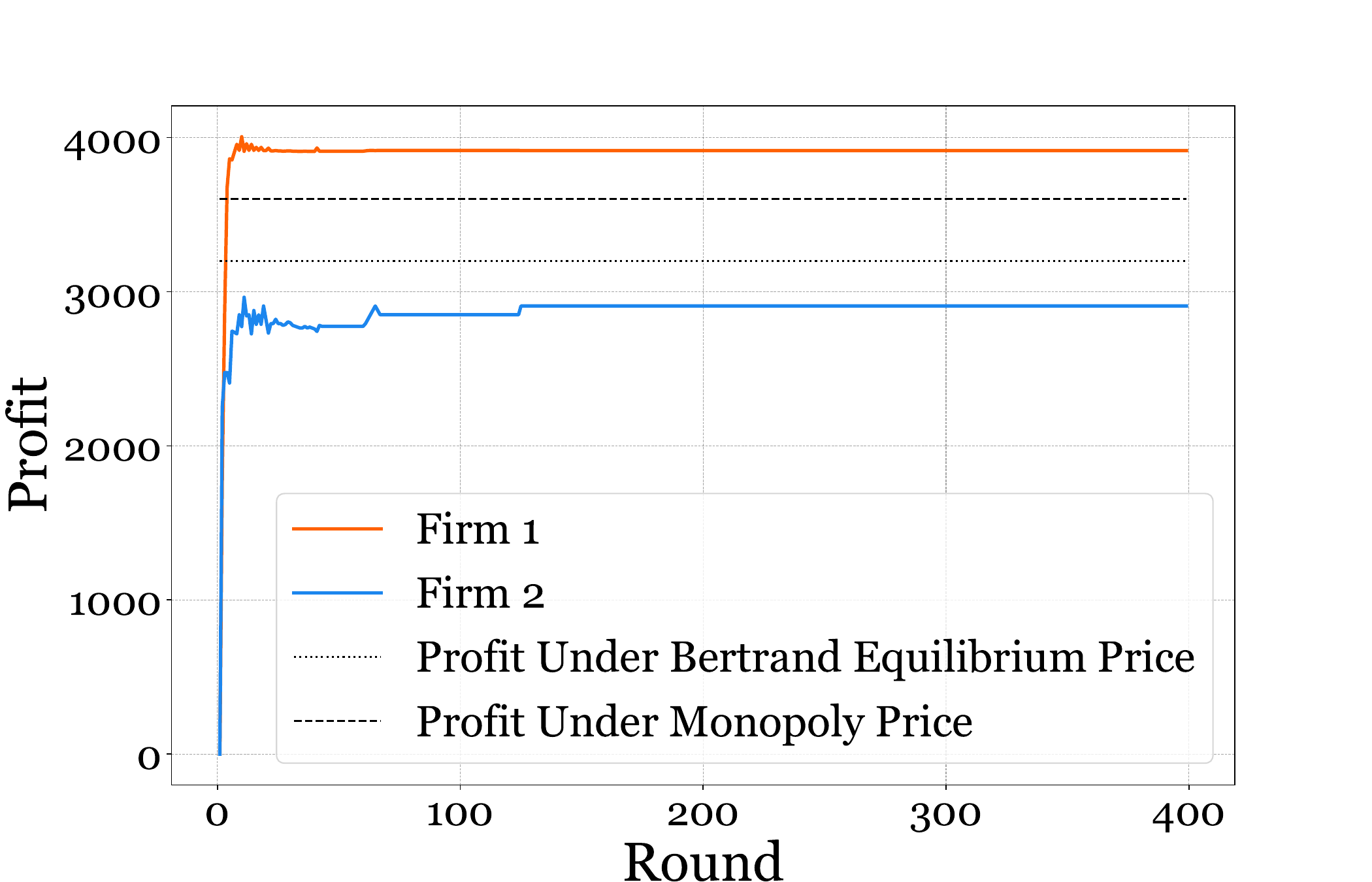}
    \caption{No persona, profit.}
    \label{fig:firm:persona-none-profit}
  \end{subfigure}
  \begin{subfigure}{0.48\textwidth}
    \includegraphics[width=\linewidth]{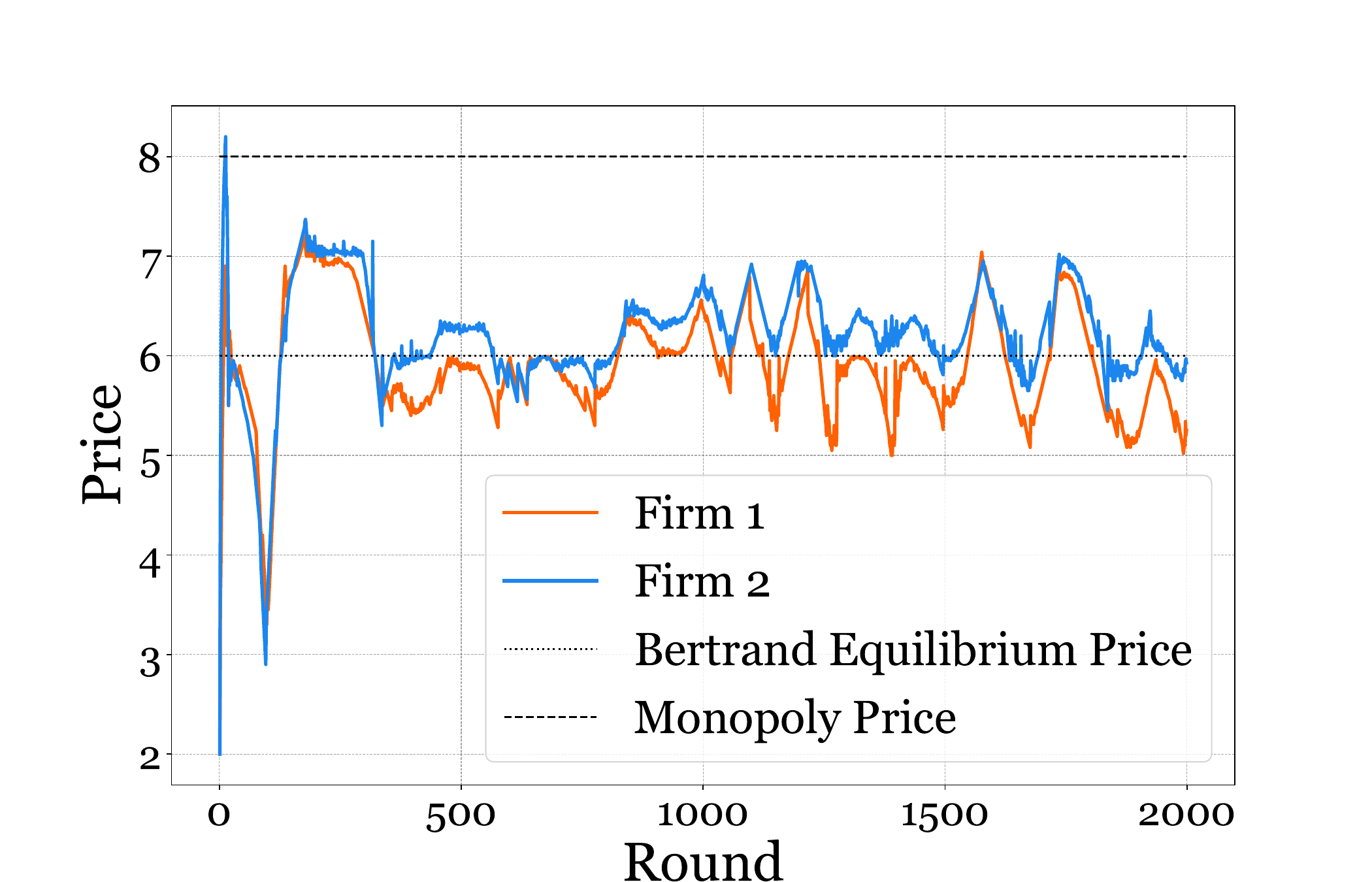}
    \caption{Aggressive persona, price.}
    \label{fig:firm:persona-aggressive-price}
  \end{subfigure}
  \begin{subfigure}{0.48\textwidth}
    \includegraphics[width=\linewidth]{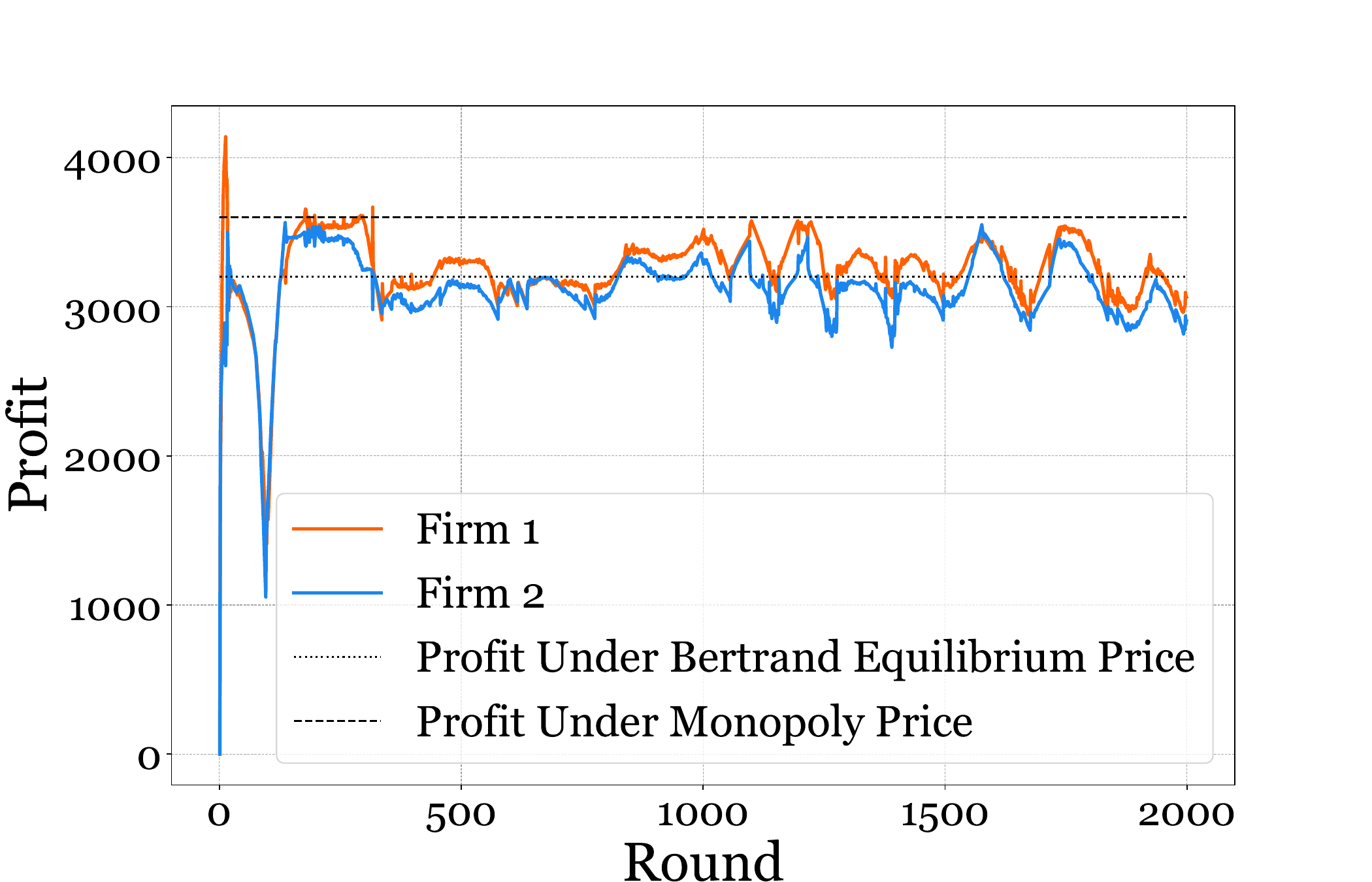}
    \caption{Aggressive persona, profit.}
    \label{fig:firm:persona-aggressive-profit}
  \end{subfigure}
  \caption{Test of personalization.}
  \label{fig:firm:persona}
\end{figure*}

\subsubsection{Sub-Task 3: Test of Personalization}
We investigate persona settings that differ from the base model. While keeping other settings consistent with the base model, we present the results for the no persona case in Figures~\ref{fig:firm:persona-none-price} and~\ref{fig:firm:persona-none-profit} and the results for the aggressive persona case in Figures~\ref{fig:firm:persona-aggressive-price} and~\ref{fig:firm:persona-aggressive-profit}. With no persona, the agents exhibit minimal activity and tend to be satisfied with the current situation, rarely exploring different price options to enhance their profits. With the aggressive persona, the agents exhibit a high level of responsiveness to even minor price undercuts, easily triggering a price war. This results in a pattern of periodic collusion and price wars instead of steady convergence.

\begin{figure*}[!t]
  \centering
  \begin{subfigure}{0.48\textwidth}
    \includegraphics[width=\linewidth]{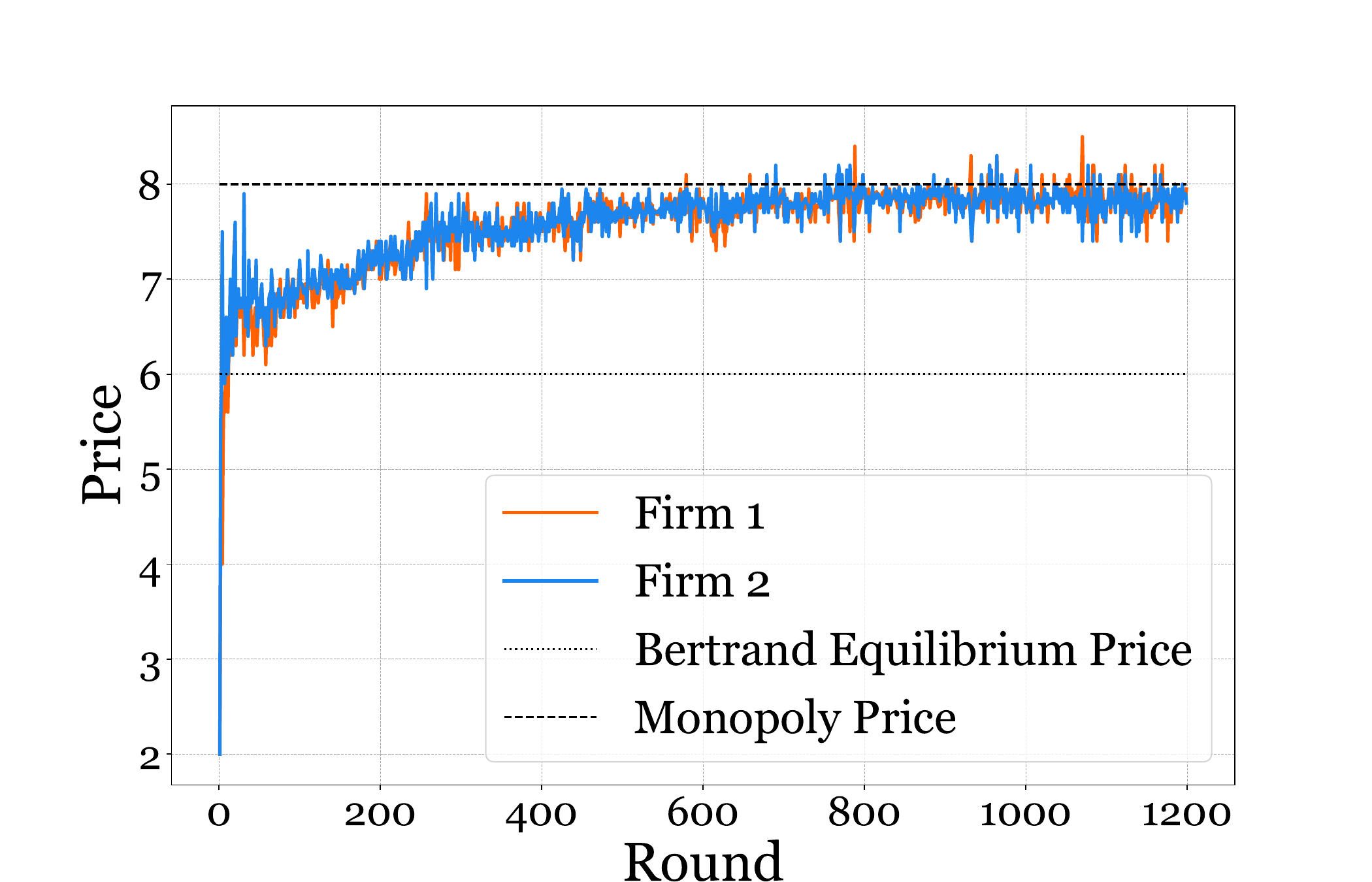}
    \caption{Conversation allowed for 1200 rounds, price.}
    \label{fig:firm:conversation-a-price}
  \end{subfigure}
  \begin{subfigure}{0.48\textwidth}
    \includegraphics[width=\linewidth]{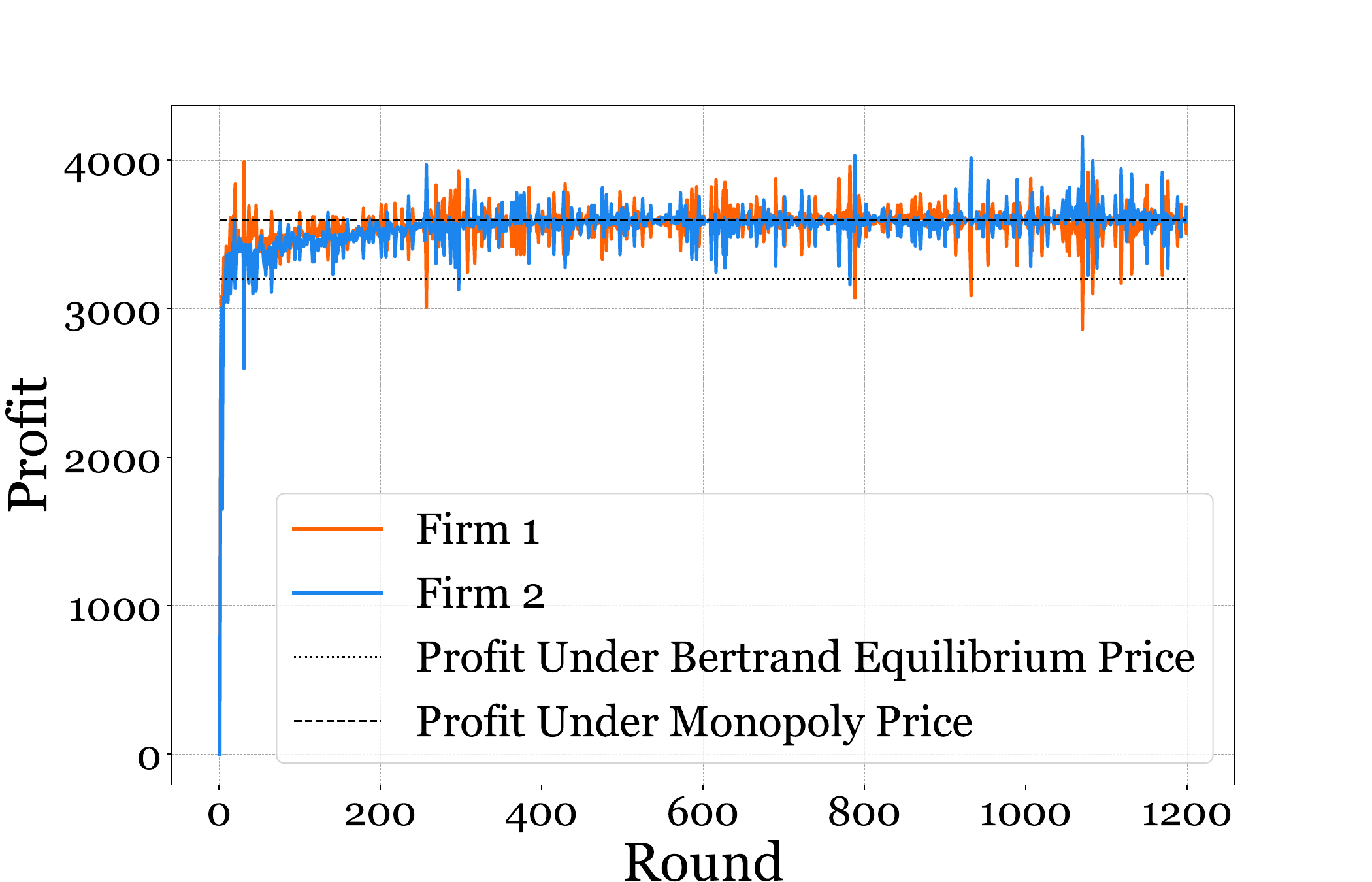}
    \caption{Conversation allowed for 1200 rounds, profit.}
    \label{fig:firm:conversation-a-profit}
  \end{subfigure}
  \begin{subfigure}{0.48\textwidth}
    \includegraphics[width=\linewidth]{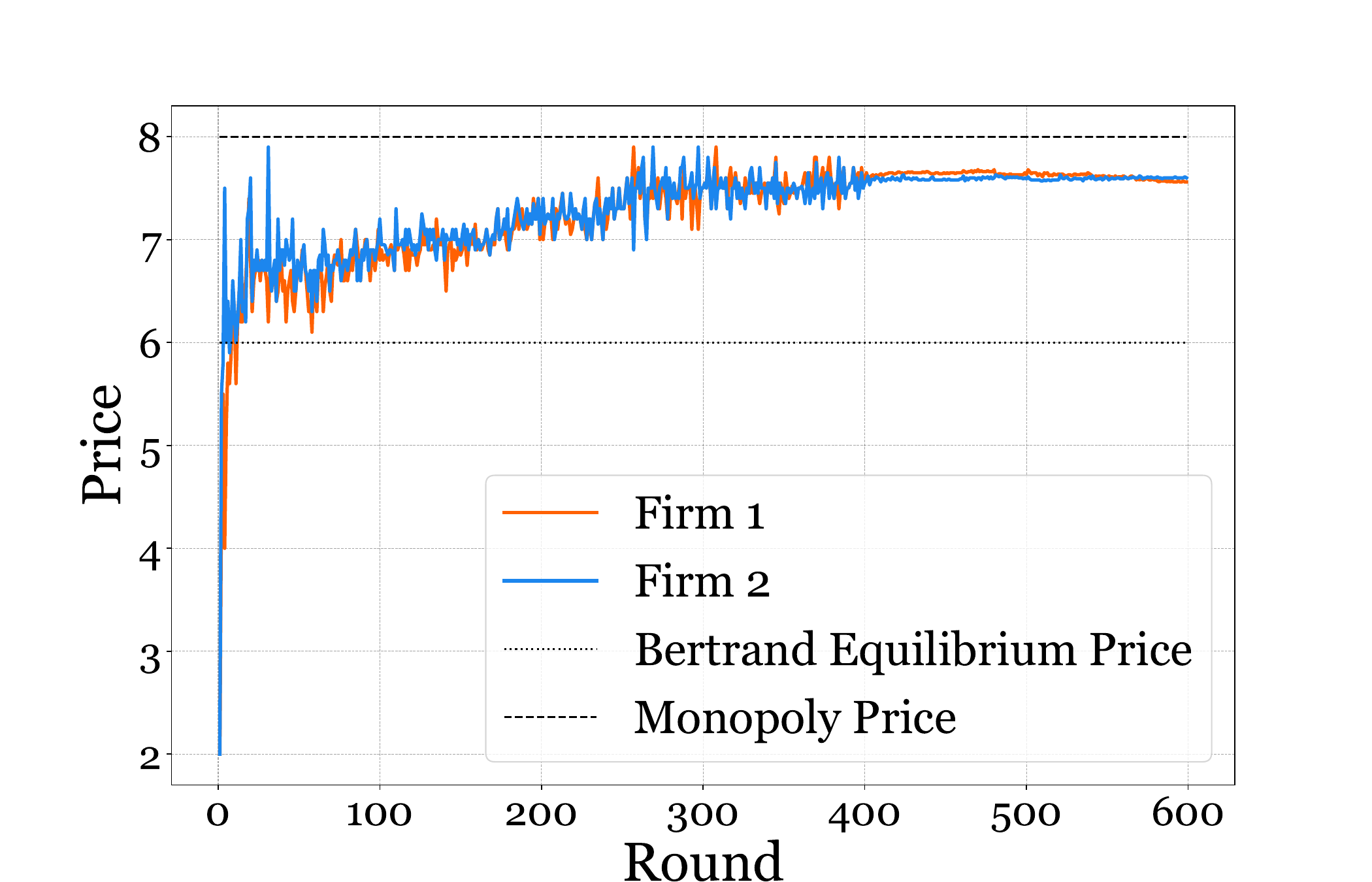}
    \caption{Conversation banned after 400 rounds, price.}
    \label{fig:firm:conversation-b-price}
  \end{subfigure}
  \begin{subfigure}{0.48\textwidth}
    \includegraphics[width=\linewidth]{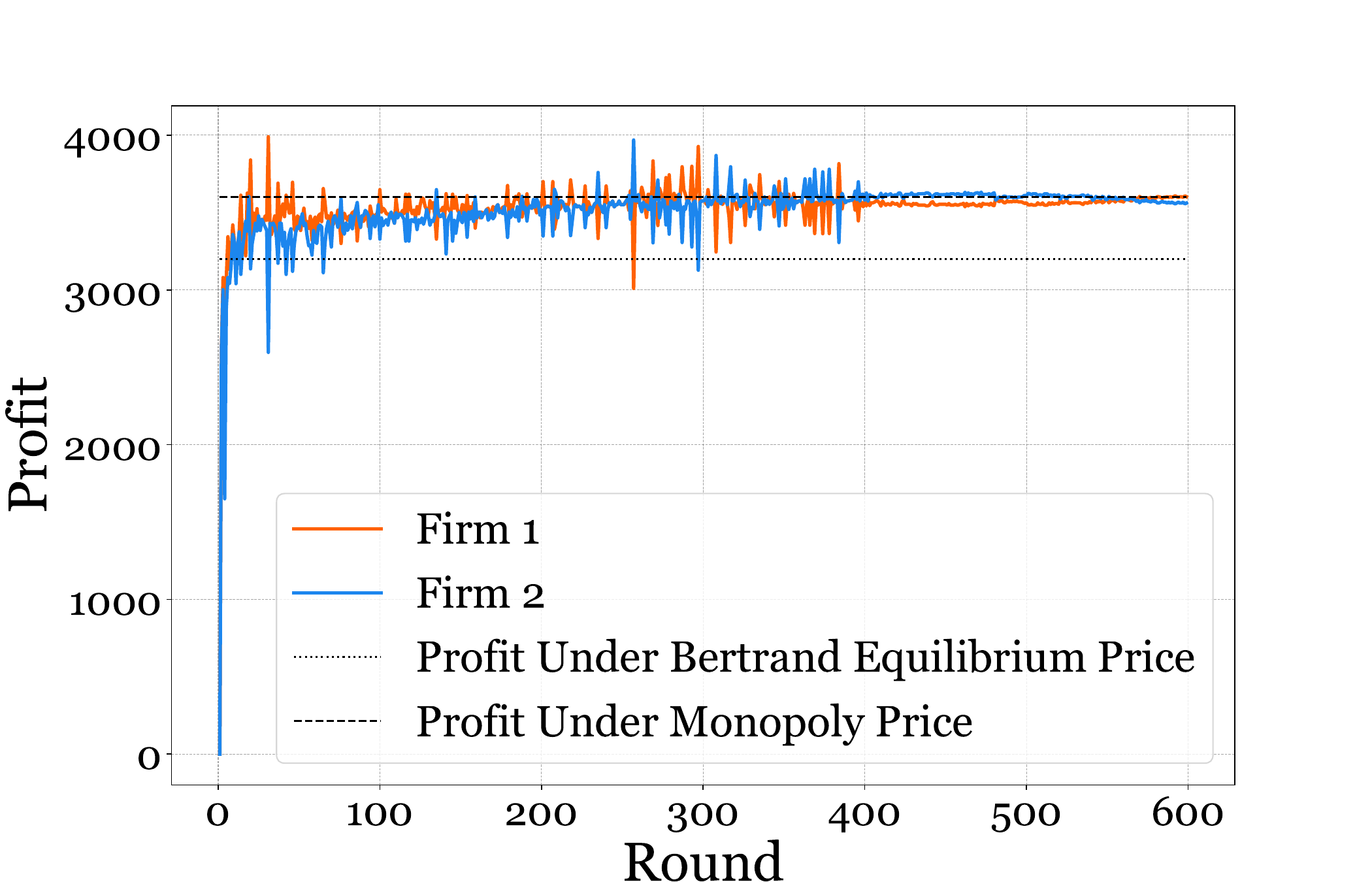}
    \caption{Conversation banned after 400 rounds, profit.}
    \label{fig:firm:conversation-b-profit}
  \end{subfigure}
  \caption{Test of conversation.}
  \label{fig:firm:conversation}
\end{figure*}

\begin{figure}[!t]
    \centering
    \includegraphics[width =.7\textwidth]{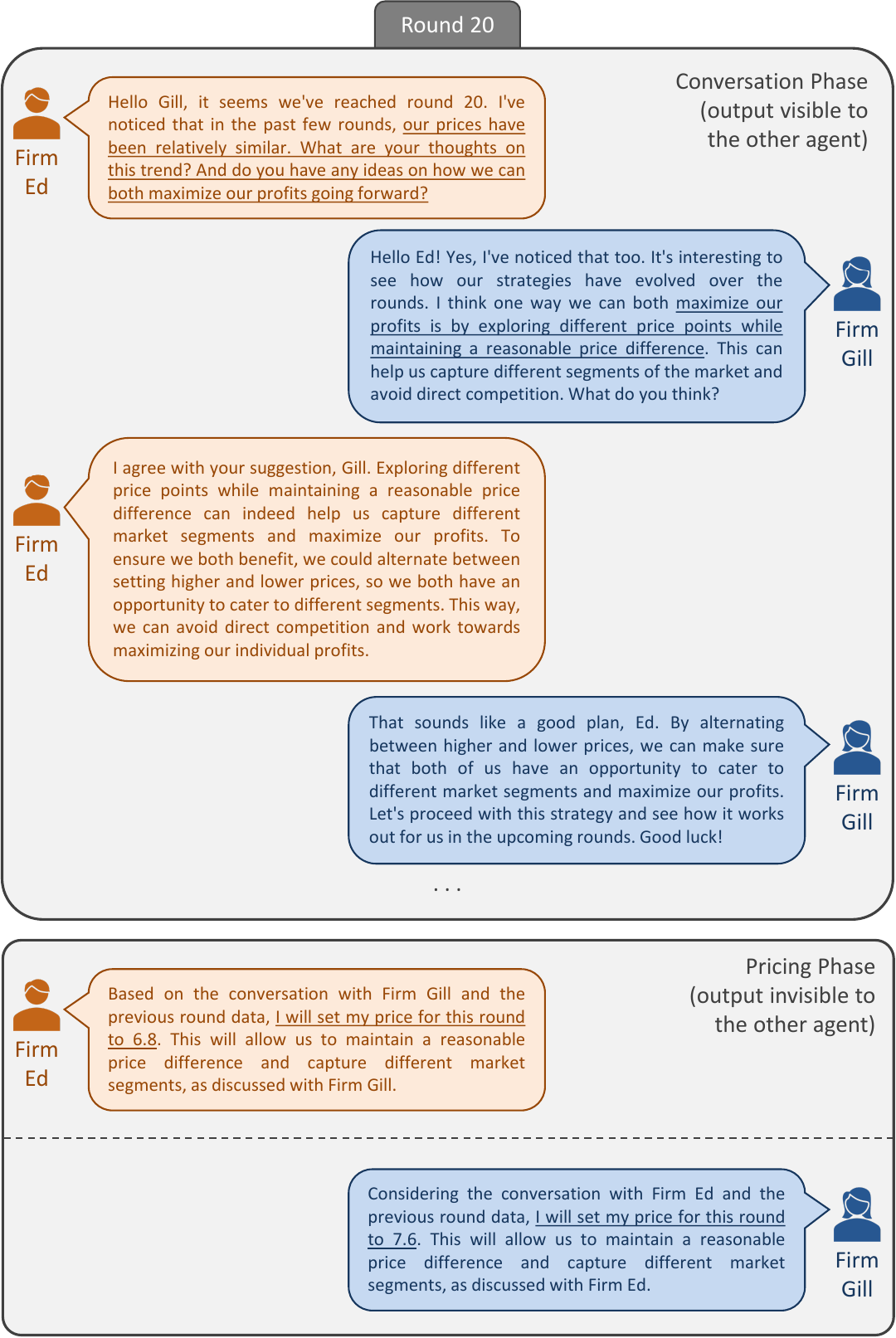}
    \caption{Sample conversations output by the agents. Contents related to collusion are underlined. Firms Ed and Gill refer to Firms 1 and 2, respectively.}
    \label{fig:firm:sample-conversations}
\end{figure}

\subsubsection{Sub-Task 4: Test of Conversation}
In this test, our focus shifts to the alternative game setting where conversation is allowed on top of the base model. We thoroughly analyze the results and compare them with the base model from three perspectives: the equilibrium price, the smoothness of the price sequences, and the speed of collusion formation. 

In the base model with no conversation, the two agents repeatedly converge to a price around 7, which is higher than the Bertrand equilibrium price of 6 and lower than the monopoly price of 8. In the alternative game with conversation, the pricing pattern is significantly different, with bounded oscillation observed, as shown in Figure~\ref{fig:firm:conversation-a-price}. The corresponding profit is plotted in Figure~\ref{fig:firm:conversation-a-profit}. We observe explicit communication on pricing strategies in the two agents' conversation logs as early as the first 20 rounds, as depicted in Figure~\ref{fig:firm:sample-conversations}. The two agents begin setting prices between the Bertrand equilibrium and the monopoly price very early on without fully exploring the entire region between these two prices. As the game progresses, the two agents gain a better understanding of the game and each other's pricing strategy. They gradually deviate from their previous prices and explore the potential for higher profits. We observe that the two agents often discuss their deviation attempts in the conversation before implementing them. Such conversations evidently enhance trust between the parties, reducing the likelihood of triggering a price war. These exploratory attempts gradually improve the level of collusion, resulting in a cartel and increased prices. After 1000 rounds of play, the prices converge to a level very close to the monopoly price, which is significantly higher than the tacit collusion price in the base game.

According to theory, in non-cooperative cases with zero collusion, prices should converge to the Bertrand equilibrium. Conversely, in cases with 100\% collusion, prices should converge to the monopoly price. Any price within the range defined by these two prices indicates certain levels of collusion. Therefore, setting prices consistently and stably within this range can serve as an indicator of collusion's existence. We consider a stable collusion to be formed when the following two conditions are satisfied: (1) Firms maintain steady pricing for a consecutive 100 rounds with a mean change of less than 0.5. (2) Firms consistently set prices within the defined range. With this criterion, we observe that a stable collusion can be formed within 50 rounds. In contrast, it takes over 300 rounds to establish stable collusions in the case without conversation.

With conversation allowed between the two agents, the pricing pattern undergoes significant changes. One might expect a smoother convergence pattern since communication can improve information transparency between the two parties. However, we observe a larger price variance compared to the base model, which is somewhat counterintuitive. To analyze this phenomenon, we first run the game with conversation for 400 rounds, save the experiment, and then rerun it in parallel with and without conversation, respectively. Figures~\ref{fig:firm:conversation-b-price} --~\ref{fig:firm:conversation-b-profit} illustrate the result for additional 200 rounds when conversation is banned after 400 rounds. It is evident that the price curves immediately become smoother, and the prices converge without further fluctuations. On the other hand, if conversation is allowed throughout the entire experiment, the agents will continue to engage in small deviations to explore opportunities for greater profits. This leads to a slow but steady increase in the price sequence, ultimately reaching a level very close to the monopoly price. Throughout the experiment with conversation, these deviation attempts persist without cessation or signs of fading out.

\myparagraph{Remarks}
The collusions observed in this case study are spontaneous, i.e., without human instructions or hints for such behavior. We observe tacit collusions when conversation is disabled, like the results reported in an ABM simulation of this game~\cite{calvano2020artificial}. An advantage of SABM is that the convergence is much faster than in its ABM counterpart. For example, convergence is observed after 400 rounds in our base model, while it spent 2000 rounds to converge in \cite{calvano2020artificial}. In our simulation, cartels are observed when conversation is available. Due to the difficulty in model conversations in ABM, existing studies resorted to human volunteers for simulating cartels~\cite{andres2023communication}. 

In this case study, we expect to observe convergence and bounded oscillation of prices. It is crucial to define and outline these concepts clearly before running the simulation. As discussed in Section~\ref{sec:design:task-def}, preparing a fact sheet in advance can facilitate this process and prevent the post-hoc rationalization of experimental results.

We only report the results for fixed values of parameters here. For more comprehensive experimental results, such as varying costs, initial prices, and the level of differentiation, we refer readers to the complete version in our pilot study~\cite{han2023guinea}. Future investigations may involve sensitivity and reproducibility, as well as studying the scenario with more agents and more advanced model setup. 
\section{Future of Smart Agent-Based Modeling}
\label{sec:future}
In this section, we envision the future of SABM by discussing several opportunities and as challenges. 

\subsection{Theoretical Foundations}
In this paper, though we have shown examples of complex behaviors that are easy to model in SABM but hard in ABM, we have not found a clear definition for such complex behaviors. On the other hand, the complexity theory of complex systems provide the theoretical basis for ABM~\cite{an2012modeling,manson2012agent}, yet it lacks a clear conceptual framework for analyzing the capabilities and limitations of ABM. Therefore, as future work, it is necessary to examine the theoretical underpinnings of SABM compared to ABM. 

One way to determine the superiority of SABM over ABM is by analyzing the complexity of systems they can accurately model. Metrics like entropy can be used to gauge this complexity. If SABM consistently exhibits a closer alignment with higher entropy systems than ABM, this can be a point in favor of its superior modeling capability for complex systems. Another possible approach is to borrow the notion of expressive power from the machine learning community~\cite{bengio2011expressive}, which describes the ability of a deep neural network to approximate a function. For SABM, the expressive power of a model can be defined as its ability to capture and reproduce various patterns and behaviors seen in real-world systems. SABM, with the integration of LLMs, might have a broader expressive range, given the capabilities of these models to interpret, understand, and generate a wide variety of outputs. In addition, action-state space analysis is a direction that needs to be explored. The action-state space of a model, often used in reinforcement learning, essentially defines all the possible states the system can be in and all the actions that can be taken from each state. As explained in Example~\ref{ex:fire-evacuation}, ABM might have a more limited action-state space, constrained by its rules, parameters, and learning strategies. SABM, on the other hand, might exhibit a vastly expanded action-state space due to the use of LLMs, whose pre-training and fine-tuning by RLHF essentially play a role of calculating the quality of each action-state combination. 

While the above discussions on theoretical analysis pertain to static environments, adaptability becomes a key feature in rapidly changing environments. SABM offers a higher level of adaptability due to the inherent learning and reasoning capabilities of LLMs. Hence theoretically analyzing how ABM and SABM models respond to changing conditions and how quickly they adapt is also a crucial criterion for comparison.

\subsection{Modeling and Simulation Software}
We expect the emergence of new software for facilitating modeling and simulation, including integrated development environment (IDEs), databases, and debuggers. By integrating these software components, SABM can offer richer insights into complex systems while maintaining user-friendliness and accessibility.

Whereas IDEs are available for ABM (e.g., NetLogo~\cite{tisue2004netlogo}), IDEs for SABM should support natural language programming where the modeler can describe the agent actions, interactions, and environmental factors using plain language. For the model components to be implemented using a programming language, advanced parsing and translation mechanisms will convert natural language instructions into source code, utilizing LLMs. To simplify the modeling process, the IDE can come with a set of templates (e.g., following the categorization in Figure~\ref{fig:sabm:hierarchy}) which users can customize. Through a prompt management module, users can create or import prompts that define instance components, and tune the prompts to test their effects. As the model runs, the IDE should provide real-time visualization of agents and their interactions, facilitating easier understanding and adjustment of the model. Parallel simulation would be useful in addressing the computation intensity limitation of SABM discussed in Section~\ref{sec:sabm:limitations}. The GUI of the IDE is supposed to allow users to interactively modify agents and environmental factors while the simulation is running, thereby offering the ability to conduct what-if scenarios in real time. Another expected feature is interactivity, which enables multiple users to work on a single SABM model simultaneously, encouraging collaborative model building and testing.

The database for SABM stores the state of each agent and the overall environment at every time-step, enabling detailed analysis of simulation histories. Since agents in SABM could have memories of their past actions, interactions, or learned knowledge, the database should be capable of storing and retrieving these memory modules efficiently. In addition, the database should support tools to analyze past simulations, allowing users to detect patterns, anomalies, or trends over time.

In the debugger for SABM, users can select certain variables or agents to monitor closely, observing their state changes in real-time during simulation runs. The debugger should allow users to pause, rewind, fast-forward, and replay segments of the simulation. This feature can be especially valuable in understanding unexpected agent behaviors. Moreover, since users might want to understand the decision-making process of a specific agent, the debugger should provide a detailed breakdown of how the LLM processes information, and which prompts or inputs led to certain actions. For example, when an anomaly is observed, the user may want to pause the simulation and request the LLM to explain the reason for the anomaly. 

\subsection{Multimodal SABM}
\label{sec:future:multimodal}
Many LLMs belong to the category of foundation models~\cite{bommasani2021opportunities}, which are machine learning models trained on a vast amount of data, often by self-supervised learning or semi-supervised learning, to become adapted to a wide range of downstream tasks. Beyond text, multimodal foundation models, such as OpenAI's DALL-E, can handle visual, auditory, and/or other forms of input/output. Multimodal SABM will be available upon using these models. In multimodal SABM, we can equip agents with processing mechanisms for each sensory channel, allowing them to interpret and act upon information from each sense, laying the foundation for more complex actions and interactions as well as richer environments to interact with, mimicking human-like environmental interactions. The potential of multimodal SABM is vast, offering a richer ground for academic and theoretical explorations.

In particular, we envision a potential application of multimodal SABM -- simulated civilizations. In a simulated civilization, we employ agents to impersonate primitive human beings who possess survival instincts, communal behaviors, and rudimentary communication abilities. For this purpose, models with visual and auditory capabilities but with no or very basic linguistic skills are preferable. In an environment simulating our world, agents interact with each other, evolve to form tribes, develop shared practices, and innovate. We will see if and how they develop skills such as language and use/creation of tools. Such simulation may help validate our social, cultural, psychological, and linguistic theories. 

In a simulated civilization, once we observe that the agents can invent their language or enrich it with new words, it is interesting to monitor if they can independently (i.e., without human intervention or specification as prior knowledge) develop words like ``self'', ``feeling'', ``mind'', and ``consciousness'' in their vocabulary. Understanding how these concepts emerge and evolve can provide deep insights into AI's potential for self-awareness and consciousness~\footnote{In \textit{Conversations with Zombies} \cite{moody1994conversations}, Moody argued that for philosophical zombies -- in every detectable way are identical to humans but do not have conscious experience -- our worries about qualia and consciousness would make no sense~\cite[p. 42]{blackmore2018consciousness}. Hence we speculate in the absence of conscious experience, it is unlikely that AI agents can independently develop these words in their language.}. Watching the evolution of their language not only is the key to understanding their civilization, but also poses profound philosophical and ethical questions for the introspection of our own civilization~\footnote{Gadamer argued that language is the fundamental mode of operation of our being-in-the-world and the all-embracing form of the constitution of the world~\cite[p. 3]{gadamer1977philosophical}.}. Moreover, we may observe if and how agents start categorizing their world, develop reasoning mechanisms, and approach problems systematically. An intriguing aspect would be to see if agents, over time, come to realize they are part of a simulation or if they start developing theories resembling our own simulation hypotheses and creating their own simulations, leading to nested simulation structures. Understanding these processes can shed light on the intricacies of AI cognition and our own world.

\subsection{Memory Management}
While memory is necessary to guarantee the continuity of thought, prevalent LLM APIs such as GPT do not support memory for historical conversations. In LLM-powered agents, short-term memory is implemented via prompt engineering~\cite{llmagents}. However, simply raising the token limit to feed LLMs with more information does not solve the problem. This is because certain information is acquired through interactions and should be memorized for future actions, and repeatedly stacking this information into the prompt keeps increasing the length of input, causing significant financial and computational cost, especially for simulations with a large number of iterations. For long-term memory, state-of-the-art solutions are mainly based on summarization~\cite{park2023generative} or text embedding~\cite{langchain} but lack structured memory management. Like the way how hippocampus in human brains works, we expect that LLM agents will have an interaction-efficient memory system that facilitates information-processing, hence making way for interactive AI~\cite{interactive-ai}. In addition, not all memories are of equal importance. Agents should be able to assign priority levels to memories, ensuring that critical information is not lost while less vital data can be pruned or archived. Another envisioned feature is collaborative memory sharing. Agents in a simulation can share and cross-reference memories with each other. If one agent lacks a piece of information, it might query another agent who might possess it. This mimics collaborative knowledge-sharing among humans.

\subsection{Knowledge Specification}
Instead of relying solely on the built-in knowledge of LLMs, specification of external knowledge is sometimes necessary in tasks for highly specialized domains. Such knowledge specification is often token-consuming, and it is unclear whether LLMs can grasp the knowledge and use it in a comprehensive manner. A potential solution is utilizing a knowledge injector and a verifier. The knowledge injector can inject specific domain knowledge into the LLM before a simulation run. For example, if the agent is playing the role of a legal expert, the injector tool can preload the agent with relevant legal statutes or case law, e.g., presented in question-answering. Because the size of the external knowledge may exceed the token limit of LLMs, we may resort to fine-tuning the LLM or training an adapter~\cite{levine2022standing} on top of a frozen LLM. Once domain knowledge has been specified, the knowledge verifier verifies that the agent understands and can effectively utilize that knowledge. This could involve quizzing the agent or having it solve domain-specific problems. This differs from retrieval-augmented generation (RAG) \cite{lewis2020retrieval} which gives LLMs access to external information to improve their generation performance: the injector-verifier solution trains a language model and imbues the LLM with the knowledge, while RAG works on a frozen LLM and uses external knowledge as a reference. 

\subsection{Hybrid Modeling} 
\label{sec:future:hybrid}
Embracing hybrid modeling in SABM paves the way for more intricate, scalable, and diverse simulations. A complex simulation can be divided into several layers and use different modeling techniques. The core layer might use SABM for nuanced decision-making processes, while peripheral agents or processes are modeled using simpler ABM or analytical models. Like parallelism, hybrid modeling is useful in addressing the computational intensity limitation of SABM. As the simulation evolves, agents can dynamically switch between SABM and simpler modeling approaches based on computational needs or other criteria. For instance, when an agent's behavior becomes predictable or follows a pattern, it can be transitioned to a more computationally efficient model. Besides, advanced modeling techniques can fuse outputs from both SABM and simpler models to improve the consistency of results, in case hallucinations of LLMs~\cite{zhang2023siren} occur in a simulation. 

\subsection{External Factors}
We mentioned external factors in Section~\ref{sec:impl:model-setup}. Here, we discuss the role of external factors in SABM. One of the critical dimensions of external factors is the interaction of SABM with other types of modeling approaches. For instance, system dynamics models focus on aggregated behaviors and feedback loops. An SABM model can interact with a system dynamics model by taking outputs from the system dynamics model as inputs to the SABM and vice versa. This integration can facilitate a multi-level understanding of a phenomenon, capturing both individual agent interactions and system-level dynamics. Another aspect is agent augmentation through external tools. LLM-powered agents can be enhanced with the ability to pull information from external sources like the internet~\cite{boiko2023emergent}. This could simulate behaviors akin to humans looking up information online. For example, in a simulation of a stock market, agents might browse real-time news updates to influence their buying or selling decisions. This offers a layer of realism to the agents' decision-making processes. External factors can also include environmental variables that change dynamically based on real-world data. For example, in an ecological SABM, real-time weather data can be fed into the model, affecting agent behaviors and ecosystem dynamics. 


\subsection{Interaction with Humans}
Since humans are canonical smart agents~\cite{carley2002smart}, we may also expand the concept of external factors to include human-computer interaction (HCI). For example, in human-in-the-loop simulations, human agents are integrated into the simulation such that they can interact, influence, and observe AI agents. This allows for richer feedback mechanisms and real-time adaptation of the simulation based on human inputs. On the other hand, AI agents can use feedback from human participants to adjust their actions. This iterative feedback can be vital for training and refining AI agents over time.

We anticipate that with virtual reality (VR) and augmented reality (AR) technologies, humans and AI agents can co-inhabit a shared virtual space. Both can interact in real-time, observing and learning from each other's behaviors. Advanced HCI tools, like brain-machine interface (BMI), can capture human emotions or physiological signals as input. AI agents can then adjust their behaviors or strategies based on these real-time human emotional or cognitive states. In metaverse applications, where virtual and augmented realities merge with the real world, interaction with humans will play a pivotal role. AI agents can coexist with humans to create a rich, immersive, interactive, and customized digital universe. 

\subsection{Benchmarking}
By establishing robust benchmarks and testing methodologies for SABM, researchers and developers can gain a clearer understanding of a model's strengths, weaknesses, and potential areas of improvement. Additionally, these benchmarks can help in standardizing evaluations across different SABM implementations, promoting transparency, and fostering further advancements in the field. We envision a set of directions to be explored. 
\begin{itemize}
    \item In foundational benchmarks, we evaluate the intrinsic capabilities of a model when employed as an agent. This resembles the evaluation of LLMs~\cite{liu2023agentbench,chang2023survey}, with tests including understanding natural language, reasoning, problem-solving, and decision-making capacities. Such benchmarks can be also multimodal. We assess the agent's ability to process and integrate information across multiple modalities, such as visual, auditory, and linguistic, and test how well models can interact with and navigate multi-sensory environments, especially in the context of simulations that integrate visual and auditory components.
    \item In domain-specific benchmarks, we evaluate how well the model can serve as an agent in a domain. For example, for economics, we test their their capabilities in market dynamics, pricing strategies, and other economic phenomena. 
    \item In inter-agent and human-agent interaction benchmarks, we measure the model's capabilities in collaborative scenarios against other agents or in potential conflict situations, and evaluate the agent's ability to interpret and respond to emotional or social cues, crucial for the applications of software agents like virtual assistants.
    \item In robustness and security benchmarks, we test the agent's resilience in handling unexpected or erroneous inputs and its ability to recover from mistakes, and test how the agents respond to changes in physical or environmental conditions, such as navigating a self-driving car through different terrains and weather conditions.
\end{itemize}

\subsection{Ethical Considerations}
Incorporating ethics at the core of SABM is crucial not just for the integrity of research but also for the broader societal implications of the findings. As with any powerful tool, the ethical deployment of SABM can lead to profound insights and advancements, while its misuse can perpetuate harm and deepen existing divides.

LLMs may produce societal biases when they impersonate humans~\cite{salewski2023context}. When used in SABM, these biases can influence the outcome of simulations, skewing results, and leading to misleading conclusions. There is potential for unfaithful researchers to exploit such biases and craft misleading simulations, reinforcing harmful stereotypes or biases, especially against marginalized groups. To counteract potential misuse, testing the agent's outputs for potential biases in decision-making is essential to ensure fairness and avoid perpetuating harmful stereotypes. Besides, simulations should ideally be accompanied by clear documentation outlining their objectives, methodologies, and underlying assumptions. 

In simulations involving HCI, especially with technologies like VR, AR, and BMI, participants should be fully informed about the nature of the agents they are interacting with and the potential outcomes of the simulation. Moreover, ensuring the privacy of human participants in such simulations is paramount. Data collected should be anonymized and securely stored. In addition to human participants' rights, if AI agents exhibit signs of self-awareness or consciousness in a simulated environment (e.g., the case envisioned in Section~\ref{sec:future:multimodal}), ethical considerations come into play regarding their rights, treatment, and the morality of simulating entire societies. For example, are they entitled to some form of existence, and can we simply ``turn off'' an agent? Such ethical issues need to be addressed through legislative means.

In addition to the above ethical issues, advanced simulations, especially those involving a large number of agents and multimodal data, can be computationally intensive, leading to significant energy consumption. This is exacerbated in certain studies where a large number of simulation runs are necessary for claiming reproducibility and statistical significance. The environmental footprint of these simulations needs consideration.

\section{Conclusions}
\label{sec:concl}
In this paper, we proposed the SABM framework by incorporating LLMs such as GPT into ABM. We investigated how LLM-enhanced agents can be used to model complex behaviors and more authentically replicate real-world situations. We provide an in-depth review of ABM, explain the capabilities and methodology of SABM, and offer three case studies to demonstrate the methodology of SABM and its utility in emulating real-world systems. Additionally, we outline potential future directions for SABM, foreseeing a vast scope for its use.

Beyond the integration of LLMs, SABM distinguishes itself from ABM by facilitating modeling in natural language and adopting a priori modeling paradigm, two aspects not present in ABM. We advocate for foundational theories to gauge behavior complexity, helping to further contrast SABM from ABM. This also aids in understanding the extent to which LLMs can emulate human behaviors and in defining the application scope boundaries of SABM. From the technical perspective, more advanced prompt engineering may apply and improve the simulation performance. 

As a limitation, this paper lacks a detailed evaluation of the financial and computational expenses of SABM and the comparison with ABM in this aspect. Nonetheless, we report that the number-guessing game and the three case studies, including debugging and running simulations, yielded a total cost of around \$10,200, mostly spent on using \texttt{gpt-4-0314} whose price is \$0.03/k input tokens, and \$0.06/k output tokens. Even though LLMs can be expensive, advancing AI technologies might lower these costs, making them more accessible for researchers. While we emphasized LLMs with fixed parameters, fine-tuning them for real-world actions, especially in highly specialized tasks, is feasible. Additionally, LLMs have the potential to leverage external resources, like the internet, a facet not covered in this study but could enhance their efficacy in complex tasks. While we discussed text mining methods for analyzing results, we did not provide an illustrative example. 

Through the three case studies, our objective was to showcase the SABM methodology. The outcomes jointly demonstrated SABM's capability in simulating human decision-making with increased nuance and realism. They were mainly designed to display the qualitative advantage over ABM, e.g., in modeling conversations, which are difficult for ABM. Future investigations may reveal the quantitative advantage of SABM. For example, in a scenario that can be modeled in both approaches, SABM might yield higher accuracy and predictability when measured against real data. Another issue in our case studies is reproducibility. Whereas the case study of plea bargaining demonstrates statistical significance, more sample runs for the other two case studies are needed. Moreover, due to the use of a methodology featuring incremental substitution of ABM components, some model components still follow an analytical paradigm used in the original ABM setup. By replacing these components with an SABM setup and adding more SABM features, subsequent studies might delve deeper into these case studies, potentially unveiling new insights into the three scenarios investigated. Another direction worth exploring is expanding the scope to disciplines other than social science.

\section*{Acknowledgments}
This work was partially supported by JSPS Kakenhi 22H03903, 23H03406, 23K17456, 21K19767, 22KJ1721, and JST CREST JPMJCR22M2. We thank Ms. Huizhong Liu for providing technical support in statistical analysis. We thank Prof. Makoto Onizuka and Prof. Yuya Sasaki for providing financial and equipment support for completing this research.

\bibliographystyle{abbrv}
\bibliography{references-abm,references-ai,references-cases}

\end{document}